\documentclass[a4paper,12]{article}
\usepackage{amsmath, amsfonts, amssymb, amsthm, bm, graphics, bbm, color}
\usepackage{graphicx}
\usepackage{subfigure}
\usepackage{mathtools}
\usepackage[table]{xcolor}
\usepackage{url}
\usepackage{mathabx}
\usepackage{cancel}
\usepackage{multicol}
\usepackage{float}
\usepackage{caption}
\usepackage{mathalfa}
\usepackage{hyperref}
\usepackage{afterpage}
\usepackage{multirow}
\DeclareMathAlphabet\mathbfcal{OMS}{cmsy}{b}{n}

\theoremstyle{definition}

\DeclareMathOperator*{\argmax}{arg\,max}

\newcommand{\im}{\mathrm{i}}
\renewcommand{\vec}[1]{\mbox{\boldmath$#1$}}

\begin{document}
\title{Identification of Metallic Objects using Spectral Magnetic Polarizability Tensor Signatures: Object Classification}

\author{B.A. Wilson$^*$, P.D. Ledger$^\dagger$  and W.R.B. Lionheart$^\ddagger$\\
$^*$Zienkiewicz Centre for Computational Engineering, Swansea University \\
$^\dagger$School of Computing \& Mathematics, Keele University \\
$^\ddagger$Department for Mathematics, The University of Manchester\\
Corresponding author: p.d.ledger@keele.ac.uk}

\maketitle

\section*{Abstract}

The early detection of terrorist threat objects, such as guns and knives, through improved metal detection, has the potential to reduce the number of attacks and improve public safety and security. To achieve this, there is considerable potential to use the fields applied and measured by a metal detector to discriminate between different shapes and different metals since, hidden within the field perturbation, is object characterisation information. The magnetic polarizability tensor (MPT) offers an economical characterisation of metallic objects and its spectral signature provides additional object characterisation information. The MPT spectral signature can be determined from measurements of the  induced voltage  over a range frequencies in a metal signature for a hidden object. With classification in mind, it can also be computed in advance for different threat and non-threat objects. In the article, we evaluate the performance of probabilistic and non-probabilistic machine learning algorithms, trained using a dictionary of computed MPT spectral signatures, to classify objects for metal detection. We discuss the importances of using appropriate features and selecting an appropriate algorithm depending on the classification problem being solved and we present numerical results for a range of practically motivated metal detection classification problems.\\

{\bf Keywords}: Finite element method; Magnetic polarizability tensor; Machine learning; Metal detection; Object classification; Reduced order model; Spectral; Validation. \\

{\bf MSC CLASSIFICATION}: 65N30; 65N21; 35R30; 35B30

\section{Introduction}

The purpose of this paper is to compare the performance of probabilistic and non-probabilistic machine learning (ML) algorithms, trained using object characterisation information, to classify objects for metal detection. Key applications include in the discrimination between threat and non-threat objects in security screening, whereby the early detection of knives and guns has the potential to reduce the number of attacks and improve safety and security. Further applications include distinguishing between metallic clutter (e.g ring-pulls, coins, shrapnel) and threat items for the improved identification of hidden anti-personnel mines and unexploded ordinance (UXO) in areas of former conflict, improving identification of metallic objects of significance in archaeological searches and treasure hunts, improving non-destructive testing as well as discriminating between real and fake coins in vending machines and automatic checkouts.

This paper builds on our earlier work~\cite{ledgerwilsonamadlion2021} and our recent \texttt{MPT-Library} open data set~\cite{bawilson94_2021_4876371}, where we have established an extensive suite of simulated  threat and non-threat object characterisations using their magnetic polarizability tensor (MPT) spectral signatures. The complex symmetric MPT 
 \begin{align}
 {\mathcal M}[ \alpha B,\omega , \sigma_*, \mu_r ]&= (
 {\mathcal M}[ \alpha B,\omega , \sigma_*, \mu_r ])_{ij} {\vec e}_i \otimes {\vec e}_j \nonumber \\
  & = ( \tilde{\mathcal R}[ \alpha B,\omega , \sigma_*, \mu_r ] + \im{ \mathcal I}[ \alpha B,\omega , \sigma_*, \mu_r ]) _{ij} {\vec e}_i \otimes {\vec e}_j ,
\end{align} 
has been shown to offer an economical characterisation of conducting permeable objects and explicit formulae for the computation of its 6 independent complex coefficients, $  {\mathcal M}[ \alpha B,\omega , \sigma_*, \mu_r ])_{ij} $, which are a function of the exciting angular frequency, $\omega$, the object's size, $\alpha$, its shape $B$ as well as its conductivity, $\sigma_*$, and  relative permeability, $\mu_r$, have been obtained~\cite{LedgerLionheart2015,LedgerLionheart2018,LedgerLionheart2016,LedgerLionheartamad2019}. In the above,
 $ \tilde{\mathcal R}[ \alpha B,\omega , \sigma_*, \mu_r ]$ and ${\mathcal I}[ \alpha B,\omega , \sigma_*, \mu_r ]$ denote the MPT's real imaginary parts, $\im: =\sqrt{-1}$ and ${\vec e}_i$ is the $i$th orthonormal unit coordinate vector. The MPT spectral signature refers to the fact that the MPT is captured as a function of frequency, which has been shown to provide important additional object characterisation information~\cite{LedgerLionheart2019}. Our suite of object characterisations has been obtained using our \texttt{MPT-Calculator} software~\cite{mpt-calculator}, which employs  a reduced order model based on proper orthogonal decomposition (POD) for efficient calculations and an a--posteriori error estimate to certify its predictions with respect to full order model solutions provided by the open source finite element library \texttt{NGSolve}~\cite{NGSolvecode,NGSolve,SchoberlZaglmayr2005,netgendet}. Our computational MPT spectral signature simulations are in excellent agreement with measured MPT spectral signatures~\cite{toykan2021,paultoykan2021} and, thus, they provide realistic object characterisations.

Measured MPT spectral signatures have previously been used in conjunction with a $k$ nearest neighbours (KNN) classification algorithm~\cite{marsh2014b} and other ML approaches~\cite{WoutervanVerre2019}.  In addition, existing examples of practical MPT classification of objects  include in airport security screening~\cite{marsh2014,marsh2014b}, waste sorting~\cite{karimian2017} and anti-personal landmine detection~\cite{rehim2016}. 
In such situations, induced voltages are measured over a range of frequencies by a metal detector from which the MPT spectral signature of the hidden object is obtained and then a classifier applied~\cite{zhao2016,zhao2014,rehim2015}. However, dictionaries of measured MPT coefficients contain unavoidable errors if the object is placed in a non-uniform background field that varies significantly over the object as well as other errors and noise associated with capacitive coupling with other low-conducting objects or soil in the background. There will also be other general noise (e.g. from amplifiers, parasitic voltages and filtering)~\cite{Makkonen2015}. This means the accuracy of the measured MPT coefficients is about 1\% to 5\%~\cite{davidsoncoins, marsh2014b,Makkonen2015}, depending on the application. Furthermore, obtaining a large dictionary is time-consuming and there may be practical challenges in acquiring a sufficiently large selection of objects.

Simulated MPT spectral signatures and a simple dictionary classifier were employed for homogenous~\cite{Ammari2015}  and inhomogeneous~\cite{LedgerLionheartamad2019} objects, respectively, but the full potential of using a dictionary of simulated MPT spectral signatures and  ML classifiers has yet to investigated.   Dictionaries of simulated MPT spectral signatures allow noise at an appropriate level to be desired application to be added and the richness of the dataset allows them to be applied to a large range of different metal detectors operating at different frequencies. Furthermore, our  \texttt{MPT-Library} can be further extended to characterise other objects of different sizes and with different conductivities using simple scaling results at negligible computational cost~\cite{ben2020}, which make it ideal for creating (very) large datasets needed for training ML classifiers that would be impractical with measured MPT spectral signatures .

In this work, we present the first systematic review of the performance of a wide range of probabilistic and non-probabilistic ML classifiers for discriminating between different classes of metallic objects using a dictionary of simulated MPT spectral signatures. The practical classification problems we consider are relevant for discriminating between threat and non-threat objects in security screening and  discriminating between real and fake coins in vending machines and automatic checkouts. However, our paper is not only of interest to metal detection experts, but is also of 
 educational value to engineers interested in learning about ML classification algorithms. In particular, the classification problem we present is of a  medium size, which cannot  be trivially solved by hand,  is different to well known image classification, and, therefore, presents an interesting test case for investigating the performance of different ML algorithms.
 
The novelties of our article are as follows:  We present a novel approach to training ML classifiers using our simulated  \text{MPT-Library} enhanced by simple scaling results to create large dictionaries of objects. We choose tensor invariants of MPT spectral signatures as novel object features and explain the benefits of our approach over eigenvalues of MPT used in 
all previous ML classifiers for this problem. We include a novel well-reasoned approach for justifying the performance of different ML classifiers for practical classification problems using uncertainty quantification, statistical analysis and ML metrics. Furthermore, we explore the ability of our classification approaches to classify unseen threat objects and also discuss the limitations of our current approach.

The presentation of the material proceeds as follows: In Section~\ref{sect:classinv} we briefly review the use of invariants of MPT spectral signatures as ML features and describe the creation of our dictionary for training ML algorithms. Then, in Section~\ref{sect:classify}, we review a range of probabilistic and non-probabilistic classifiers that will be considered in this work and describe a simple approach for investigating uncertainty. Section~\ref{sect:metrics} discusses a range of ML metrics for assessing the performance of classifiers. This is followed in Section~\ref{sect:results} by the application of the classifiers to a series of practically motivated classification problems, where each dataset is analysed and the suitability of ML classifiers investigated and their subsequent performance  critically analysed.
We finish with some concluding remarks.

\section{MPT Spectral signature invariants for object classification} \label{sect:classinv}

Bishop~\cite{bishopbook} describes the process of classification  as taking an input vector ${\mathbf x}$ and assigning it to one of $K$ discrete classes $C_k$, $k=1,\ldots,K$. For example, in security screening, the simplest form of classification with $K=2$ involves only the classes {\em threat object} ($C_1$) and {\em non-threat object} ($C_2$), and one with a higher level of fidelity might include the classes  of metallic objects such as key ($C_1$), coin ($C_2$), gun ($C_3$), knife ($C_4$) ... where the class numbers are assigned as desired. He recommends that it is convenient to use a 1-of-$K$ coding system in which the entries in a  vector ${\mathbf t}\in {\mathbb R}^K$ take the form
\begin{equation}
{\rm t}_i := \left \{ \begin{array}{ll} 1 & \text{if $i=k$} \\
0 & \text{otherwise} \end{array} \right . ,
\nonumber 
\end{equation}
 if the correct class is $C_k$.
 Requiring that we always have $\sum_{k=1}^K {\rm t}_k=1$, then this approach has the advantage that ${\rm t}_k$ can be interpreted as the probability that the correct class is $C_k$.  
 
 In~\cite{ledgerwilsonamadlion2021} we have considered different choices for the $F$  features in the input vector ${\mathbf x}\in{\mathbb R}^F$, which are associated with either the eigenvalues, principal invariants or deviatoric invariants of $\tilde{\mathcal R}[ \alpha B,\omega , \sigma_*, \mu_r ]$ and ${\mathcal I}[ \alpha B,\omega , \sigma_*, \mu_r ]$, respectively, evaluated at different frequencies $\omega=\omega_m$,  $m=1,\ldots, M$. In this paper we restrict our focus to the situation where
 \begin{equation}
{\rm x}_i = \left \{ \begin{array}{ll} I_j ( \tilde{\mathcal R} [\alpha B,\omega_m,  \sigma_*,\mu_r  ]), & i= j+ (m-1)M  \\
I_j ( {\mathcal I} [\alpha B,\omega_m,  \sigma_*, \mu_r   ]), & i=  j+ (m+2)M
 \end{array} \right . , \label{eqn:featinv}
\end{equation}
with $j=1,2,3$ , $m=1,\ldots , M$ and, for a rank 2 tensor ${\mathcal A}$,
\begin{subequations}\label{eqn:tensorinvarients}
\begin{align}
I_1 ({\mathcal A}):= & \text{tr} ({\mathcal A}) , \\
I_2 ({\mathcal A}):= & \frac{1}{2} \left ( \text{tr} ({\mathcal A})^2 - \text{tr}({\mathcal A}^2) \right ),  \\
I_3 ({\mathcal A}):= & \text{det}({\mathcal A})  ,
\end{align}
\end{subequations}
are the
principal invariants,
where $\text{tr}(\cdot)$ denotes the trace and $\text{det}(\cdot)$ the determinate. We define $\mathbf x$ in this way for the following reasons:
\begin{enumerate}
\item Using features that are invariant to object rotation is important as both a hidden object's shape and its orientation are unknown. Using either eigenvalues $\lambda_i( \tilde{\mathcal{R}} )$,  $\lambda_i({\mathcal I})$, $i=1,2,3$  or the principal tensor invariants overcomes this issue as both are invariant to an object's unknown orientation and, hence, simplifies the classification problem. 
 \item Invariants overcome the ordering issue that is associated with assigning the eigenvalues  as the invariants are independent of how the eigenvalues are assigned.
 \item The invariants can be computed as either products or sums from the tensor coefficients. Hence, they are smooth functions of the tensor coefficients. On the other-hand, determining the eigenvalues involves finding roots and may involve singularities if eigenvalues are equal. The root finding process may result in a loss of accuracy compared to just simple sums or products.
\item The (probabilistic) ML classification algorithms we will consider are better at capturing an underlying relationship between features and the likelihood of class  that is smooth, albeit, with noisy data. The root-finding process involved in finding eigenvalues may lead to a greater entanglement between class and features  that a classifier might need to unravel compared to using invariants. 
 \end{enumerate}

\begin{figure}[!h]
\centering
\hspace{-1.cm}
$\begin{array}{cc}
\includegraphics[scale=0.5]{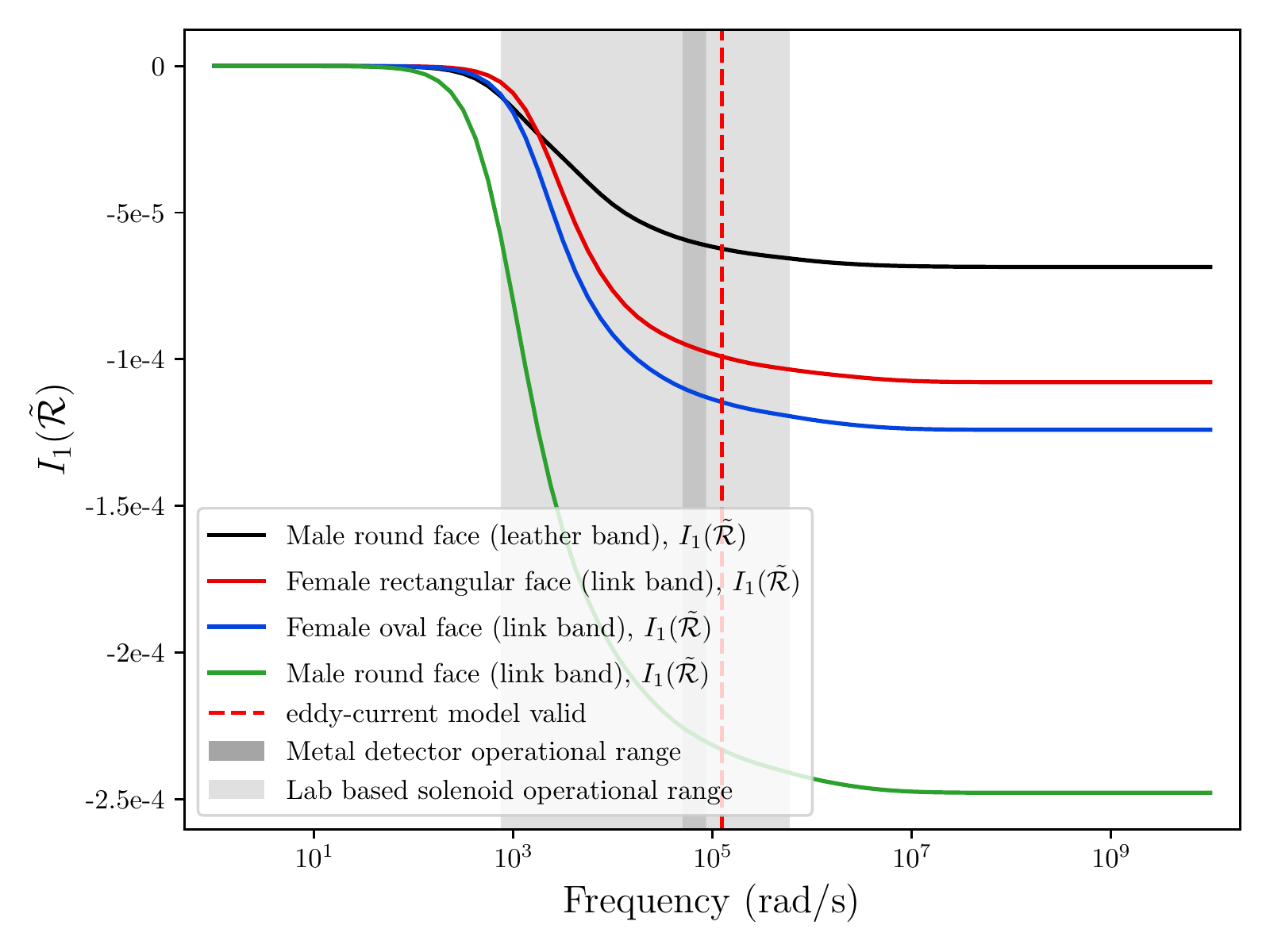}  &
\includegraphics[scale=0.5]{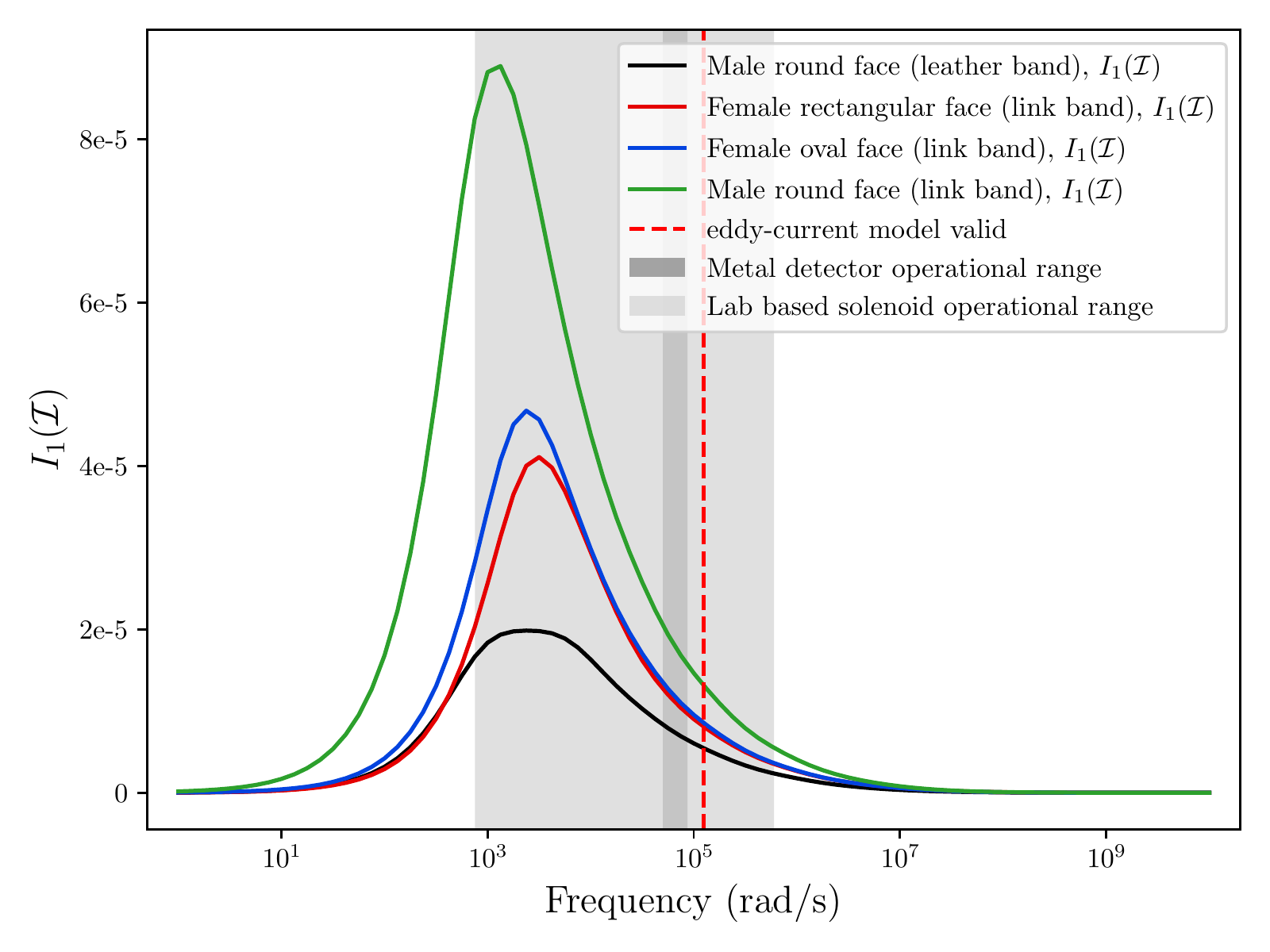}  \\
\text{(a) } I_{1} ( \tilde{\mathcal{R}} ) & 
\text{(b) } I_{1} ( \mathcal{I} )  \\
\includegraphics[scale=0.5]{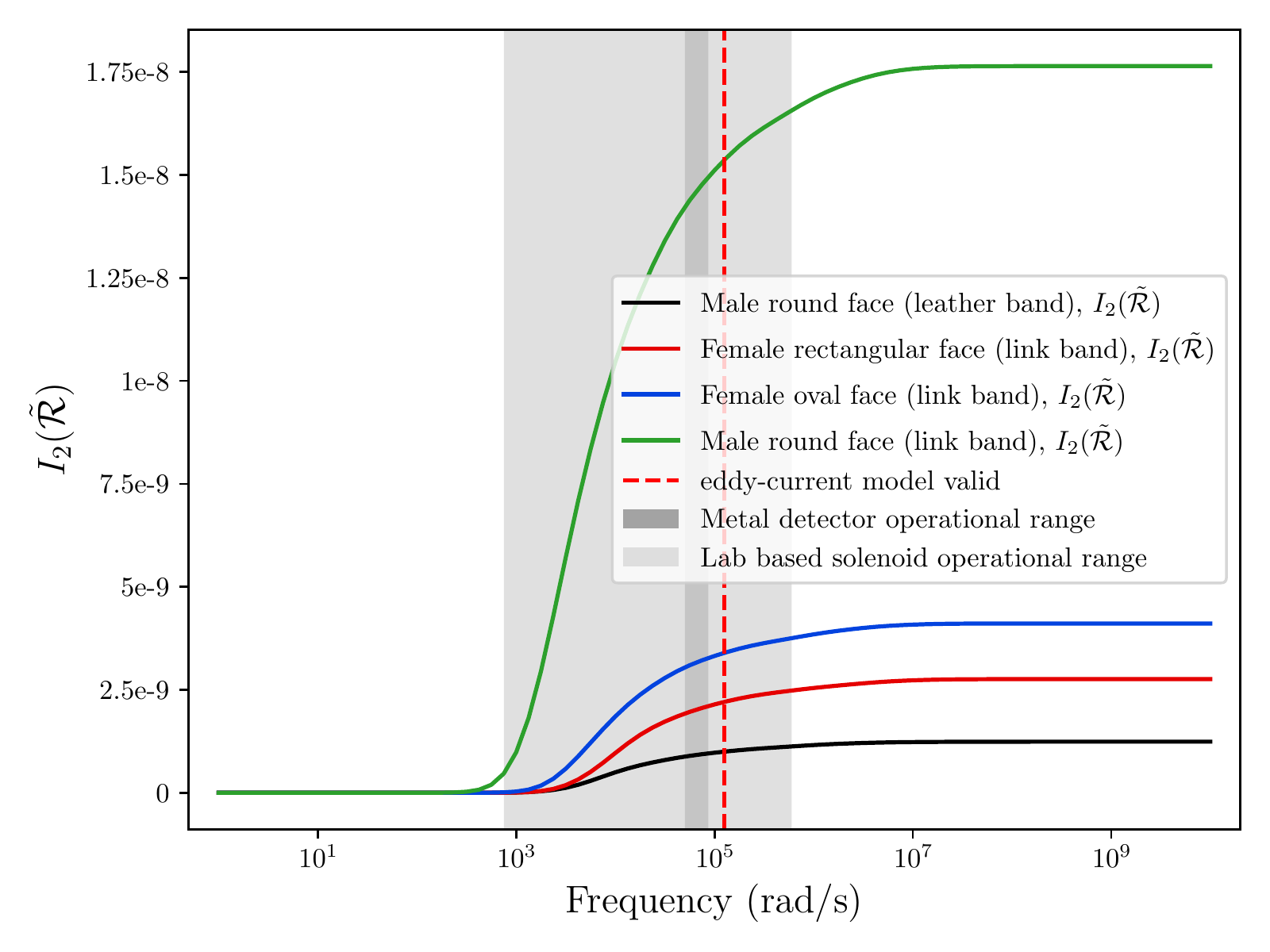}  &
\includegraphics[scale=0.5]{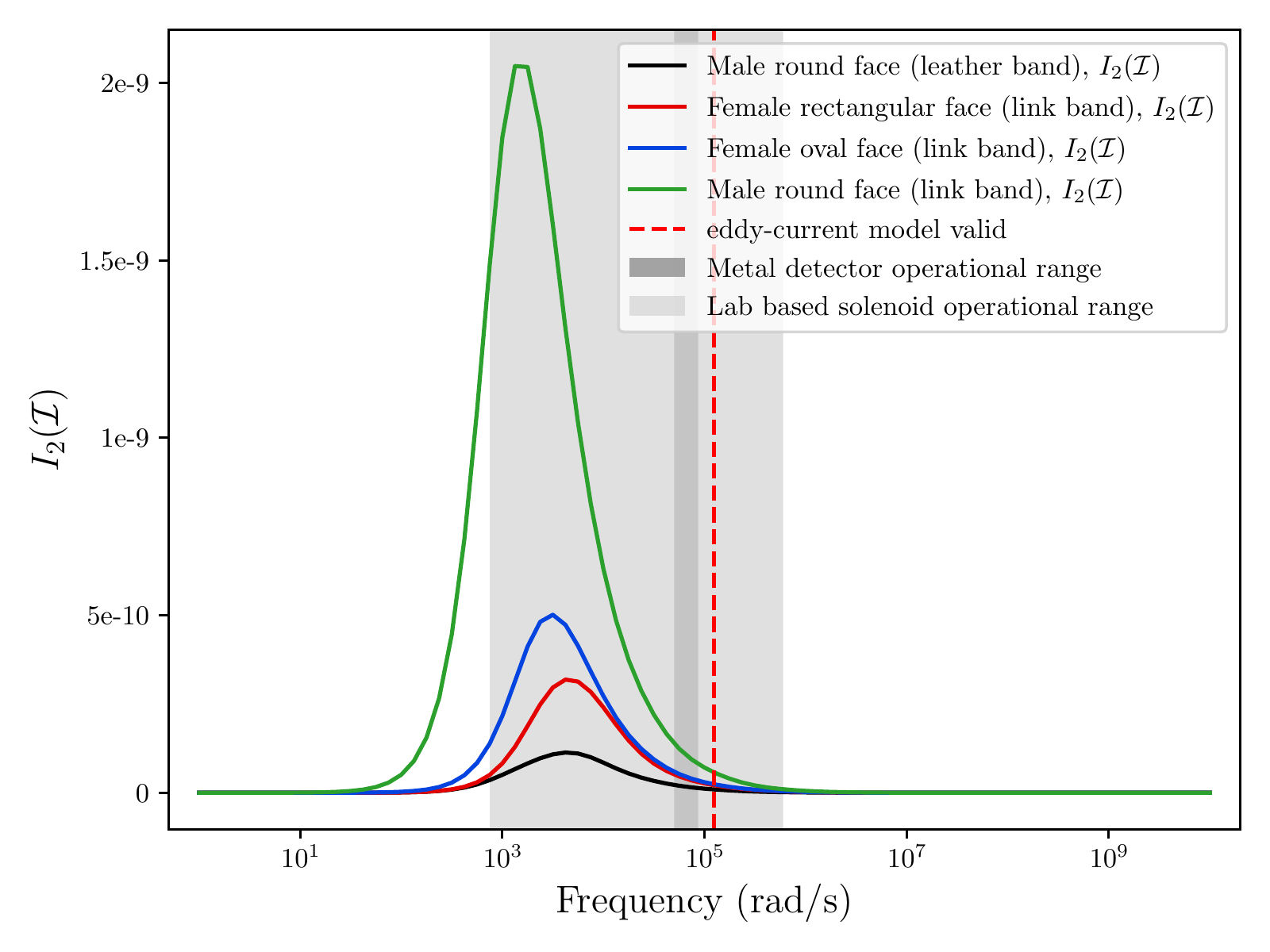}  \\
\text{(c) } I_{2} ( \tilde{\mathcal{R}} ) & 
\text{(d) } I_{2 } ( \mathcal{I} )  \\
\includegraphics[scale=0.5]{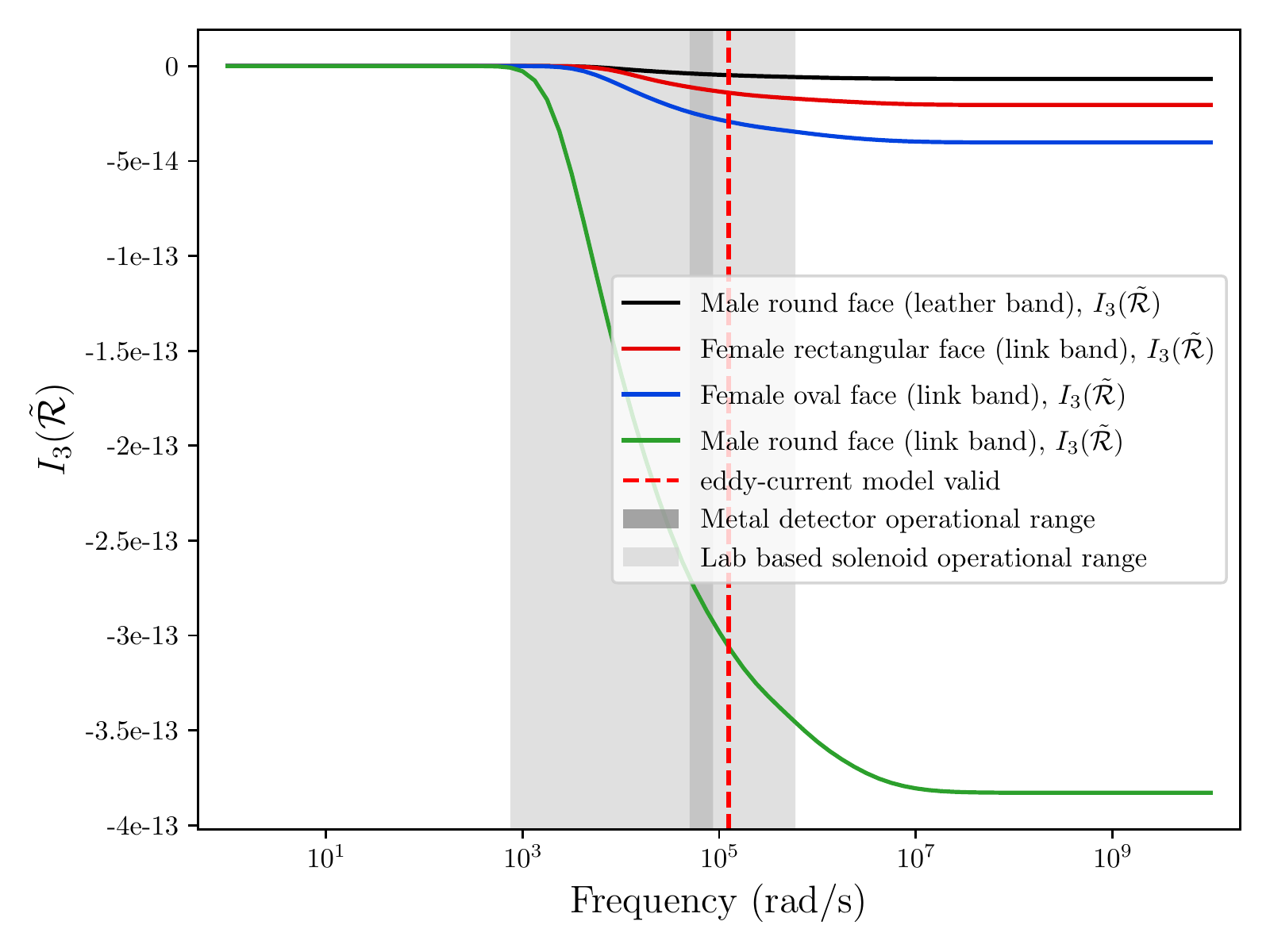}  &
\includegraphics[scale=0.5]{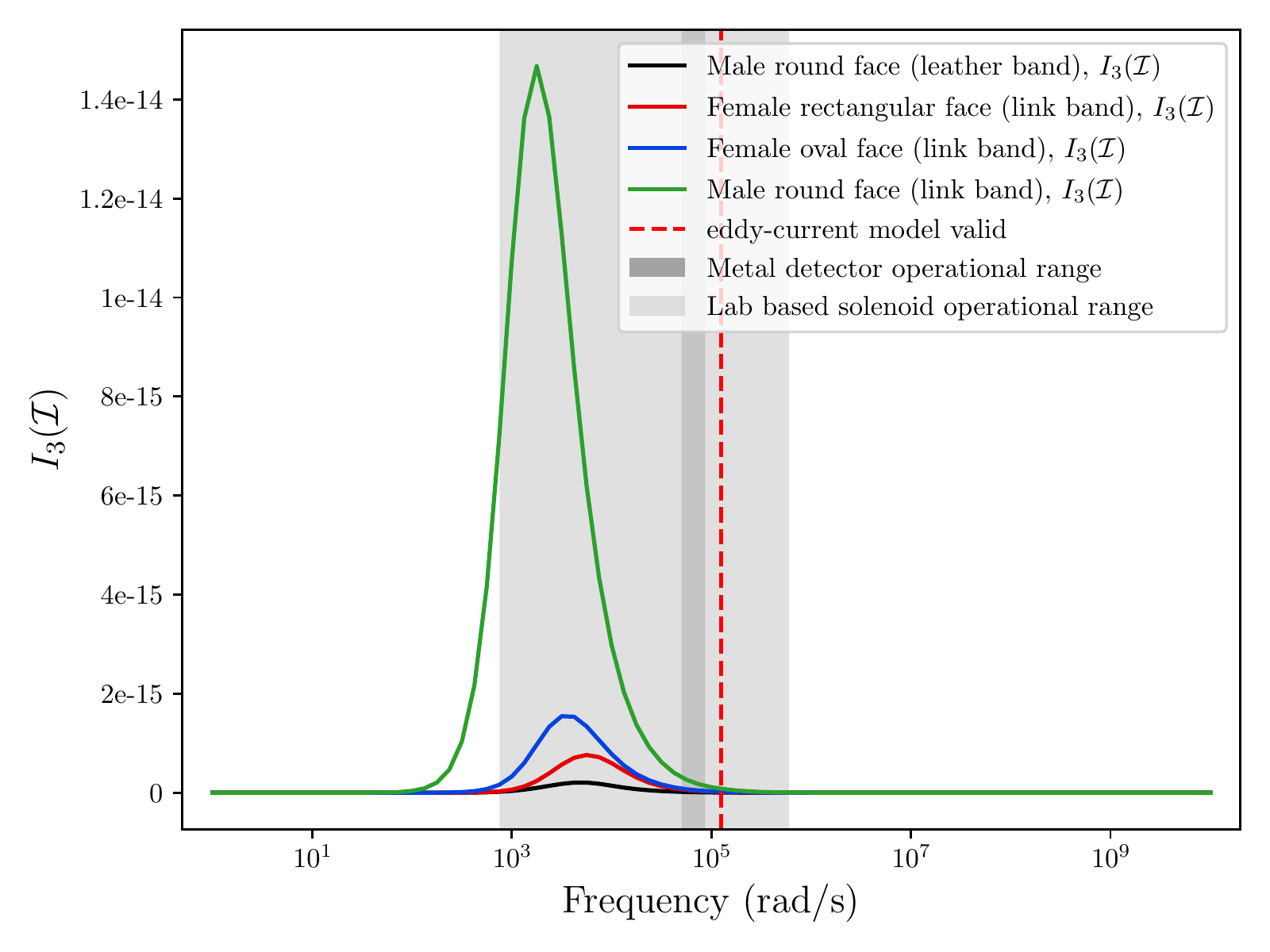}  \\
\text{(e) } I_{3} ( \tilde{\mathcal{R}} ) &
\text{(f) }  I_{3} ( \mathcal{I} )  
\end{array}$
  \caption{Set of watches: Comparison of tensor invariants. (a) $I_{1} ( \tilde{\mathcal{R}} ) $, (b) $I_{1} ( \mathcal{I} ) $
  (c) $I_{2} ( \tilde{\mathcal{R}} ) $, (d) $I_{2} ( \mathcal{I} ) $,
  (e) $I_{3} ( \tilde{\mathcal{R}} ) $ and (f)  $I_{3} ( \mathcal{I} ) $.}
        \label{fig:Watch_Invarients}
\end{figure}

 As an example, Figure \ref{fig:Watch_Invarients} shows a comparison of the principal tensor invariants for a selection of 4 different metallic watch styles computed by the \texttt{MPT-Calculator} software. These calculations were performed in a similar manner to those  for other geometries in~\cite{ledgerwilsonamadlion2021}. The object dimensions are in mm so $\alpha =0.001$m and the results shown are for where the material is gold, so that $\sigma_* = 4.25 \times10^7 $ S/m and $\mu_r=1$ (\texttt{MPT-Library} also includes MPT spectral signatures for watches made of platinum and silver). An unstructured mesh of tetrahedra is used to discretise each object and the truncated unbounded region which surrounds it, resulting in meshes ranging from 14935 to 175217 elements. In each case, the truncated boundary for the non-dimensional transmission problem is $[-1000,1000]^3$. Order  $p=4$ elements were applied on the meshes and snapshot  solutions obtained at 13 logarithmically spaced frequencies over the range $1  \text{ rad/s}\leq\omega\leq 1 \times 10^{10}\text{ rad/s}$. The MPT spectral signature for each object was produced using the
 projected proper orthogonal decomposition (known as PODP) method~\cite{ben2020} using  a relative singular value truncation of $10^{-4}$. Also shown is a vertical line, which indicates the value of $\omega$ that the eddy current model assumption is likely to become inaccurate for this geometry~\cite{ledgerwilsonamadlion2021,schmidteddycurrent}. Finally, we include a grey window corresponding to the frequency range  $5.02\times 10^4 \text{ rad/s} \le \omega \le 8.67 \times 10^4 \text{ rad/s}$, where measurements taken by a commercial walk through metal detector~\cite{marsh2014b}, and the greater range $7.53 \times 10^2 \text{ rad/s} \le \omega\le 5.99 \times 10^5 \text{ rad/s}$, where measurements are taken using recent MPT measurement system~\cite{toykan2021}, the latter being able to capture more information from the signature.
  These spectral signatures will form part of the dictionary for object classification, which will discussed later in Section~\ref{sect:multi}.

 \subsection{Construction of the dictionary} \label{sect:dict}
 Each class $C_k$ may  be comprised of $G^{(k)}$ geometries and, in addition, we consider $V^{(k)}$ variations in object size and object materials so that each class is comprised of $P^{(k)}$ different samples. In total, over all the classes, we have $P=\sum_{k=1}^KP^{(k)}$ samples.

Given the information $\alpha$, $B$, $\sigma_*$, $\mu_r$  the MPT spectral signature described by  $\tilde{\mathcal R}[\alpha B,\omega,\sigma_*,\mu_r]$ and ${\mathcal I}[\alpha B,\omega,\sigma_*,\mu_r]$  can be obtained, as described above, and  then invariants then  follow from (\ref{eqn:featinv}).  
We can repeat this process for each of the geometries $B^{(g_k)}$, $g_k=1,\ldots,G^{(k)}$ that makes up the class.   To take account of the $V^{(k)}$ different object sizes and materials, we draw physically motivated samples $\alpha\sim N(m_\alpha, s_\alpha)$ and $\sigma_* \sim N(m_{\sigma_*},s_{\sigma_*})$,
where $m_\alpha$ and $m_{\sigma_*}$ denotes means and $s_\alpha$ and $s_{\sigma_*}$ standard deviation, respectively, and $N(m,s)$ denotes a normal distribution with mean $m$ and standard deviation $s$. While it would be possible to also obtain the MPT spectral signature using the \texttt{MPT-Calculator} software for each sample, instead, we reduce the computational cost of obtaining these spectral signatures by using the the simple scaling results we have derived in Lemmas 2 and 3 of~\cite{ben2020}. This means that the spectral signatures for each of these additional samples can be generated at negligible computational cost. The  invariants then again follow from (\ref{eqn:tensorinvarients}). Finally, by setting the class labels, the pairs (${\mathbf x}_p  \in {\mathbb R}^{F}$, ${\mathbf t}_p \in {\mathbb R}^K $), $p=1\ldots,P^{(k)}$~\footnote{Note that the entries of ${\mathbf t}_p$ are all $0$ except for $({\mathbf t}_p)_{C_k}=1$ corresponding to the $k$th class} for all the samples that make up the class $C_k$ are obtained. 
Repeating this process for each of the classes gives rise to the the general dictionary
 \begin{equation}
 D = ( ({\mathbf x}_1, {\mathbf t}_1) ,  ({\mathbf x}_2, {\mathbf t}_2) , \ldots, ({\mathbf x}_P, {\mathbf t}_P) ) , \label{eqn:dictionary}
 \end{equation}
  or alternatively,
\begin{equation}
D=(D^{(1)},D^{(2)},...,D^{(K)}),
\end{equation}
where,
\begin{equation}
D^{(k)}=((\mathbf{x}_1,\mathbf{t}_1),(\mathbf{x}_2,\mathbf{t}_2),...,(\mathbf{x}_{P^{(k)}},\mathbf{t}_{P^{(k)}})), \label{eqn:dictionaryk}
\end{equation}
is the dictionary associated with class $C_k$ and consists of $P^{(k)}$ observations.

 In practice, we split the dictionary stated in (\ref{eqn:dictionary})  as $D = (D^{\text{(train)}},D^{\text{(test)}})$ where $D^{\text{(train)}}$ is the training  and  $D^{\text{(test)}}$ is the testing dataset, respectively. The purpose of this splitting is to enable the classifier to be trained on given data $D^{\text{(train)}}$ and then tested on previously unseen data $D^{\text{(test)}}$. Although there is no optimal choice for the ratio of training to testing, we will employ a ratio  of  3:1 throughout, which is commonly used in ML classification.

\subsection{Noise} 

When MPT spectral signatures for hidden objects are measured by a metal detector they will contain un-avoidable errors, as pointed out in~\cite{ben2020}. For example, if an  object is placed in a non-uniform background magnetic field that varies significantly over the object there is a modelling error since the background field in the rank 2 MPT model assumes the field over the object is uniform. There are other errors and noise associated with capacitive coupling with other low-conducting objects or soil, if the object is buried,  as well as other generals noise (e.g. from amplifiers, parasitic voltages and filtering)~\cite{Makkonen2015}.  The accuracy of the signature can be improved by repeating the measurements and applying averaging filters, at the cost of spending more time to take the measurements. However, in a practical setting, there is trade-off to be made in terms of improving the accuracy against the meaurement time 
 and, consequently, the accuracy of the measured MPT coefficients is about 1\% to 5\%~\cite{davidsoncoins, marsh2014b,Makkonen2015}, depending on the application.

The MPT spectral signature coefficients we will use have been produced numerically using our \texttt{MPT-Calcu} \texttt{lator} software tool~\cite{ben2020,ledgerwilsonamadlion2021}. This means that the MPT coefficients are obtained with higher accuracy than can currently be achieved from measurements,  since the spectral signature is accurately computed for a large frequency range (up to the limit of the eddy current model) rather than noisy measurements being taken at a small number of discrete frequencies.
The advantage of this  is it allows a much larger library of objects and variations of materials to be considered, which is all highly desirable for achieving greater fidelity and accuracy when training a ML classifier. But, for practical classification, noise appropriate to the system must be added.

After filtering and averaging, the noise remaining in an MPT spectral signature measured by a metal detector can be well approximated by Gaussian additive noise, with a level dependent on the factors identified above. 
Hence, in this work, we add noise to our simulated MPT spectral signatures in the following way: We specify a signal noise ratio (SNR) in decibels and use this to determine the amount of noise to add to each of the complex tensor coefficients $({\mathcal M}[\alpha B^{(p)} , \omega,\sigma_*,\mu_r] )_{ij}   $ as a function of frequency for each object $\alpha B^{(p)} $ in the dictionary. Considering each of the $i,j$th MPT coefficients individually, we introduce 
\begin{equation}
{\rm v}_m := ({\mathcal M}[\alpha B^{(p)} , \omega_m ,\sigma_*,\mu_r] )_{ij}, \nonumber
\end{equation}
and calculate a noise-power measure as
\begin{equation}
\text{noise} = \frac{\overline{\rm v}_m  {\rm v}_m}{10^{\text{SNR}/10}}. \nonumber
\end{equation}
The noisy coefficients are then specified as
\begin{equation}
 ({\mathcal M}[\alpha B^{(p)} , \omega_m ,\sigma_*,\mu_r] )_{ij}  +e_{ij} \qquad \text{where} \qquad e_{ij} = \sqrt{\frac{\text{noise}}{2}} (u + \im v)
\nonumber
\end{equation}
with $u,v\sim N(0,1)$. The above process is  repeated for the 6 independent coefficients of the complex symmetric MPT  and for each frequency in the spectral signature. SNR values of 40, 20 and 10dB lead to values of $|\frac{e_{ij}}{\mathcal{M}_{ij}}|=0.01,0.10$ and 0.32 on average, which is equivalent to 1\%, 10\% and 32\% noise, respectively. One realisation of the effect of the added noise on $I_1( \tilde{\mathcal R})$ and $I_1({\mathcal I})$ for a  British one penny coin can be seen in Figure ~\ref{fig:coins_noise_levels}.  Note that our model for noise results in a noise power measure that varies over the MPT spectral signature according to $\overline{\rm v}_m {\rm v}_m$. If the physical system behaves differently, this can be taken in to account by applying an appropriate model for the noise at this stage.  Once the noisy ${\mathcal M}$ at each $\omega_m$ are found, the principal invariants of the real and imaginary parts of ${\mathcal M}$ at each $\omega_m$ easily follow. Hence, an entry $({\mathbf x},{\mathbf t}) \in D$ is replaced with  $({\mathbf x}^{\text{noise}},{\mathbf t}) $.  By repeating this for all objects  leads to the updated dictionary $D$.

 \section{Classification} \label{sect:classify}
 
 \subsection{Probabilistic versus non-probabilistic classification}

Applied to classification problems, Bayes' theorem can be expressed in the form~\cite{bishopbook}
 \begin{equation}
 p (C_k | {\mathbf x}) = \frac{ p({\mathbf x}| C_k ) p( C_k) }{ p({\mathbf x})},\qquad k=1,\cdots,K  , \label{eqn:bayes}
 \end{equation}
which relates  the posterior probability density function $ p (C_k | {\mathbf x}) $ to the  likelihood probability density function $p({\mathbf x}| C_k )$ and   the prior probability density function $p( C_k)$ where, for classification,
 \begin{equation}
p({\mathbf x}) =  \sum_{n=1}^K  p({\mathbf x}| C_n ) p( C_n) ,
\nonumber 
\end{equation}
is easily  explicitly obtained as the normalising constant. 
 
In the inference stage of probabilistic classification, one seeks to design 
a classifier $\gamma_k({\mathbf x})$ that provides a probabilistic output, which approximates  $p (C_k | {\mathbf x})$. 
On the hand,  non-probabilistic classifiers either predict a class $C_k$ with certainty or, more commonly, have a statistical interpretation that provides a  frequentist approximation $\gamma_k({\mathbf x})$ to  $p (C_k | {\mathbf x})$. One measure of accuracy of classification is the mean squared error (MSE)
\begin{equation}
\text{MSE} (\gamma_k) = E_{\mathbf x} [ \gamma_k ({\mathbf x}) - p (C_k | {\mathbf x})]^2,
\end{equation}
where $E_{\mathbf x}$ is the expectation with respect to $p({\mathbf x})$~\cite{manningbook}[pg 309.]. If desired, this can be summed over the classes $k=1,\ldots,K$ or considered for each class. Other metrics are considered in Section~\ref{sect:metrics}. 

Given a classifier $\gamma_k({\mathbf x})\approx p (C_k | {\mathbf x})$, $k=1,\ldots,K$,  the class decision is typically achieved using the maximum a-posterior (MAP) estimate
$\argmax_{k\in K}(p(C_k|\mathbf{x}))$ i.e. the MAP estimate corresponds to the class  $C_n$ with $n$ such that
\begin{equation}
n=\argmax_{k\in K}(p(C_k|\mathbf{x})). \label{eqn:map}
\end{equation}
 However, the MAP may have drawbacks if the are several similar probabilities and/or if the data is noisy.
Hence, understanding the uncertainty in the approximations of  $ p (C_k | {\mathbf x})$ are also important. We consider this in Section~\ref{sect:uncertain}.

\subsection{Bias and variance}
The classifiers we will consider are based on ML algorithms.  For a ML method $\Gamma$, which takes ${\mathbb D} = D^{\text{train}}$ as the input and returns a learned classifier $\gamma_k$, $k=1,\ldots,K$, Manning, Raghavan and {Sch\"utze}~\cite{manningbook}[pg 309-312]
  define the $ \text{Learning-error}(\Gamma) = E_{\mathbb D}[ \text{MSE}( \Gamma_{\mathbb D})]$ as a measure of accuracy of the classifier, which is to be minmised, and they show that it can be expressed as
  \begin{align}
 \text{Learning-error}(\Gamma) := &  
 E_{\mathbf x} [ \text{bias} (\Gamma, {\mathbf x}) + \text{variance} (\Gamma, {\mathbf x}) ] , \nonumber  \\
 \text{bias} (\Gamma, {\mathbf x}) := & [ p(C_k  | {\mathbf x} ) - E_ {\mathbb D}\Gamma_{\mathbb D} ({\mathbf x}) ]^2  \nonumber  ,\\
 \text{variance} (\Gamma, {\mathbf x}) := &E_ {\mathbb D}[ \Gamma_{\mathbb D}({\mathbf x}) -   E_ {\mathbb D}\Gamma_{\mathbb D} ({\mathbf x}) ]^2  ,\nonumber
  \end{align}
  where $\Gamma_{\mathbb D}({\mathbf x}) $ implies that the ML method is applied to data set ${\mathbb D}$ and outputs  $\gamma_k({\mathbf x}) \approx p(C_k|{\mathbf x})$ for data ${\mathbf x}$.
  
Bias measures the difference between  the true $p(C_k  | {\mathbf x} ) $ and the prediction $\Gamma_{\mathbb D} ({\mathbf x})$ averaged over the training sets~\cite{manningbook}[pg 311]. Bias is large if the classifier is consistently wrong, which may stem from erroneous assumptions. A small bias may indicate several things and not just that the classifier is consistently correct, for further details see~\cite{manningbook}. 
Related to high bias is underfitting, which refers to a classifier that is unable to capture the relationship between the input and output variables correctly and produces large errors on both the training and testing data sets.

Variance is the variation of the prediction of learned classifiers and is the average difference between $\Gamma_{\mathbb D}({\mathbf x})$ and its average  $E_ {\mathbb D}\Gamma_{\mathbb D} ({\mathbf x})$~\cite{manningbook}[pg 311]. Variance is large if different training sets give rise to different classifiers and  is small if the choice of training set has only a small influence on the classification decisions, for further details see~\cite{manningbook}.
Related to high variance is overfitting, which is the opposite of underfitting, and refers to a classifier that has too much complexity and also learns from the noise resulting in high errors on the test data.

The ideal classifier would be a classifier having low bias and low variance, however, the two are inextricably linked and there is therefore a trade off between two~\cite{bishopbook,manningbook}. While the performance of classifiers  is very application dependent, linear classifiers tend to have a high bias and low variance and non-linear classifiers tend to have a low bias and high variance~\cite{manningbook}.
The ML methods $\Gamma$ that we will consider are commonly found in established ML libraries such as \texttt{scikit-learn}, which is the implementation that we will use. The finer details of the different methods can be found in~\cite{bishopbook,Hastie2009elements,geron2019hands} among many others, although we give a brief summary of the methodologies for those less familiar with the approaches and to set notation. We start with non-probabilistic methods and then proceed to probabilistic classifiers.

 \subsection{Non-probabilistic classifiers}\label{sect:noprob}
 
 \subsubsection{Decision trees}\label{sect:Decision_Tree}
Tree based algorithms can simply be interpreted as making a series of  (binary) decisions that ultimately lead to the prediction of a class. For $F=2$, this results in the partition of the feature space into a series of $K$ rectangular regions corresponding to the different classes. To establish the regions, and, hence the classes,  a tree is constructed where, at each node, a binary decisions about a component of ${\mathbf x}$,  and  the process is terminated with a decision at the leaf-node (for example, if ${\rm x}_1  \le t_1$ and ${\rm x}_2 \le t_2$ then the class is $C_1$, whereas if ${\rm x}_1  \le t_1$ and ${\rm x}_2 > t_2$ the class is $C_2$ etc).  
   In order to grow a tree, a greedy algorithm is applied to decide how to split the variables, the split points and the topology of the tree ~\cite{Hastie2009elements}[pg. 308].
  Growing a tree that is too large may overfit the data, while a small tree may not capture the structure. To overcome this, a larger than needed tree  is usually grown and is then pruned. 
To understand how this works, consider a tree, then by  applying the rules of the tree, a  subset of the training data for its $m$  node is obtained. It then follows that $\hat{p}_{mk}$ is the proportion of this data that has class $C_k$ and the associated class is determined as $k(m)=\text{arg max}_k\, \hat{p}_{mk}$. 
 The pruning is then achieved by applying a cost-complexity optimisation based on the node impurity measures   {\em misclassification error},  {\em Gini index} or {\em cross-entropy}, which  can each be written in terms of ((non-linear) functions and/or sums of) $\hat{p}_{mk}$. For further details see~\cite{Hastie2009elements}[pg 309]~\cite{bishopbook}[pg 666].  Decision trees are often used as a base classifier for more advanced ensemble methods such as gradient boost and random forests discussed below.

\subsubsection{Random forests}
The random forests classifier is an example of what is known as  ``a bootstrap aggregating (bagging) algorithm'' that attempts to reduce the variance of the typically high-variance low-bias decision tree algorithm~\cite{Hastie2009elements}[pg587]. As Hastie, Tibshirani and Friedman~\cite{Hastie2009elements}[pg587]  continue to describe, the idea behind bagging is to average many noisy, but unbiased models to reduce their variance. Trees are notoriously noisy, and hence benefit from averaging. Since each tree generated by bagging is identically distributed then expectation of an average of such trees is the same as the expectation of any of them. This means that bias unchanged, but random forests offer the hope of improvement by variance reduction. For details of their implementation we refer to~\cite{Hastie2009elements}[Chapt. 15].

 \subsubsection{Support vector machine}
A support vector machine (SVM) classifier generalises simple linear classifiers (e.g. Fisher's linear discriminant analysis) by producing non-linear, rather than linear, classification boundaries. These are obtained  by constructing a linear boundary in a transformed version of the feature space, which becomes non-linear in the feature space ~\cite{bishopbook}[pg. 325], ~\cite{Hastie2009elements}[Chapt 12], ~\cite{cristianini2000introduction}. So, after making a transformation ${\vec \psi} ({\mathbf x}): {\mathbb R}^F \to {\mathbb R}^{\tilde{F}}$ with $\tilde{F}\ge F$ being possibly infinite dimensional~\footnote{We will return their properties shortly}, the goal of the classifier is to learn how to  determine ${\mathbf w}$ and ${\rm w}_0$ in
\begin{equation}
G({\mathbf x})= {\mathbf w}^T {\vec \psi} ({\mathbf x}) + {\rm w}_0,
\end{equation}
with $\| {\mathbf w} \| =1 $ such that it predicts  $C_1  (say)$ if $G({\mathbf x})<0$ and $C_2 $ if  $G({\mathbf x})>0$.  For the separable case, the idea behind  SVM is to find the hyperplane that creates the biggest margin, defined by $2M$ between the training data describing the two classes.  Given $N$ training points $({\mathbf x}_1,{\rm y}_1), ({\mathbf x}_2,{\rm y}_2),\ldots, ({\mathbf x}_N,{\rm y}_N)$ with ${\rm y}_i\in \{ -1,1\} $ indicating the class label, then this problem can be framed as the optimisation problem
\begin{align}
\text{max}_{{\mathbf w}, {\rm w}_0,\| {\mathbf w} \| =1 }  &M  \nonumber \\
\text{subject to } {\rm y}_i ({\mathbf w}^T{\vec \psi}( {\mathbf x}_i) + {\rm w}_0) & \ge M, \qquad i=1,\ldots,N
\end{align}
which can also be rephrased as a convex optimisation problem (quadratic criterion and linear constraints). In the case of the non-separable case, slack variables $s_1,\ldots,s_N$ are introduced to deal with points that lie on the wrong side of the margin and the linear constraint is replaced by ${\rm y}_i ({\mathbf w}^T{\vec \psi}( {\mathbf x}_i) + {\rm w}_0) \ge M (1- s_i)$. For further details, and its computational implementation using Lagrange multipliers, see~\cite{Hastie2009elements}[pg. 420]. This practical implementation involves the introduction of symmetric positive definite or symmetric positive semi definite kernel functions $k({\mathbf x},{\mathbf x}') = {\vec \psi} ({\mathbf x}')^T {\vec \psi} ({\mathbf x})$~\cite{Hastie2009elements}[pg. 424], which avoids the  introduction of   ${\vec \psi} ({\mathbf x})$ itself, but imposes limitations on their choice in order that optimisation problem remains convex. Typically kernel types include polynomial, Gaussian, and radial basis function kernels, however, we limit our investigation to the latter in this paper.

To apply SVM to multi-class problems, the problem is reduced into a series of binary classification problems. This is done by employing either an \textit{ovo} (one versus one) or an \textit{ovr} (one versus rest) strategy, we have chosen the former, where an SVM is trained for each possible binary classification. This leads to $\frac{K(K-1)}{2}$ models being trained, for example in a 3-class problem ($K=3$) we would train a classifier to separate the pairs $(C_1,C_2)$, $(C_1,C_3)$ and $(C_2,C_3)$, we then create a voting scheme based on these classifiers.

 \subsection{Probabilistic classification} \label{sect:prob}

We will consider problems with $K>2$  and, in this case,
(\ref{eqn:bayes}) is often written in the form of the softmax function (also known as the normalised exponential)
 \begin{equation}
 p (C_k | {\mathbf x}) = \sigma({\mathbf x}): = \frac{ \exp a_k   }{ \sum_{k=1}^K \exp a_k },
 \label{eqn:bayes2}
 \end{equation}
where $a_k = \ln (p({\mathbf x}| C_k ) p(C_k))$.

If $p({\mathbf x}|C_k)$ has a simple Gaussian form,  the evaluation of $ p (C_k | {\mathbf x})$ for given ${\mathbf x}$ becomes explicit. However, in many practical cases, $p({\mathbf x}|C_k)$ will have a complicated form and this will dictate the use of classifier  $\gamma_k ({\mathbf x})$  that approximates $ p (C_k | {\mathbf x})$ instead.
Nonetheless, all probabilistic classifiers benefits from establishing (approximations of) the likelihood of each of the classes $C_k$, $k=1,\ldots,K$ rather than just a single output. 

We explore some alternative ML methods $\Gamma$, which provide probabilistic classifiers, below.

\subsubsection{Logistic regression}\label{sect:logreg}
In the case that $p({\mathbf x}|C_k)$ has a simple Gaussian form, a suitable linear classifier is logistic regression~\cite{bishopbook}[Chapt. 4]. This is based on the following assumptions
\begin{enumerate}
\item The likelihood probability distribution is Gaussian:
\begin{equation}
p({\mathbf x}| C_k ) = \frac{1}{(2 \pi)^{F/2} | {\mathbf \Sigma}|^{1/2}} \exp \left (  - \frac{1}{2} ( {\mathbf x}- {\mathbf m}_k)^T {\mathbf \Sigma}^{-1} ( {\mathbf x}- {\mathbf m}_k)  \right ) \nonumber,
\end{equation}
where $\mathbf{m}_k$  is the mean of all ${\mathbf x}_i$, $({\mathbf x}_i,{\mathbf t}_i)\in D^{(k)}$, that are associated with class $C_k$ and ${\mathbf \Sigma}$ is a covariance matrix.
\item The covariance matrix ${\mathbf \Sigma}$ is common to all the classes.

\end{enumerate}

In this case, the evaluation of $p(C_k|{\mathbf x})$ is explicit  with $a_k$ in  (\ref{eqn:bayes2})  replaced with the rescaled $\tilde{a}_k$ for $K>2$~\cite{bishopbook} 
\begin{equation}
\tilde{a}_k =  \tilde{\mathbf w}_k^T{\mathbf x} + \tilde{{\rm w}}_{k0} , \nonumber 
\end{equation}
where 
\begin{align}
\tilde{\mathbf w}_k & = {\mathbf \Sigma}^{-1} {\mathbf m}_k , \nonumber \\
\tilde{{\rm w}}_{k0} & = - \frac{1}{2} {\mathbf m}_k  {\mathbf \Sigma}^{-1} {\mathbf m}_k+ \ln p(C_k) . \nonumber 
\end{align}
In the generative approach, the learning involves first computing ${\mathbf m}_k$ and ${\mathbf \Sigma}$ directly from the training data $D^{(\text{train})}$ using while in discriminative approach, the $(K-1)(F+1)$ coefficients of $ \tilde{\mathbf w}_k$ and $ {\rm w}_{k0}$ compared to the $KF +F^2/2$ coefficients needed otherwise are found by numerical optimisation from  $D^{(\text{train})}$. When applied to other data sets, this results in an approximate classifier $\gamma_k ({\mathbf x}) \approx p (C_k |{\mathbf x})$. Note, that only $\tilde{\mathbf w}_k$ and $ \tilde{{\rm w}}_{k0}$  for $k=1,\cdots, K-1$ need to be determined since we know $\sum_{k=1}^K p (C_k |{\mathbf x})=1$, which allows $\gamma_K({\mathbf x}) \approx  (C_K |{\mathbf x})$  to be found from the others.
 
 \subsubsection{Multi-layer perceptron}\label{sect:mlp}
  
If $p({\mathbf x}|C_k)$ does not have a simple form, and $p(C_k|{\mathbf x})$ is not explicit, then the  multi-layer perceptron (MLP) neural network can be applied in an attempt to approximate $p(C_k|{\mathbf x})$~\cite{bishopbook}[Chapt. 5]. 
  For example, for a $K>2$ class problem using 3-layers (with 1 input, 1 hidden and 1 output) and $J$ internal variables (neurons) in the hidden layer, the approximation takes the form.
 \begin{equation}
 \gamma_k( {\mathbf x}, {\mathbf w}) = \sigma  \left  (\sum_{j=1}^J {\rm w}^{(2)}_{kj} \sigma \left  ( \sum_{i=1}^F {\rm w}^{(1)}_{ji}  ({\mathbf x})_i +  {\rm w}^{(1)}_{j0}  \right ) + {\rm w}^{(2)}_{k0}  \right ) \approx p(C_k|{\mathbf x})\label{eqn:mlayer}
  \end{equation}
  where ${\rm w}^{(1)}_{ji}$, ${\rm w}^{(2)}_{kj}$ are the $J(F+1+K)+K$ coefficients of ${\mathbf w}$ to be found from network training and $\sigma(\cdot)$ is the soft-max activation function  defined in (\ref{eqn:bayes2}).
 In our case, the input layer comprises of the features ${\mathbf x} \in{\mathbb R}^ F$ and output layer are the approximations of the posterior probabilities $\gamma_k( {\mathbf x} ,{\mathbf w}) $, $k=1\ldots,K$.
  If the number of internal variables in each hidden layer is fixed at $J$, and there are $L$ hidden layers, then the total number of parameters,  ${\rm w}_{ji}^{(1)}, \, {\rm w}_{ji}^{(2)}, \, \ldots$, which describe the network, that need to be found are $J^2(L-1) +J(F+L+K) +K$. 
Given $N$ training points $({\mathbf x}_1,{\mathbf t}_1), ({\mathbf x}_2,{\mathbf t}_2),\ldots, ({\mathbf x}_N,{\mathbf t}_N)$, and following a maximum likelihood~\cite{bishopbook}[pg. 232] approach,  the parameters are found by optimising  a logloss error function
      evaluated over the training data set
     \begin{equation}
     E({\mathbf w}) = - \sum_{n=1}^N \sum_{k=1}^K ({\mathbf t}_n)_k \ln \gamma_k ({\mathbf x}_n, {\mathbf w})\label{eqn:error},
     \end{equation}
          or, alternatively, they can be found  by a Bayesian approach~\cite{bishopbook}[pg. 277].

  As remarked by Richard and Lippmann~\cite{richard1991}, MLP can provide good estimates of $p(C_k|{\mathbf x})$ if sufficient training data is available, if the network is complex enough and if the classes are sampled with the correct a-prori class probabilities in the training data.
Nonetheless, designing appropriate networks, with the correct number of hidden layers and neurons, can be challenging. Furthermore, a complex network with a large number of neurons can require a large amount of training data to avoid overfitting.

   \subsubsection{Gradient boost}

Gradient boost is an example of what is known as a ``boosting algorithm'' ~\cite{friedman,elith2008working}. Boosting attempts to build a stronger classifier by combining the results of weaker base classifiers through a weighted majority vote~\cite{Hastie2009elements}[Chapt 10] and,  in the case of gradient boost, this is achieved through optimisation using steepest descent.
As described by  Friedman~\cite{friedman}, gradient boost can be applied to 
approximating $p(C_k |{\mathbf x})$  in
probabilistic classification.  
By again considering  (\ref{eqn:error}), with $y_k({\mathbf x}_n, {\mathbf w})$ replaced by (\ref{eqn:bayes2}) with ${\mathbf x}= {\mathbf x}_n$, and choosing the parameters ${\mathbf w} ={\mathbf w}({\mathbf x})$ as $({\mathbf w}({\mathbf x}))_k={\rm w}_k({\mathbf x})=a_k ({\mathbf x})$, then
the $k$th component of the negative gradient of the loss function  is
\begin{equation}
r _k= -\frac{\partial E( {\mathbf w}) }{\partial w_k} = \sum_{n=1}^N ({\mathbf t}_n)_k - \gamma_k ( {\mathbf x}_n,{\mathbf w}). \nonumber
\end{equation}
Starting with an initial guess $a_k^{[0]} ({\mathbf x})=0$, $k=1, \ldots, K$,  then, for a given ${\mathbf x}$, an iterative procedure is used to improve the estimate $ \gamma_k({\mathbf x}, {\mathbf w}^{[m]}) \approx p(C_k |{\mathbf x})  $  at iteration $m$. In this procedure, $K$ decision trees are trained at each iteration  to predict $r_k$, $k=1,\ldots,K$, and the leaf nodes of the tree are then used to update $a_k^{[m]}({\mathbf  x})$ until a convergence criteria is reached. For details of the practical implementation see~\cite{friedman}.

\subsection{Understanding uncertainty in classification}\label{sect:uncertain}
For the majority of the classifiers we will consider,  an ML algorithm  $\Gamma_{\mathbb D}$ trained on dictionary 
${\mathbb D}$ produces a classifier $\Gamma_{\mathbb D} ({\mathbf x}) = \gamma_k ({\mathbf x}) \approx p(C_k|{\mathbf x})$, which provides an indication of the likelihood of the class $C_k$ being correct. We then base the decision as to the correct class based on the MAP estimate. When this process is repeated for different pairs $({\mathbf x}_i, {\mathbf t}_i) \in{\mathbb T} =  D^{\text{(test)}}$,  $ \gamma_k ({\mathbf x}_i)   $ may be different for each ${\mathbf x}_i$. It is useful to explore how sensitive $ \gamma_k  ({\mathbf x}_i)  $
 is to changes in ${\mathbf x}_i$ when it is evaluated for different  $({\mathbf x}_i, {\mathbf t}_i) \in  {\mathbb T}_\ell = D^{\text{(test},(\ell))}$ associated with the test data for one class $C_\ell$. To do this, we consider confidence intervals for the average $ E_{{\mathbb T}_\ell} \gamma_k({\mathbf x})$. 

A first approach might be to use the sample mean  and sample variance
\begin{equation}
\overline{\gamma_k } = \frac{1}{P^{ (\text{test},(\ell) )}   } \sum_{i=1}^{ P^{ (\text{test},(\ell) )}     } \gamma_k ( {\mathbf{x}}_i) \qquad \text{and}
 \qquad S_k=
 \sqrt { \frac{\sum_{i=1}^{ P^{(\text{test},(\ell) ) }  } ( \gamma_k ( \mathbf{x}_i)-\overline{\gamma_k})^2}{ P^{ (\text{test},(\ell) )}  }}, \nonumber
\end{equation}
to construct an interval in the form 
\begin{align}
\overline{\gamma_k } - CV \frac{S_k}{\sqrt{ P^{ (\text{test},(\ell) )  } }      } & \le 
 E_{{\mathbb T}_\ell} \gamma_k({\mathbf x})
\le \overline{\gamma_k } + CV \frac{S_k}{\sqrt{ P^{ (\text{test},(\ell) )}    }}.\label{eqn: true bounds}
\end{align}
In the above, $CV$ is a critical value based on a t-test and the confidence level chosen.
However, in practice, if $\overline{\gamma}_k\to 0.5$ as the sample size $P^{(\text{test},(\ell))}\to \infty$ and, we have small confidence bounds, we might wrongly conclude that $\gamma_k(\mathbf{x})=0.5$ with a high degree of confidence. Instead, this may also indicate that half the observations are predicting $\gamma_k ({\mathbf x}_i )\approx 0$  and the half are predicting $\gamma_k ({\mathbf x}_i)\approx 1$, which has the same sample mean. This can occur, since at most $S_k=0.5$, and, for  large $P^{(\text{test},(\ell))}$, we find the confidence bounds produced by (\ref{eqn: true bounds}) are narrow due to division by this quality in computation of the bounds.
Hence, $\overline{\gamma_k } $ and (\ref{eqn: true bounds})  do not give any insight into the variation within the different observations $\gamma_k( {\mathbf x}_i)$. 

Instead, ordering $\gamma_k ({\mathbf x}_i)$
 as
\begin{equation}
O(C_k) =  ( \gamma_k \mathbf{x}_1),\gamma_k(\mathbf{x}_2),...,\gamma_k(\mathbf{x}_{P^{(\text{test},(\ell))}})) ,\quad\text{such that}\quad \gamma_k(\mathbf{x}_i)\leq \gamma_k(\mathbf{x}_{i+1}),
\end{equation}
we define the $y^{\text{th}}$-percentile 
\begin{equation}
\gamma_{k,y} = ( O(C_k))_{\frac{y}{100}P^{(\text{test},(\ell))} }, \label{eqn:percentile}
\end{equation}
and use interpolation between neighbouring values if $\frac{y}{100}$  is not an integer. We then consider the median value of $\gamma_k ( {\mathbf x} )$, given by $\gamma_{k,50}$, as the average and use the percentiles corresponding to $Q_1\equiv \gamma_{k,25} $, $Q_3\equiv \gamma_{k,75}$
 and $ \gamma_{k,5}$,  $\gamma_{k,95}$
to understand uncertainty in the predictions.

\section{Evaluating the performance of classifiers} \label{sect:metrics}

%We first recall the definitions of bias, variance and under/overfitting and set them in  the context of our classification problem, as these will be important to consider when evaluating the applicability and performance of classifiers.
% We  then describe metrics for assessing performance are applicable to both probabilistic and non-probabilistic classifiers given $D^\text{(train)}$  and $D^\text{(test)}$ data sets.

%We first recall the definitions of bias, variance and under/overfitting and set them in  the context of our classification problem, as these will be important to consider when evaluating the applicability and performance of classifiers.
We describe metrics for assessing performance are applicable to both probabilistic and non-probabilistic classifiers given $D^\text{(train)}$  and $D^\text{(test)}$ data sets.

\subsection{Metrics}

\subsubsection{Confusion matrices, precision, sensitivity and specificity} \label{sect:confmat}

We begin by recalling the definitions of true positive, false positive, true negative and false negative for a given class $C_k$ (see e.g.~\cite{powers2007evaluation}).
\begin{itemize}
\item \textbf{True positive (TP)}, the case where the classifier predicts $\mathbf{x}$ \underline{belongs} to $C_k$ and is \underline{correct} in its prediction.
\item \textbf{False positive (FP)} (type 1 error), the case where the classifier predicts $\mathbf{x}$ \underline{belongs} to $C_k$ and is \underline{incorrect} in its prediction.
\item \textbf{True negative (TN)}, the case where the classifier predicts $\mathbf{x}$ \underline{does not belong} to $C_k$ and is \underline{correct} in its prediction.
\item \textbf{False negative (FN)} (type 2 error), the case where the classifier predicts $\mathbf{x}$ \underline{does not belong} to $C_k$ and is \underline{incorrect} in its prediction.
\end{itemize}

Following the training of a classifier, its performance can be evaluated on the test data set $D^{(\text{test})}$. Applying the classifier to each sample $({\mathbf x}_n,{\mathbf t}_n) \in D^{(\text{test},(i))}$, where the true class label is $C_i$, the number of predictions of  each class $C_j$, $j=1,\ldots,K$ can be counted and the result recorded in the $(\mathbf{C})_{ij}$th element of a confusion matrix $\mathbf{C} \in {\mathbb R}^{K \times K}$. Repeating this process for $i=1,\ldots,K$ leads to the complete matrix~\footnote{Note that we follow the convention used by \texttt{scikit-learn} for ${\mathbf C}$, other references use a different convention where the rows and columns are swapped.}. 
 The 4 cases (TP, FP, TN, FN) for each class $C_k$ can be defined in terms of $(\mathbf{C})_{ij}$ as~\cite{powers2007evaluation,geron2019hands}
\begin{align}
\text{TP}(C_k)&:=(\mathbf{C})_{kk},  &&\text{FN}(C_k):=\sum_{\substack{j=1\\j\neq k}}^{K}(\mathbf{C})_{kj},\nonumber\\
\text{FP}(C_k)&:=\sum_{\substack{i=1\\i\neq k}}^{K}(\mathbf{C})_{ik},  &&\text{TN}(C_k):=\sum_{\substack{i=1\\i\neq k}}^{K}\sum_{\substack{j=1\\j\neq k}}^{K}(\mathbf{C})_{ij}, \nonumber
\end{align}
and the precision, sensitivity and specificity for each of the classes $C_k$ using~\cite{powers2007evaluation}
\begin{align*}
\text{precision}(C_k)&:=\frac{\text{TP}(C_k)}{\text{TP}(C_k)+\text{FP}(C_k)} := \frac{\text{TP}(C_k)}{\#   \text{predicted positives for } C_k}, \\
\text{sensitivity}(C_k)&:=\frac{\text{TP}(C_k)}{\text{TP}(C_k)+\text{FN}(C_k)} :=
\frac{\text{TP}(C_k)}{\# \text{actual positives for }C_k},
 \\
\text{specificity}(C_k)&:=\frac{\text{TN}(C_k)}{\text{TN}(C_k)+\text{FP}(C_k)}=\frac{\text{TN}(C_k)}{\# \text{predicted negatives for }C_k}.
\end{align*}
The precision and sensitivity (also known as the true positive rate or recall) are measures  of the proportion of positives that are correctly identified and specificity (also called the true negative rate)  measures the proportion of negatives that are correctly identified.

The entries in confusion matrices are often presented as frequentist probabilities (i.e.  $(\mathbf{C})_{ij}$ is normalised by $\sum_{p=1}^K\sum_{q=1}^K  (\mathbf{C})_{pq} $), which, as the sample size $P^{(k)}$ becomes large, provides an approximation to $p(C_j | {\mathbf x})$ with $({\mathbf x}, {\mathbf t}) \in D^{(\text{test},(i))}$. We will also present confusion matrices in this way.

\subsubsection{$\kappa$ Score}\label{sect: F1}
 Possible choices for
a metric which provides an overall score of the performance of the classifier
 include accuracy, the $F_1$ score and Cohen's $\kappa$ score~\cite{f1metrics,kappametrics,powers2007evaluation,cohen1960coefficient,sammut2011encyclopedia}. Vairants of the commonly used $F_1$ score include the macro-averaged $F_1$ score (or macro $F_1$ score), the weighted-average $F_1$ score (or weighted $F_1$ score) and the micro-averaged $F_1$ score (micro $F_1$ score). However, the $F_1$ score can sometimes lead to an incorrect comparison of classifiers \cite{powers2015f,f1metrics}. As Powers' \cite{powers2015f} notes,  the macro $F_1$ score is not normalised, which is overcome by the weighed $F_1$ score and the $F_1$ score is not symmetric with respect to positive and negative cases. 
Some of these drawbacks are taken in to account by using the micro $F_1$ score, however, the $\kappa$ score also takes into account chance agreement \cite{cohen1960coefficient,kappametrics}. This is useful when comparing problems with both, differing numbers of instances per class and differing numbers of classes as it takes the chance a naive classifier has into account. For these reasons, we will use the $\kappa$ score defined as
\begin{equation}
\kappa :=  \frac{\text{accuracy} - \text{random accuracy}}{1-\text{random accuracy}} \label{eqn:kappa}
\end{equation}
for comparing classifiers where 
\begin{align}
\text{accuracy} :=  & \frac{\sum_{k=1}^K\text{TP}(C_k)}{\sum_{k=1}^K\text{TP}(C_k)+\text{FN}(C_k)},\nonumber \\
\text{random accuracy}:= & \sum_{k=1}^K\frac{(\text{TP}(C_k)+\text{FN}(C_k))\cdot(\text{TP}(C_k)+\text{FP}(C_k))}{(\text{TP}(C_k)+\text{FP}(C_k)+\text{TN}(C_k)+\text{FN}(C_k))^2}.\nonumber
\end{align}

%%%%%%%%%%%%%%%%%%%%%%%%%%%%%%%%%%%%%%%%%%%%%%%%%%%%%%
\subsection{Validation methods}\label{sect: Validation}
Evaluating the performance of different classifiers can be considerably enhanced by  employing  cross validation \cite{kuhn2013applied}. This is particularly important if $D^{\text{(test)}}$ is small and, otherwise,  may lead to inaccurate predictions of a classifier's performance.   
We have chosen to employ the Monte Carlo cross validation (MCCV) technique (also known as ``Leave group out cross validation'')\cite{kuhn2013applied}[pg. 71]. This involves performing  $\ell$ iterations where, for each iteration, the dataset $D$ is split into training and testing $D=(D^{\text{(test)}}, D^{\text{(train)}})$ with $D^{\text{(train)}}$ and $D^{\text{(test)}}$ being drawn differently from $D$ each time,  irrespective of the splittings in previous iterations. 
Other variants of cross validation include k-fold cross validation, repeated k-fold cross validation and bootstrapping, for further details see Kuhn and Johnson~\cite{kuhn2013applied}. Kuhn and Johnson explain that no resampling method is uniformly better than another and that the differences between the different methodologies is small for larger samples sizes, which further motivates that actually performing cross validation is more important than the method chosen for doing so.

\section{Results} \label{sect:results}

\subsection{Classification of British coins}\label{sect:coins}
\subsubsection{Construction of the coin dictionary} \label{sect:coindata}

\begin{figure}[h]
\begin{center}
$\begin{array}{cc}
\includegraphics[scale=0.5]{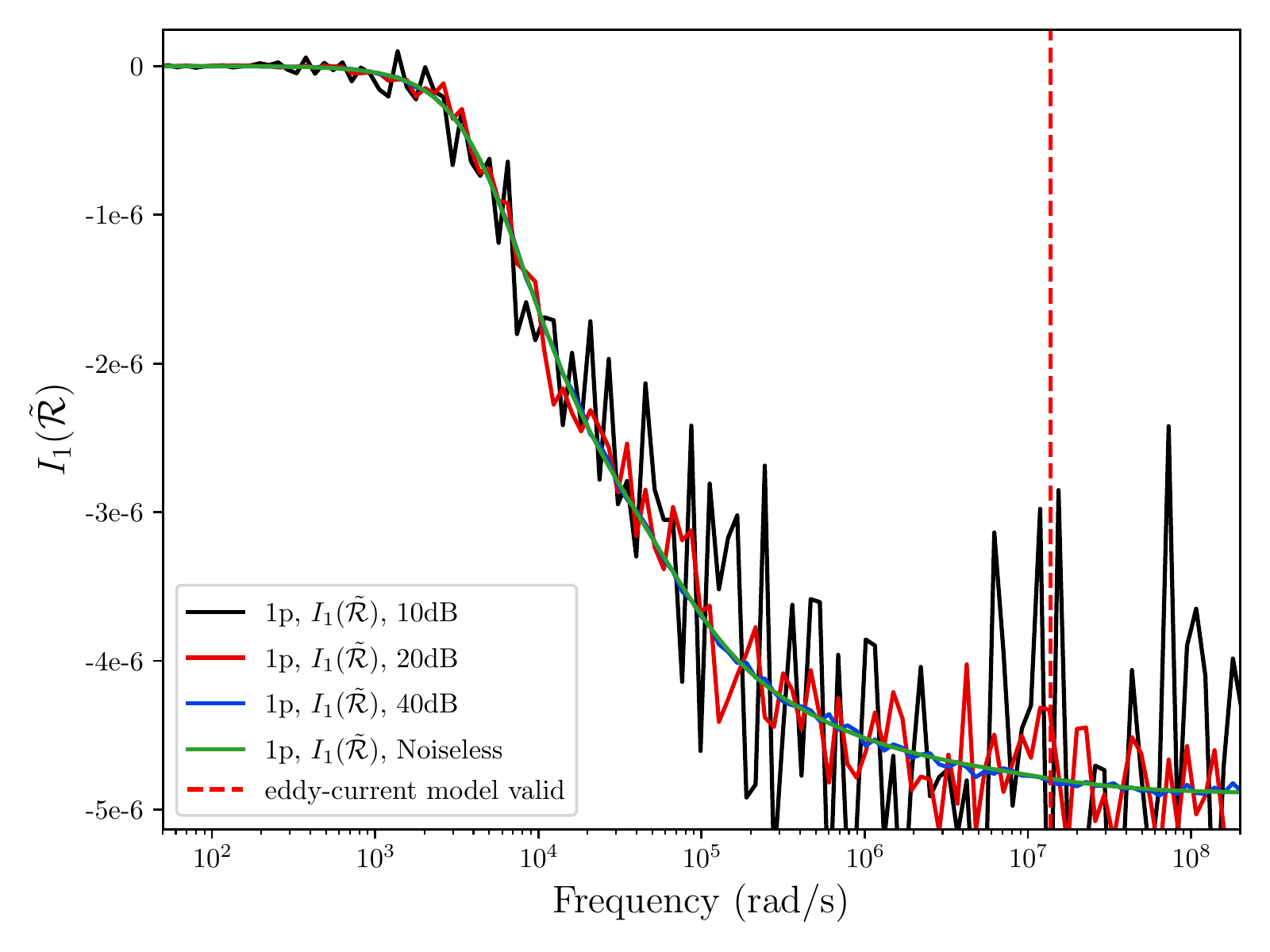} &
\includegraphics[scale=0.5]{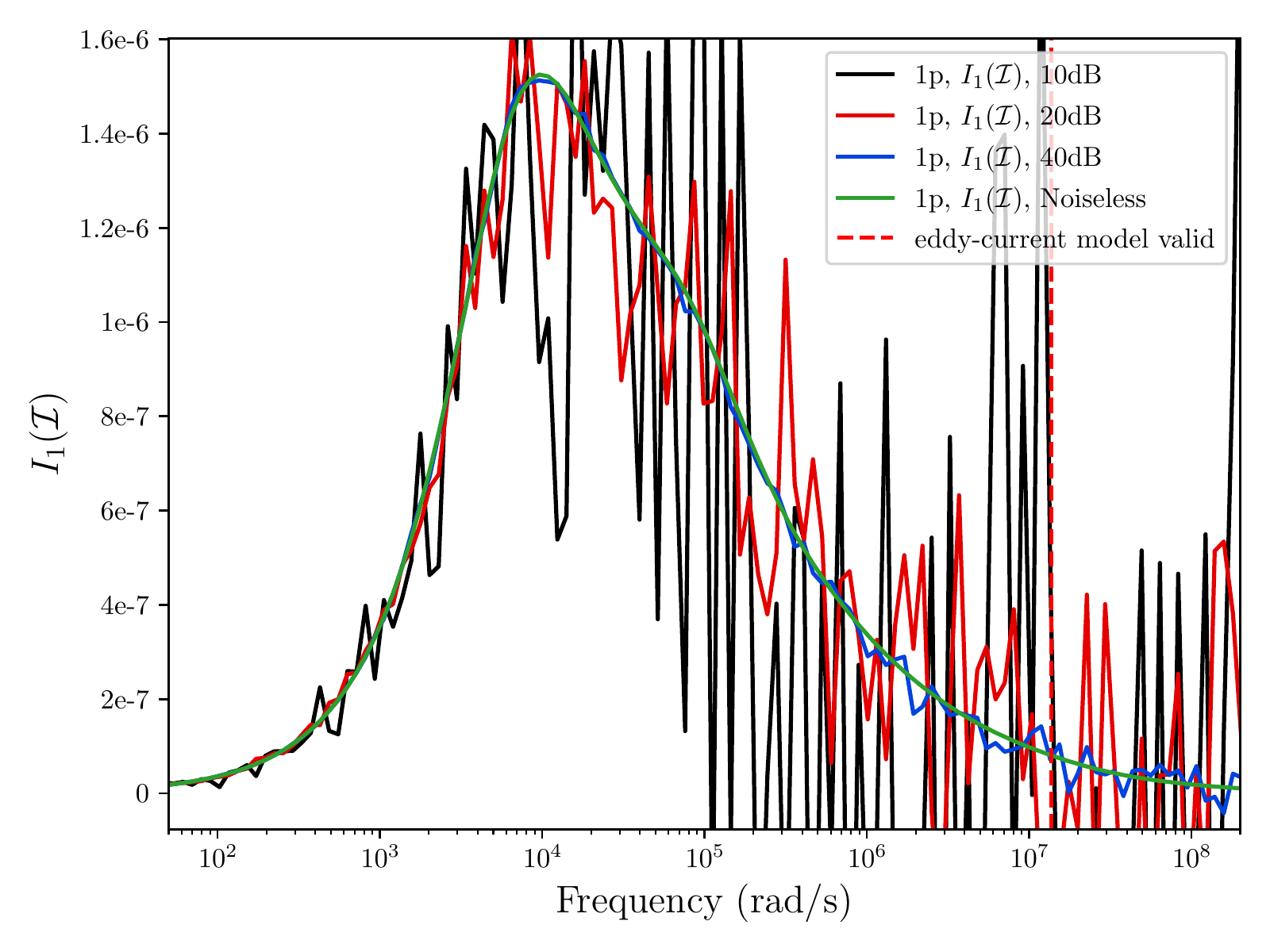}  \\
\text{(a) } I_1 (\tilde{\mathcal{R}}) & \text{(b) } I_1 (\mathcal{I})
\end{array}$
\end{center}
\caption{Set of British coins: 1p coin $\alpha B$, $\alpha=0.001$m, $\mu_r=1$ and $\sigma=4.03\times10^7$ S/m showing the spectral signatures for noiseless and realisations of noise with SNRs of 40dB,  20dB and 10dB (a) $I_1 (\tilde{\mathcal R}[\alpha B ,\omega,\sigma_*,\mu_r])$ and (b) $I_1 ( {\mathcal I}[\alpha B,\omega,\sigma_*,\mu_r])$.}
\label{fig:coins_noise_levels}
\end{figure}

To create the coin dictionary, we follow the approach described in Section~\ref{sect:dict}. 
We choose the $k$th class $C_k$, $k=1,...,K=8$, to correspond to the $k$th British coin denomination one penny (1p), two pence (2p), five pence (5p), ten pence (10p), twenty pence (20p), fifty pence (50p), one pound (\pounds 1), two pounds (\pounds 2),  respectively, in order of increasing value. The 
coins have different geometries, and, in some cases different materials~\cite{ledgerwilsonamadlion2021}, although within each class we restrict ourselves to a single geometry $G^{(k)}=1$ and consider a fixed $P^{(k)}=P/K$ number of samples for each class so that $P^{(k)}=V^{(k)}$ in this case. We use $P^{(k)}=2000$ for coin classification unless otherwise stated.
  The shape $B^{(k)}$ of  the $k$th coin class is  as described in the specification of Table 1 in \cite{ledgerwilsonamadlion2021} and in our \texttt{MPT-Library}~\cite{bawilson94_2021_4876371} we have  previously obtained the MPT spectral signatures for each coin geometry. 
From these, we choose the signatures evaluated at $M$ equally spaced $\omega_m$ such that $5.02\times 10^4 \text{ rad/s}  \le \omega_m \le 8.67 \times 10^4 \text{ rad/s}$, although we have also considered the larger frequency range of $7.53 \times 10^2 \text{ rad/s} \le \omega_m \le 5.99 \times 10^5 \text{ rad/s}$, which leads to very comparable results~\cite{thesisben}  to those presented here.
   To account for the fact that the measured MPT coefficients will be noisy, we illustrate in Figure~\ref{fig:coins_noise_levels} 
   realisations of noise being added to the spectral signatures  of $I_1 (\tilde{\mathcal{R}})$ and $I_1 (\mathcal{I})$ for the 1p  coin.  The curves in this figure correspond to the cases of  no noise and noise with SNR values of  40dB, 20dB and 10dB. 
 
The $V^{(k)}$ variations account for the fact that within each class there can be: 
 \begin{enumerate}
\item Variation in the object size $\alpha$, such that the volume in the different observations of coins of a certain denomination can change.  While it is expected that a coins size may be fairly uniform when they leave the mint, they are likely to become increasingly bashed and dented  once they enter circulation, hence their object size may change over time. Hence we choose  the object size to be $\pm2.52\%$ of the coins specification. We set $m_\alpha=0.001$m for each coin and set  $s_\alpha = 0.001(1.0252-1)/3=8.4\times10^{-6}$m i.e. to be $1/3$ the difference of the upper limit and the respective mean.
\item Variation in the object conductivity/conductivities $\sigma_*({\bm x})$ to account for variations in the manufacturing process, such that the conductivity in the different observations of coins of a certain denomination can change. In most cases the coins are assumed to be homogeneous conductors, but for \pounds 1 and \pounds 2 coin denominations the objects are each an annulus.  As the coins dominant composition material is copper, using  \cite{copper1982copper,ho1983electrical} we find an upper limit for conductivity to be $\pm7.09\%$ of the coins specification. We set $m_{\sigma_*}$ corresponding to the conductivities  of each coin denomination, so for a 1p coin, for example, $m_{\sigma_*} = 4.307 \times 10^7 $S/m and set $s_{\sigma_*}$ in a similar way to $s_\alpha$, so that  $s_{\sigma_*}=4.307\times10^7 (1.0709-1)/3=9.52\times10^5$S/m.
\item Note that the object's permeability will be fixed as $\mu_r=1$ as we assume all the coins considered are non-magnetic ~\cite{ledgerwilsonamadlion2021}.
\end{enumerate}

Given that  $p\left (-3 \le \frac{\alpha-m_\alpha}{s_\alpha}\le 3 \right) = 0.9974$, we expect 99.74\% of the object sizes generated to fall within our prescribed variation in object sizes due to being bashed and dented in circulation. Similarly, we expect 99.74\% of the $\sigma_*$ values generated to fall within our prescribed variation in $\sigma_*$. Overall, this means that $0.9974 \cdot 0.9974 =0.9948$ or 99.48\% of the values generated are representative of genuine currency.

The effect of these samples on the MPT spectral signature is illustrated in Figure \ref{fig:One_p_histograms} for the 1p coin class ($C_1$) and noiseless data. We now explore this further: 
\begin{figure}[!h]
\centering
\hspace{-1.cm}
$\begin{array}{cc}
\includegraphics[scale=0.5]{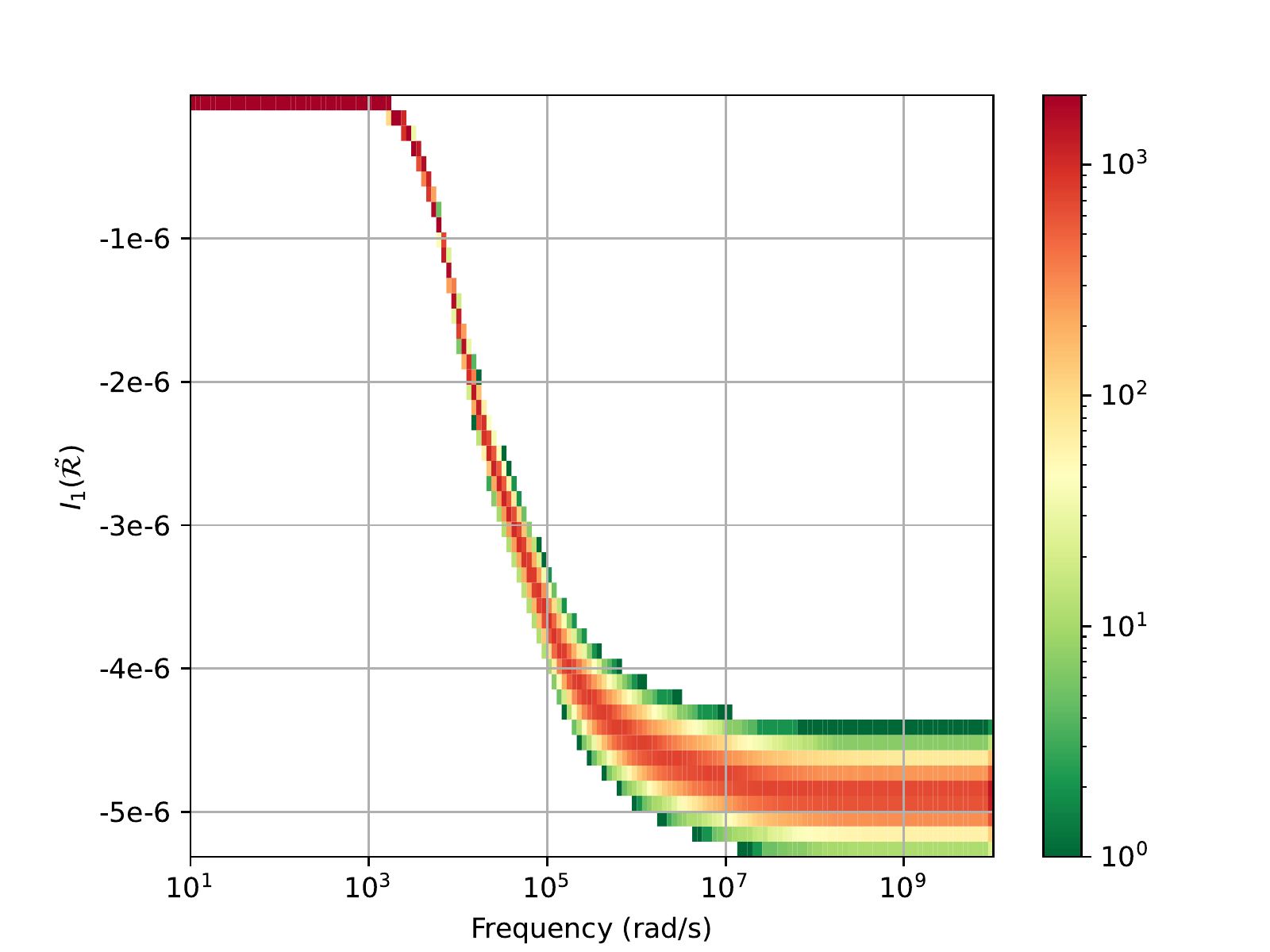}  &
\includegraphics[scale=0.5]{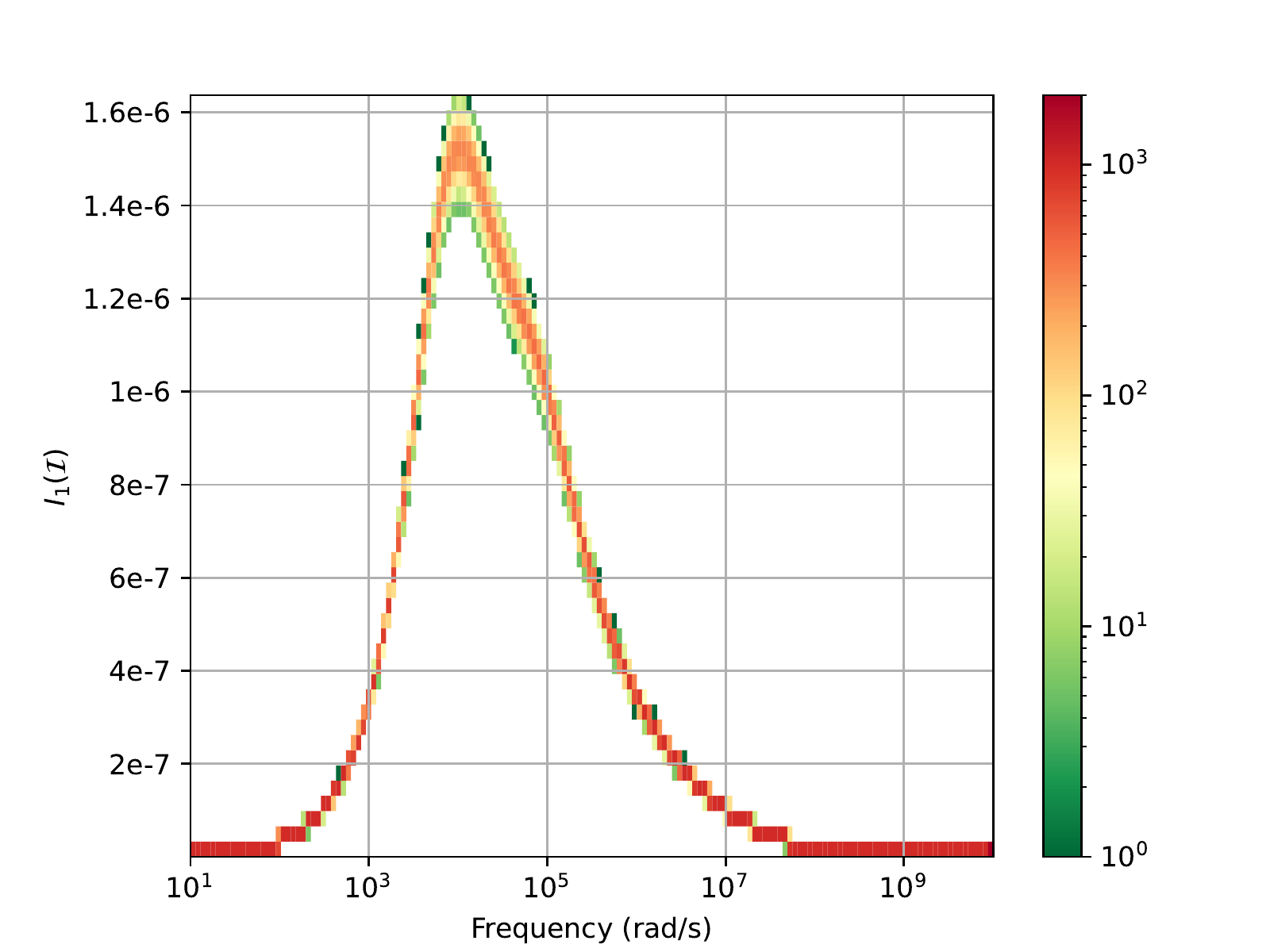}  \\
\text{(a) } I_1 (\tilde{\mathcal{R}}) & \text{(b) } I_1 (\mathcal{I})
\end{array}$
  \caption{Set of British coins: 1p coin (class $C_1$) with $P^{(k)}=P/K=2000$, with $\alpha \sim N ( 0.001, 8.4\times10^{-6})$ m and $\sigma_* \sim N (4.03\times10^7, 9.52\times10^5)$ S/m showing the histograms for the distribution of the spectral signatures (a) $I_1 (\tilde{\mathcal R}[\alpha B^{(1)},\omega,\sigma_*,\mu_r])$ and (b) $I_1 ( {\mathcal I}[\alpha B^{(1)},\omega,\sigma_*,\mu_r])$.}
        \label{fig:One_p_histograms}
\end{figure}
\begin{figure}[!h]
\centering
\hspace{-1.cm}
$\begin{array}{cc}
\includegraphics[scale=0.5]{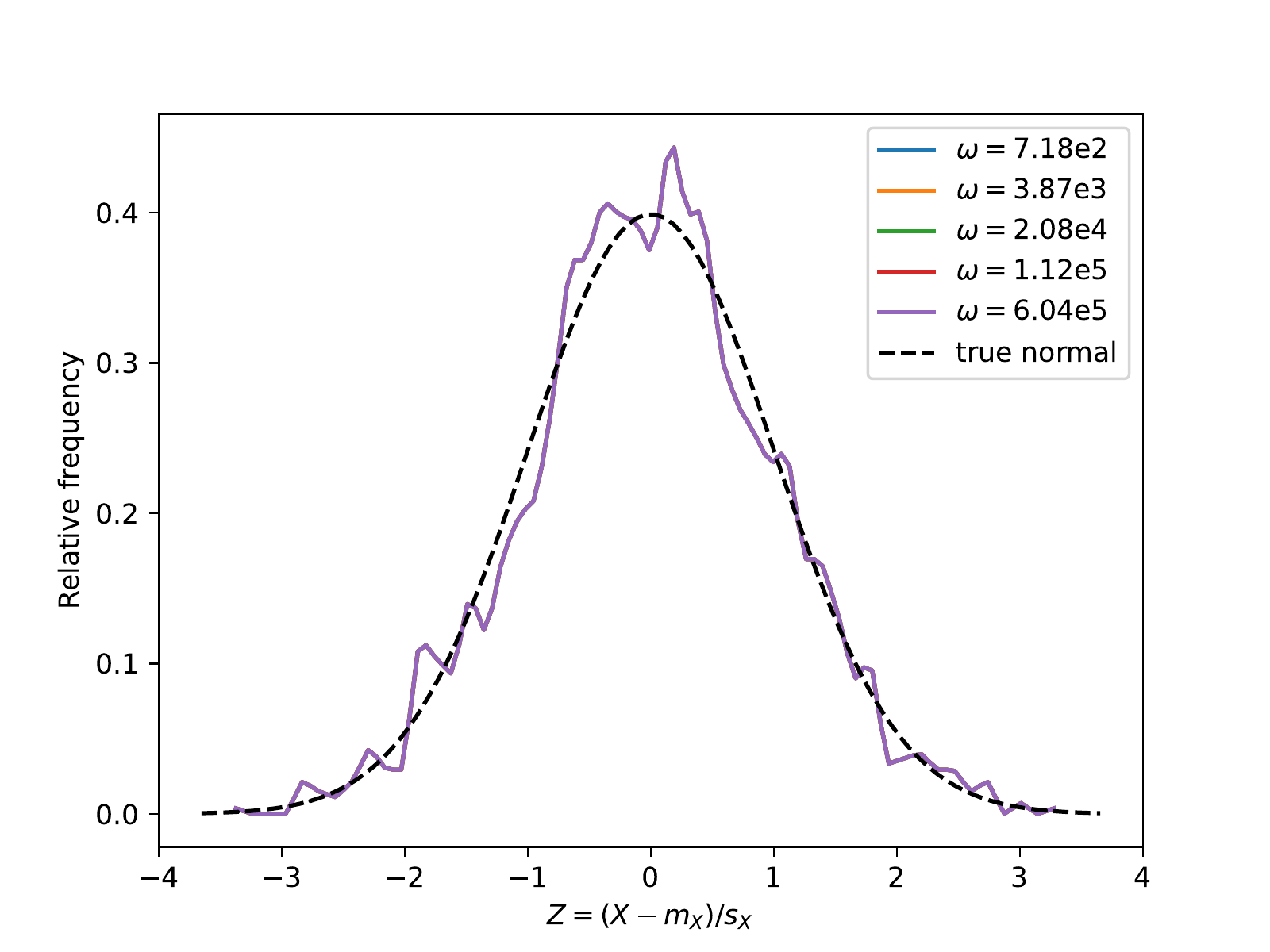}  &
\includegraphics[scale=0.5]{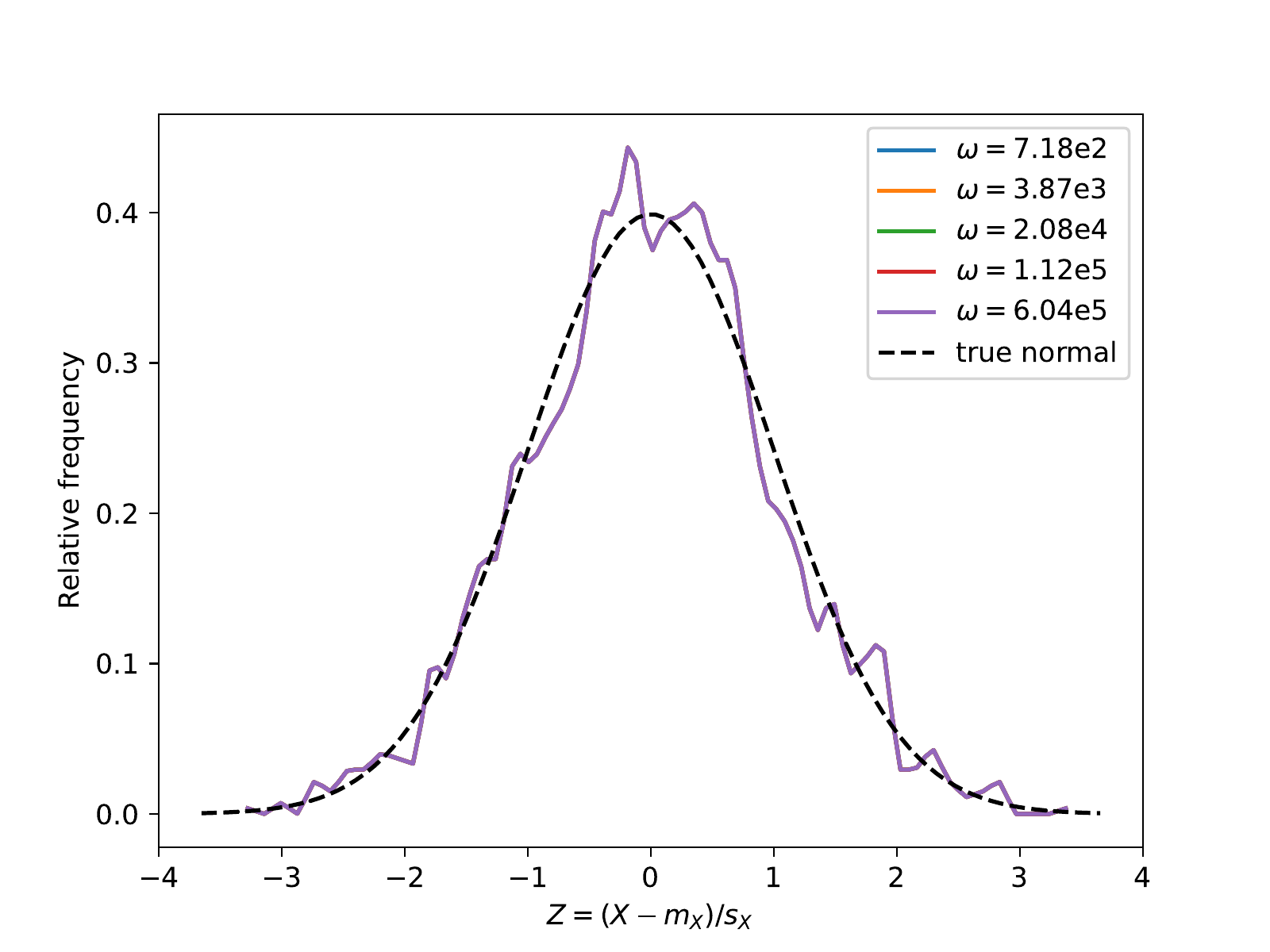}  \\
\text{(a) } I_1 (\tilde{\mathcal{R}}) & \text{(b) } I_1 (\mathcal{I})\\
\includegraphics[scale=0.5]{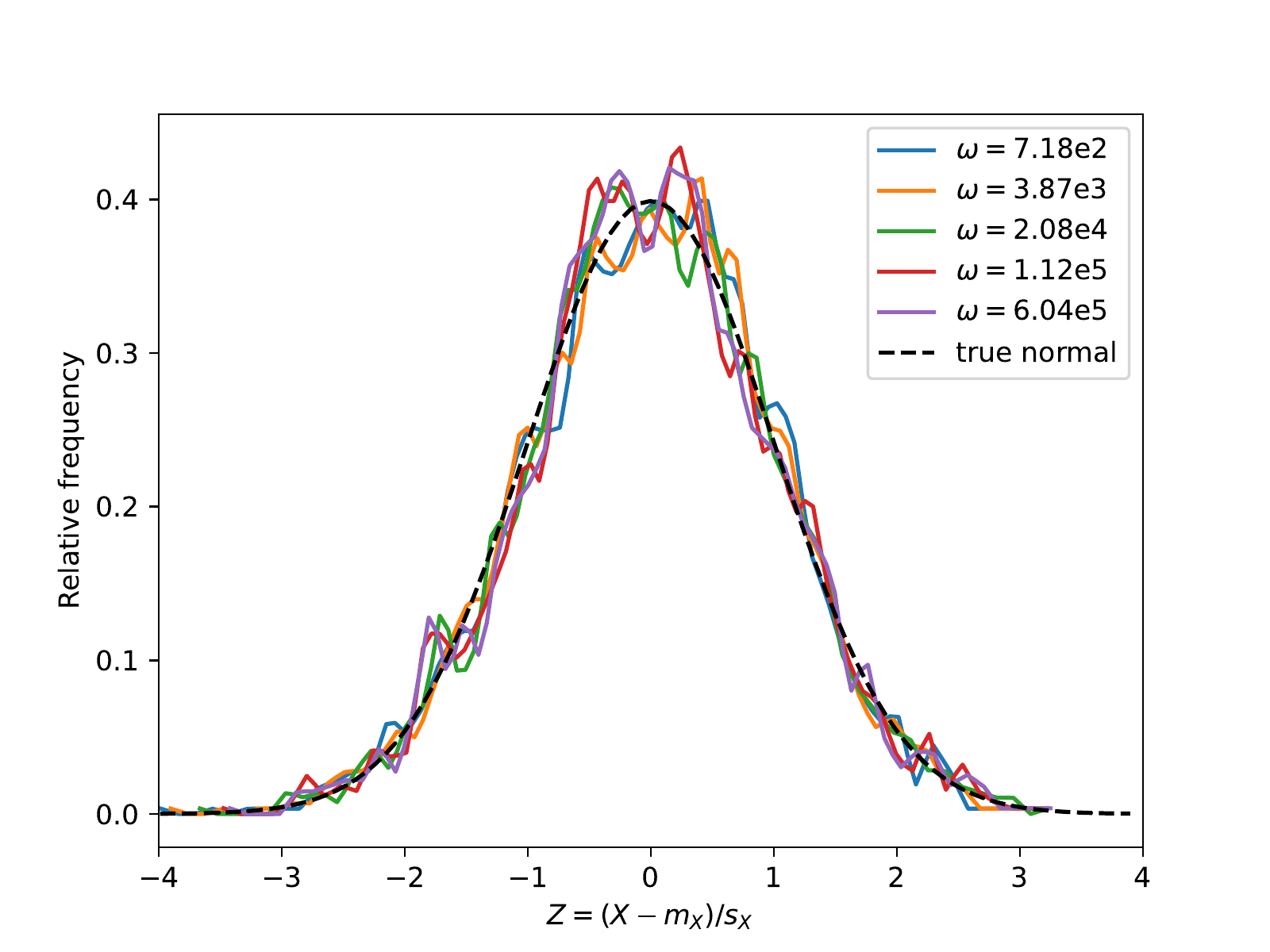}  &
\includegraphics[scale=0.5]{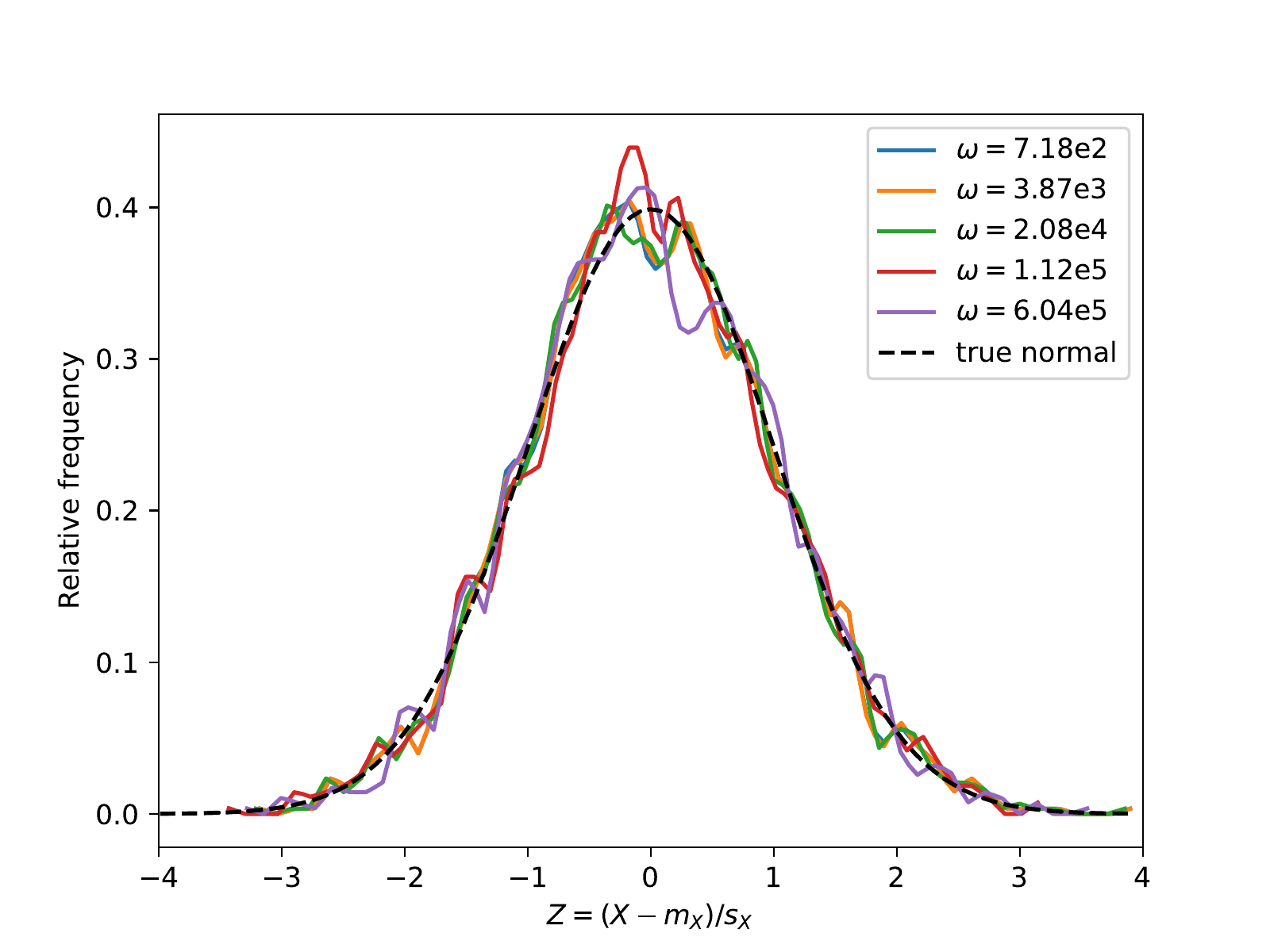}  \\
\text{(c) } \lambda_1 (\tilde{\mathcal{R}}) & \text{(d) } \lambda_1 (\mathcal{I})
\end{array}$
  \caption{Set of British coins: 1p coin (class $C_1$) with $P^{(k)}=P/K=2000$, with $\alpha \sim N ( 0.001, 8.4\times10^{-6})$ m and $\sigma_* \sim N (4.03\times10^7, 9.52\times10^5)$ S/m showing the normalised histograms of $(Z-m_X)/s_X$ where $X$ is instances of the following (a) $I_1 (\tilde{\mathcal R}[\alpha B^{(1)},\omega_m,\sigma_*,\mu_r)$, (b) $I_1 ( {\mathcal I}[\alpha B^{(1)},\omega_m,\sigma_*,\mu_r])$, (c) $\lambda_1 (\tilde{\mathcal R}[\alpha B^{(1)},\omega_m,\sigma_*,\mu_r])$ and (d) $\lambda_1 ( {\mathcal I}[\alpha B^{(1)},\omega_m,\sigma_*,\mu_r])$ at distinct frequencies $\omega_m$. }
        \label{fig:One_p_histograms_dist}
\end{figure}
Given $\alpha \sim N ( 0.001, 8.4\times10^{-6})$ m, $\sigma_* \sim N (4.03\times10^7, 9.52\times10^5)$ S/m, 
drawing  $P^{(k)}$ samples of $\alpha$   and $\sigma_*$ and applying the scaling results in Lemmas 2 and 3 of~\cite{ben2020}, we obtain 
the histograms of the random variables $X =I_1 (\tilde{\mathcal R}[\alpha B^{(1)},\omega_m,\sigma_*,\mu_r)\sim p({\rm x}_{1+(m-1)M}  |C_1) $ and $X=I_1 ( {\mathcal I}[\alpha B^{(1)},\omega_m,\sigma_*,\mu_r]) \sim p({\rm x}_{1+(m+2)M}  |C_1)$  shown in Figure~\ref{fig:One_p_histograms}. Then, by taking cross section sections at selected frequencies $\omega_m$ we obtain the histograms shown in Figure~\ref{fig:One_p_histograms_dist}.
The corresponding distributions obtained by sampling $X=\lambda_1 (\tilde{\mathcal R}[\alpha B^{(1)},\omega_m,\sigma_*,\mu_r])$ and $X=\lambda_1 ( {\mathcal I}[\alpha B^{(1)},\omega_m,\sigma_*,\mu_r])$ are also included in this figure. In each case, these distributions have been normalised using the transformation $Z=(X - m_X)/s_X$, where $m_X$ and $s_X$ indicates the mean and standard deviation of $X$~\footnote{The sample mean and sample standard deviation are used as an approximation to $m_X$ and $s_X$, respectively.} and a curve of best fit made through the histogram. As a comparison, the standard normal distribution is included. Notice that the normalised sample distributions of $X=I_1 (\tilde{\mathcal R}[\alpha B^{(1)},\omega_m,\sigma_*,\mu_r])$  for different $\omega_m$ are identical, as are  those for $X=I_1 ( {\mathcal I}[\alpha B^{(1)},\omega_m,\sigma_*,\mu_r])$  for different $\omega_m$, and have a close fit to the standard normal distribution. The conclusion is similar for the other invariants and other coins. This outcome can be explained by the central limit theorem, which implies, given a large enough sample size, we expect the samples of $I_1 (\tilde{\mathcal R}[\alpha B^{(1)},\omega_m,\sigma_*,\mu_r])$  and  $I_1 ( {\mathcal I}[\alpha B^{(1)},\omega_m,\sigma_*,\mu_r])$ to follow a normal distribution even if the parent distribution is not normal~\footnote{Although we chosen $\alpha\sim N ( m_\alpha, s_\alpha)$ and $\sigma_* \sim N (m_{\sigma_*}, s_{\sigma_*})$  we should not expect the parent distributions $p({\rm x}_{1+(m-1)M}  |C_1) $ and $p({\rm x}_{1+(m+2)M}  |C_1)$,  for $(\tilde{\mathcal R}[ \alpha B,\omega , \sigma_*, \mu_r ])_{ij}$ or $({\mathcal I}[ \alpha B,\omega , \sigma_*, \mu_r ])_{ij}$, respectively, to be normally distributed, due to the powering operation involved in the scaling in Lemma 3 of~\cite{ben2020}, which, for large $s_\alpha$, can have significant effect. However, the invariant $I_1$ only involves summation and will not change the distribution further for independent variables. The invariants $I_2$ and $I_3$ do involve products and so are likely to further change the parent distribution, but, compared to the root finding in eigenvalues, these are much smoother operations and so their effects are expected to be smaller.}.
 The normalised sample distributions of  $X=\lambda_1 (\tilde{\mathcal R}[\alpha B^{(1)},\omega_m,\sigma_*,\mu_r])$ and $X=\lambda_1 ( {\mathcal I}[\alpha B^{(1)},\omega_m,\sigma_*,\mu_r])$ also follow a normal distribution, but the fit is not as good as for the invariants. The results are similar for other eigenvalues and other coins. 
 For our chosen $\alpha \sim N ( 0.001, 8.4\times10^{-6})$ m, $\sigma_* \sim N (4.03\times10^7, 9.52\times10^5)$ S/m  and smaller sample sizes, the distributions of $X =I_1 (\tilde{\mathcal R}[\alpha B^{(1)},\omega_m,\sigma_*,\mu_r)$, $X=I_1 ( {\mathcal I}[\alpha B^{(1)},\omega_m,\sigma_*,\mu_r])$, $X=\lambda_1 (\tilde{\mathcal R}[\alpha B^{(1)},\omega_m,\sigma_*,\mu_r])$ and $X=\lambda_1 ( {\mathcal I}[\alpha B^{(1)},\omega_m,\sigma_*,\mu_r])$ still approximately follow a normal distribution with the fit being superior for the invariants. 
By considering different instances of noise, similar histograms to those shown in Figure~\ref{fig:One_p_histograms_dist} can be obtained and again similar conclusions about the resulting distributions of the eigenvalues and invariants at each $\omega_m$ apply.

\subsubsection{Classification results}\label{sect:Classification_Results}

\begin{figure}[h]
$$\begin{array}{cc}
\includegraphics[scale=0.5]{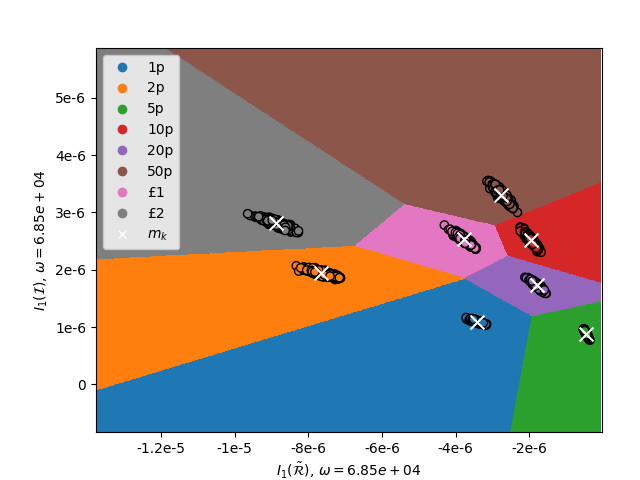} &
\includegraphics[scale=0.5]{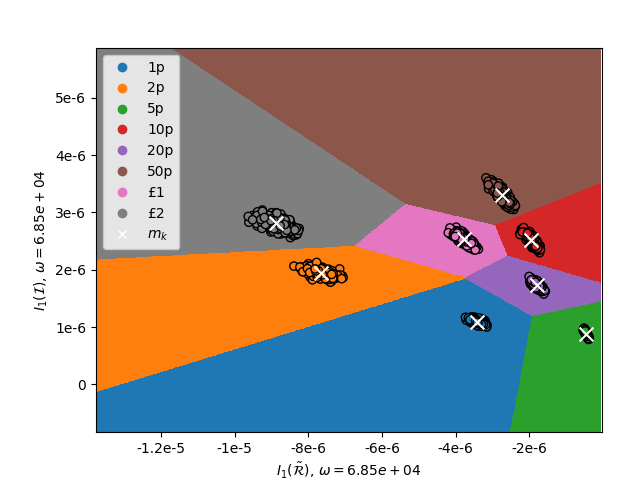}\\
\textrm{\footnotesize{(a) Noiseless}} & \textrm{\footnotesize{(b) SNR =40dB }}\\
\includegraphics[scale=0.5]{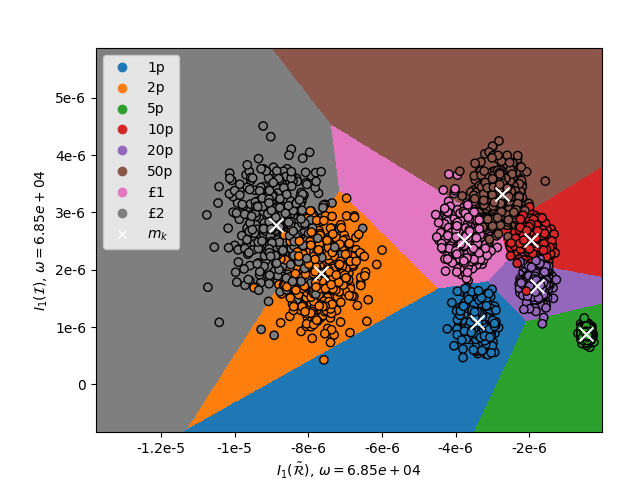} &
\includegraphics[scale=0.5]{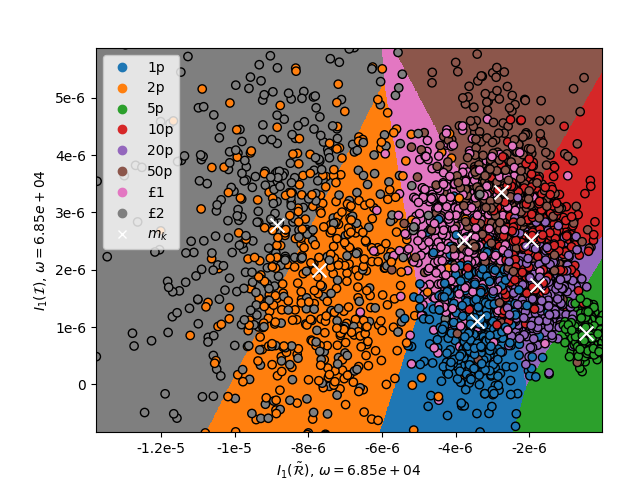}\\
\textrm{\footnotesize{(c) SNR=20dB}} & \textrm{\footnotesize{(d) SNR=10dB}}\\
\end{array}$$
\caption{Set of British coins: linear feature space splitting for the classes $C_k$, $k=1,\ldots,K,$ for a simplified case of $F=2$ features based on $\max_k \gamma_k ({\mathbf x})$ for logistic regression and $P^{(k)} = P/K=2000$ for (a) noiseless and SNR of (b) 40dB, (c) 20dB, (d) 10dB. }
\label{fig:coins_feature_space}
\end{figure}

For the coin classification problem, we restrict consideration to the logistic regression classifier  and the default settings of \texttt{scikit-learn},  as Figure~\ref{fig:One_p_histograms_dist} indicates that a normal distribution is a good approximation for the sample distributions of  $I_i (\tilde{\mathcal R}[\alpha B^{(k)},\omega_m,\sigma_*,\mu_r)$ and $I_i (\mathcal I[\alpha B^{(k)},\omega_m,\sigma_*,\mu_r)$, $i=1,2,3$, for a sufficiently large sample size. We have also observed  that the feature space can be separated linearly. To illustrate this, we begin by examining the simplest case of just $F=2$ features,  $I_1(\tilde{\mathcal R}[\alpha B^{(k)},\omega_1,\sigma_*,\mu_r])$ and $I_1( {\mathcal I}[\alpha B^{(k)},\omega_1,\sigma_*,\mu_r])$, and $M=1$ with $\omega_1=6.85 \times 10^4 $ rad/s. 
 Figure~\ref{fig:coins_feature_space} shows the class boundaries when the MAP estimate (\ref{eqn:map}) is applied for different levels of noise, the crosses indicate the locations of the means ${\mathbf m}_k$ for each class obtained from $D^{(\text{train})}$ and the circles indicate the samples from $D^{(\text{test})}$ assuming a 3:1 training-testing $D=(D^{(\text{train})},D^{(\text{test})})$ splitting and MCCV with $\ell =100$ (as described in Section~\ref{sect: Validation}), which we employ throughout.
 From this figure, we observe the class boundaries change only slightly if noise with SNR of 40dB is added and, with greater noise, the changes to the boundaries are only moderate. It is also possible to see that the number of misclassifications is very small for SNR= 40dB and 20dB and still modest for 10dB, which has a $\kappa=0.66$ score using (\ref{eqn:kappa}). Furthermore, and importantly, the locations of the means ${\mathbf m}_k$ for each class do not change significantly
 for $P^{(k)}=P/K=2000$ since using this large number of instance per class has the effect of largely averaging out the effects of noise and the object variations that we previously illustrated in Figures~\ref{fig:coins_noise_levels} and~\ref{fig:One_p_histograms}.
While  this figure indicates that the samples form a $p({\mathbf x}|C_k)$ that is normally distributed, especially for noiseless and noisy data with SNR of 40dB, 20dB, which is consistent with the assumptions of this classifier, described in Section~\ref{sect:logreg}, we observe that the assumption of a common covariance matrix between the classes does not hold for coin data set. The variance between the features  is anisotropic for each cluster, as indicated by different sized and different orientated ellipses, which also becomes increasingly apparent for increased noise levels. While logistic regression typically has a high-bias and low-variance, we expect its bias to be lower for this problem than others given the above.

\begin{figure}[h]
$$\begin{array}{cc}
\includegraphics[scale=0.5]{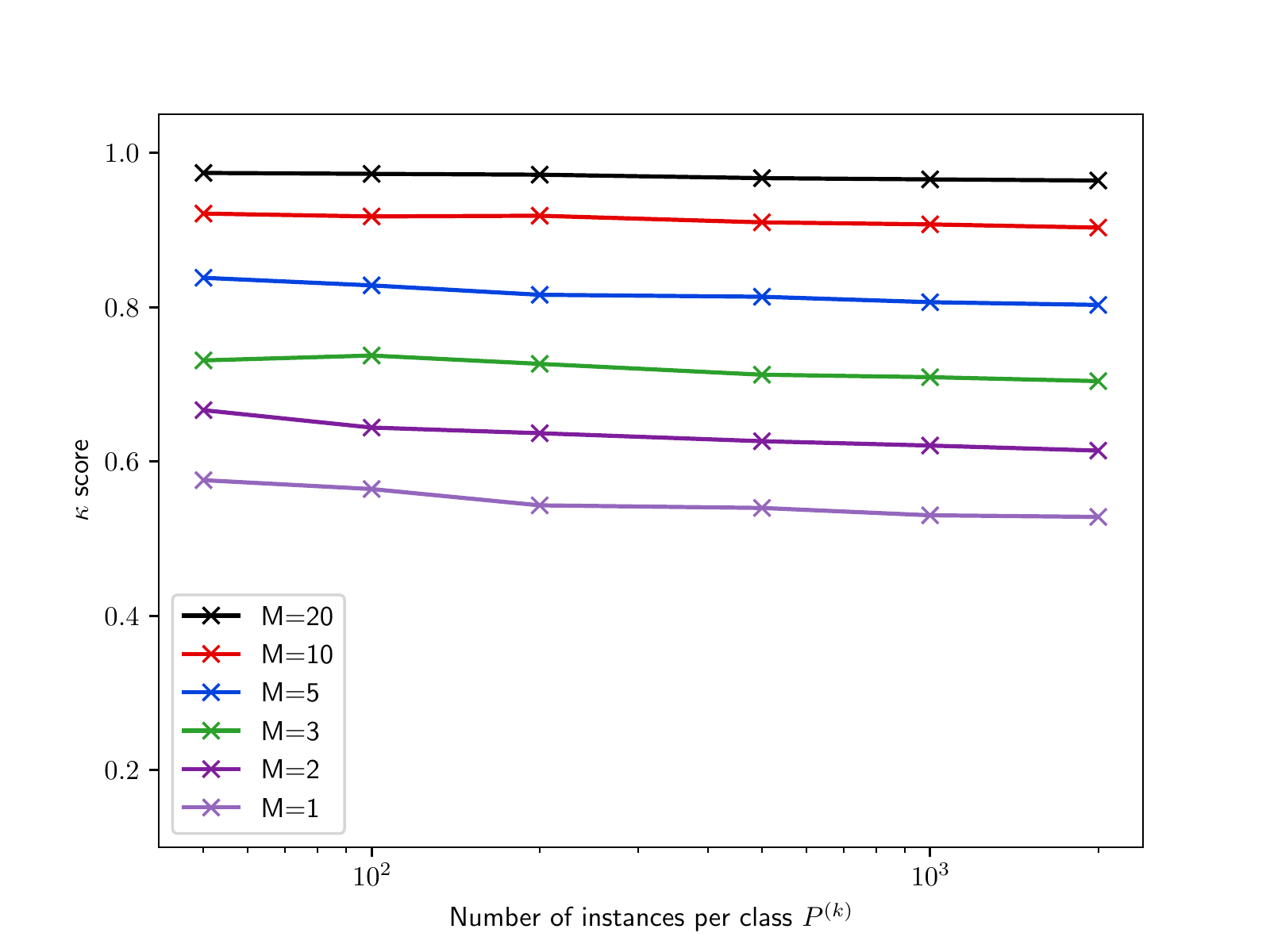} &
\includegraphics[scale=0.5]{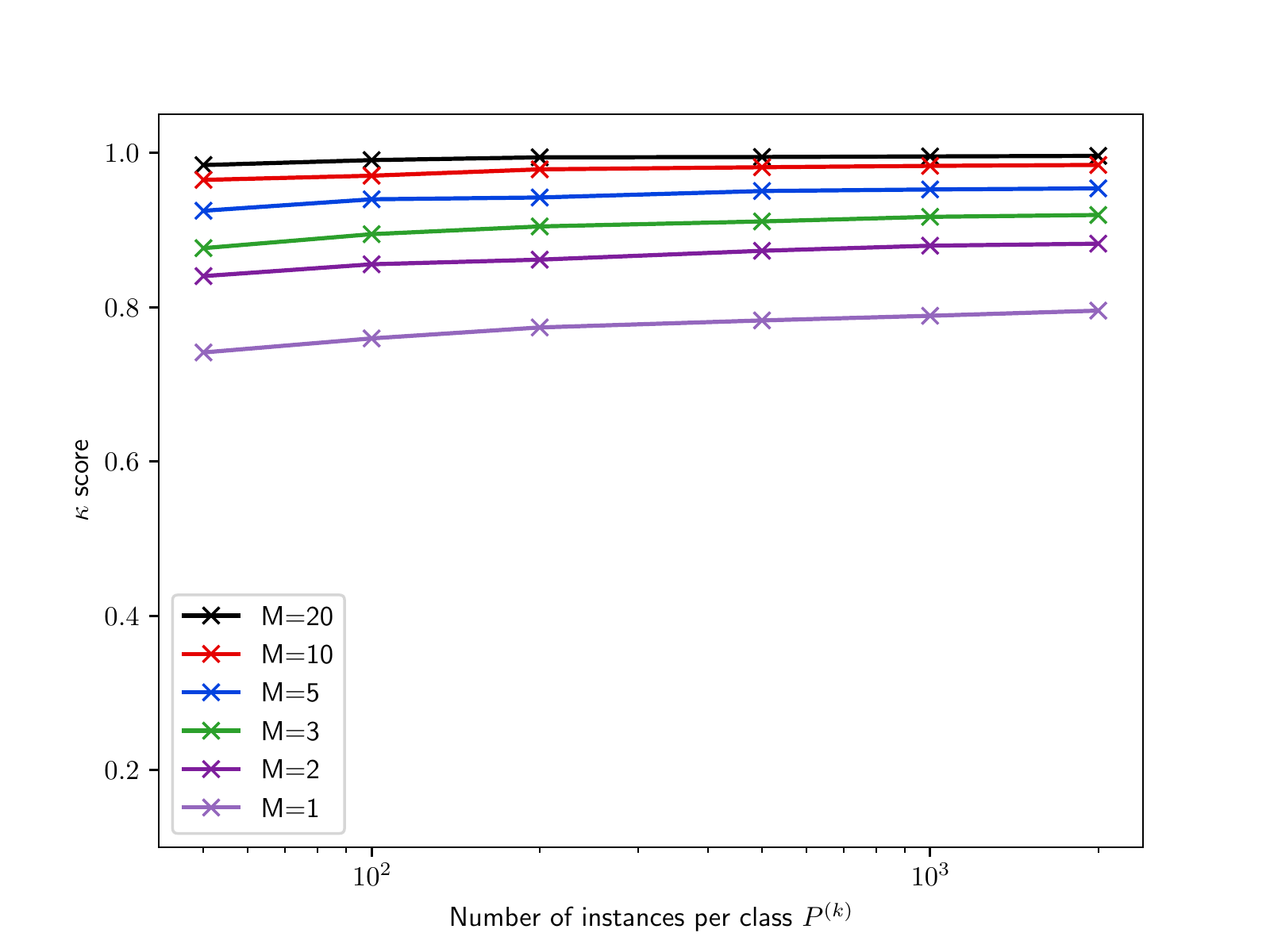}\\
\textrm{\footnotesize{(a) Noiseless training data}} & \textrm{\footnotesize{(b)SNR=10dB training data}}
\end{array}$$
\caption{Set of British coins: Overall performance of logistic regression classifier as a function of $P^{(k)}$ and different numbers of frequencies $M$ using the $\kappa$ score (\ref{eqn:kappa}) for a testing noise SNR=10dB, for (a) noiseless training data and (b) training data with SNR=10dB.} \label{fig:noiselvlvsinstance}
\end{figure}

This behaviour also carries over when we use $F=6M$ and greater $M$. In Figure~\ref{fig:noiselvlvsinstance} we illustrate the overall performance of the classifier using the $\kappa$ score (\ref{eqn:kappa}) as a function of $P^{(k))}$ for test data with SNR=10dB noise and $(a)$ for noiseless training data and $(b)$ noisy training data with SNR=10dB. The different curves correspond to $M=1,2,3,5,10,20$ frequencies. The curve for $M=1$ corresponds to the same frequency considered in Figure~\ref{fig:coins_feature_space}, but has $F=6$ features instead of $F=2$.
Increasing $M$ also increases $F$ and, for  either noiseless or noisy training data,  the classifier's performance is improved for fixed $P^{(k)}$ as more feature information is available in  ${\mathbf x}\in {\mathbb R}^F$ for each $({\mathbf x},{\mathbf t})\in D^\text{(train)}$ and, hence, it becomes easier for the classifier to find relationship between the features and classes and, in the decision stage, partitioning according to (\ref{eqn:map}) becomes easier for larger $F$. % This  all  reduces the classifier's bias further. 
On the other hand, increasing $P^{(k)} = P/K$, for a fixed $M$ and noiseless training data, reduces the $\kappa$ score and increases the variability as the classifier becomes increasingly overfitted to the training data and experiences more misclassifications as $P^{(k)}$ is increased.
  For noisy  training data, the classifier is exposed to more noisy data in  $D^\text{(train)}$ as $P^{(k)} $ is increased and, hence, its performance improves and its variability decreases. 
The relatively high accuracy of logistic regression for the coin classification problem,  even with an SNR of 10dB, can in part be attributed to how well the assumptions of logistic regression hold in practice for this problem and the normalisation of the data that is performed prior to training.

\begin{center}
\begin{figure}[h]
\begin{center}
\includegraphics[scale=0.5]{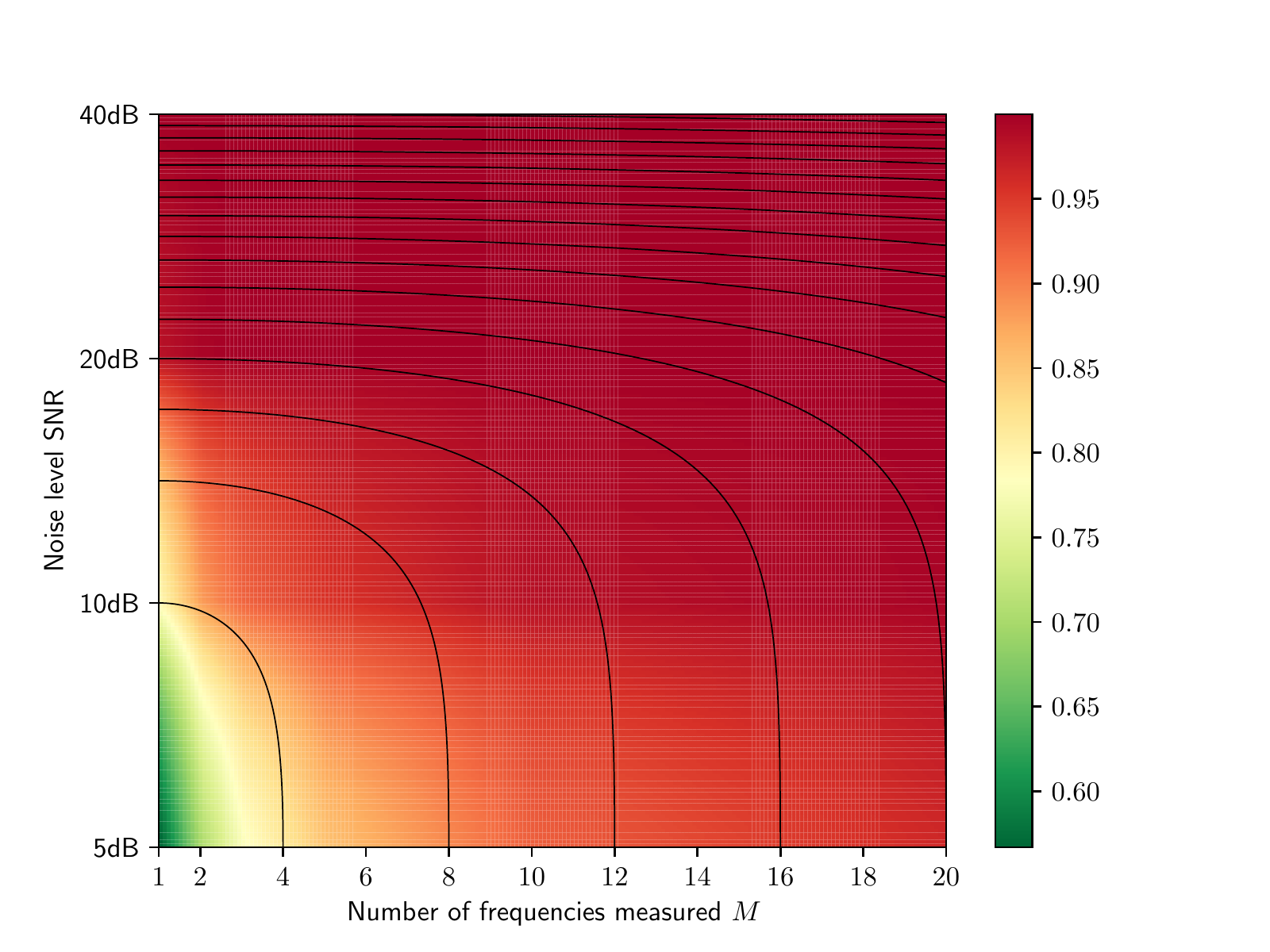}
\caption{Set of British coins: Overall performance of logistic regression classifier with $P^{(k)}=2000$, comparing $M$ and SNR using the $\kappa$ score (\ref{eqn:kappa}).} \label{fig:Contour Kappa}
\end{center}
\end{figure}
\end{center}

We  consider further the relationship  between noise level, number of frequencies and classifier performance 
in Figure~\ref{fig:Contour Kappa}. This figure shows the noise level against $M$, with the contours indicating the resulting $\kappa$ score  for fixed $P^{(k)}$.  The concentric curves correspond to all the systems with a $M$ and a noise level that achieve the same accuracy.  As is to be expected, results for the classifier can be improved by increase both/ either $M$ or SNR. This figure is of practical value as it allows practitioners to choose $M$, given an SNR, in order to achieve a desired level of accuracy.

\begin{figure}[h]
\begin{center}
$\begin{array}{cc}
\includegraphics[scale=0.5]{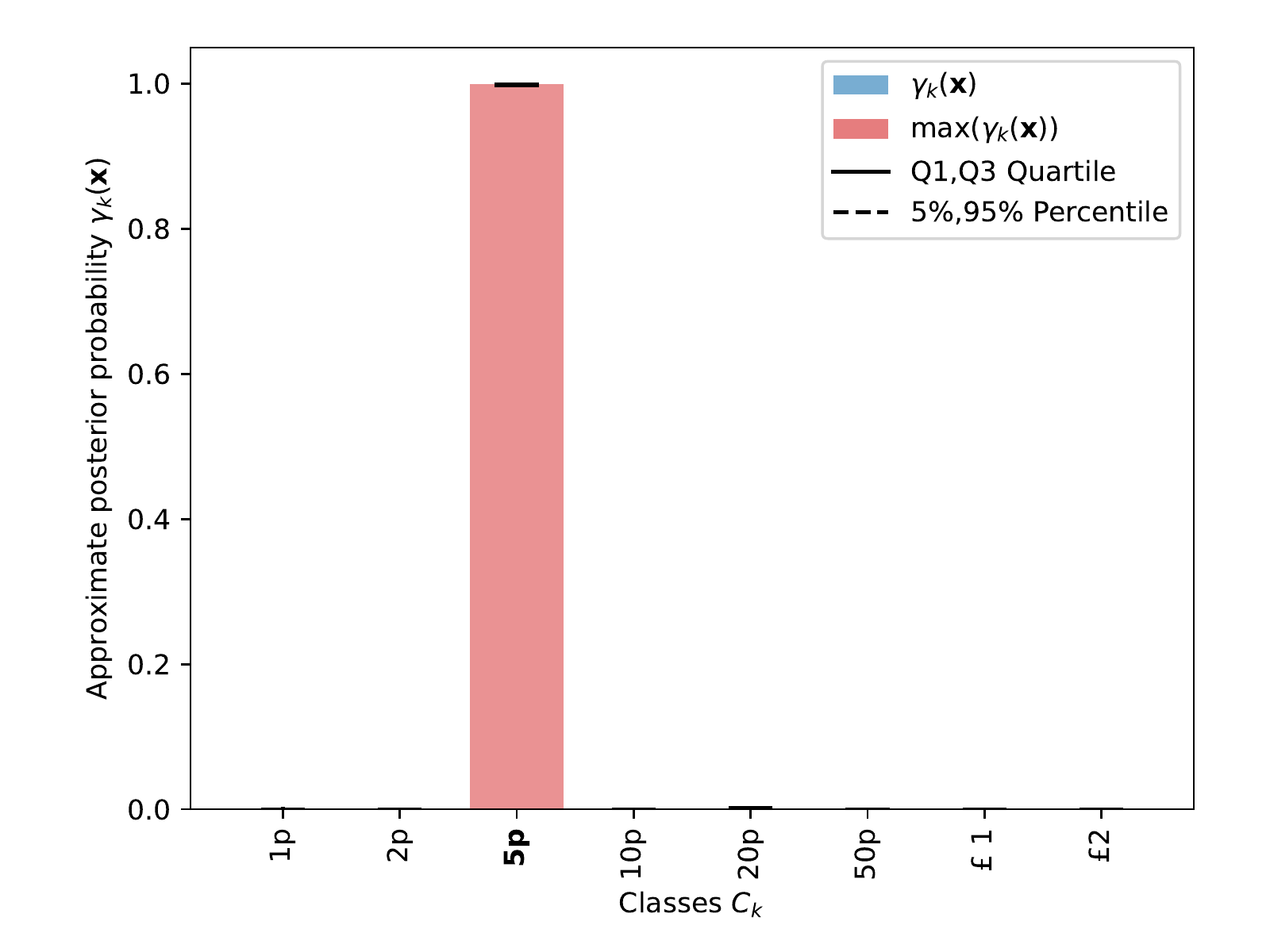} &
\includegraphics[scale=0.5]{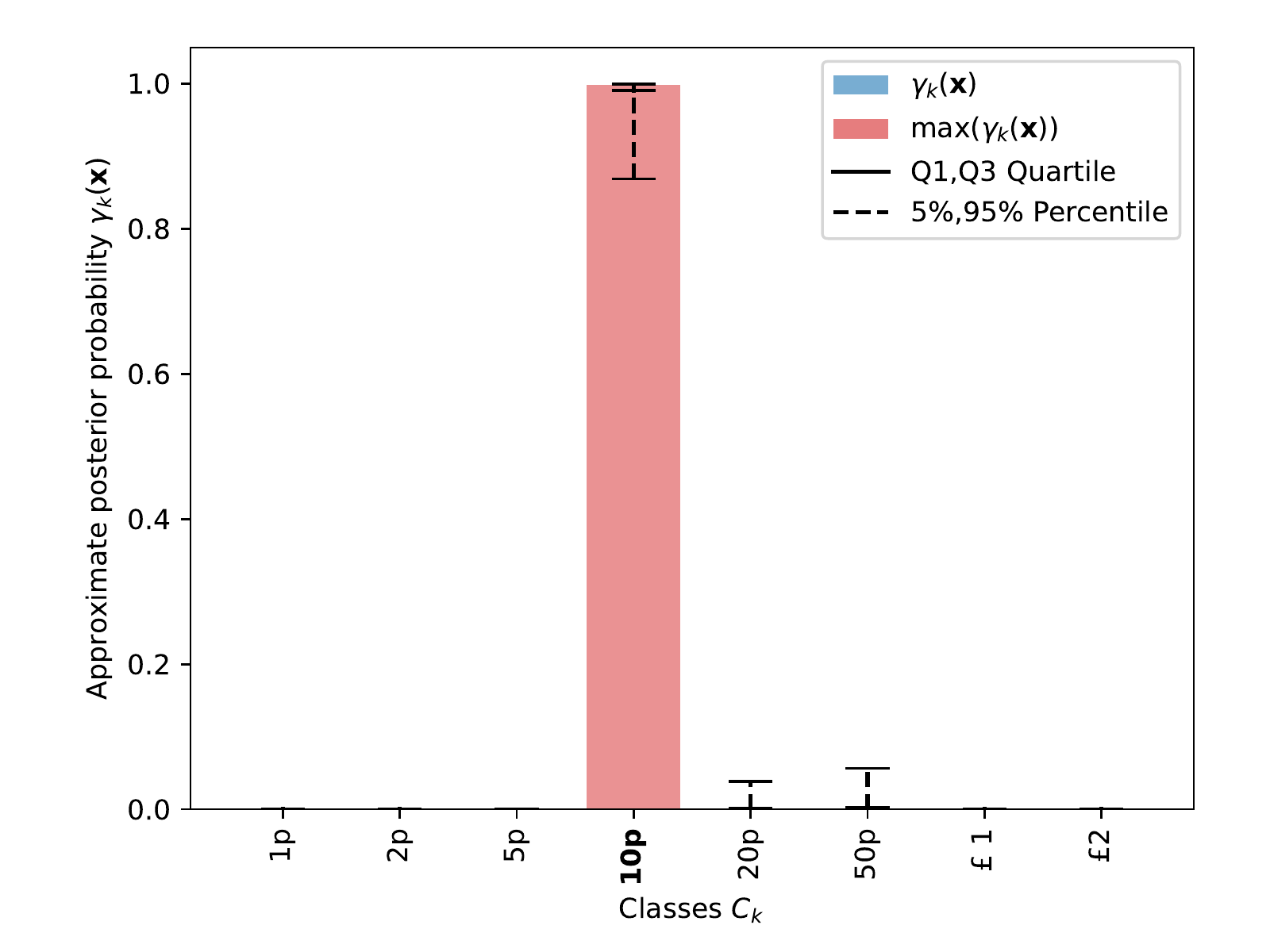} \\
\textrm{\footnotesize{(a) 5p}} & \textrm{\footnotesize{(b) 10p }}
\end{array}$
\end{center}
\caption{Set of British coins: Approximate posterior probabilities $\gamma_k({\mathbf x}) \approx p(C_k|{\mathbf x})$, $k=1,\ldots,K$, using the  logistic regression classifier using $P^{(k)}=2000$ where (a) $({\mathbf x}, {\mathbf t})\in D^{(\text{test},(3))}$ and (b) $({\mathbf x}, {\mathbf t})\in D^{(\text{test},(4))}$. }
\label{fig:postprobcoins}
\end{figure}

Some illustrative approximate posterior probability distributions 
$\gamma_k ({\mathbf x}) \approx p(C_k|{\mathbf x})$, $k=1\ldots,K,$ that are obtained for SNR=10dB are illustrated in Figure ~\ref{fig:postprobcoins}. For each $({\mathbf x}, {\mathbf t})\in D^{(\text{test})}$,  a potentially different distribution can be expected and, in the cases shown, we have chosen $({\mathbf x}, {\mathbf t})\in D^{(\text{test},(3)}$ and $({\mathbf x}, {\mathbf t})\in D^{(\text{test},(4))}$
 so that correct classifications should be $C_3$ (a 5p coin) and  $C_4 $ (a 10p coin), respectively. Additionally, the bars we show are for the median value $\gamma_{k,50} $, obtained by considering all the samples  $({\mathbf x}, {\mathbf t})\in D^{(\text{test},(3))}$ and   $({\mathbf x}, {\mathbf t})\in D^{(\text{test},(4))}$, respectively, and we also indicate the $Q_1$, $Q_3$ quartiles as well as $\gamma_{k,5}$ and $\gamma_{k,95}$ percentiles, which have been obtained using (\ref{eqn:percentile}). The cases shown correspond to the best and worst cases among all $D^{(\text{test},(k))}$ for this level of noise. A common trait of logistic regression is that it gives a strong $\gamma_k( {\mathbf x})$ for one class and low values for the other classes and the results we obtain also exhibit this. Comparing $\gamma_k( {\mathbf x})$, $k=1,\ldots,K$,  for $({\mathbf x}, {\mathbf t})\in D^{(\text{test},(3)}$ and $({\mathbf x}, {\mathbf t})\in D^{(\text{test},(4))}$, we find the most likely classes correspond to the 5p and 10p coins, respectively. For the 5p coin, the inter quartile and inter percentile ranges are small and so we have high confidence in this prediction and a low variability. For the 10p coin, they are larger indicating we have less confidence in the prediction and a higher variability.

\begin{figure}[h]
$$\begin{array}{cc}
\includegraphics[scale=0.5]{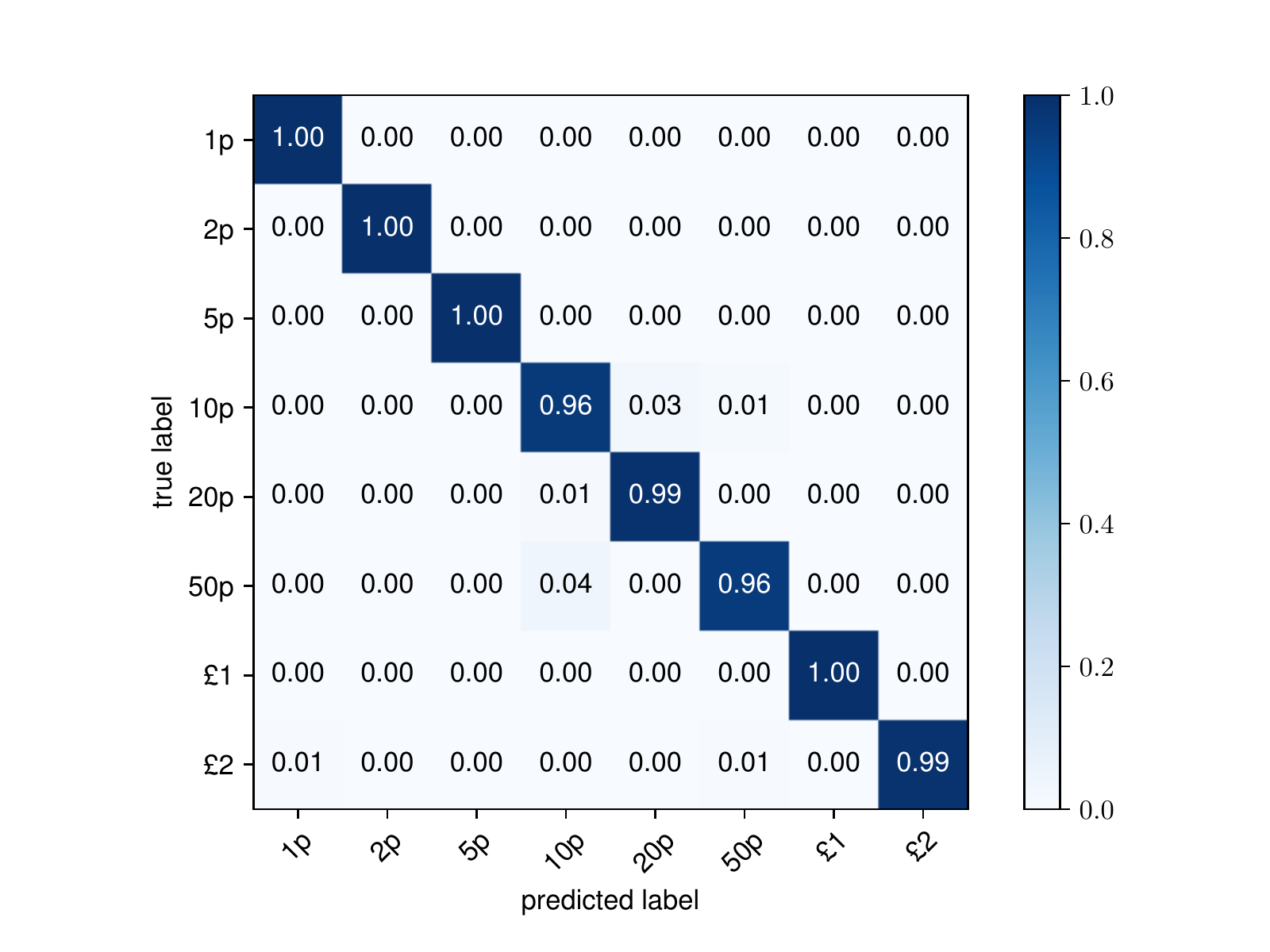} &
\includegraphics[scale=0.5]{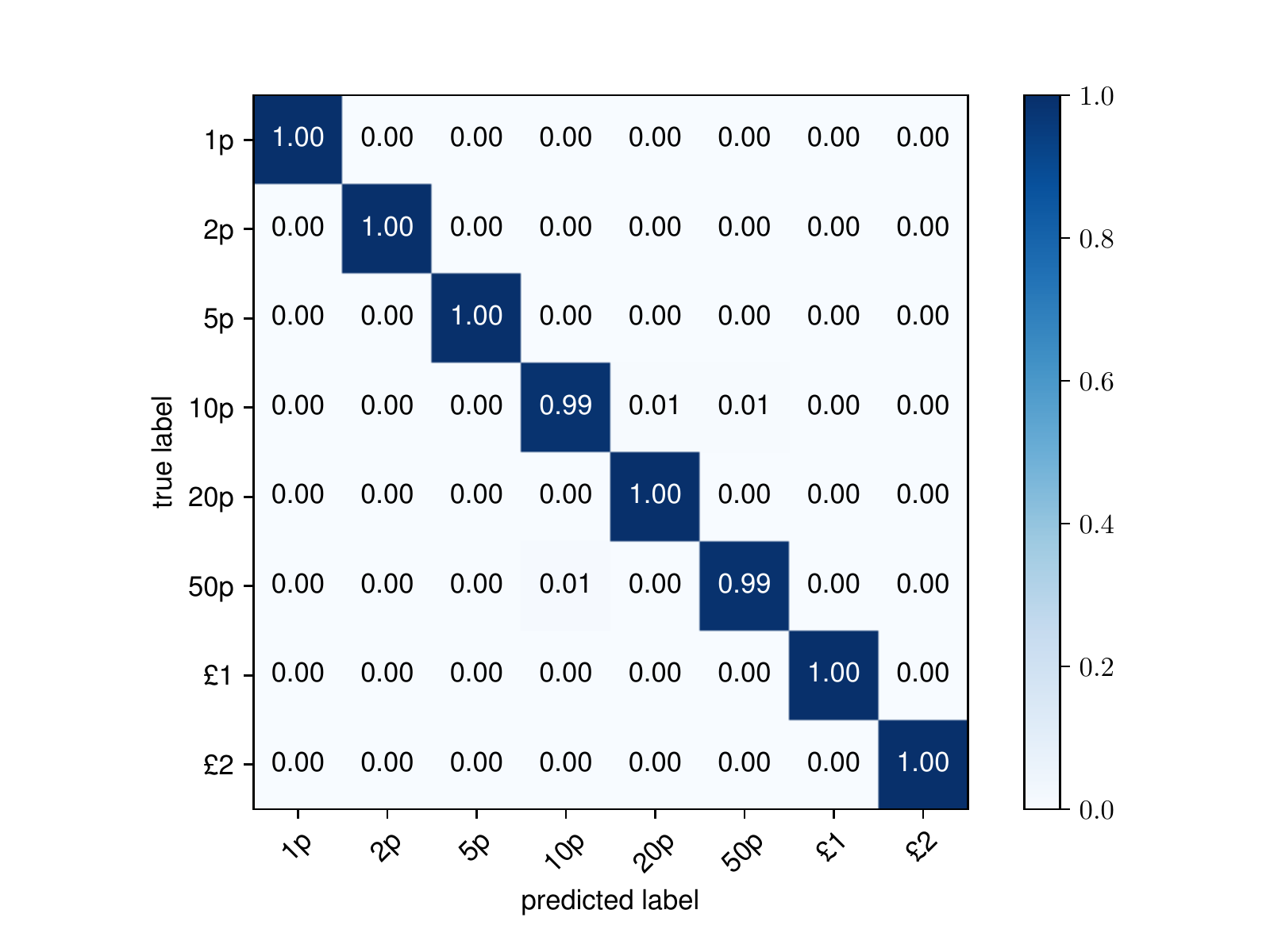} \\
\textrm{\footnotesize{(a) $P^{(k)}=50$}} & \textrm{\footnotesize{(b) $P^{(k)}=2000$}} 
\end{array}$$
\caption{Set of British coins: Confusion matrices for noise corresponding to SNR=10dB showing  showing the effect of different numbers of instance per class (a) $P^{(k)}=50$ and (b) $P^{(k)}=2000$.} \label{fig:confmatrixcoins}
\end{figure}

Next, we consider the frequentist approximations to $p(C_j |{\mathbf x})$ for $({\mathbf x},{\mathbf t})\in D^{(\text{test},(i))}$ presented in the form of a confusion matrix with entries $({\mathbf C})_{ij}$, $i,j=1,\ldots,K$,  for the cases of SNR=20dB and SNR=10dB in Figure~\ref{fig:confmatrixcoins}, using the approach described in Section~\ref{sect:confmat}.
We consider the case with SNR=10dB and compare the performance of the classifier using $P^{(k)}=50$ and $P^{(k)}=2000$ instances per class. There are only a small number of misclassifications for the $P^{(k)}=50$ case and these are further reduced by using $P^{(k)}=2000$.

\subsection{Multi-class problem}\label{sect:multi}
\subsubsection{Construction of the multi-class dictionary} \label{sect:multidata}
\begin{figure}[!h]
\begin{center}
$\begin{array}{cc}
\begin{array}{cc}
\includegraphics[width=0.2\textwidth, keepaspectratio]{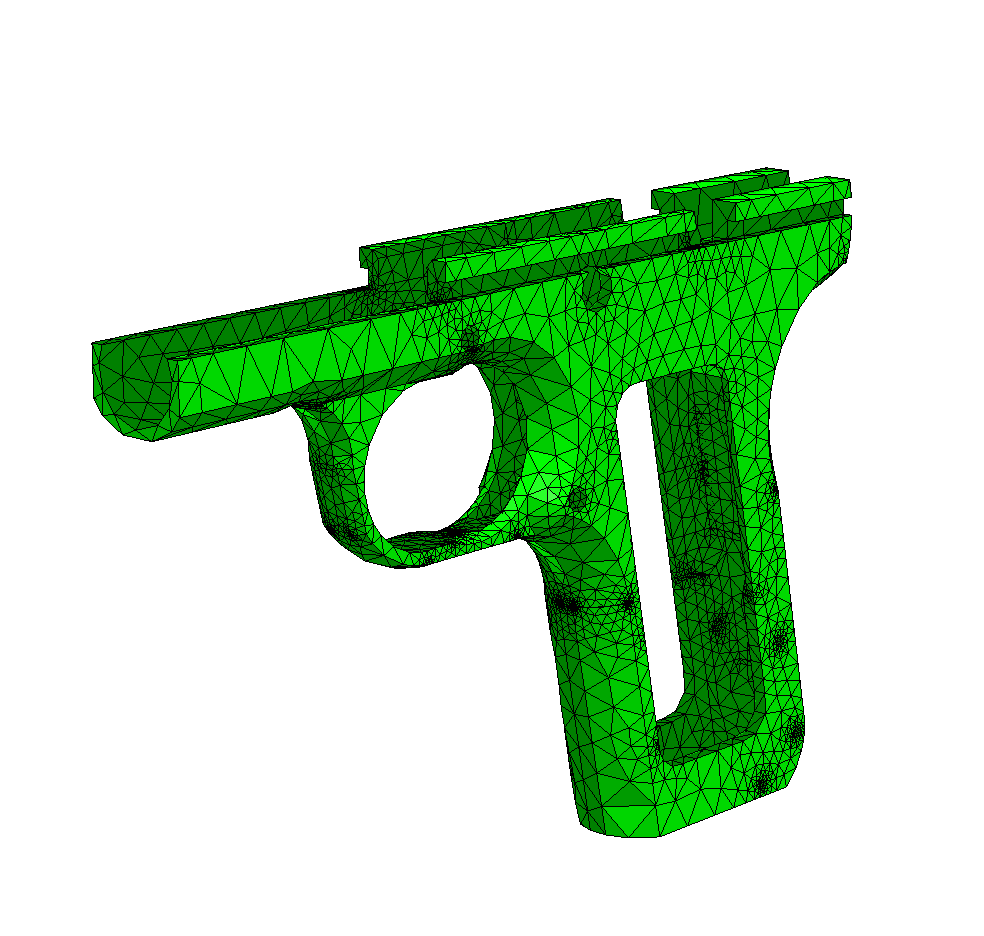} &
\includegraphics[width=0.2\textwidth, keepaspectratio]{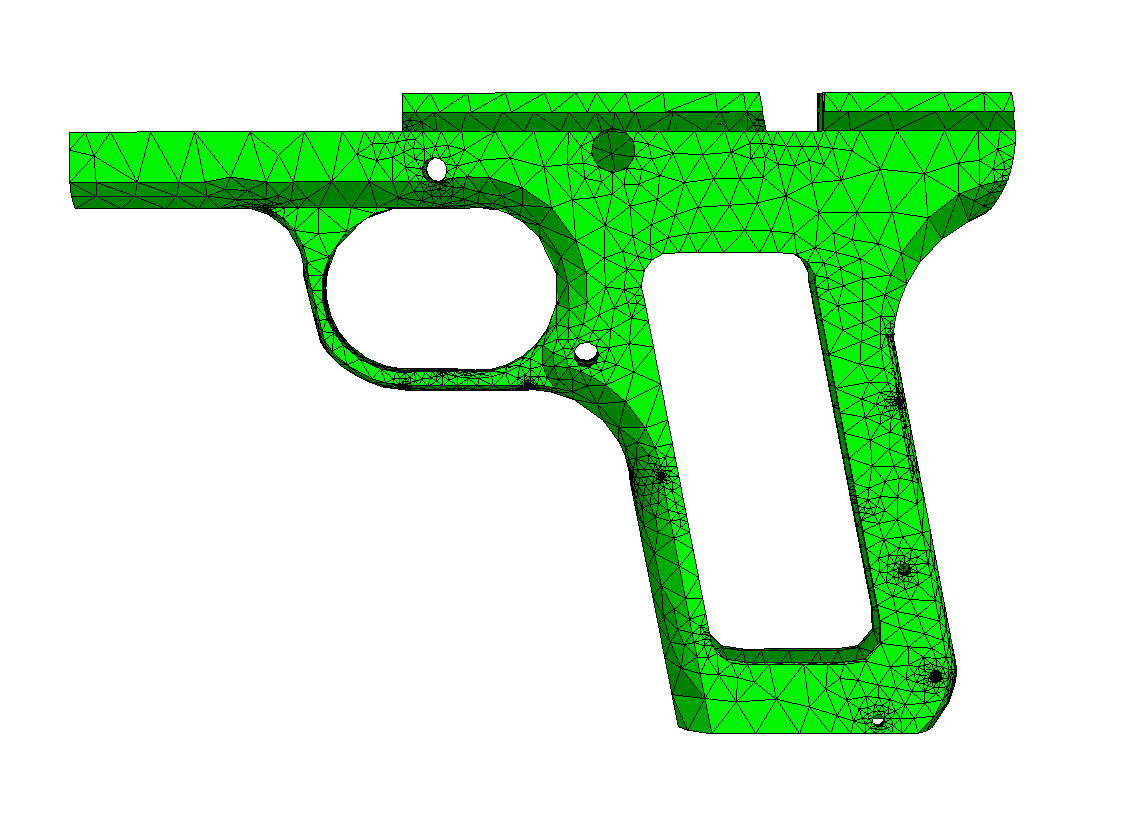}
\end{array} &
\begin{array}{ccc}
\includegraphics[width=0.15\textwidth, keepaspectratio]{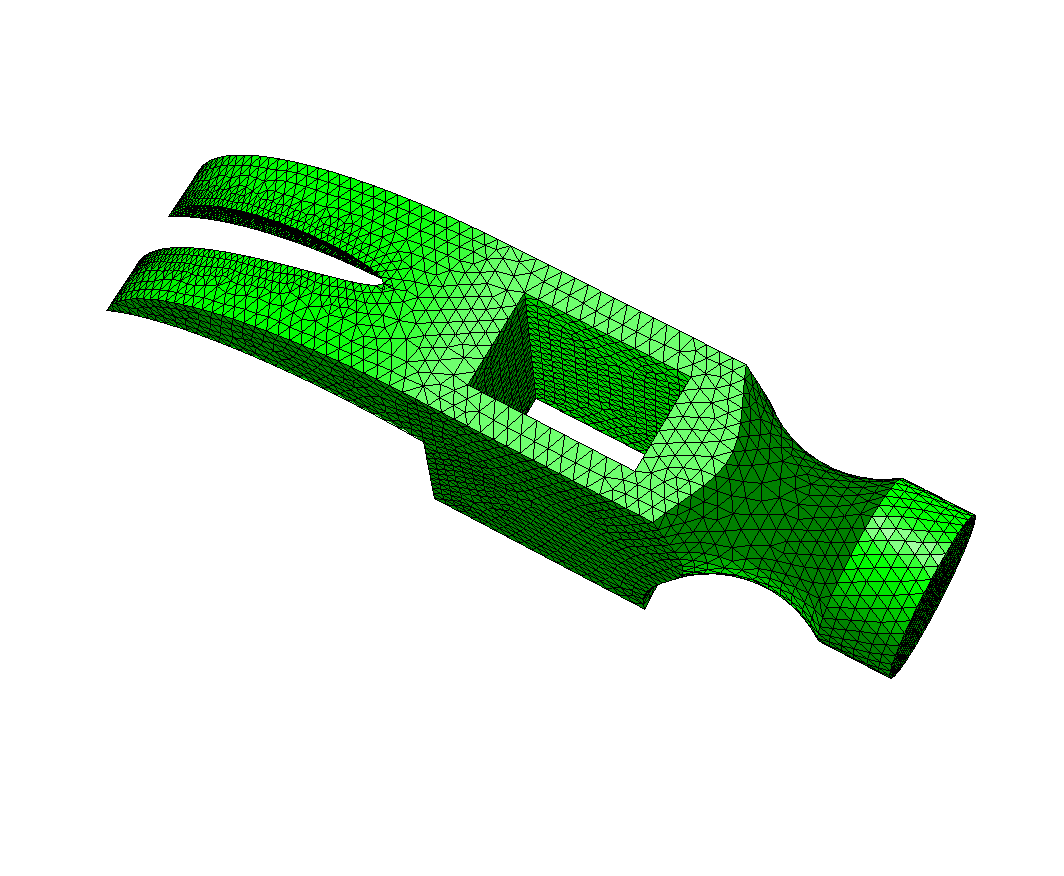} &
\includegraphics[width=0.15\textwidth, keepaspectratio]{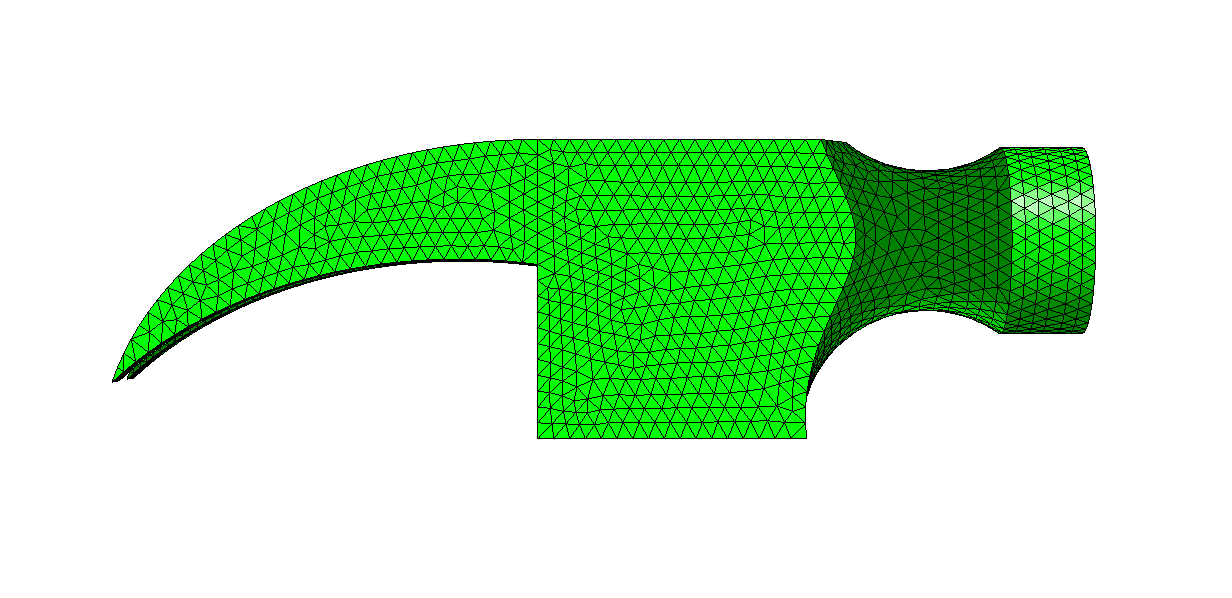} &
\includegraphics[width=0.15\textwidth, keepaspectratio]{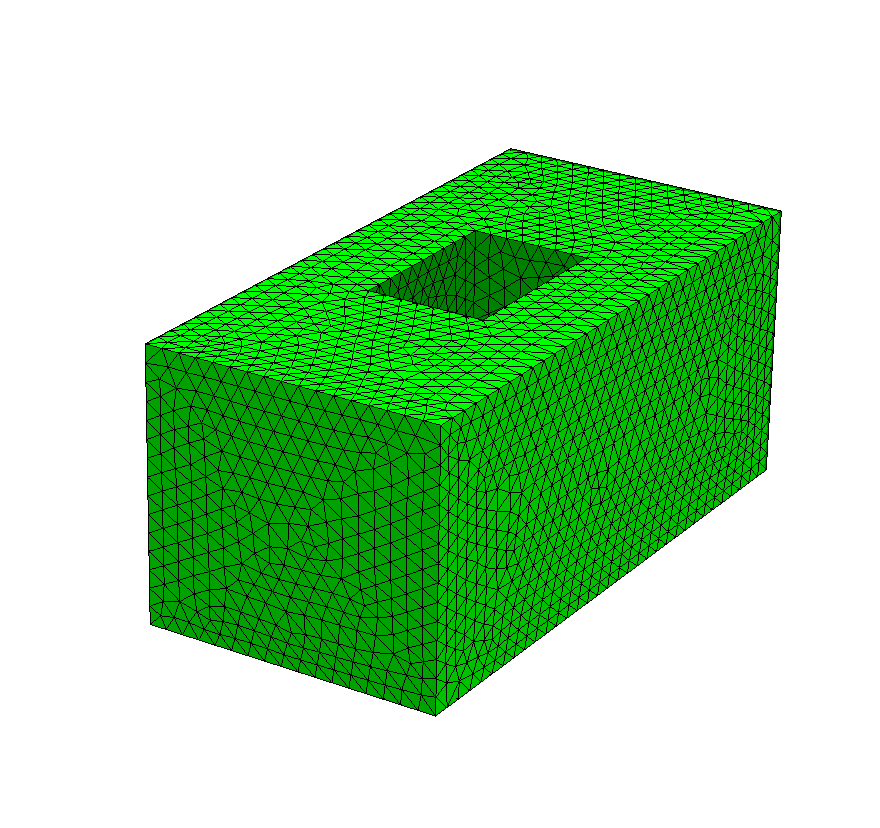} 
\end{array}\\
\text{Gun model } (C_1) & \text{Hammer heads } (C_2) \end{array}$
\end{center}
\begin{center}
$\begin{array}{ccccc}
  \includegraphics[width=0.15\textwidth, keepaspectratio]{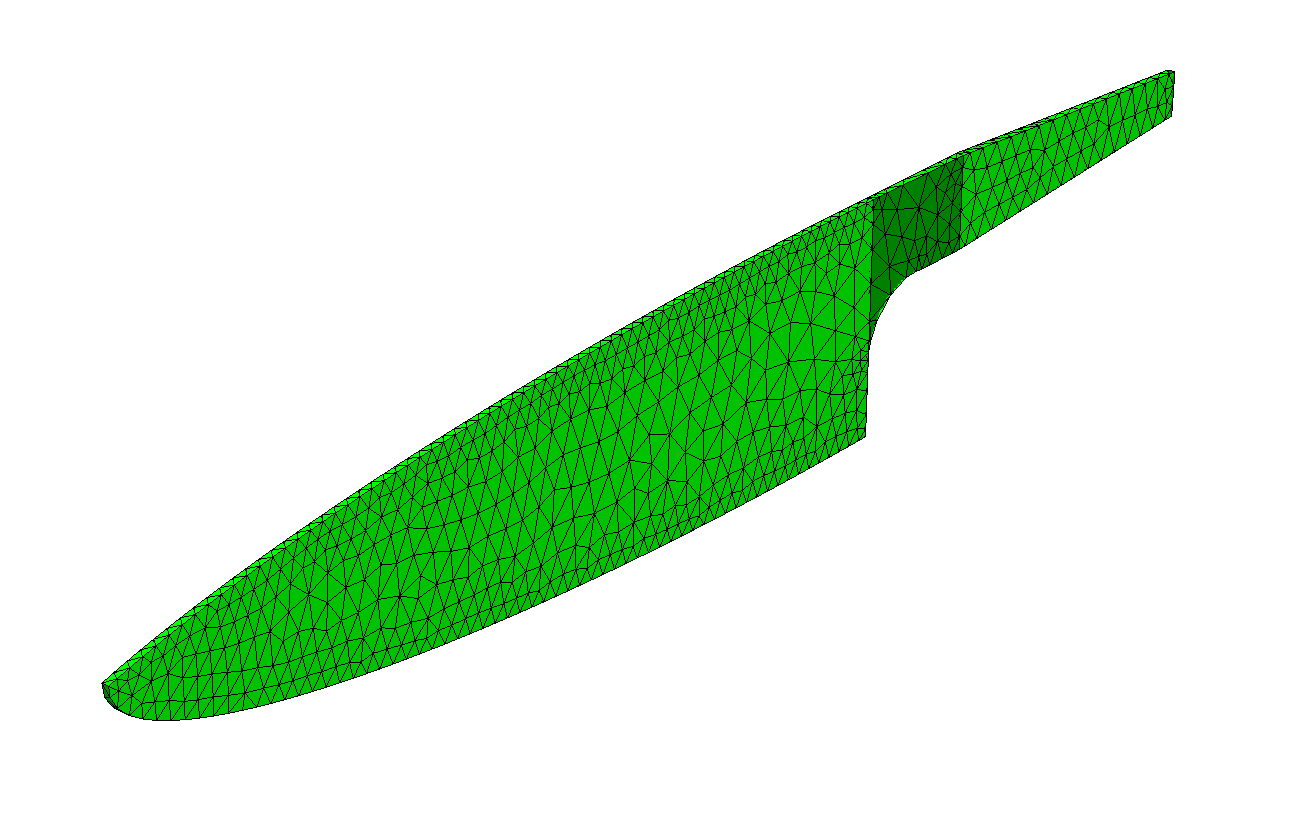} &
 \includegraphics[width=0.15\textwidth, keepaspectratio]{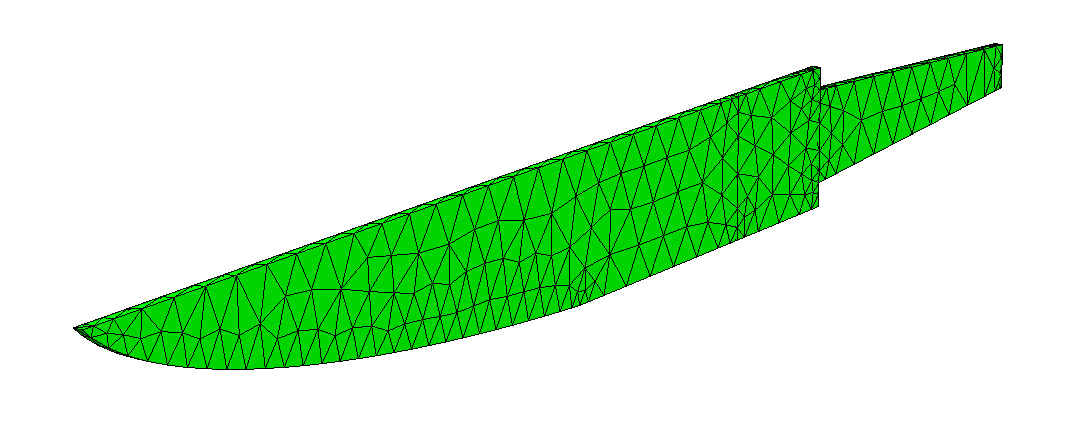} &
 \includegraphics[width=0.15\textwidth, keepaspectratio]{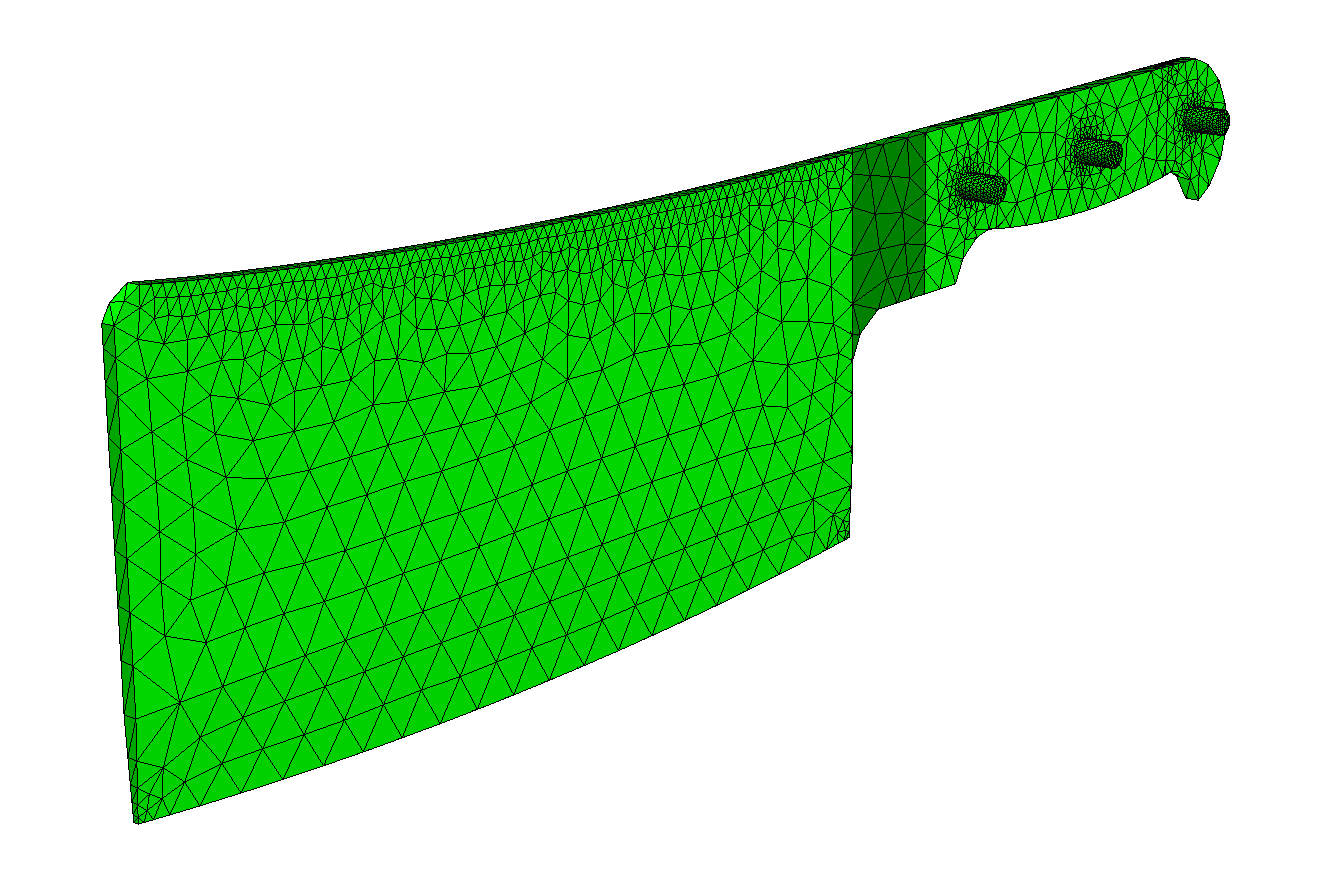} &
  \includegraphics[width=0.1\textwidth, keepaspectratio]{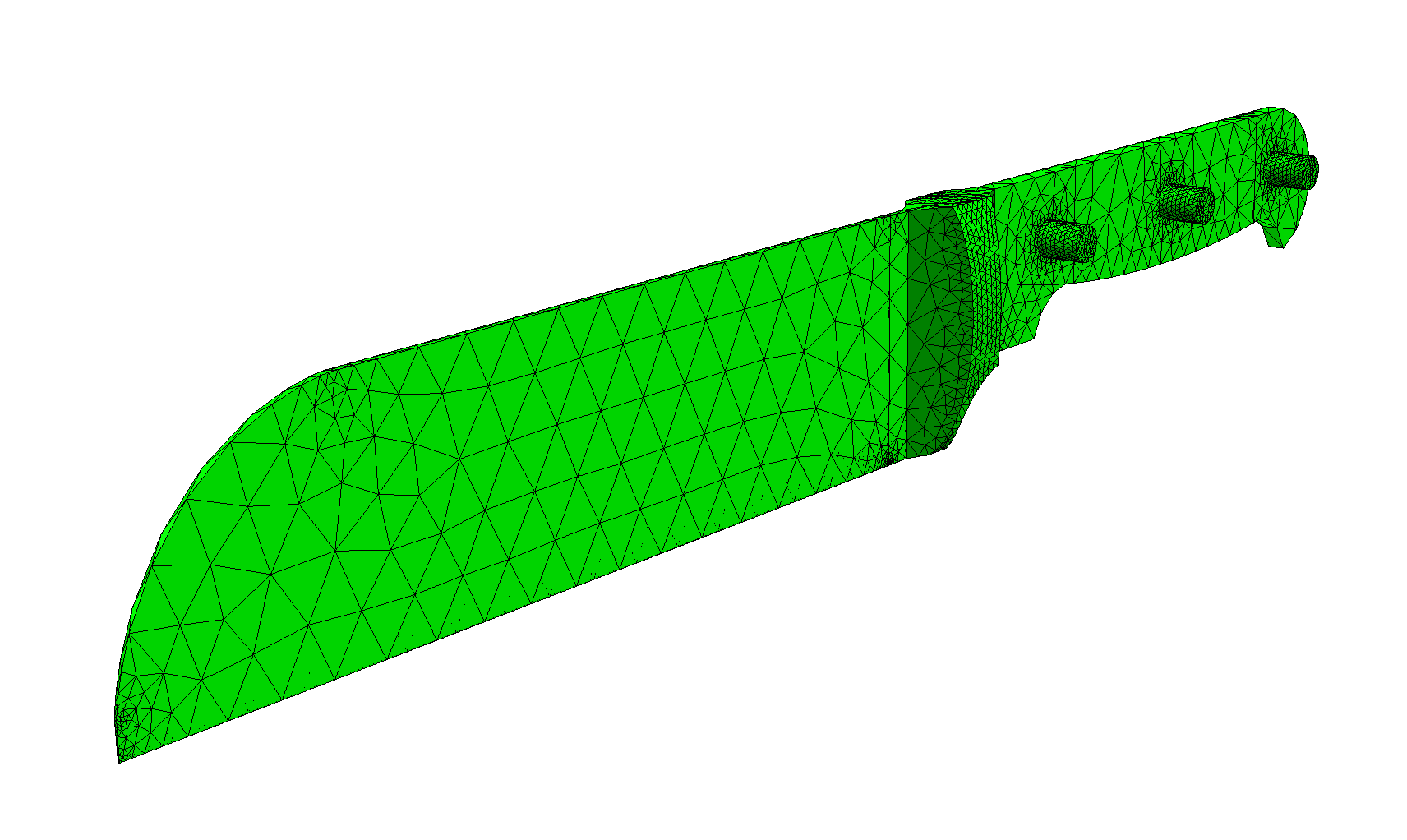} &
 \includegraphics[width=0.1\textwidth, keepaspectratio]{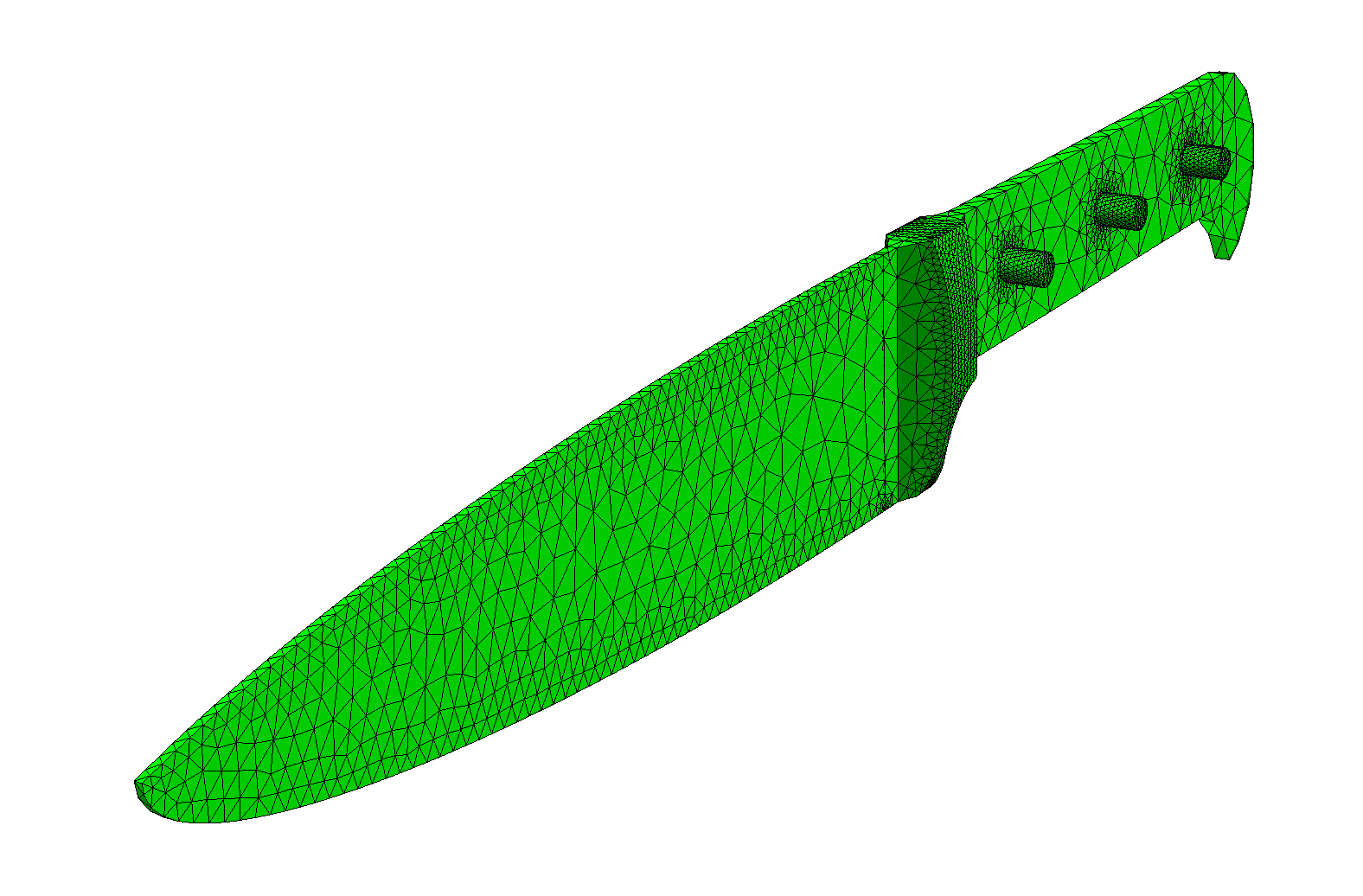} 
 \end{array}$\\ 
 Knife models $(C_3)$
 \end{center}
\begin{center}
$\begin{array}{cc}
\includegraphics[width=0.15\textwidth, keepaspectratio]{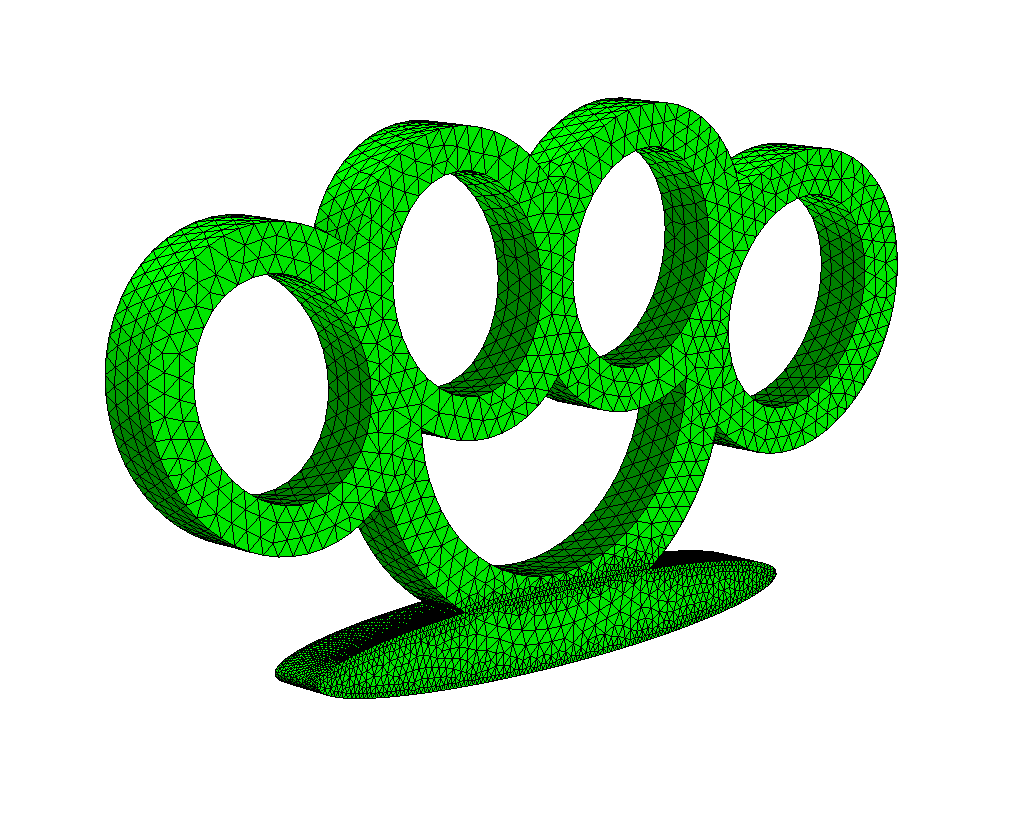} &
\includegraphics[width=0.15\textwidth, keepaspectratio]{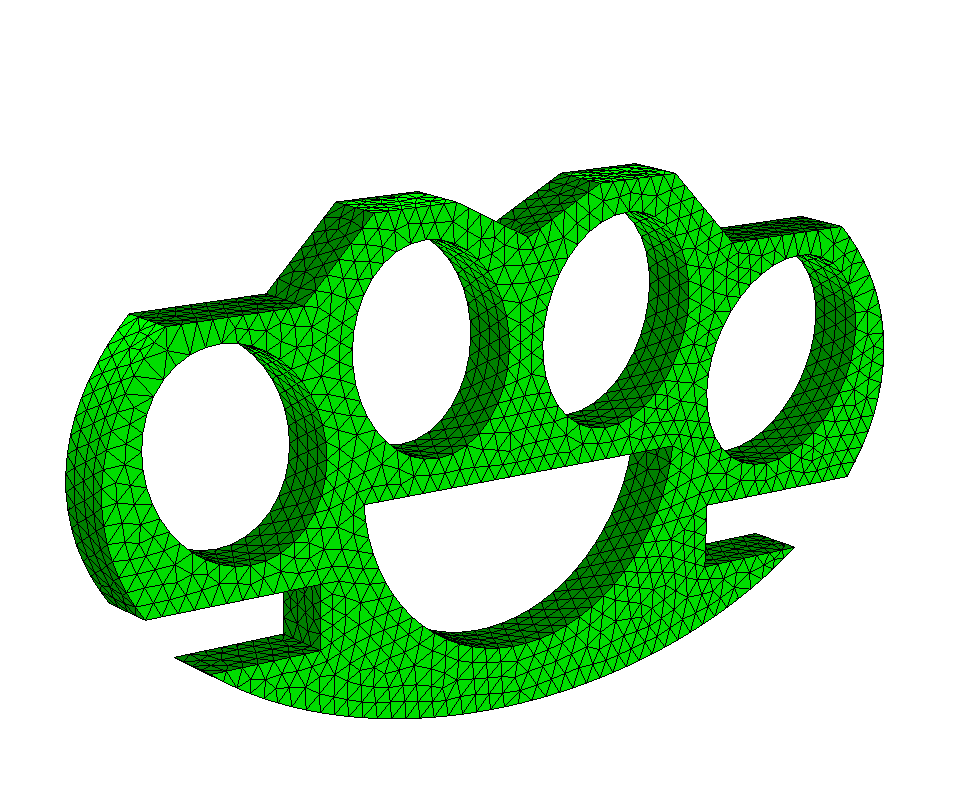} \end{array}$\\
Knuckle dusters $(C_4)$ 
\end{center}
\begin{center}
$\begin{array}{cc}
\includegraphics[width=0.2\textwidth, keepaspectratio]{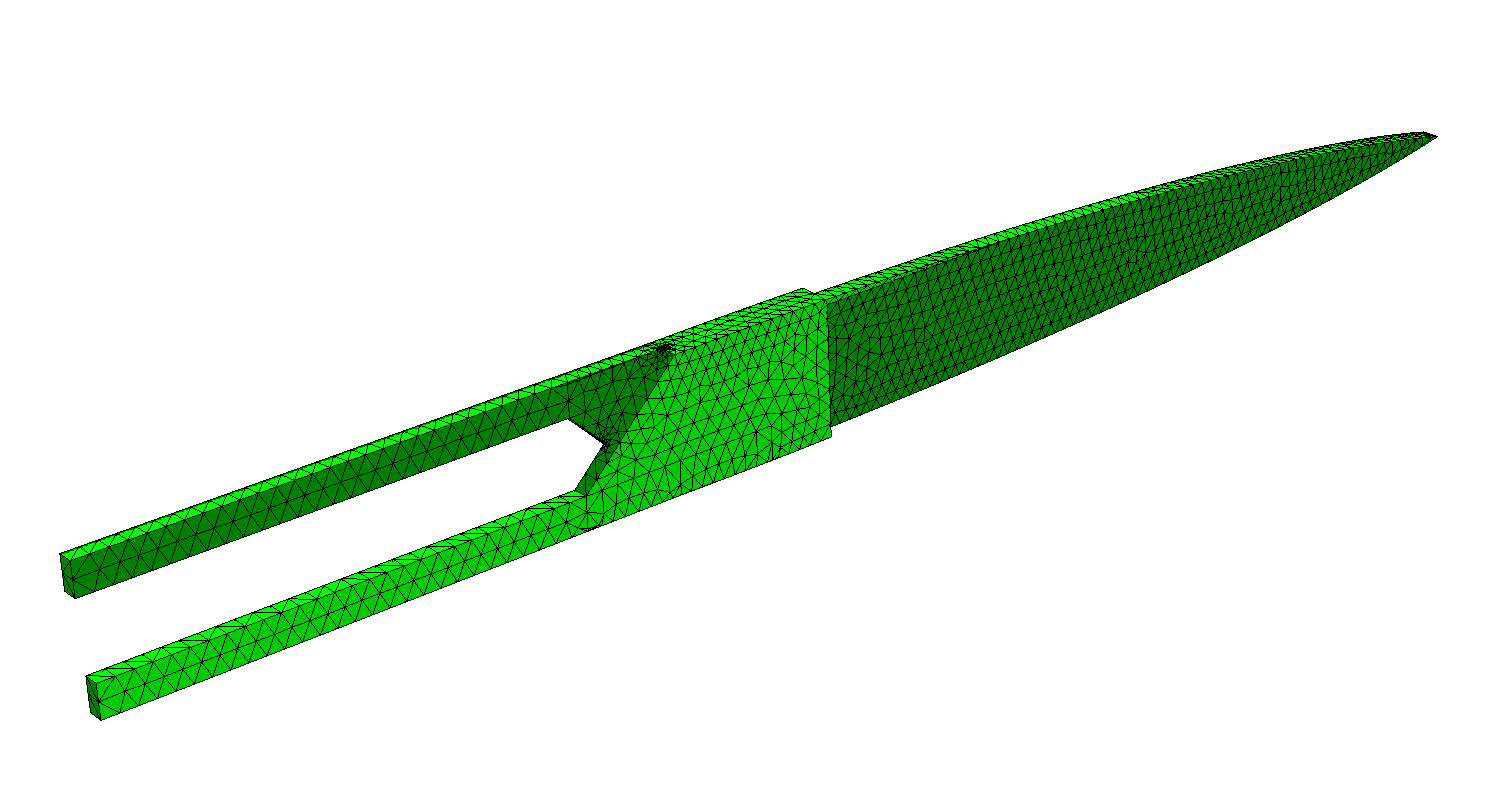} & {\includegraphics[width=0.2\textwidth, keepaspectratio]{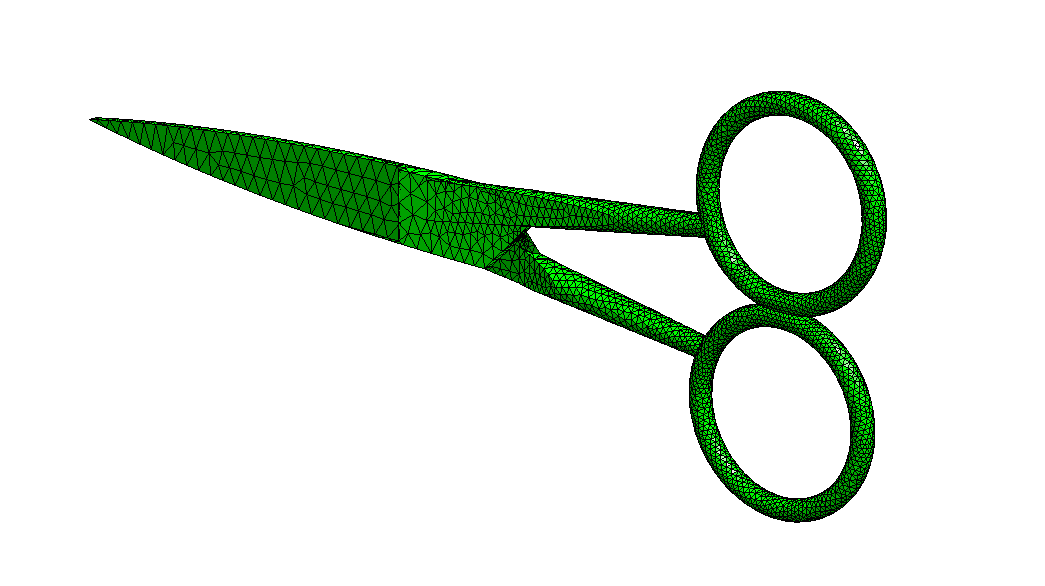}}
\end{array}$\\
Pairs of scissors  $(C_5)$
\end{center}
\begin{center}
$\begin{array}{cccc}
\includegraphics[width=0.15\textwidth, keepaspectratio]{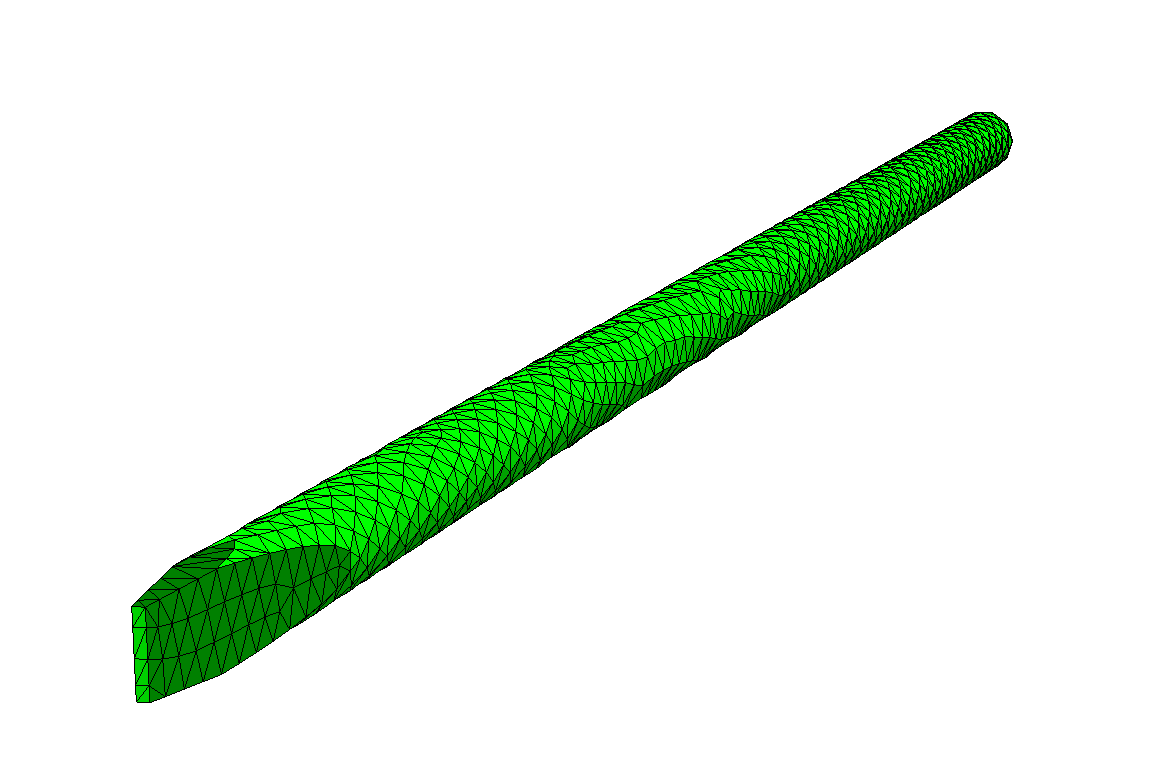} &
\includegraphics[width=0.15\textwidth, keepaspectratio]{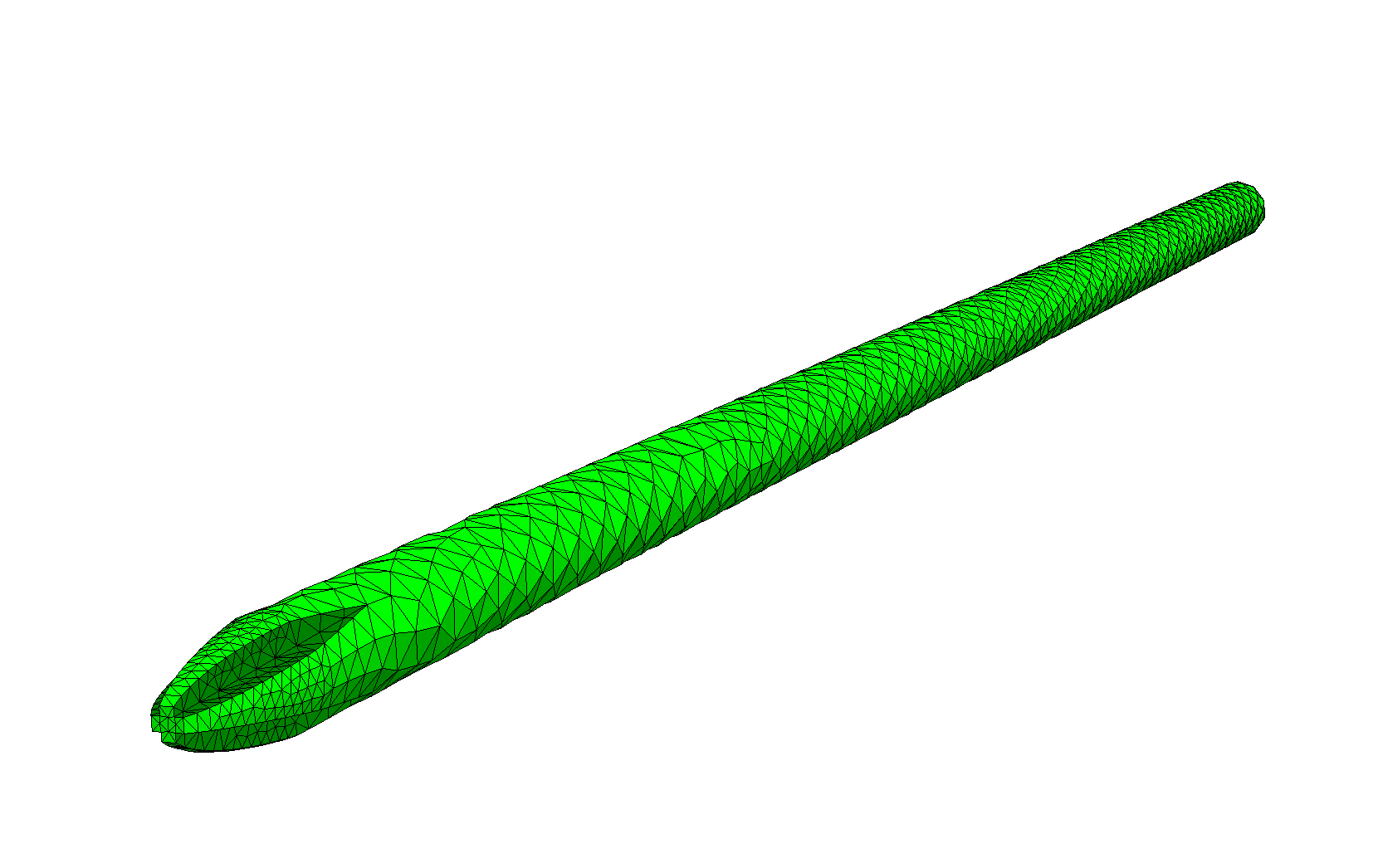} &
\includegraphics[width=0.1\textwidth, keepaspectratio]{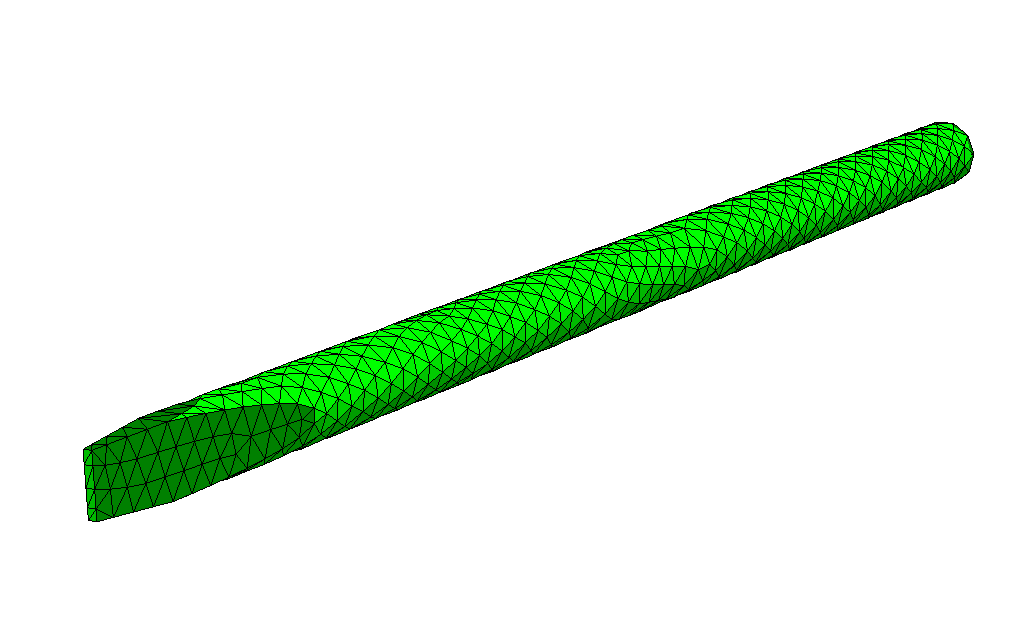} &
\includegraphics[width=0.1\textwidth, keepaspectratio]{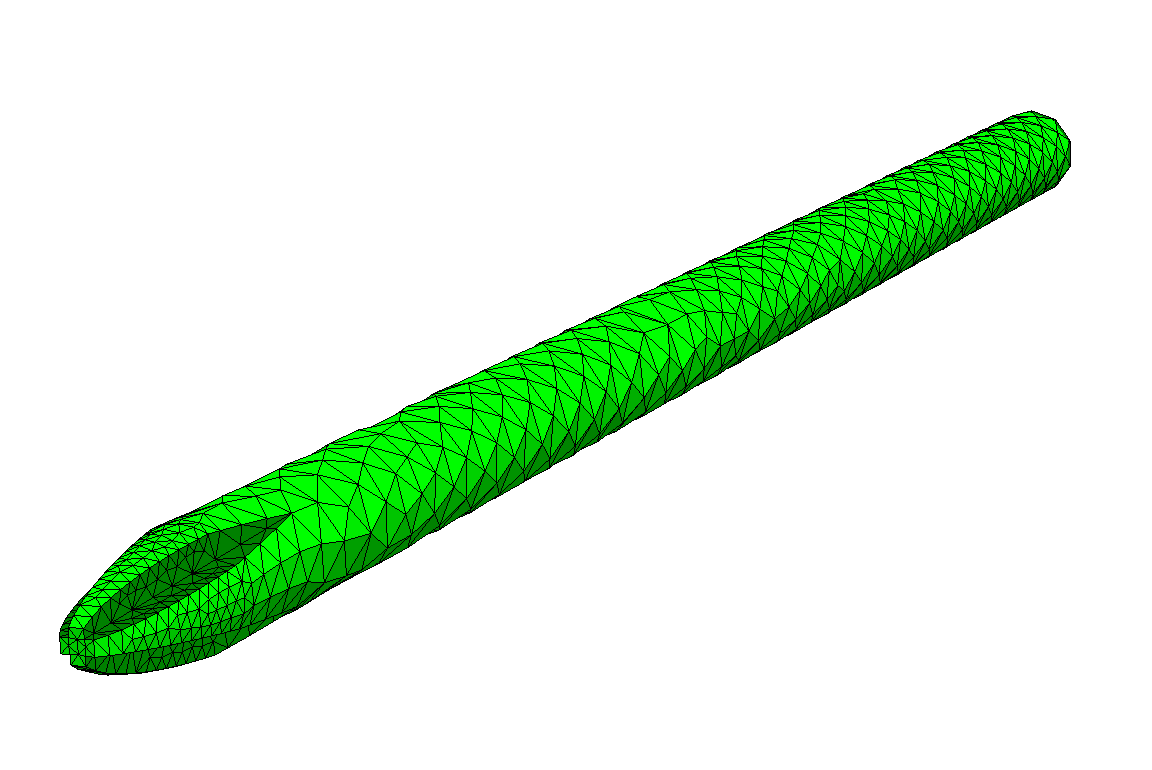} 
\end{array}$\\
Screw drivers $(C_6)$
\end{center}
\caption{Set of multiple threat and non-threat objects: Sample illustrations of some of the different threat object geometries considered  (not to scale).} \label{fig:egthreat}
\end{figure}

\begin{figure}[!h]

 \begin{center}
$\begin{array}{cc}
\begin{array}{cccc}
\includegraphics[width=0.05\textwidth, keepaspectratio]{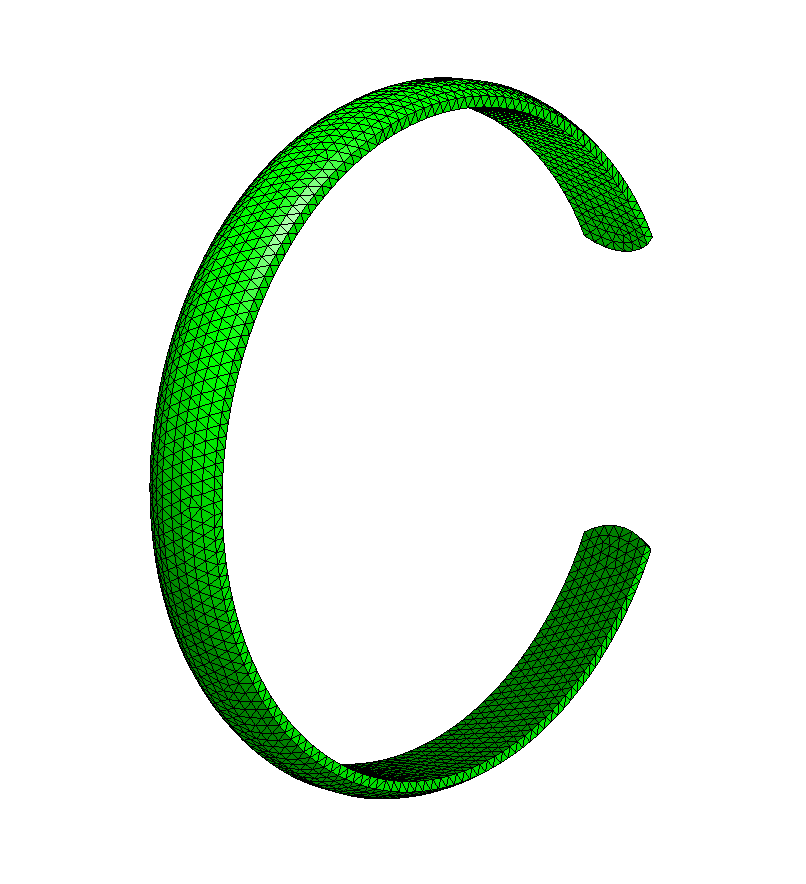} &
\includegraphics[width=0.05\textwidth, keepaspectratio]{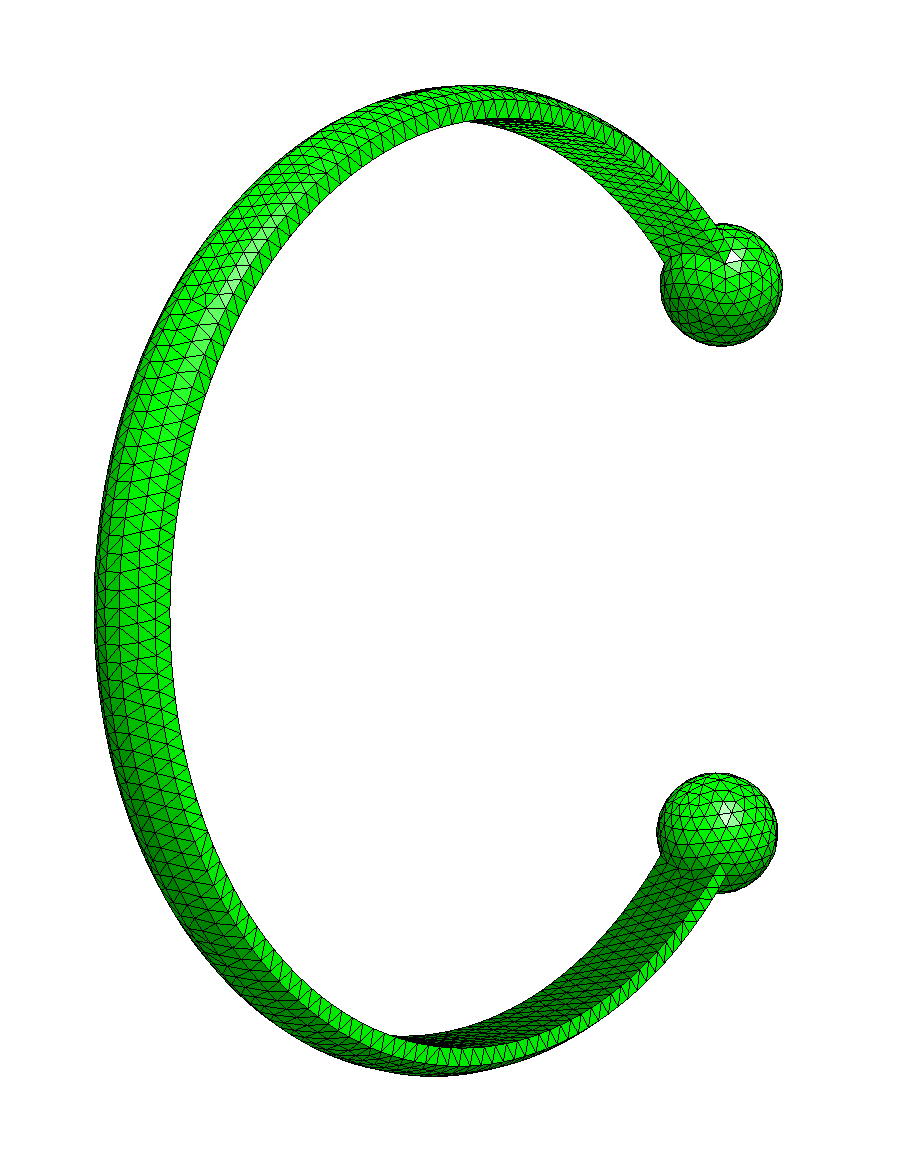} &
\includegraphics[width=0.05\textwidth, keepaspectratio]{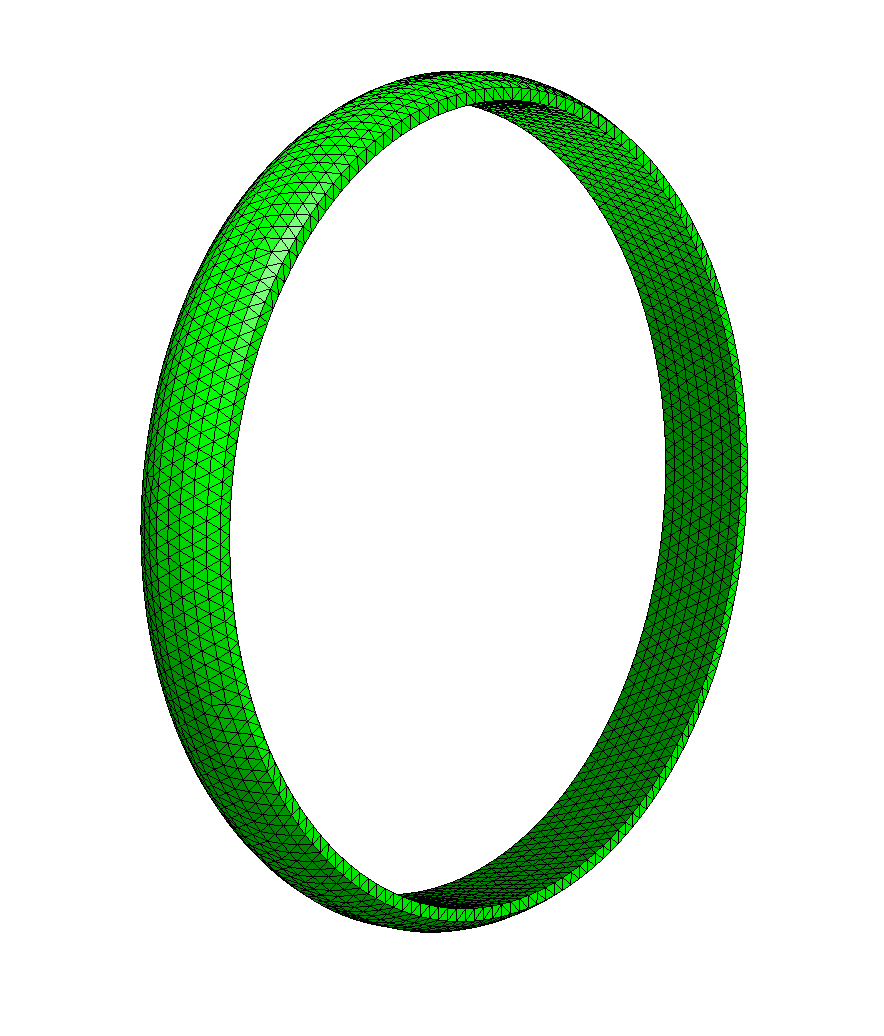} &
\includegraphics[width=0.05\textwidth, keepaspectratio]{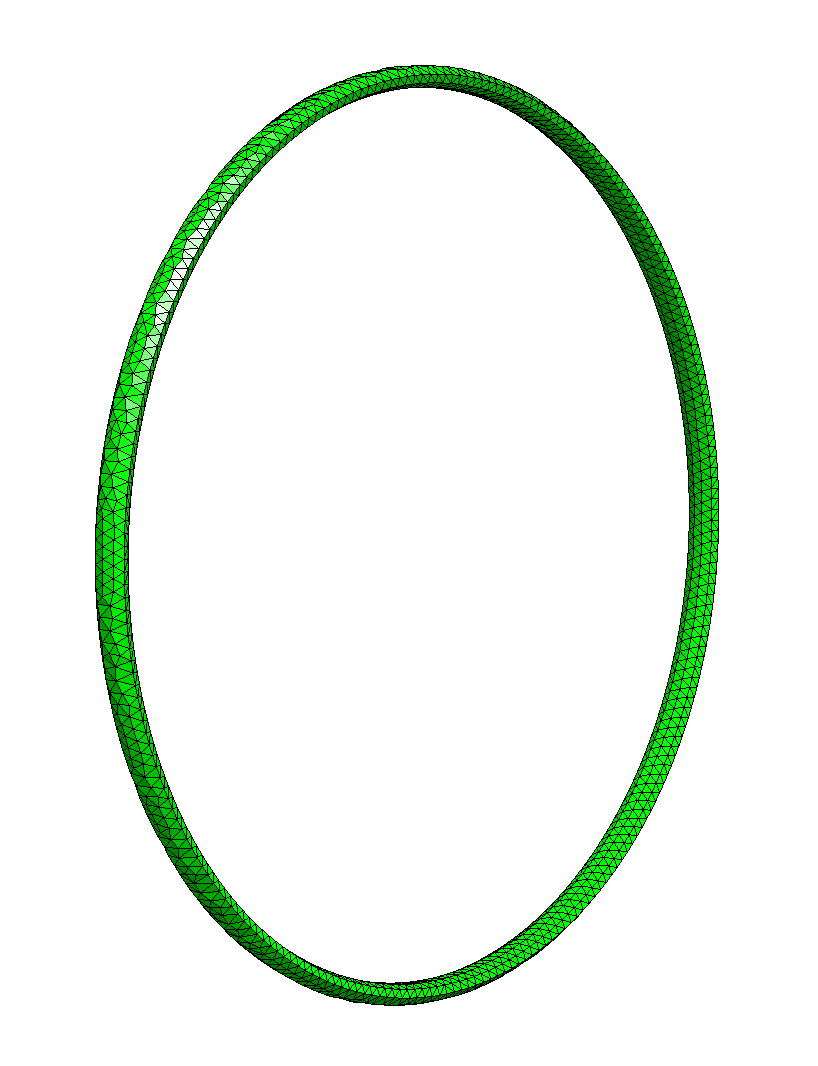} 
\end{array} &
\begin{array}{ccc}
\includegraphics[width=0.1\textwidth, keepaspectratio]{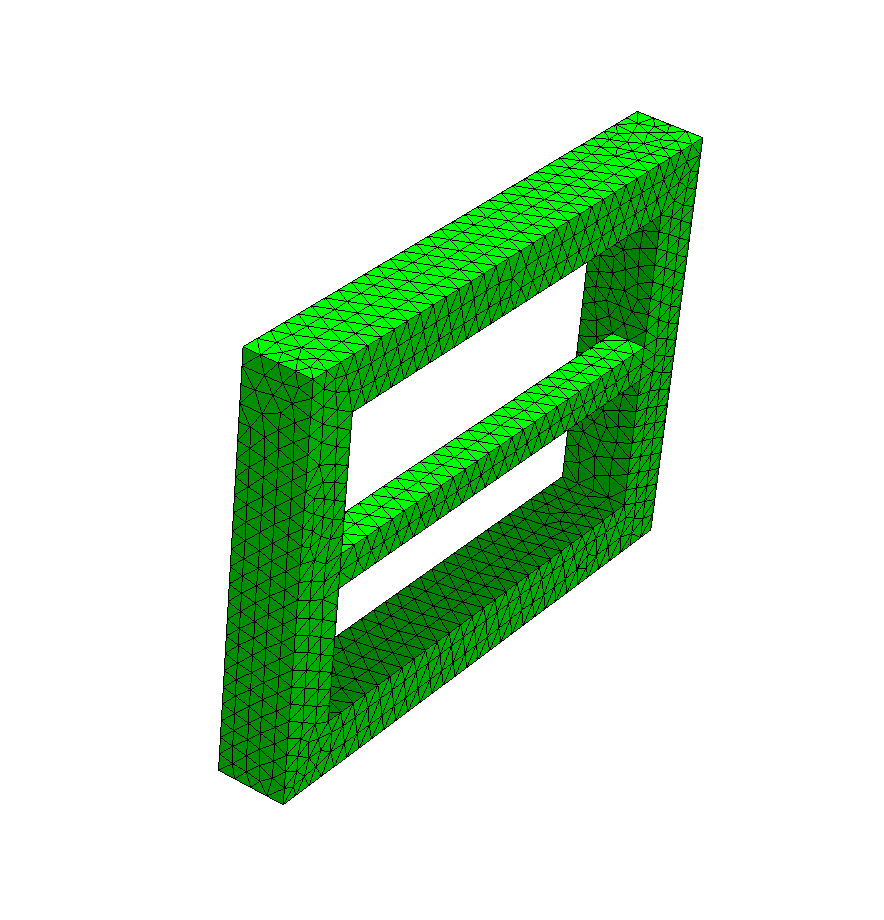} &
\includegraphics[width=0.1\textwidth, keepaspectratio]{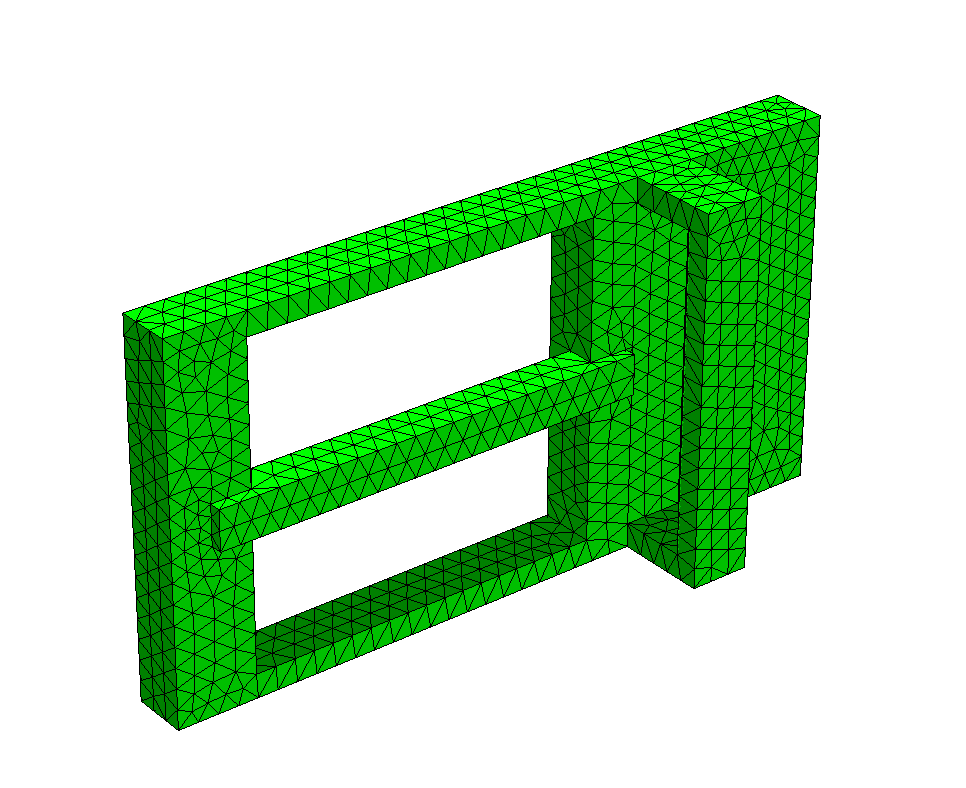} &
\includegraphics[width=0.05\textwidth, keepaspectratio]{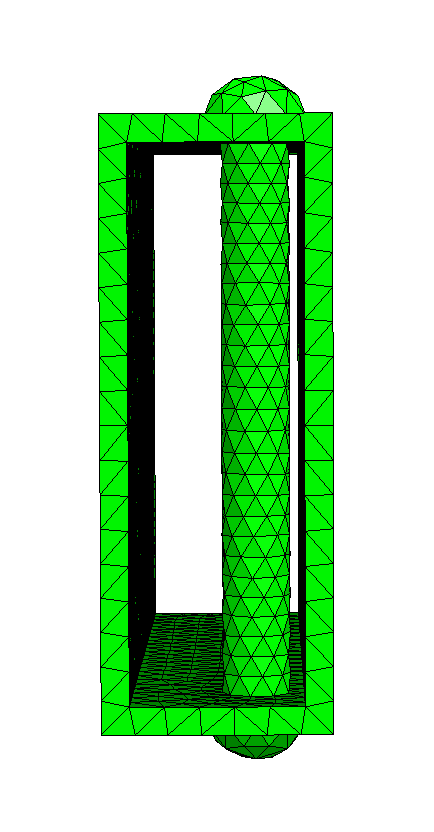}
\end{array}\\
\text{Bracelets } (C_7) &
\text{Belt buckles } (C_8) \end{array}$
\end{center}
 \begin{center}
$\begin{array}{cccccccc}
\includegraphics[width=0.05\textwidth, keepaspectratio]{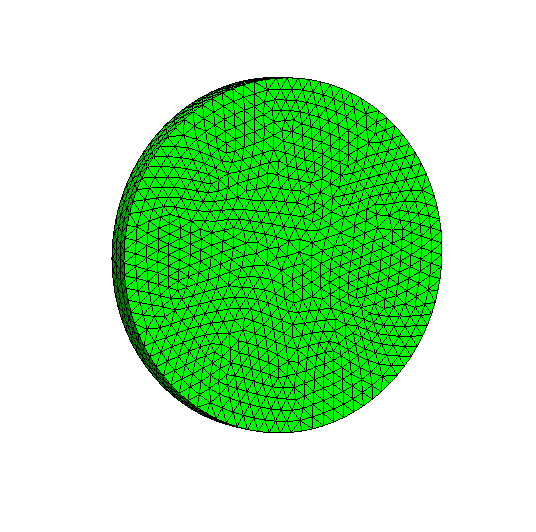} &
\includegraphics[width=0.06\textwidth, keepaspectratio]{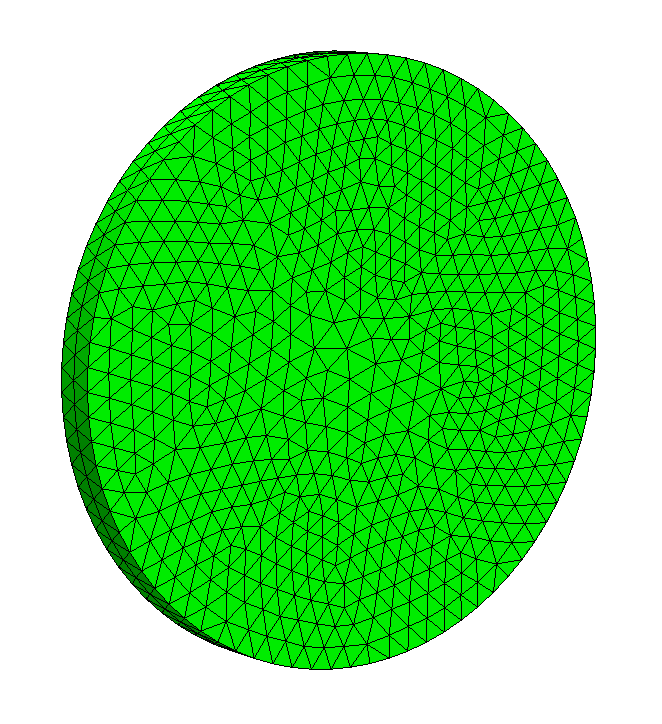} &
\includegraphics[width=0.075\textwidth, keepaspectratio]{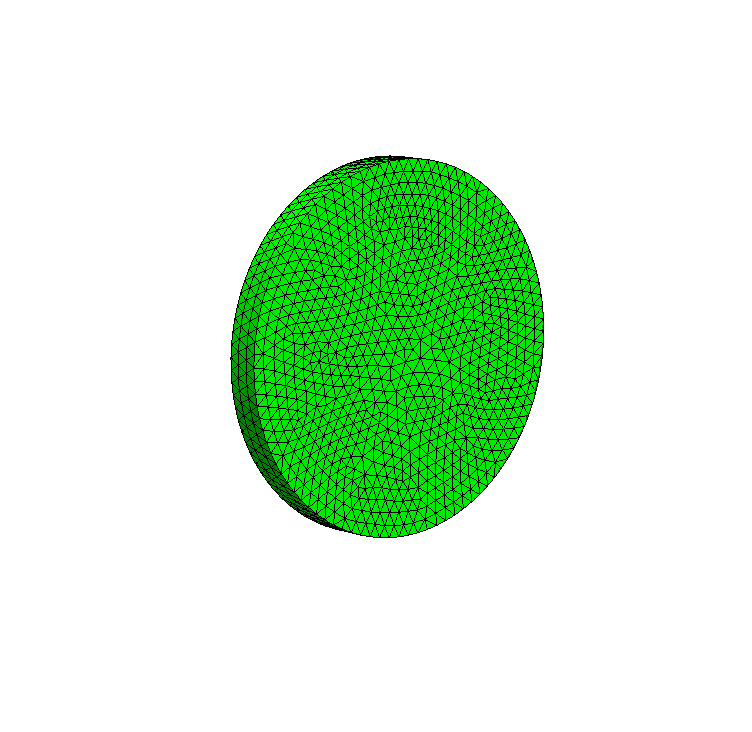} &
\includegraphics[width=0.06\textwidth, keepaspectratio]{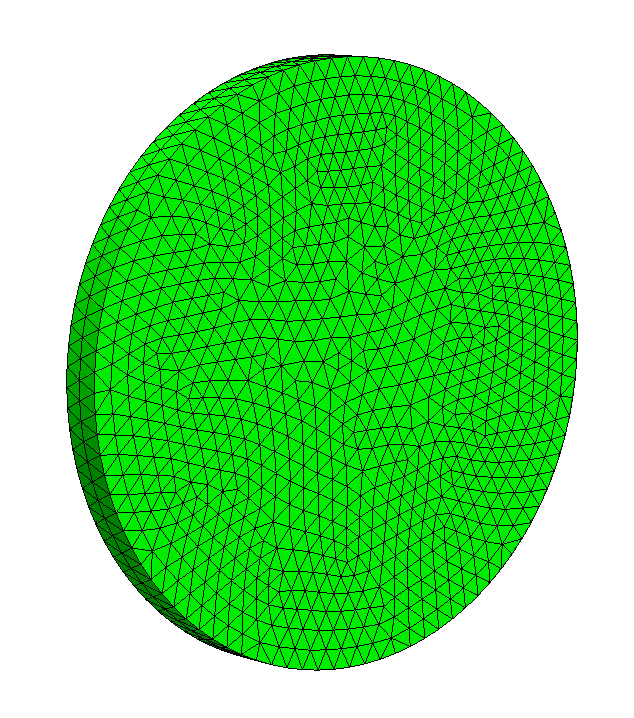} &
\includegraphics[width=0.05\textwidth, keepaspectratio]{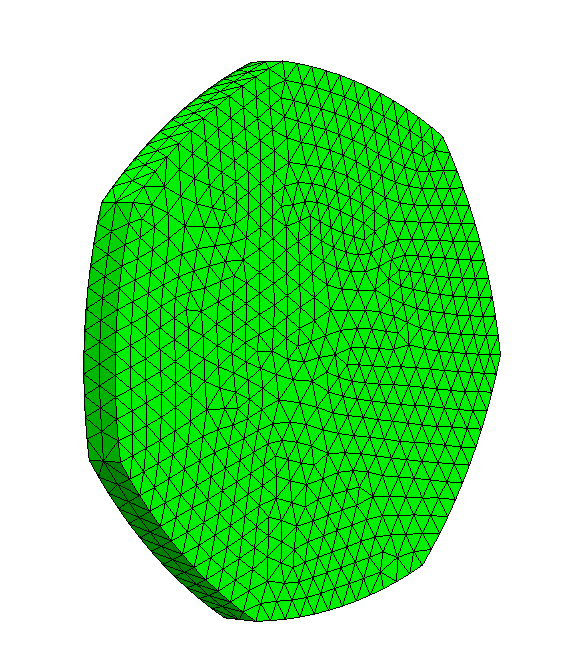} &
\includegraphics[width=0.1\textwidth, keepaspectratio]{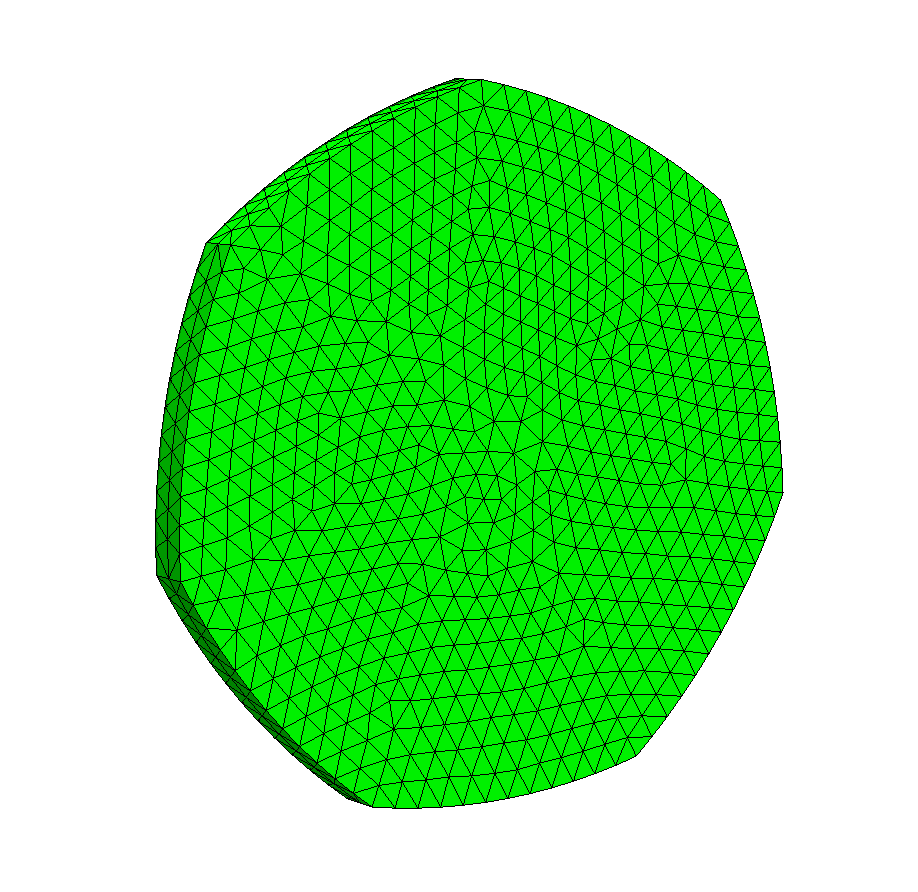} &
\includegraphics[width=0.1\textwidth, keepaspectratio]{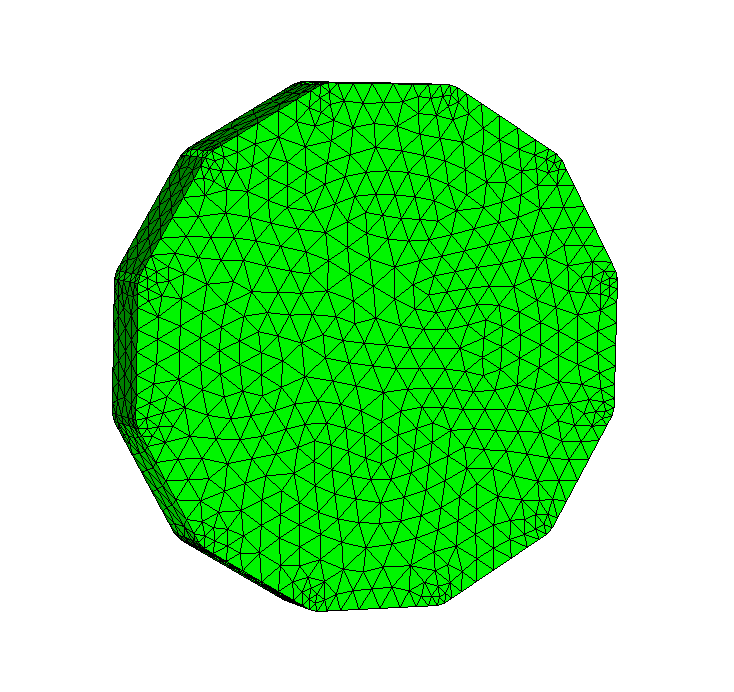} &
\includegraphics[width=0.1\textwidth, keepaspectratio]{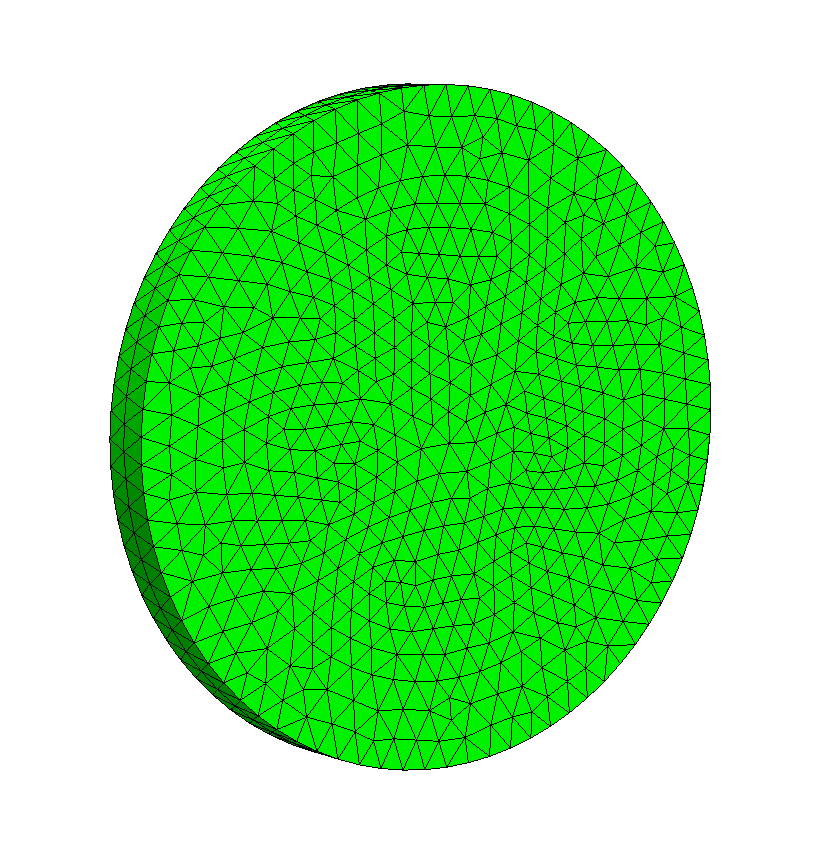}  
\end{array}$\\
Complete set of UK coin denominations  $(C_9)$
\end{center}
\begin{center}
$\begin{array}{cccccccc}
\includegraphics[width=0.025\textwidth, keepaspectratio]{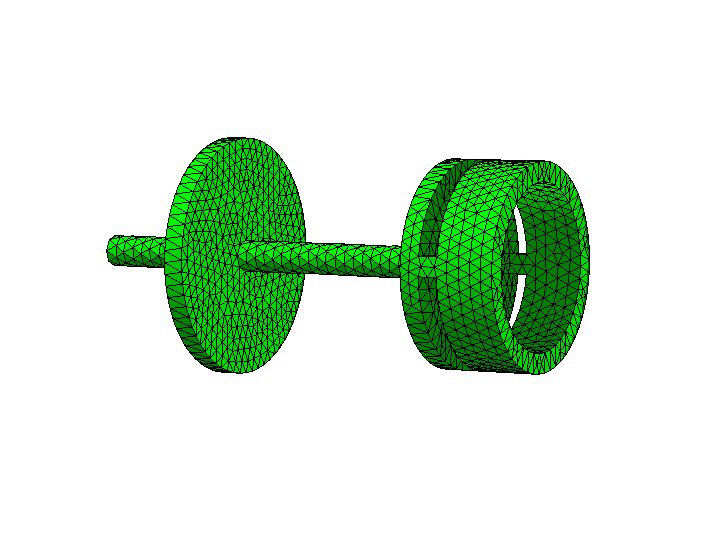} &
\includegraphics[width=0.025\textwidth, keepaspectratio]{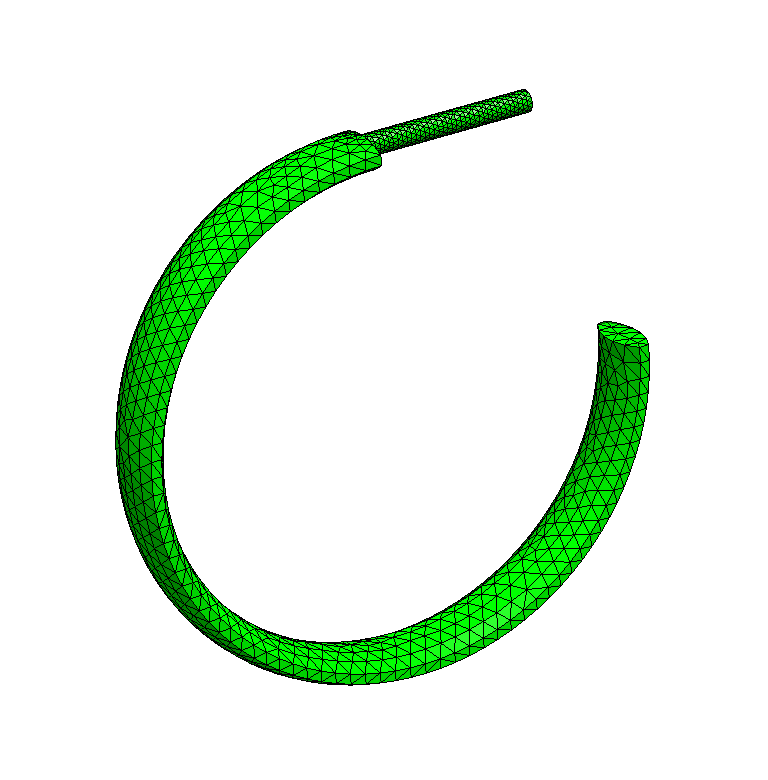} &
\includegraphics[width=0.025\textwidth, keepaspectratio]{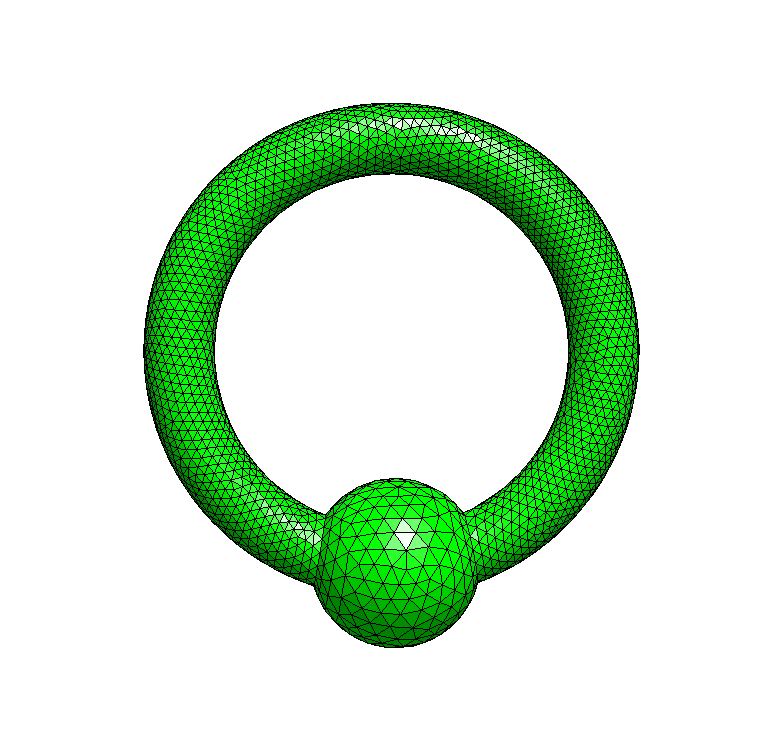}&
\includegraphics[width=0.025\textwidth, keepaspectratio]{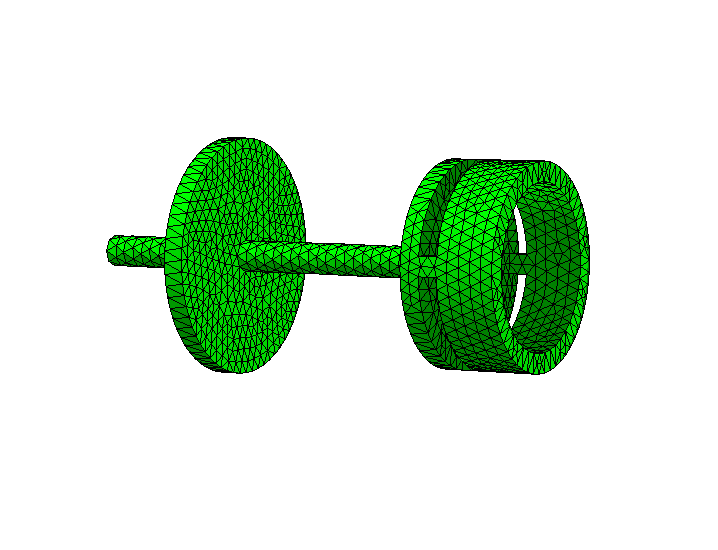} &
\includegraphics[width=0.025\textwidth, keepaspectratio]{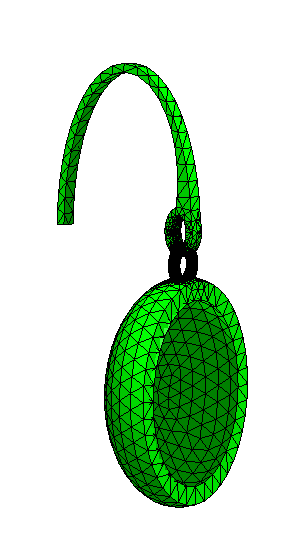} &
\includegraphics[width=0.025\textwidth, keepaspectratio]{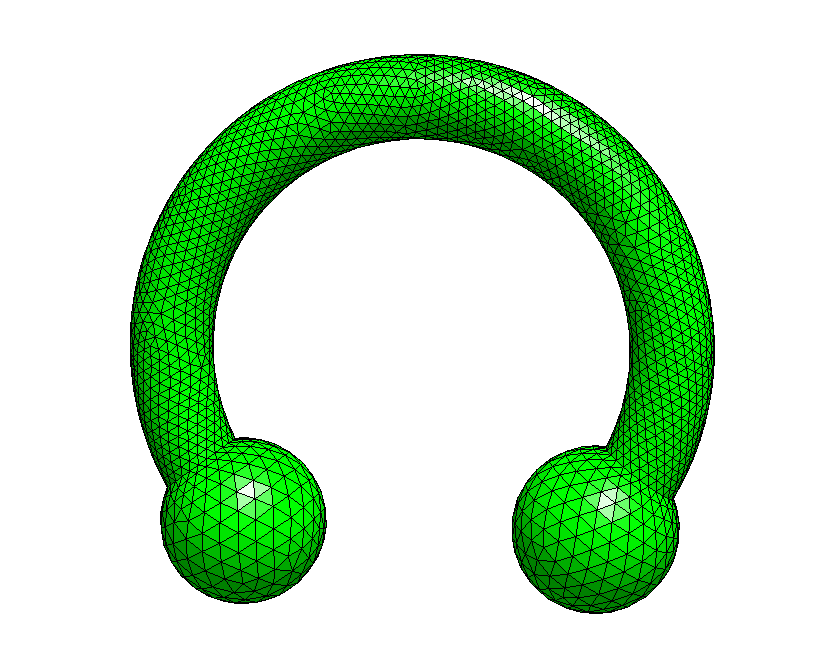} &
\includegraphics[width=0.025\textwidth, keepaspectratio]{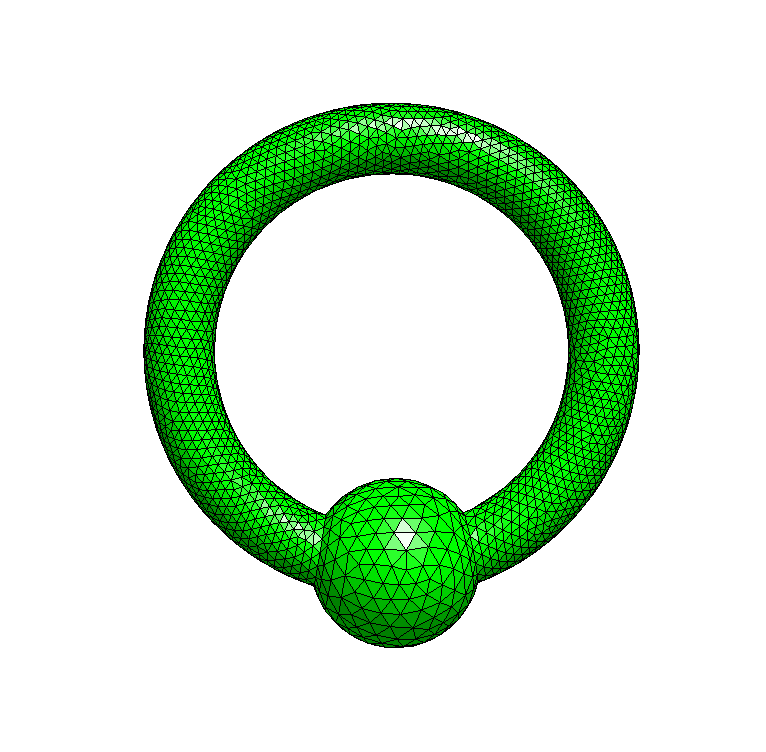}
\end{array}$\\
Earrings / Piercings  $(C_{10})$
\end{center}
\begin{center}
$\begin{array}{cccc}
\includegraphics[width=0.15\textwidth, keepaspectratio]{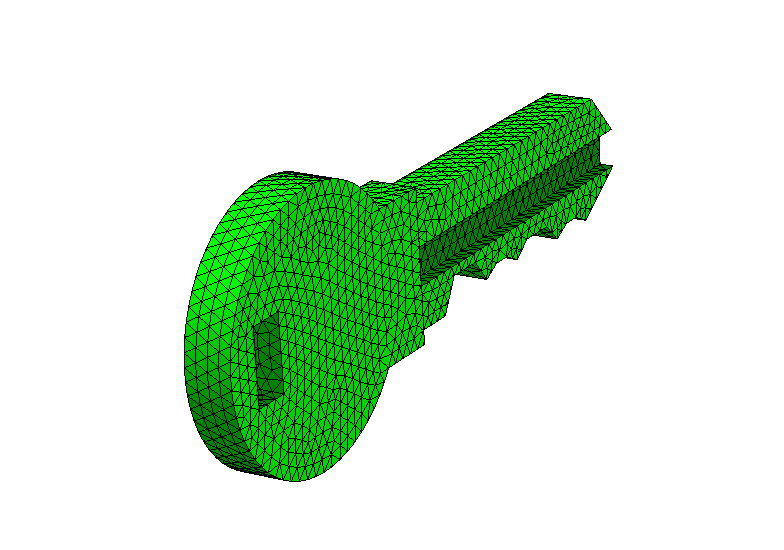} &
\includegraphics[width=0.15\textwidth, keepaspectratio]{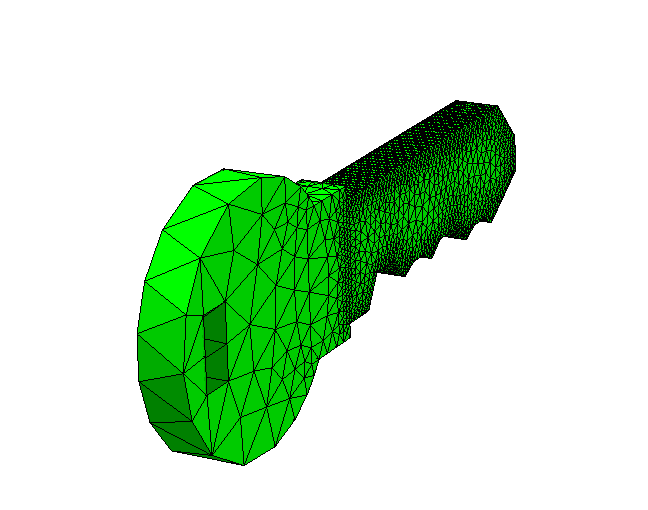} &
\includegraphics[width=0.15\textwidth, keepaspectratio]{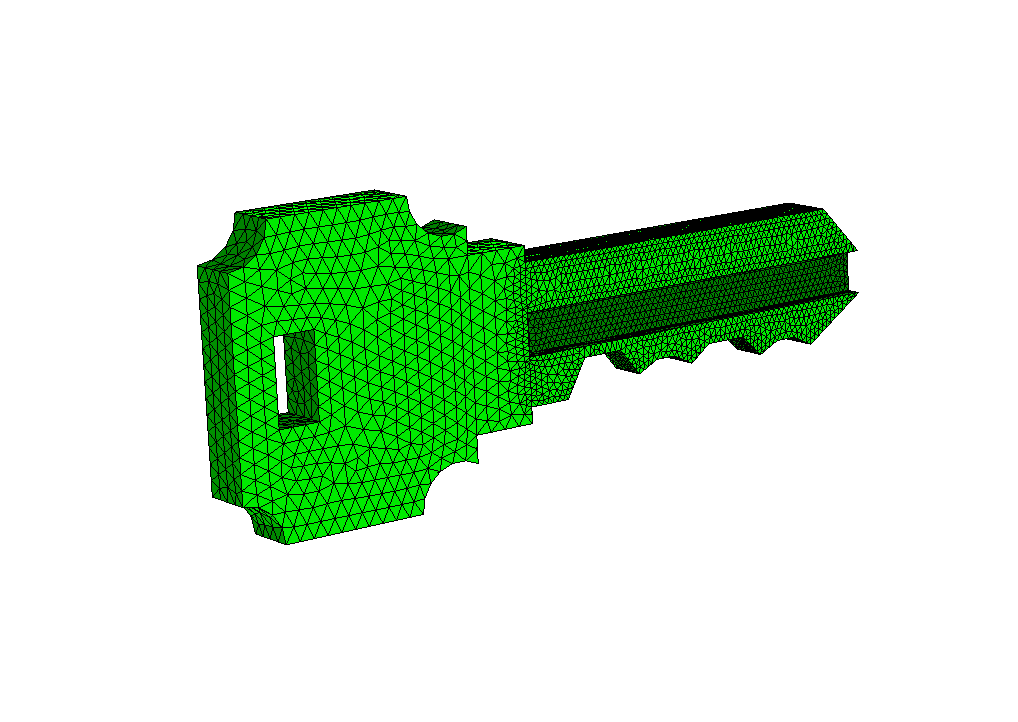} &
\includegraphics[width=0.15\textwidth, keepaspectratio]{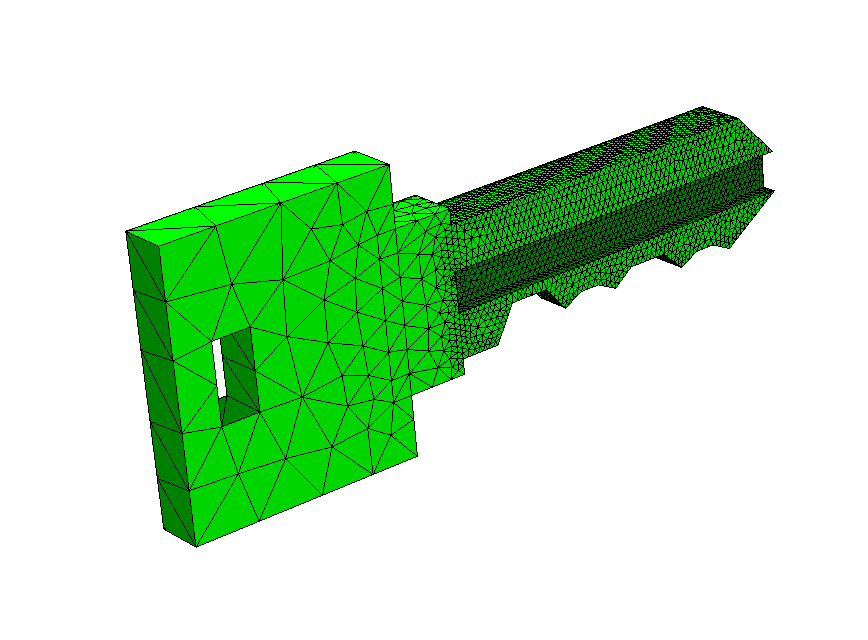} \\
\end{array}$\\
Keys  $(C_{11})$
\end{center}
\begin{center}
$\begin{array}{cc}
\begin{array}{cccccc}
\includegraphics[width=0.025\textwidth, keepaspectratio]{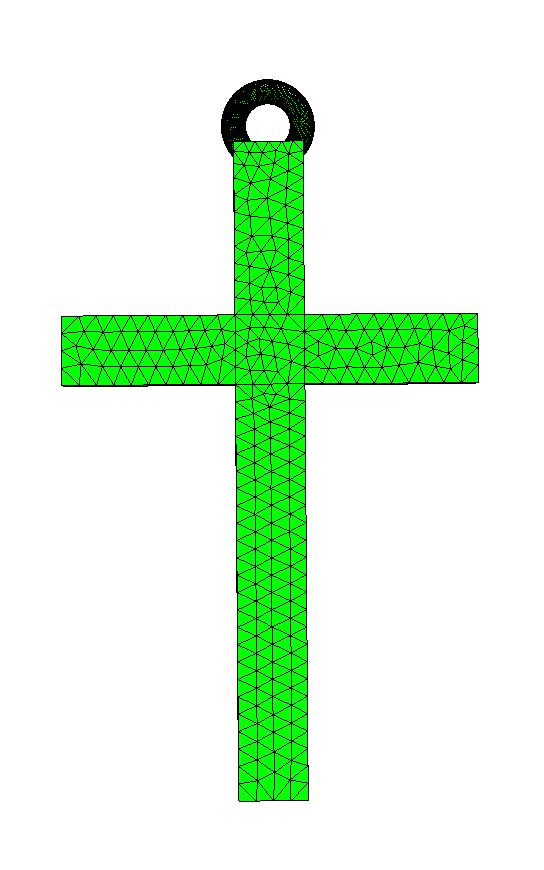} &
\includegraphics[width=0.025\textwidth, keepaspectratio]{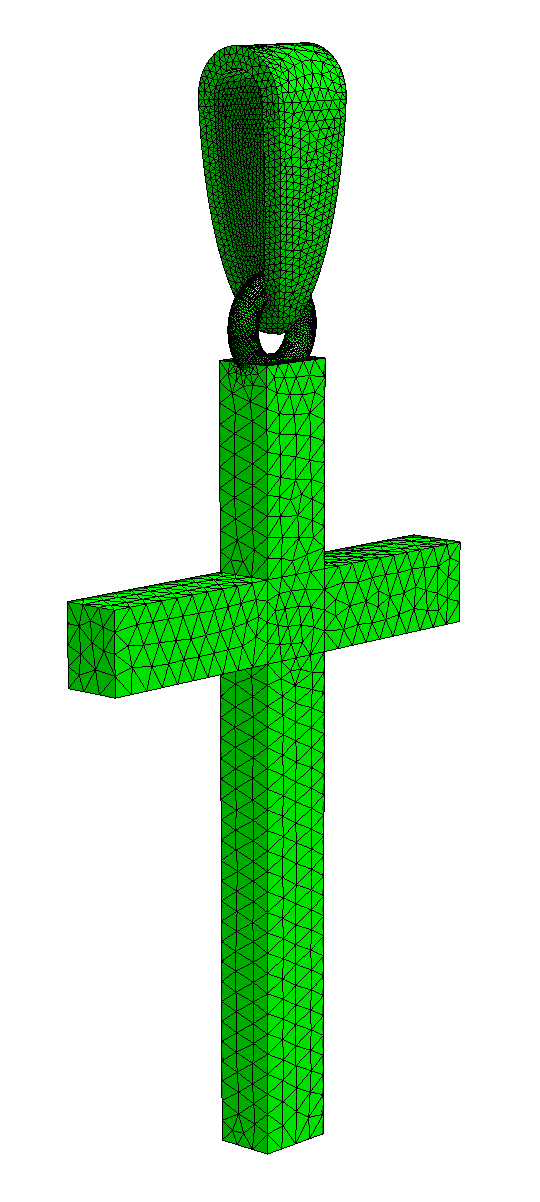} &
\includegraphics[width=0.025\textwidth, keepaspectratio]{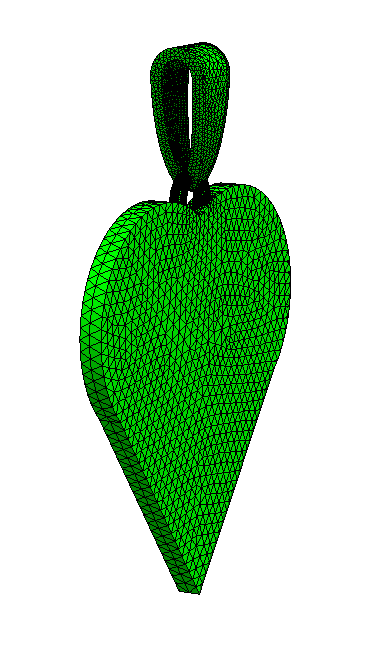} &
 \includegraphics[width=0.025\textwidth, keepaspectratio]{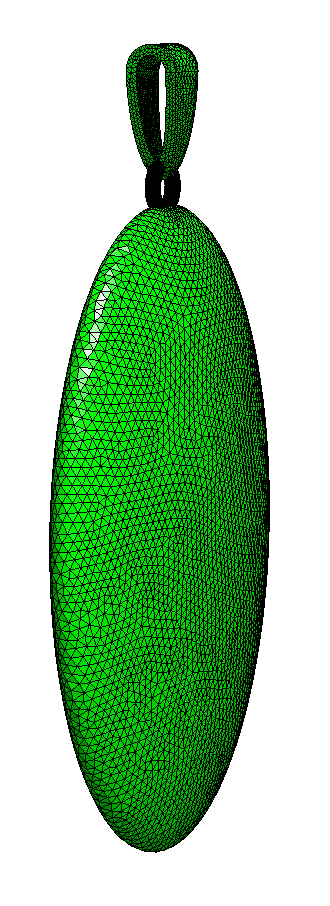} &
\includegraphics[width=0.025\textwidth, keepaspectratio]{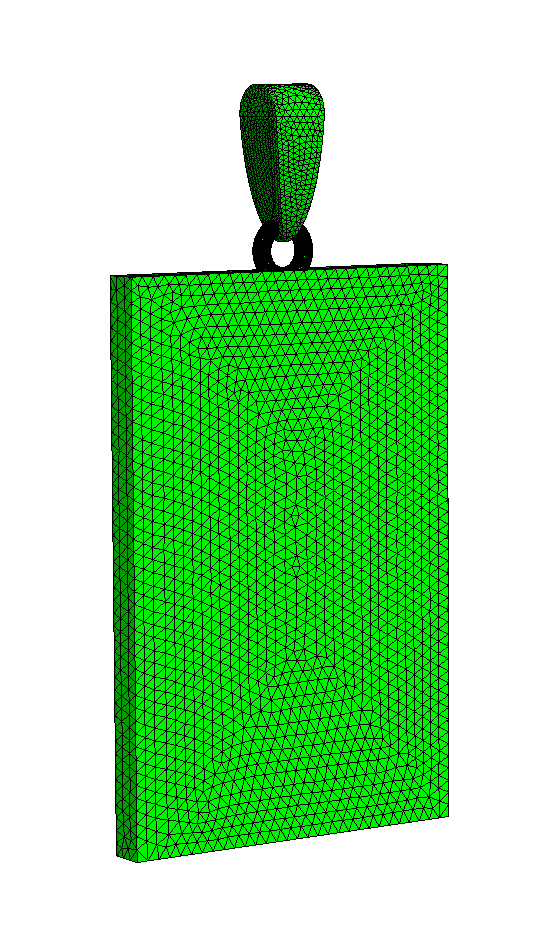} &
\includegraphics[width=0.025\textwidth, keepaspectratio]{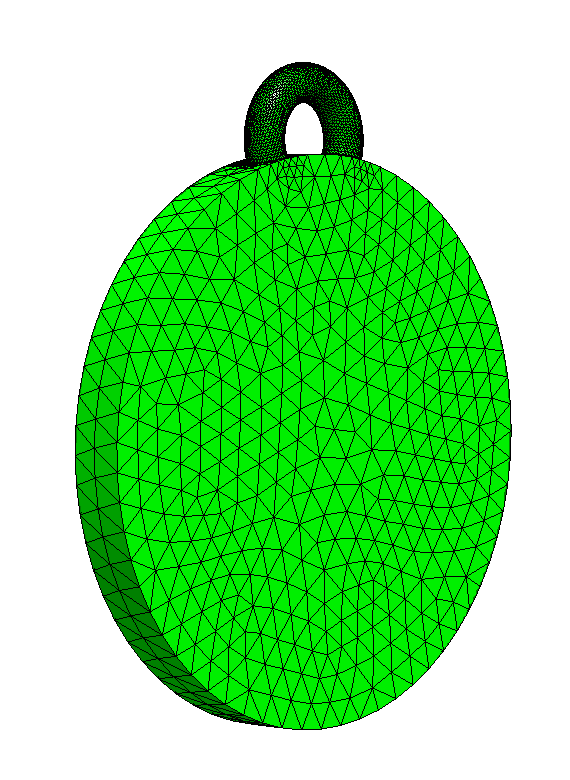}
\end{array}&
\begin{array}{ccccc}
\includegraphics[width=0.05\textwidth, keepaspectratio]{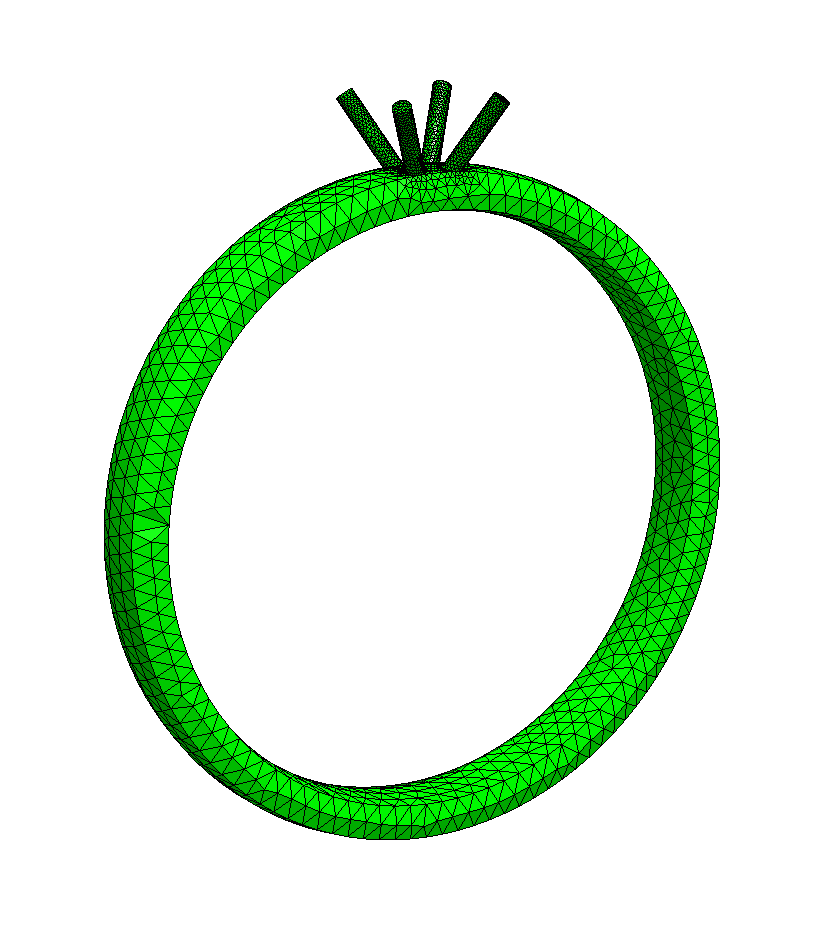} &
\includegraphics[width=0.05\textwidth, keepaspectratio]{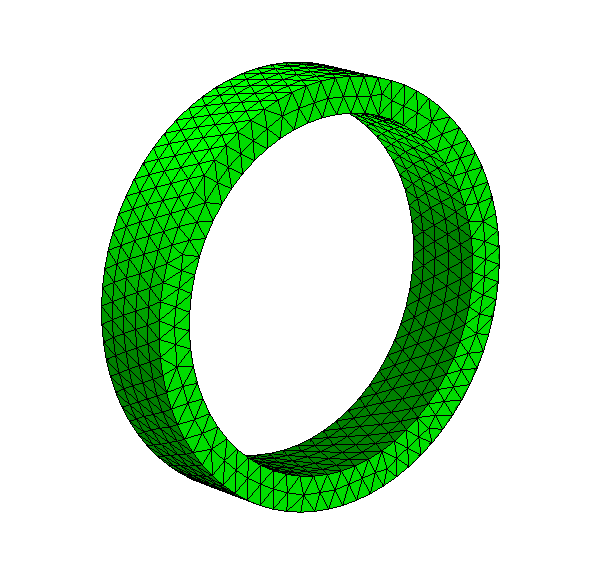} &
\includegraphics[width=0.05\textwidth, keepaspectratio]{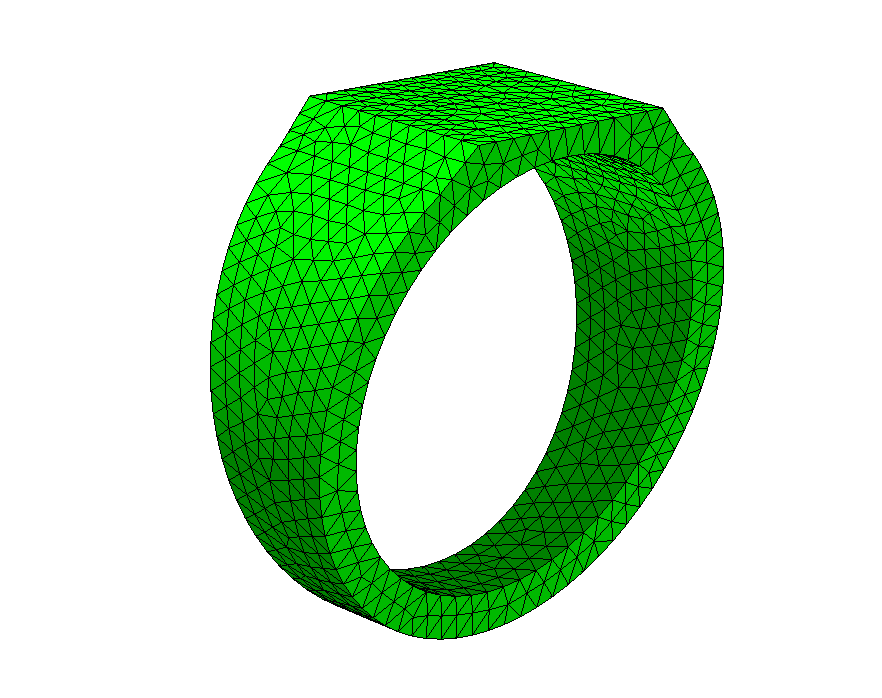} &
\includegraphics[width=0.05\textwidth, keepaspectratio]{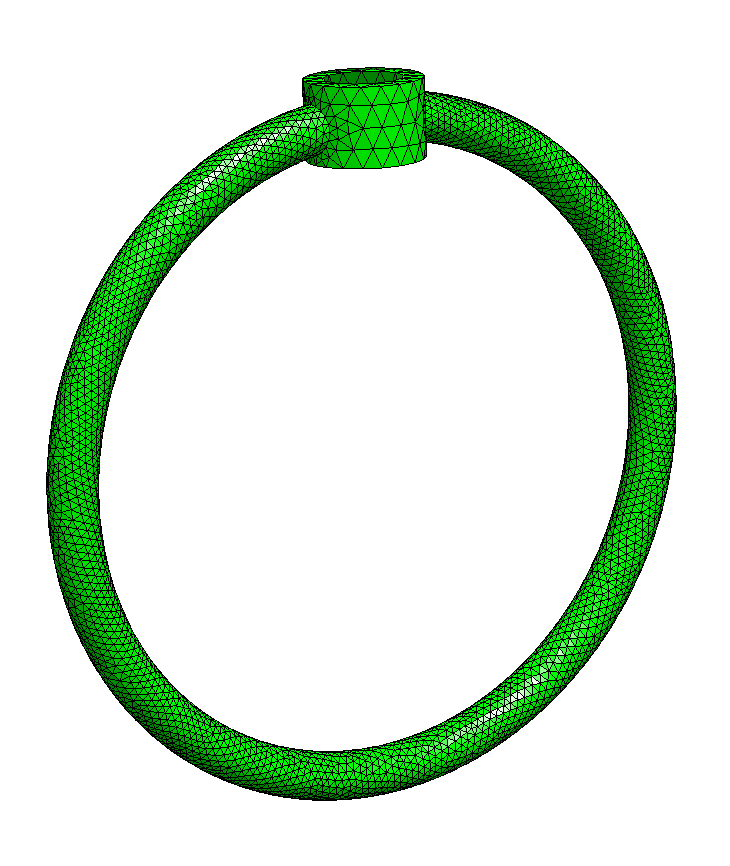} &
\includegraphics[width=0.05\textwidth, keepaspectratio]{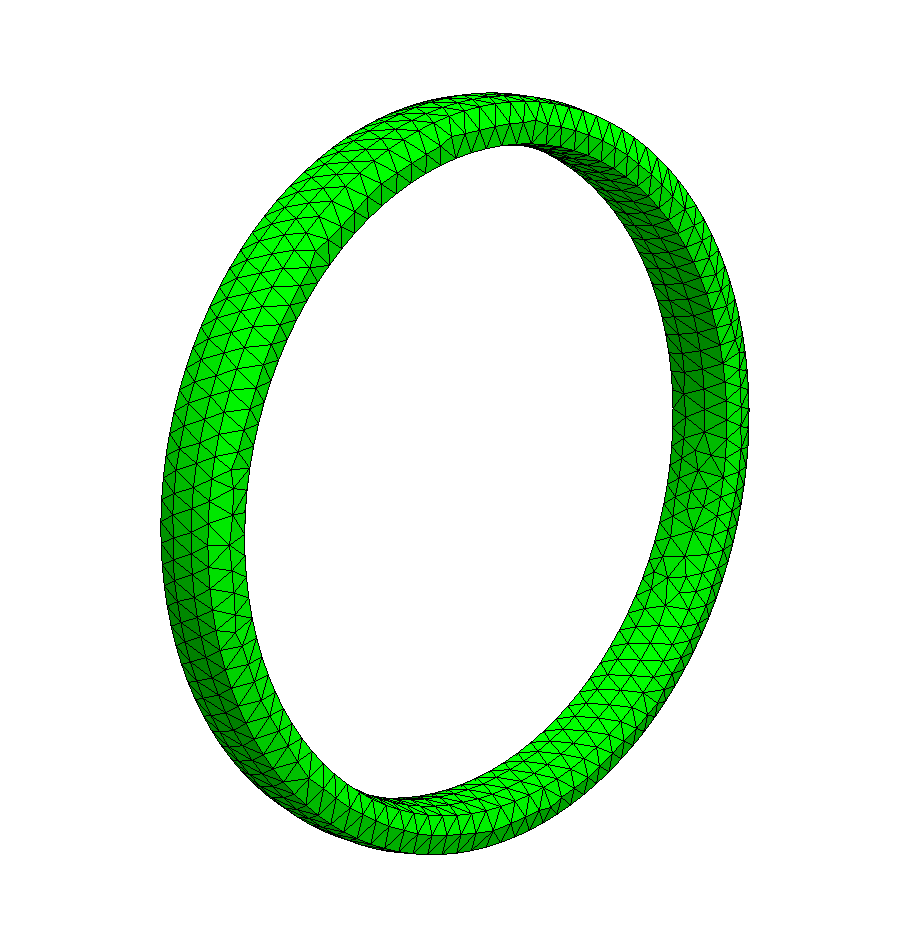} 
\end{array}\\
\text{Pendents }  (C_{12}) & 
\text{Rings} (C_{13}) \end{array}$
\end{center}
\begin{center}
$\begin{array}{ccc}
\includegraphics[width=0.2\textwidth, keepaspectratio]{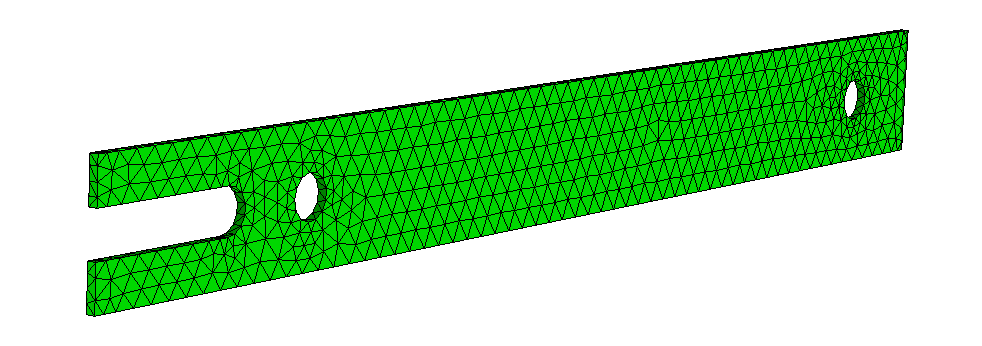} &
\includegraphics[width=0.2\textwidth, keepaspectratio]{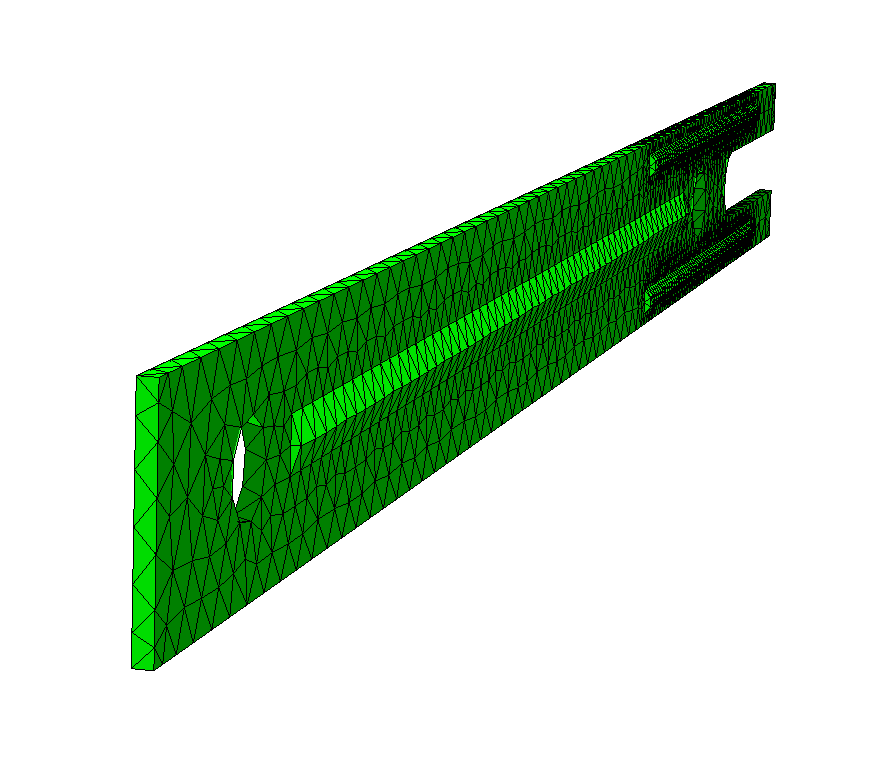} &
\includegraphics[width=0.2\textwidth, keepaspectratio]{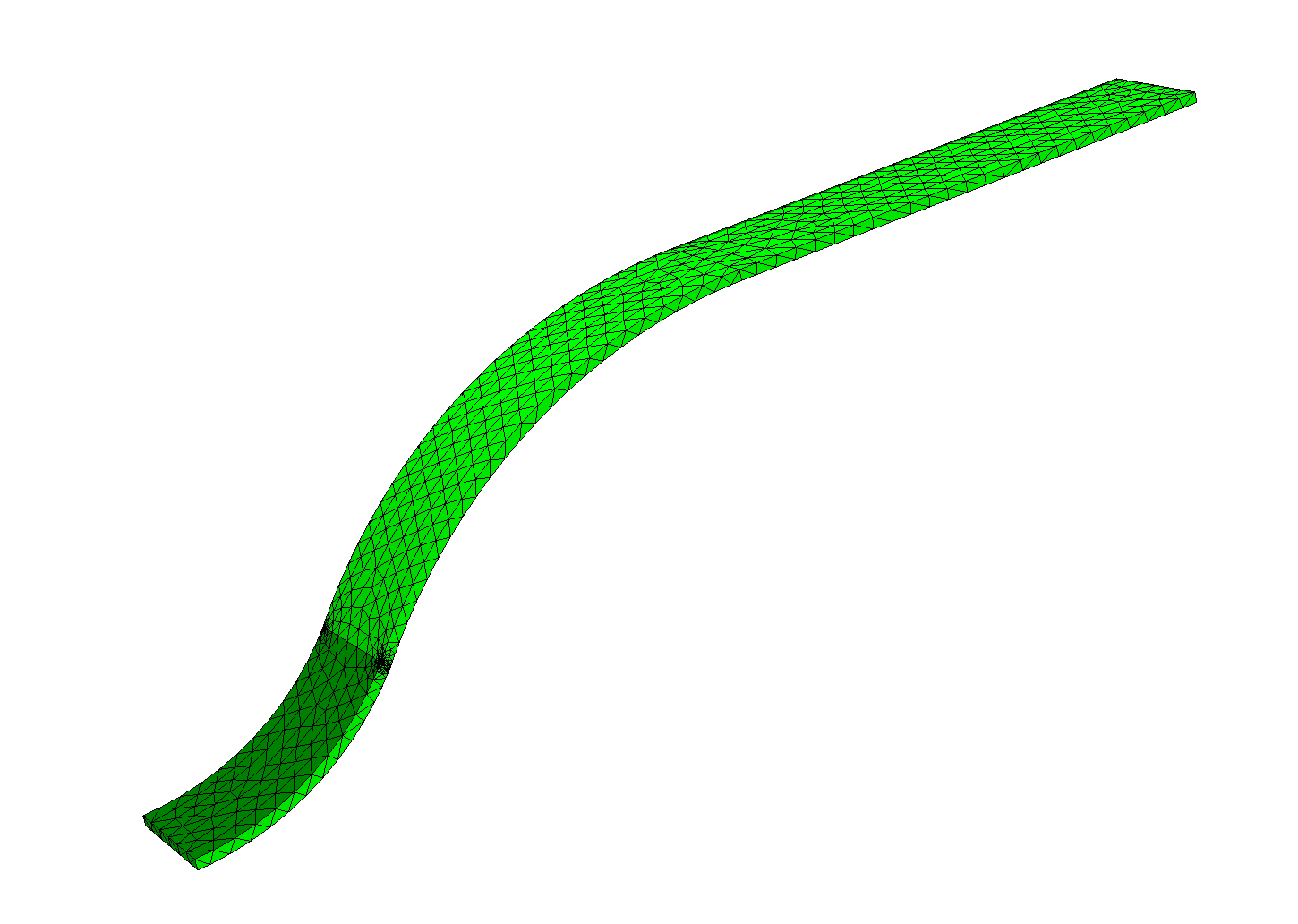}
\end{array}$\\
Shoe shanks  $(C_{14})$
\end{center}
\begin{center}
$\begin{array}{ccc}
\includegraphics[width=0.1\textwidth, keepaspectratio]{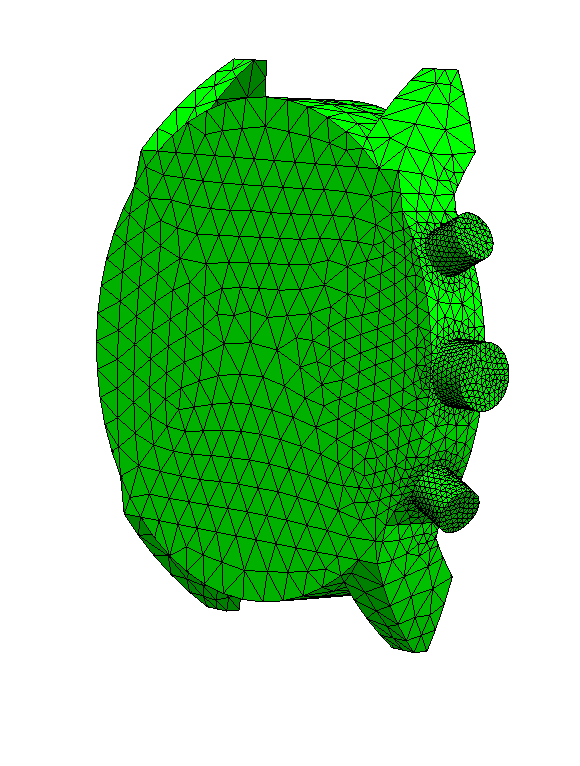} &
\includegraphics[width=0.1\textwidth, keepaspectratio]{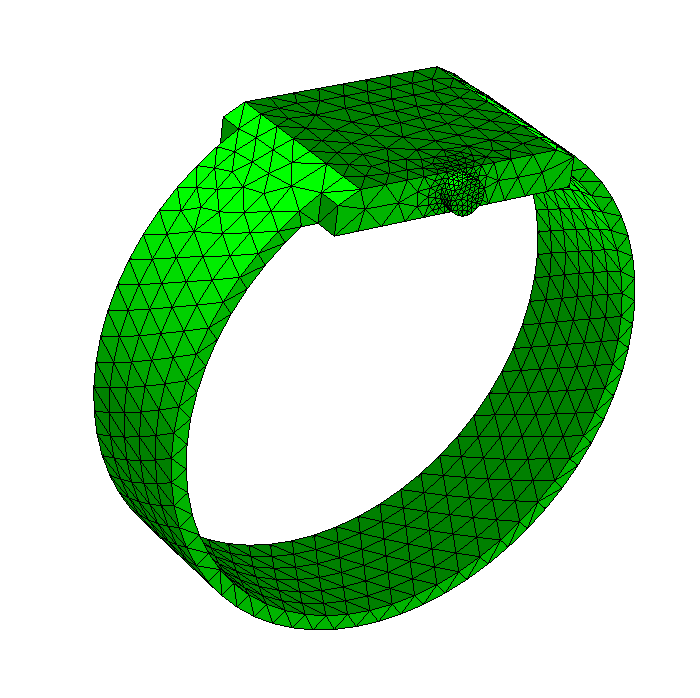} &
\includegraphics[width=0.1\textwidth, keepaspectratio]{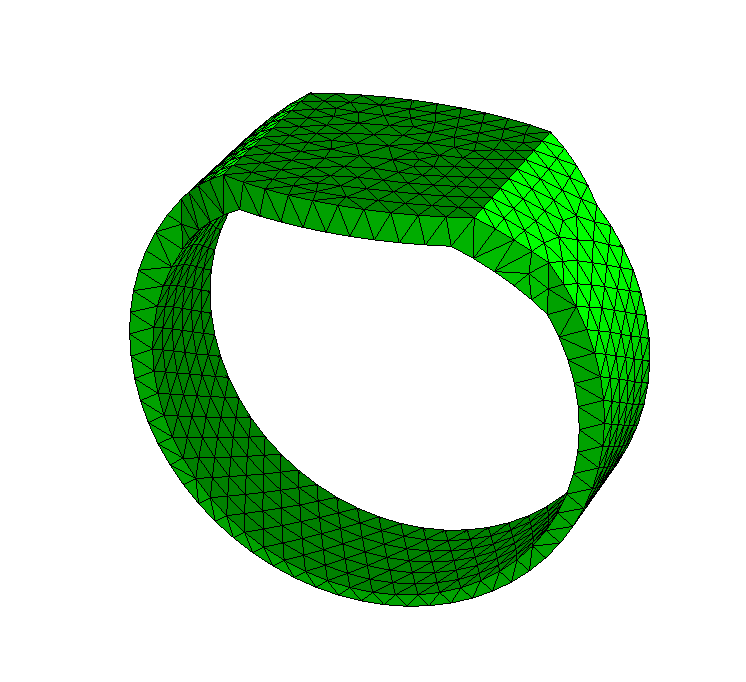}
\end{array}$\\
Watch/ watch with metallic strap  $(C_{15})$
\end{center}
\caption{Set of multiple threat and non-threat objects: Sample illustrations of some of the different non-threat object geometries considered (not to scale).}\label{fig:egnonthreat}
\end{figure}

\begin{figure}[!h]
\begin{center}
$\begin{array}{ccc}
\includegraphics[width=0.25\textwidth, keepaspectratio]{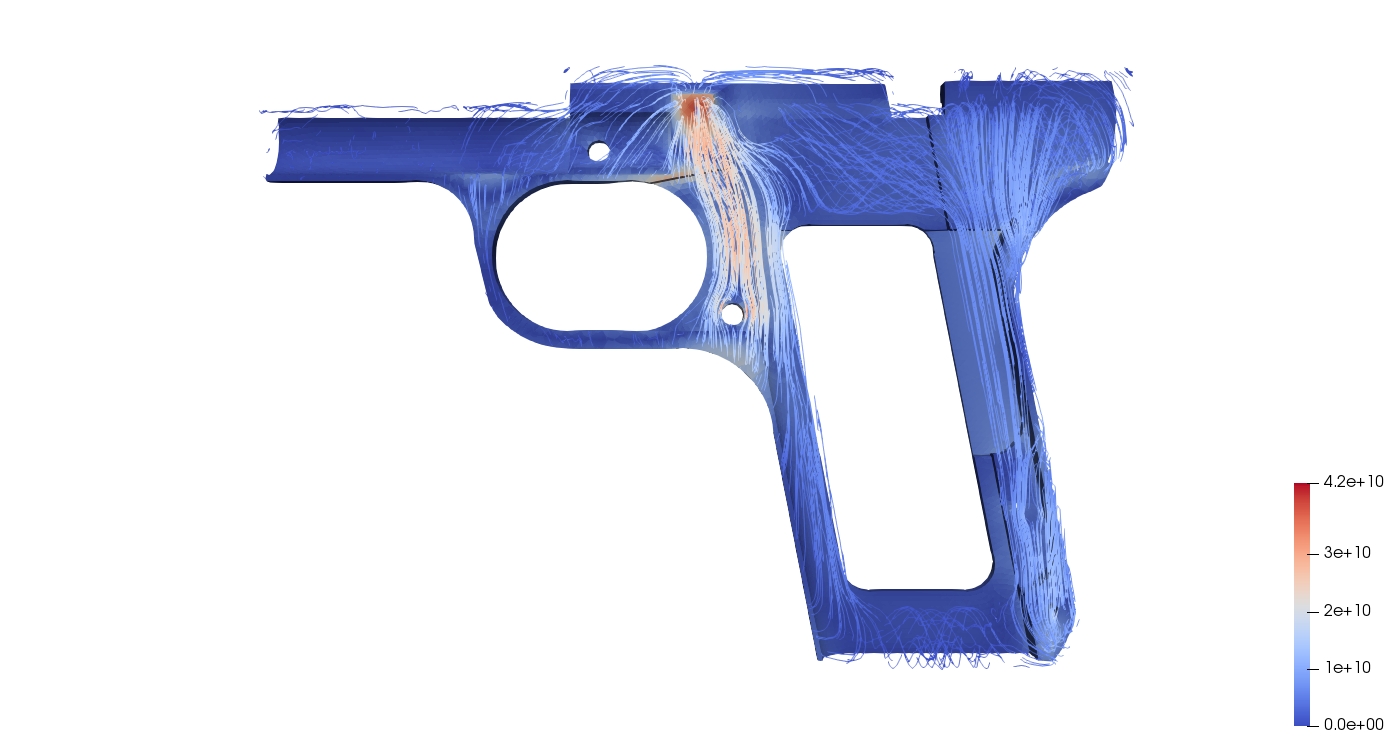} &
\includegraphics[width=0.25\textwidth, keepaspectratio]{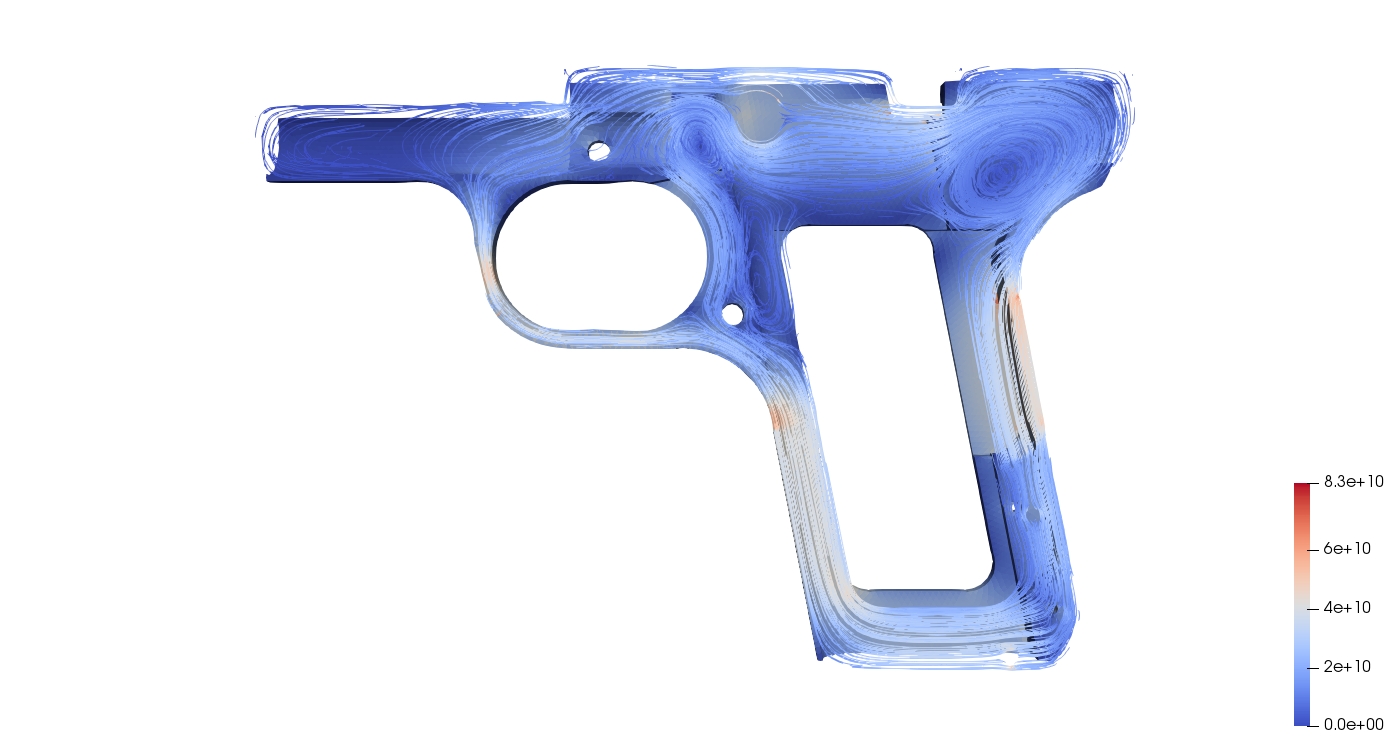} &
\includegraphics[width=0.25\textwidth, keepaspectratio]{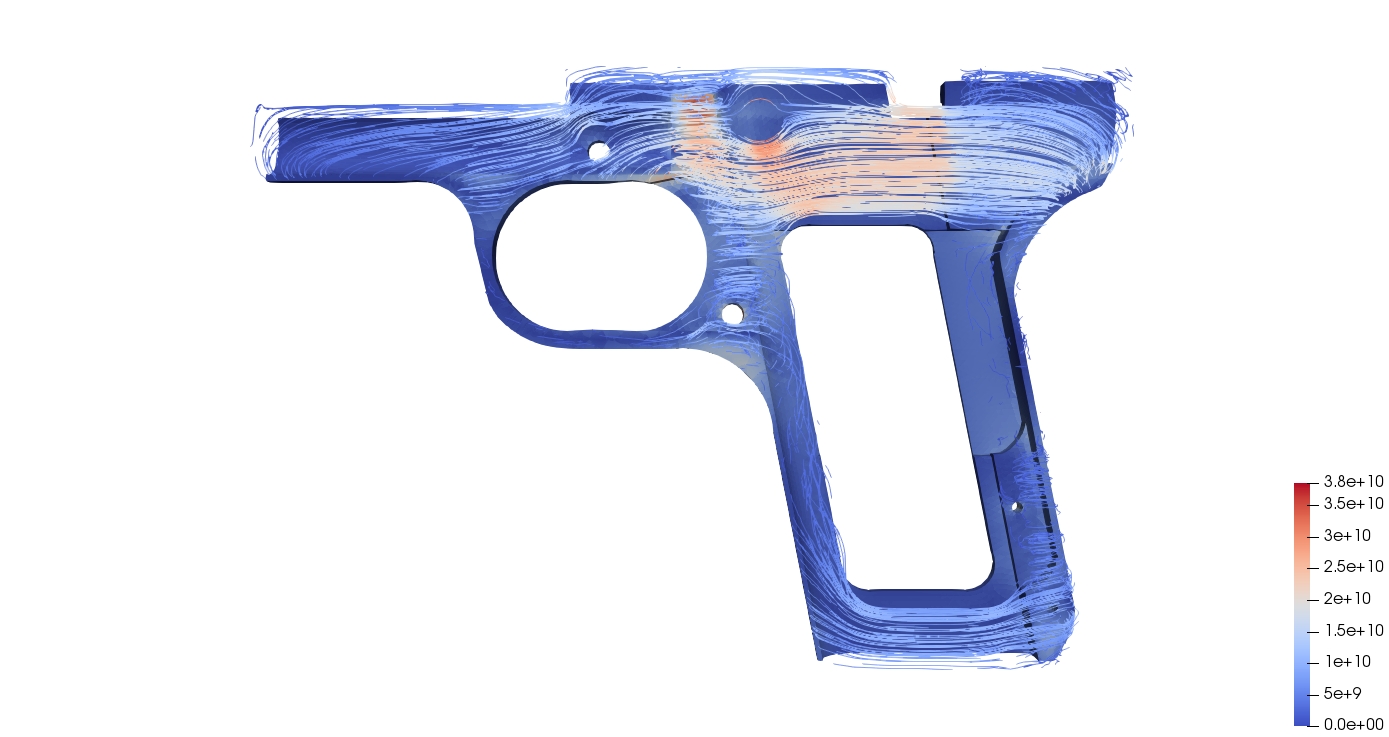} \\
\end{array}$\\
$\begin{array}{ccc}
\includegraphics[width=0.2\textwidth, keepaspectratio]{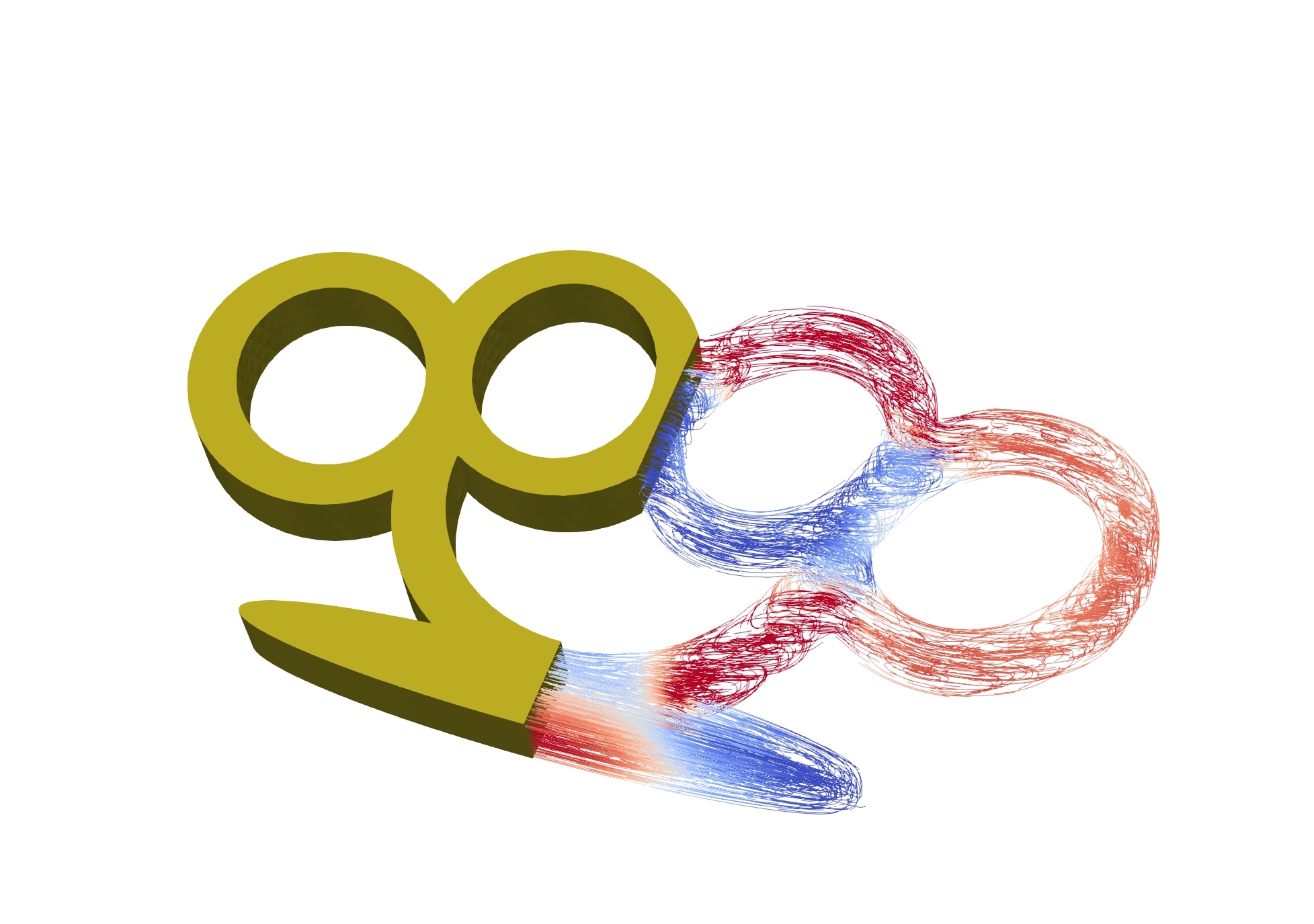} &
\includegraphics[width=0.2\textwidth, keepaspectratio]{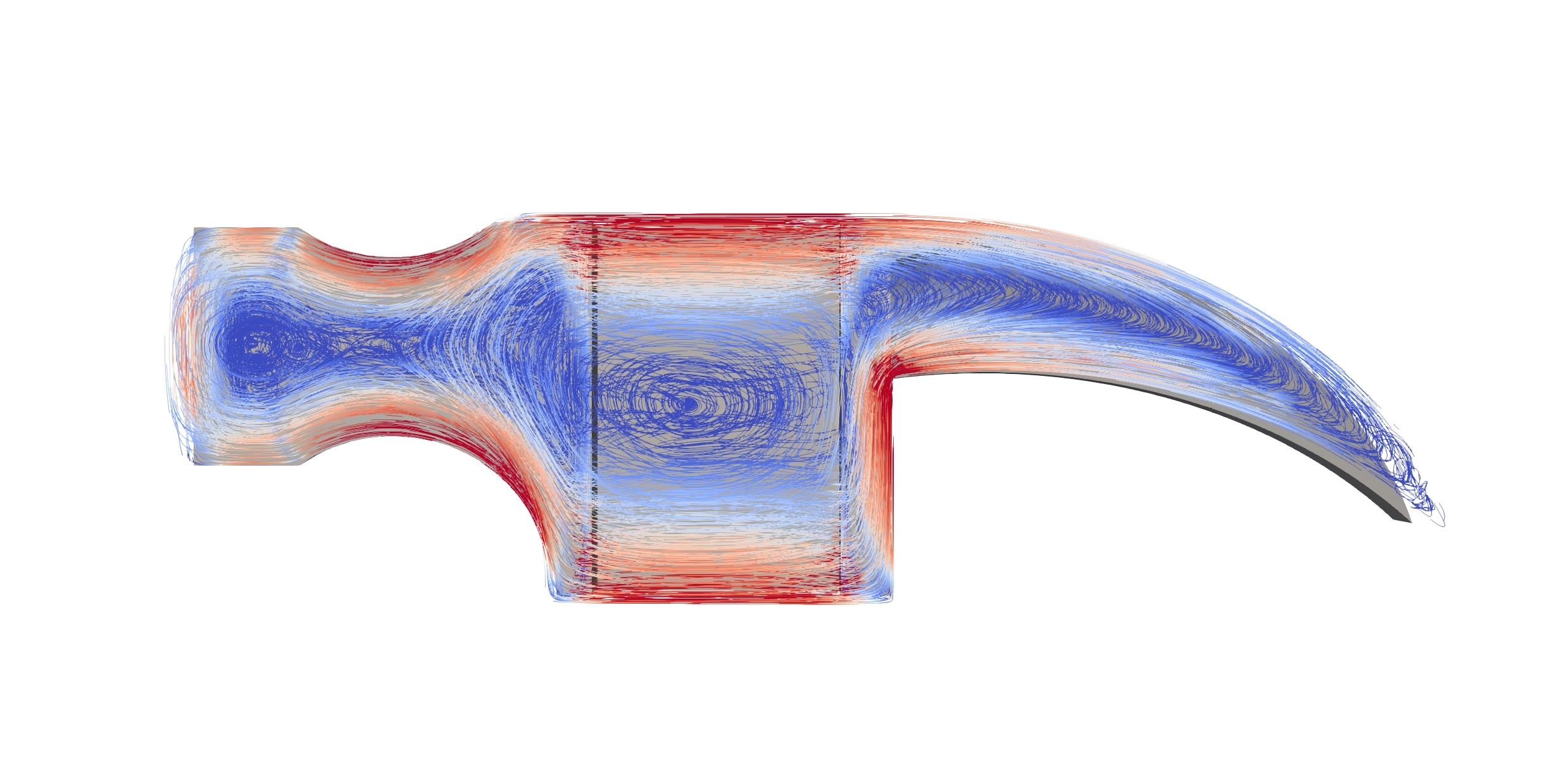} &
\includegraphics[width=0.2\textwidth, keepaspectratio]{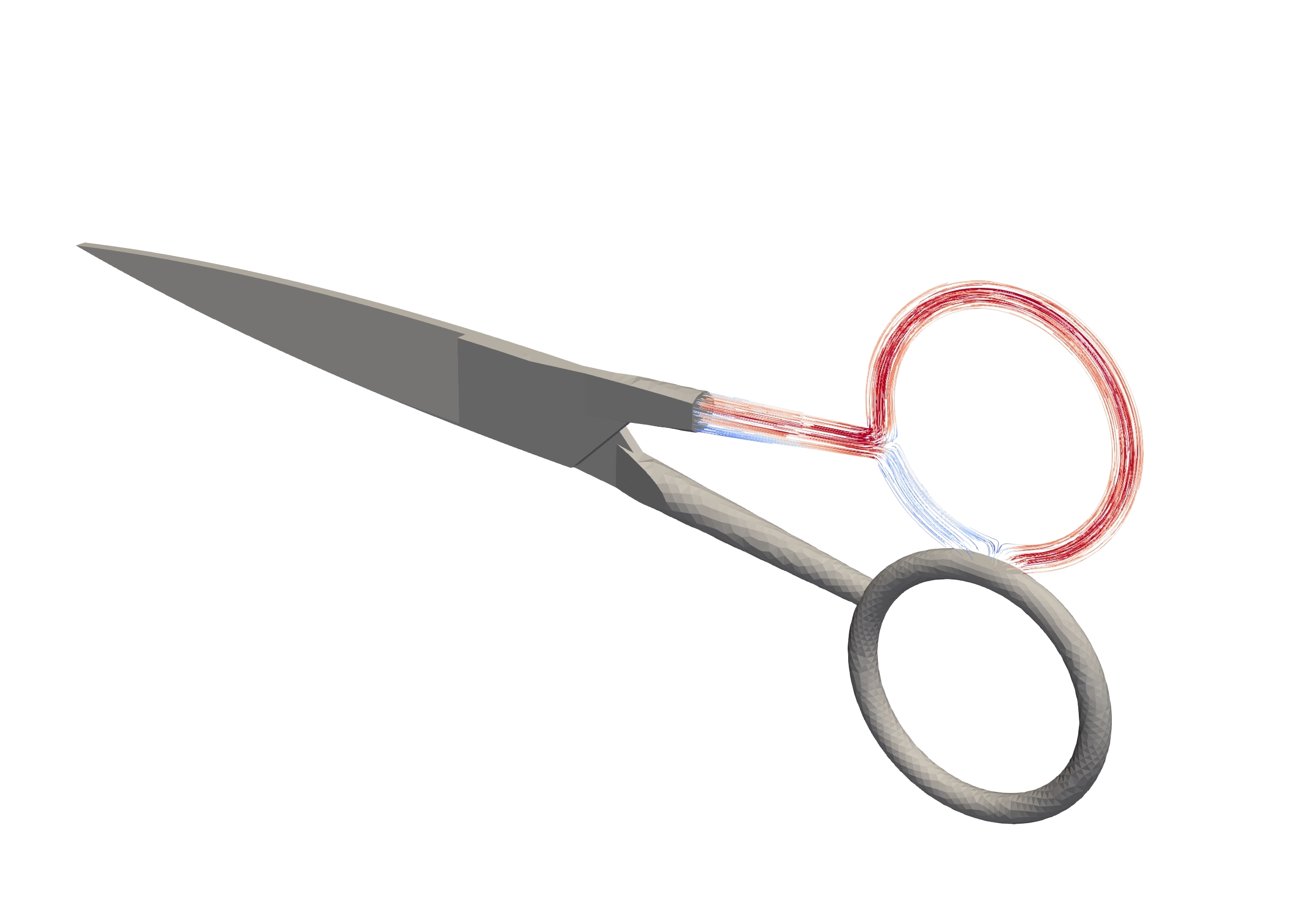} 
\end{array}$\\
$\begin{array}{cc} 
\includegraphics[width=0.075\textwidth, keepaspectratio]{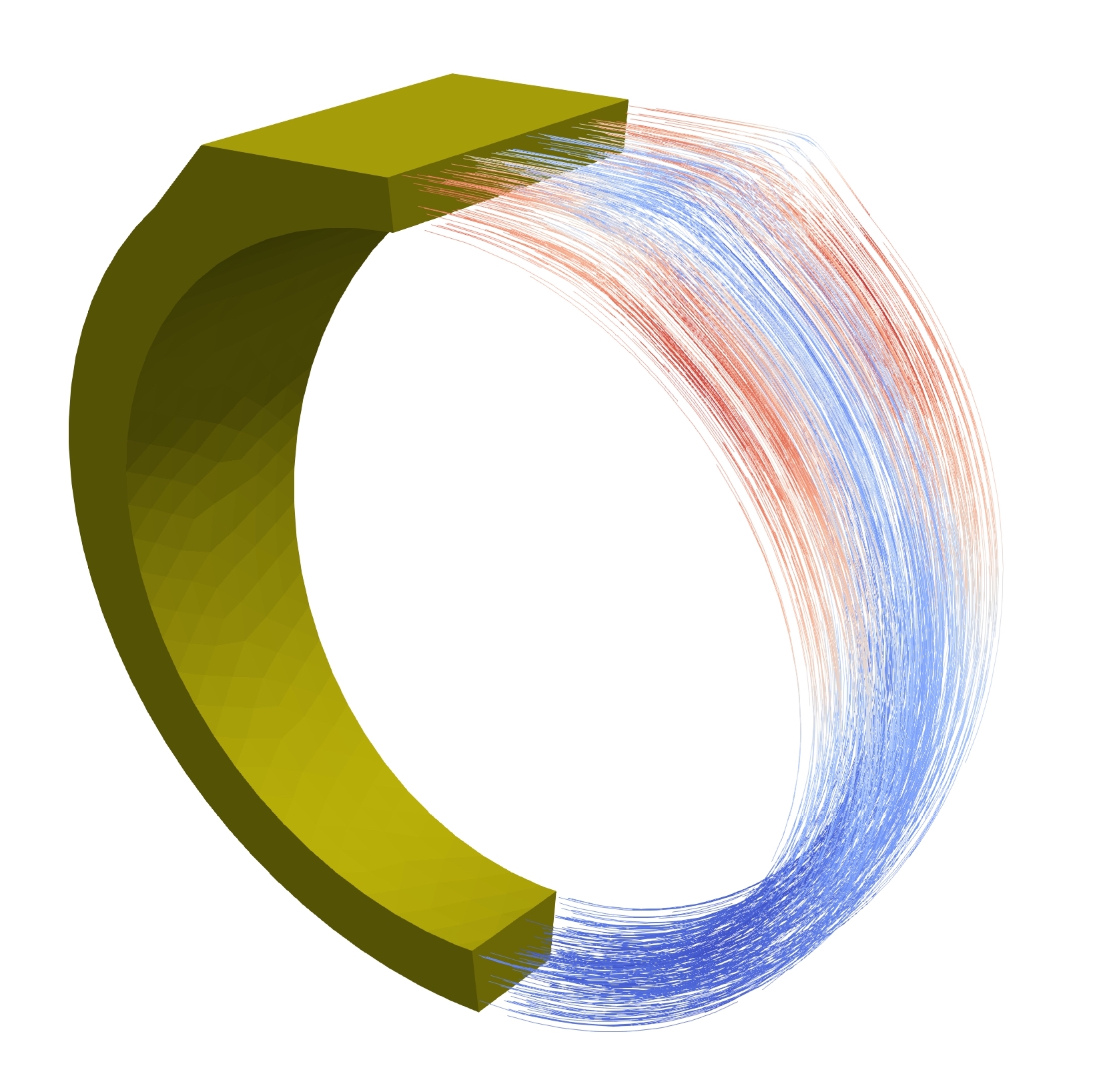} &
\includegraphics[width=0.2\textwidth, keepaspectratio]{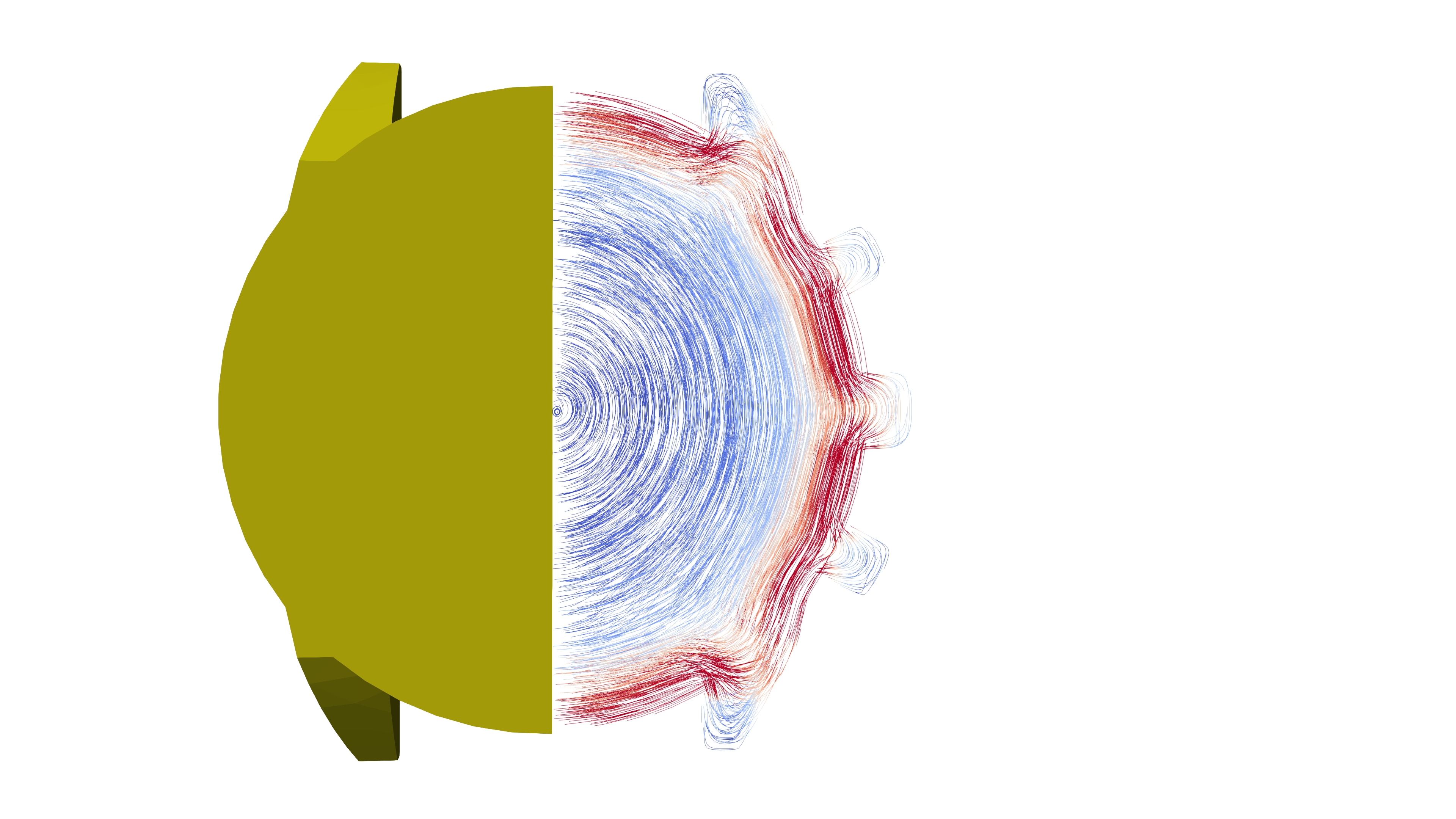}
\end{array}$
\end{center}
\caption{Set of multiple threat and non-threat objects: Illustrative contours of ${\bm J}_i^e = \im \omega \sigma_* {\bm \theta}_i$ at $\omega =1\times 10^4$ rad/s for a sample of the threat and non-threat objects.} \label{fig:eddycurrent}
\end{figure}

To create the multi-class dictionary, we follow the approach described in Section~\ref{sect:dict} where, in
 the most general setting, we choose  the classes $C_k$, $k=1,...,K$, to correspond to the different threat and non-threat type objects listed in Table~\ref{tab:threatandnothreat}. Unlike the coins, each class is comprised of objects of different geometries, as well as different sizes and materials, so that $G^{(k)}\ne1$ in general. However, when creating the classes, we have assembled  geometries that have (physical) similarities. For example, the coins class $C_1$ includes the $G^{(1)}=8$ different denomination of British coins described in the previous section.
Furthermore, in Figures~\ref{fig:egthreat} and~\ref{fig:egnonthreat}, we  illustrate the surface finite element meshes corresponding to exemplar threat and non-threat object geometries, respectively, within each of these different classes and Table~\ref{tab:objectsummarymat} gives an overall summary of the materials and object sizes. In the case of the coins, guns, keys and knives, the simulated MPT spectral signatures  are those presented in~\cite{ledgerwilsonamadlion2021} and we provide the complete set of MPT spectral signatures for all objects  in our \texttt{MPT-Library}  dataset~\cite{bawilson94_2021_4876371}. These simulated spectral signatures were generated in a similar way to those described in~\cite{ledgerwilsonamadlion2021} and  in Figure~\ref{fig:eddycurrent} we show an illustration of some the contours of ${\bm J}_i^e = \im \omega \sigma_* {\bm \theta}_i$ at $\omega =1\times 10^4$ rad/s that are obtained as part of this process.
 In total, we have $\sum_{k=1}^KG^{(k)} = 67 $ different distinct geometries and $ 158 $ including different material variations. In Table~\ref{tab:threatandnothreat} we give the relationship between $P^{(k)}$, $V^{(k)}$ and $G^{(k)}$ for each class $C_k$ and choose
  $V^{(k)}$ times so that we have an approximately equal number of samples $P^{(k)}\approx P/K$ for each class of object. We employ
  $P^{(k)}=5000$ in the following unless otherwise stated.
While   $m_{\sigma_*}$ is object specific, we set $ m_\alpha = 0.001$m, to fix the object size, and choose $s_\alpha = 0.0084m_\alpha$ and $s_{\sigma_*} = 0.024m_{\sigma_*}$, to account for manufacturing imperfections. We consider a fixed number of number of $M= 28$ linearly spaced frequencies, such that $7.53 \times 10^2 \text{ rad/s} \le \omega_m \le 5.99 \times 10^5 \text{ rad/s}$, although we also give some comments about the performance using $5.02 \times 10^4 \text{rad/s} \le \omega_m \le 8.67 \times 10^4 $ rad/s. In a similar manner to the coin classification problem, noise corresponding to SNR values of  40dB, 20dB is added. We do not consider an SNR of 10dB as this represents a very high level of 32\% noise, which, of course, performs worse than 20dB noise.
Using the information above, two different types of dictionaries were formed. Firstly, $D_{15}$  for the complete set of  $K=15 $ classes and, secondly, $D_8$ comprising of $K= 8$ different classes. The grouped classes for the $D_8$  are described in Table~\ref{tab:8classproblem}. 

\begin{table}[!h]
\begin{center}
\begin{tabular}{|c|c|c|c|c|}
\hline
Class &  \# Geometries & \#Materials & \# Additional & Total \\
 {} & ($G^{(k)}$) &  per geometry &  variations  ($V^{(k)}$) & ($P^{(k)}$)\\
\hline
Guns ($C_1$) & {1} & {1} & $V^{(1)}$  & {$V^{(1)}$}  \\
\hline
Hammers  ($C_2$) & {2} & {3} & {$V^{(2)}$} & {$6 V^{(2)}$} \\
\hline
Knives  ($C_3$) & {5} & {1} & {$V^{(3)}$}  & {$ 5 V^{(3)}$} \\
\hline
Knuckle dusters  ($C_4$) & {2} & {1} & {$V^{(4)}$}   & {$ 2 V^{(4)}$}\\
\hline
Screw drivers  ($C_5$) & {6} & {3} & {$V^{(5)}$}  & {$ 18 V^{(5)}$} \\

\hline
Scissors  ($C_6$) & {2} & {3} & {$V^{(6)}$}  & {$ 6 V^{(6)}$}  \\
\hline
\hline
Bracelets  ($C_7$) & {4} & {3} & {$V^{(7)}$} & {$12V^{(7)}$} \\
\hline
Belt buckles  ($C_8$) & {3} & {4} & {$V^{(8)}$} & {$12 V^{(8)}$}\\
\hline
Coins  ($C_9$) & {8} & {1} & {$V^{(9)}$} & {$8 V^{(9)}$}\\
\hline
Earrings  ($C_{10}$) & {9} & {3} & {$V^{(10)}$} & {$18 V^{(10)}$}\\
\hline
Keys ($C_{11}$) & {4} & {1} & {$V^{(11)}$} & {$4 V^{(11)}$}\\
\hline
 Pendents  ($C_{12}$) & {7} & {3} & {$V^{(12)}$} & {$21 V^{(12)}$}\\
\hline
 Rings  ($C_{13}$) & {7} & {3} & {$V^{(13)}$} & {$21 V^{(13)}$}\\
\hline
 Shoe shanks  ($C_{14}$) & {3} & {1} & {$V^{(14)}$} & {$3 V^{(14)}$}\\
 \hline
 Watches  ($C_{15}$) & {4} & {3} & {$V^{(15)}$} & {$12 V^{(15)}$}\\
 \hline
\end{tabular}
\end{center}
\caption{Set of multiple threat and non-threat objects: Full list of 15 threat and non-threat object classes detailing the number of geometries in each class,  $G^{(k)}$, the number of materials per geometry, the number of additional variations to account for geometrical and material variations and the total number in each class, $P^{(k)}$.} \label{tab:threatandnothreat}
\end{table}

\begin{table}[!h]
\begin{center}
\begin{tabular}{|l|c|c|c|c|c|c|}
\hline
Class &  $\min(\sigma_*)$ & $\max(\sigma_*)$ & $\min(\mu_r)$  & $\max(\mu_r)$ & $\min(\alpha^3|B|)$  & $\max(\alpha^3|B|)$ \\
{} & S/m & S/m & {} & {} & m$^3$ & m$^3$ \\
\hline
Guns ($C_1$) & {$6.2\times10^6$}  & {$6.2\times10^6$} & {5} & {5} & {$3.3\times10^{-5}$} & {$3.3\times10^{-5}$} \\
\hline
Hammers  ($C_2$) & {$1.3\times10^6$}  & {$1.6\times10^6$} & {1.02} & {5} & {$4.3\times10^{-5}$} & {$2.0\times10^{-4}$}  \\ 
\hline
Knives  ($C_3$) & {$1.5\times10^5$}  & {$5.8\times10^7$} & {1} & {5} & {$3.3\times10^{-6}$} & {$6.5\times10^{-5}$}   \\ 
\hline
Knuckle  & {$1.5\times10^7$}  & {$1.5\times10^6$} & {1}  & {1} & {$1.7\times10^{-5}$} & {$1.8\times10^{-5}$}  \\
dusters ($C_4$)  & {} & {} & {}& {} & {} & {} \\
\hline
Screw  & {$1.3\times10^6$}  & {$1.6\times10^6$} & {1.02} & {5} & {$1.1\times10^{-6}$} & {$3.4\times10^{-6}$}  \\ 
drivers  ($C_5$) & {} & {} & {}& {} & {} & {} \\
\hline
Scissors  ($C_6$) & {$1.3\times10^6$}  & {$1.6\times10^6$} & {1.02} & {5} & {$2.4\times10^{-6}$} & {$9.5\times10^{-6}$}  \\ 
\hline
\hline
Bracelets  ($C_7$)& {$9.4\times10^6$}  & {$6.3\times10^7$}  & {1} & {1} & {$5.5\times10^{-7}$} & {$2.1\times10^{-6}$}  \\ 
\hline
Belt  & {$5.6\times10^5$}  &  {$1.5\times10^7$} & {1} & {5} & {$1.5\times10^{-5}$} & {$2.1\times10^{-5}$}  \\
buckles  ($C_8$) & {} & {} & {}& {} & {} & {} \\
\hline
Coins  ($C_9$) & {$2.9\times10^6$}  & {$4.0\times10^7$} & {1} & {1} & {$4.3\times10^{-7}$} & {$1.9\times10^{-6}$}  \\
\hline
Earrings  ($C_{10}$) & {$9.4\times10^6$}  & {$6.3\times10^7$} & {1} & {1} & {$1.0\times10^{-8}$} & {$2.3\times10^{-7}$}  \\
\hline
Keys ($C_{11}$) & {$2\times10^7$}  & {$1.5\times10^7$} & {1} & {1} & {$6.3\times10^{-7}$} & {$6.7\times10^{-7}$}  \\
\hline
 Pendents  ($C_{12}$) & {$9.4\times10^6$}  & {$6.3\times10^7$} & {1} & {1} & {$8.0\times10^{-8}$} & {$1.6\times10^{-6}$}  \\
\hline
 Rings  ($C_{13}$) & {$9.4\times10^6$}  &  {$6.3\times10^7$} & {1} & {1} & {$7.0\times10^{-8}$} & {$9.2\times10^{-7}$}  \\
\hline
 Shoe  & {$6.2\times10^6$}  & {$6.2\times10^6$} & {5} & {5} & {$8.9\times10^{-7}$} & {$1.4\times10^{-6}$}  \\
 shanks  ($C_{14}$) & {} & {} & {}& {} & {} & {} \\
 \hline
 Watches  ($C_{15}$) & {$9.4\times10^6$} & {$6.3\times10^7$} & {1} & {1} & {$4.7\times10^{-6}$} & {$3.3\times10^{-5}$}  \\
\hline
\end{tabular}
\end{center}
\caption{Set of multiple threat and non-threat objects: Full list of 15 threat and non-threat object classes detailing composition of different materials and different object sizes in each class.} \label{tab:objectsummarymat}
\end{table}

\begin{table}[!h]
\begin{center}
\begin{tabular}{|c|c|c|}
\hline
Class & Composition & Total $P^{(k)}$  \\
\hline
Tools $(C_1)$ & Hammers   & $6V^{(2)} + 6V^{(6)} + 18V^{(5)}$ \\
	      & Scissors & {} \\
	       &  Screwdrivers & {} \\
	       \hline
Weapons $(C_2)$ & Guns   & $V^{(1)} + 2 V^{(4)} +5V^{(3)} $\\
	      & Knuckle dusters & {} \\
	       &  Knives & {} \\
\hline
\hline
Clothing $(C_3)$ &  Belt buckles  & $ 12V^{(8)} +3V^{(14)}$ \\
	      & Shoe shanks & {} \\
	      \hline
Earrings $(C_4)$ & Earrings   & $18V^{(10)}$\\
\hline
Pendants $(C_5)$ & Pendants   & $21V^{(12)}$\\
\hline
Pocket items $(C_6)$ &  Coins  & $8V^{(9)} + 4V^{(11)}$\\
	      & Keys & {} \\
	      \hline
Rings $(C_7)$ & Rings   & $21 V^{(13)}$\\
\hline
Wrist items $(C_8)$ &  Bracelets  &$21V^{(7)} + 12V^{(15)} $ \\
	      & Watches & {} \\
\hline
\end{tabular}
\end{center}
\caption{Set of multiple threat and non-threat objects: Amalgamated list of $K=8$  threat and non-threat object classes detailing their composition and total number in each class $P^{(k)}$.} \label{tab:8classproblem}
\end{table}

\begin{figure}[!h]
\centering
\hspace{-1.cm}
$\begin{array}{cc}
\includegraphics[scale=0.5]{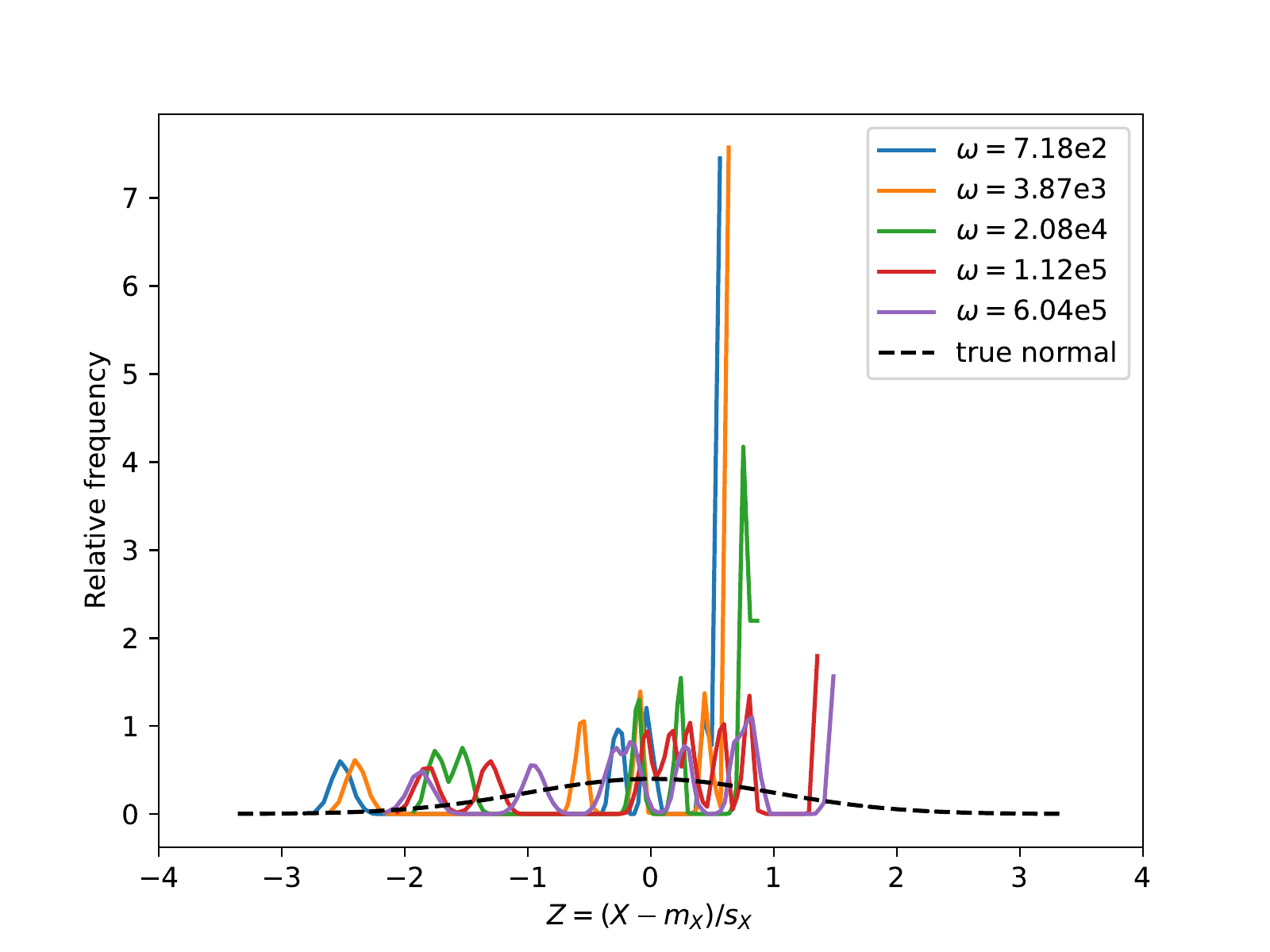}  &
\includegraphics[scale=0.5]{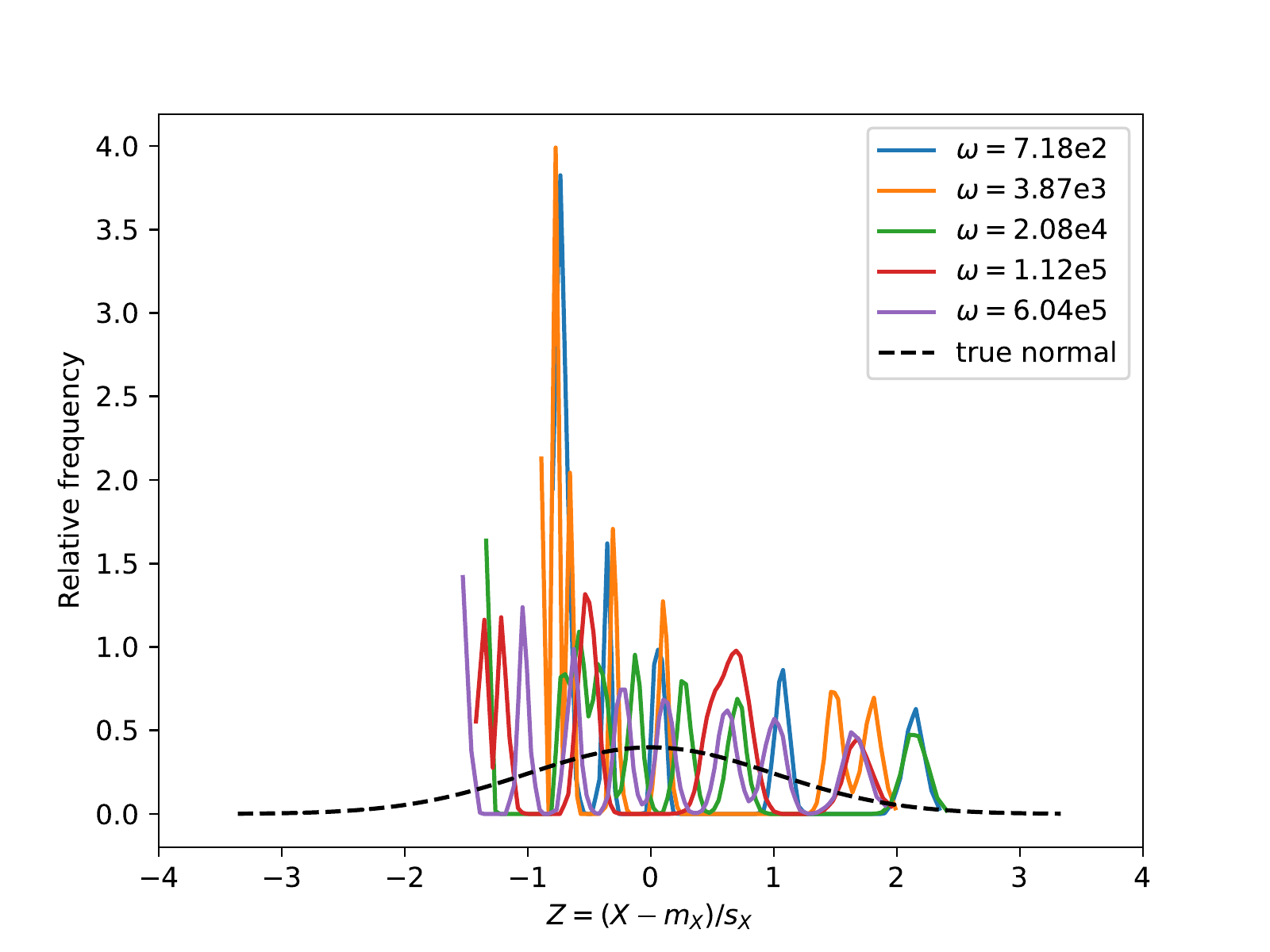}  \\
\text{(a) } I_1 (\tilde{\mathcal{R}}) & \text{(b) } I_1 (\mathcal{I})
\end{array}$
  \caption{Set of multiple threat and non-threat objects: British coins (class $C_1$) with $P^{(k)}=P/K=5000$, with $\alpha \sim N ( 0.001, 8.4\times10^{-6})$ m and $\sigma_* \sim N (m_{\sigma_*}, 0.024m_{\sigma_*})$, where $m_{\sigma_*}$ is determined by the material $B^{(k)}$, showing the normalised histograms of $(Z-m_X)/s_X$ where $X$ is instances of the following (a) $I_1 (\tilde{\mathcal R}[\alpha B^{(1)},\omega_m,\sigma_*,\mu_r)$ and (b) $I_1 ( {\mathcal I}[\alpha B^{(1)},\omega_m,\sigma_*,\mu_r])$ at distinct frequencies $\omega_m$.}
        \label{fig:Coins_p_histograms_dist}
\end{figure}

For these dictionaries, we have $G^{(k)} \ne 1$ in the majority of cases. Considering 
$\alpha\sim   N ( m_\alpha, s_\alpha)$, $\sigma_* \sim N (m_{\sigma_*}, s_{\sigma_*})$ and  $g_k=1,\ldots, G^{(k)}$,  the parent distributions  of  the variables $X =I_i (\tilde{\mathcal R}[\alpha B^{(g_k)},\omega_m,\sigma_*,\mu_r) \sim p(x_{i+(m-1)M}  |C_k) $ and $X=I_i( {\mathcal I}[\alpha B^{(g_k)},\omega_m,\sigma_*,\mu_r]) \sim p(x_{i+(m-1)M}  |C_k) $ will be far from normal. For $P^{(k)}=P/K=5000$ samples and the class $C_1$, comprised of the $G^{(1)}=8$ different denominations of UK coins, the normalised distributions are shown in Figure~\ref{fig:Coins_p_histograms_dist}. Even with a sample size of $P^{(k)}=P/K=5000$ the sample distributions are also far from normal and a very large sample is expected to be needed in order for the central limit theorem to apply in this case.

In the following we will start with classification using the dictionary  $D_8$ and then proceed to present results for  $D_{15}$.

\subsubsection{Classification results using $D_8$}\label{sect:Classification_Results8}
From the observations in Figure~\ref{fig:Coins_p_histograms_dist},  we do not expect logistic regression to perform well using the  $D_8$ dictionary and for it to have a high bias. Instead, we will consider the full range of probabilistic and non-probabilistic classifiers described in Sections~\ref{sect:noprob} and~\ref{sect:prob} and retain logistic regression for comparison.  
The default hyper parameters from \texttt{scikit-learn} were employed in each case apart from the following:
For SVM, rather than the default {\em ovr} strategy, we employ \texttt{decision\_function\_shape=`ovo'}, this is due to the performance of kernel based methods not scaling in proportion with the size of the training dataset.  For MLP, rather than the default settings of \verb#hidden_layer_sizes=(100)#, which means a network with $L=1$ hidden layer and $J=100$ neurons, we have used $L=3$ hidden layers with $J= 50$ neurons in each layer, a choice which we will also justify shortly. In addition, we use \verb#max_iter=300# rather than \verb#max_iter=200# to allow an increased number of iterations to be performed to ensure convergence.  For gradient boost we use \verb#n_estimators=100# and \verb#max_depth=3# and later show the effects of varying the number of trees within the ensemble and the maximum depth of each tree.

\begin{figure}[!h]
$$\begin{array}{cc}
\includegraphics[scale=0.5]{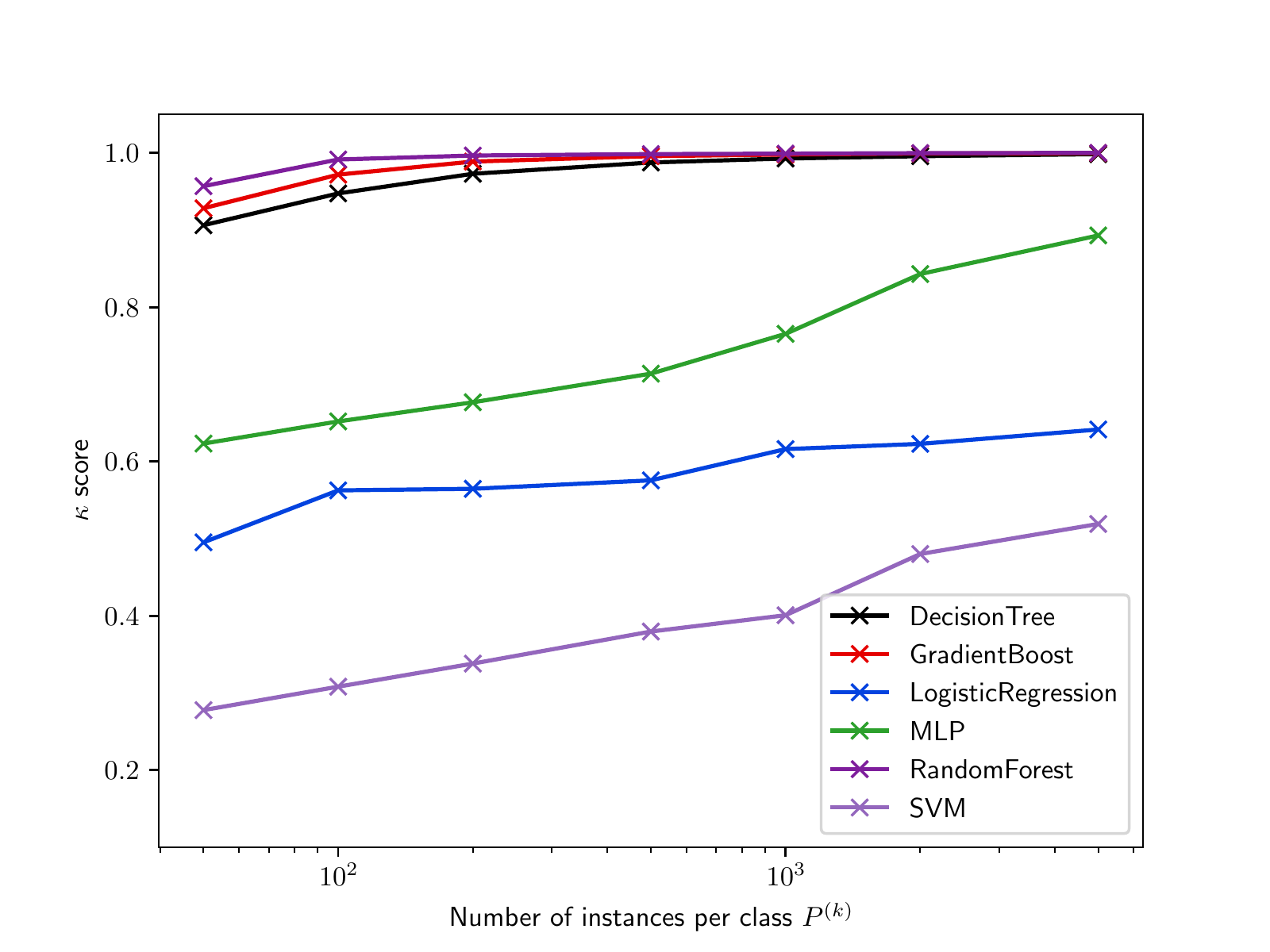} & \includegraphics[scale=0.5]{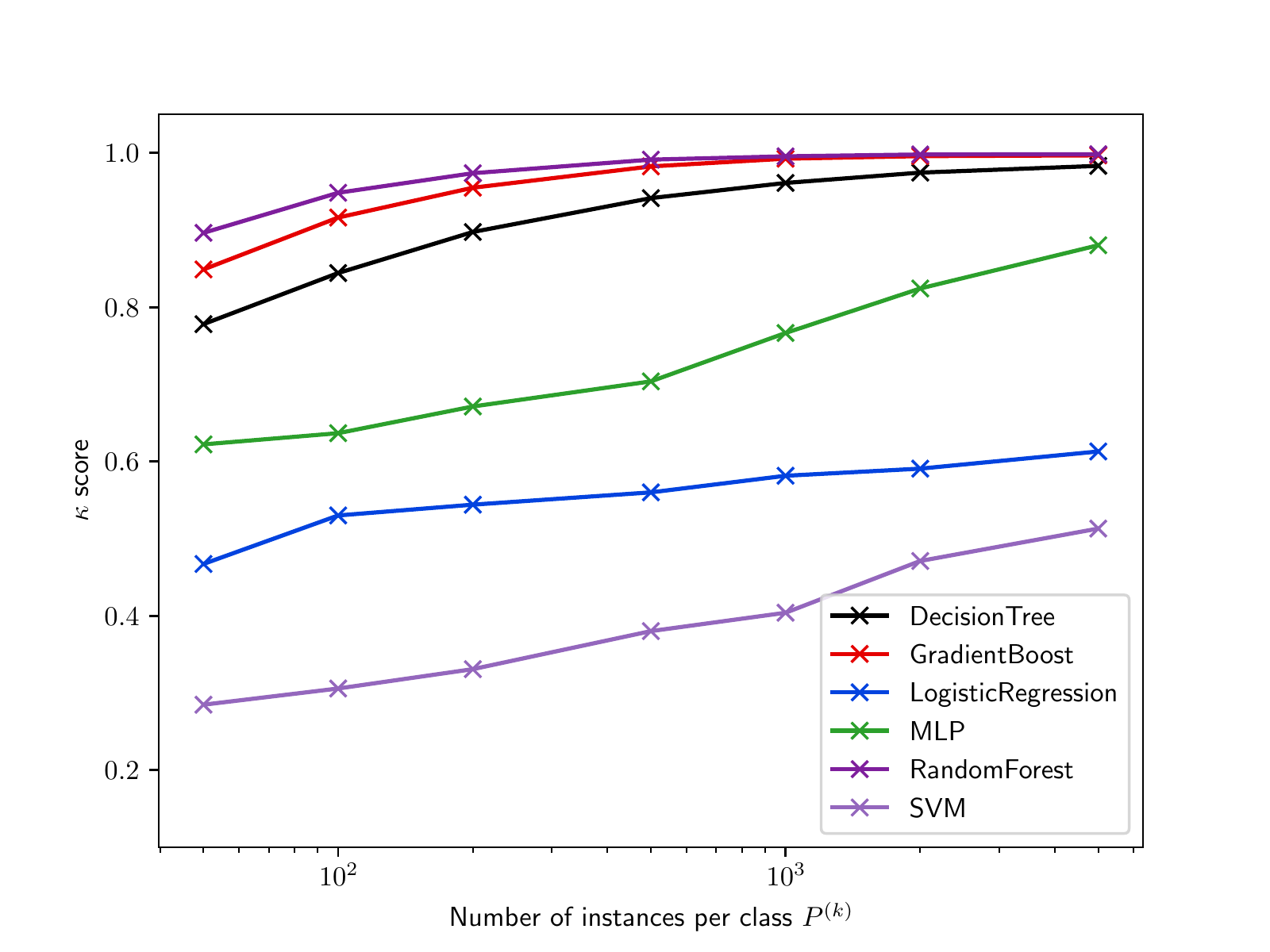}\\
\textrm{\footnotesize{(a) SNR=40dB}} & \textrm{\footnotesize{(b) SNR=20dB}}
\end{array}$$
\caption{Set of multiple threat and non-threat objects:  
Overall performance of different classifiers as a function of $P^{(k)}$ when $K=8$ using the $\kappa$ score (\ref{eqn:kappa}) showing (a) SNR=40dB and (b) SNR=20dB.} \label{fig:noiselvlvsinstance8}
\end{figure}

 In  Figure~\ref{fig:noiselvlvsinstance8}, we show the overall performance of the classifiers with different levels of noise. We use the $\kappa$ score (\ref{eqn:kappa}) to assess the performance of the classification due to the variations within the classes. In each case, we observe that increasing $P^{(k)}\approx P/K$ generally leads to an improved performance of the classification in all cases, since the classifier is exposed to more noisy data in $D_8^{(\text{train})}$ and its variability decreases.
 The figure shows that, in both noise cases, the best performing classifier is random forests, although, for large $P^{(k)}$, the performance of random forest, gradient boost and decision trees are all very similar  with $\kappa\approx 1$ indicating a low bias and low variance. As random forest is a bagging algorithm and gradient boost is a boosting algorithm we expect them to perform well. However, the good performance of decision trees is surprising. 
  The second best probabilistic classifier is MLP, which shows a significant benefit for large $P^{(k)}$.  We do not expect any signifiant improvement of logistic regression for different choices of hyper parameters, although SVM could possibly be improved. Comparing SNR=40dB and SNR=20dB we see a slight reduction in accuracy for a given $P^{(k)}$ using SNR=20dB, although, by increasing $P^{(k)}$, the effects of noise can be overcome.
 Also, although not included, the corresponding results for $5.02 \times 10^4 \text{ rad/s} \le \omega_m \le 8.67 \times 10^4 $ rad/s  using $M=20$ are not as good as those for $7.53 \times 10^2 \text{ rad/s} \le \omega_m \le 5.99 \times 10^5 \text{ rad/s}$ using $M=28$, with those shown offering at least a 5\% improvement for the best performing classifiers, small $P^{(k)}$ and SNR=20dB. Interestingly, logistic regression improves by 25\% when the larger frequency range is used.  
We focus on the two best performing probabilistic classifiers, gradient boost and MLP, in the following.
 
\begin{figure}[!h]
$$\begin{array}{cc}
\includegraphics[scale=0.5]{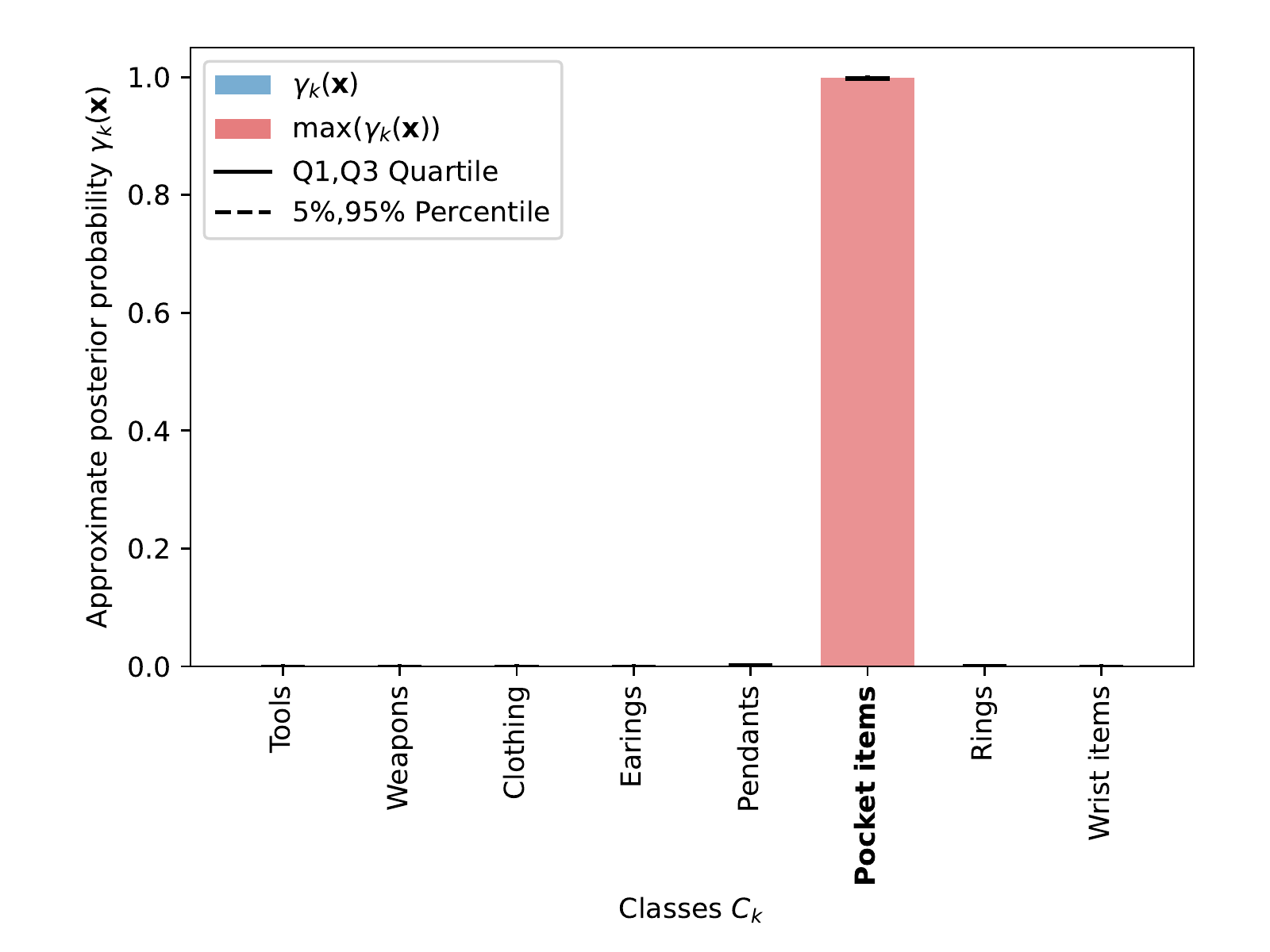} & \includegraphics[scale=0.5]{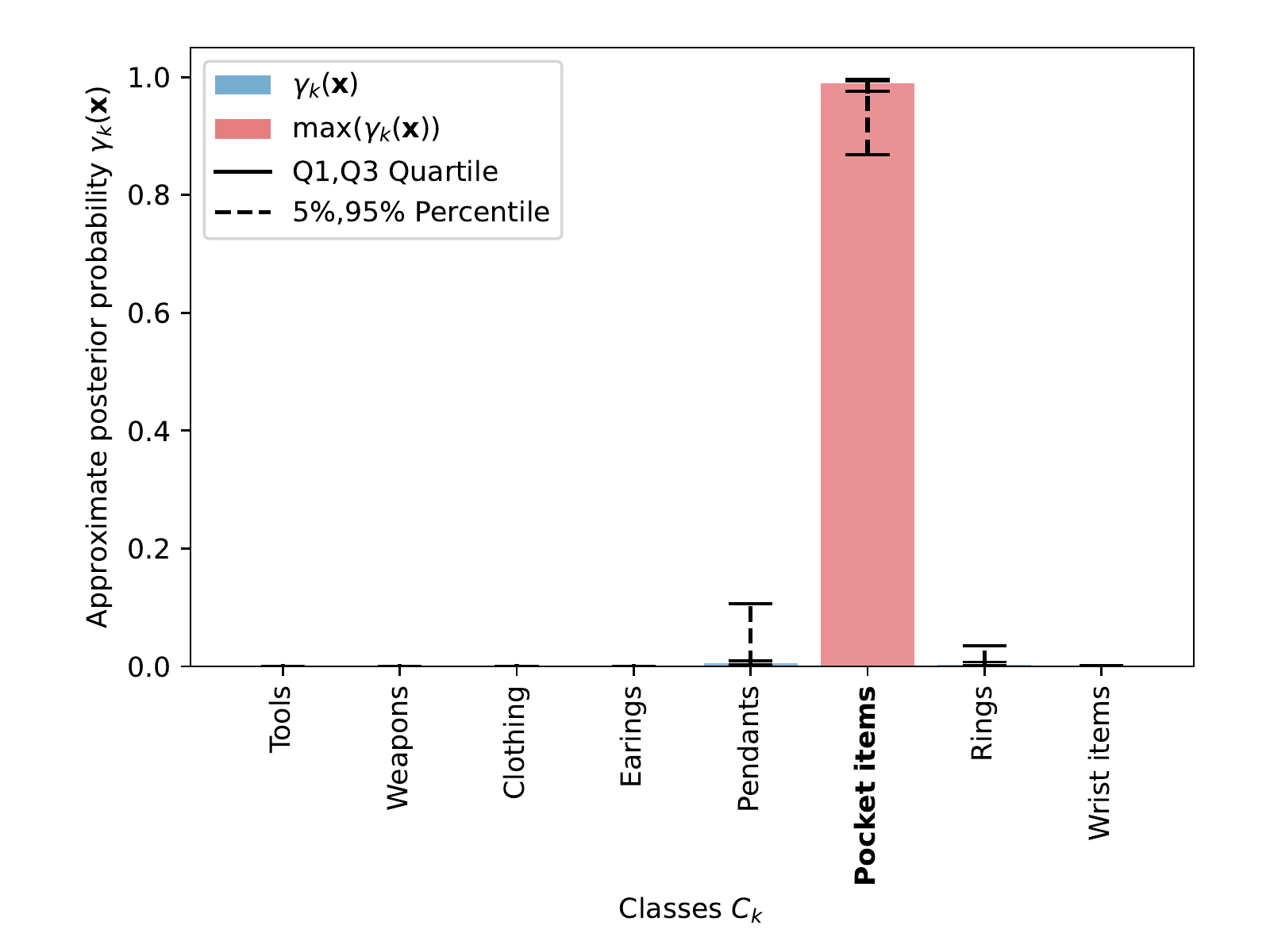}\\
\textrm{\footnotesize{(a) Gradient boost SNR=40dB}} & \textrm{\footnotesize{(b) Gradient boost SNR=20dB}}\\
\includegraphics[scale=0.5]{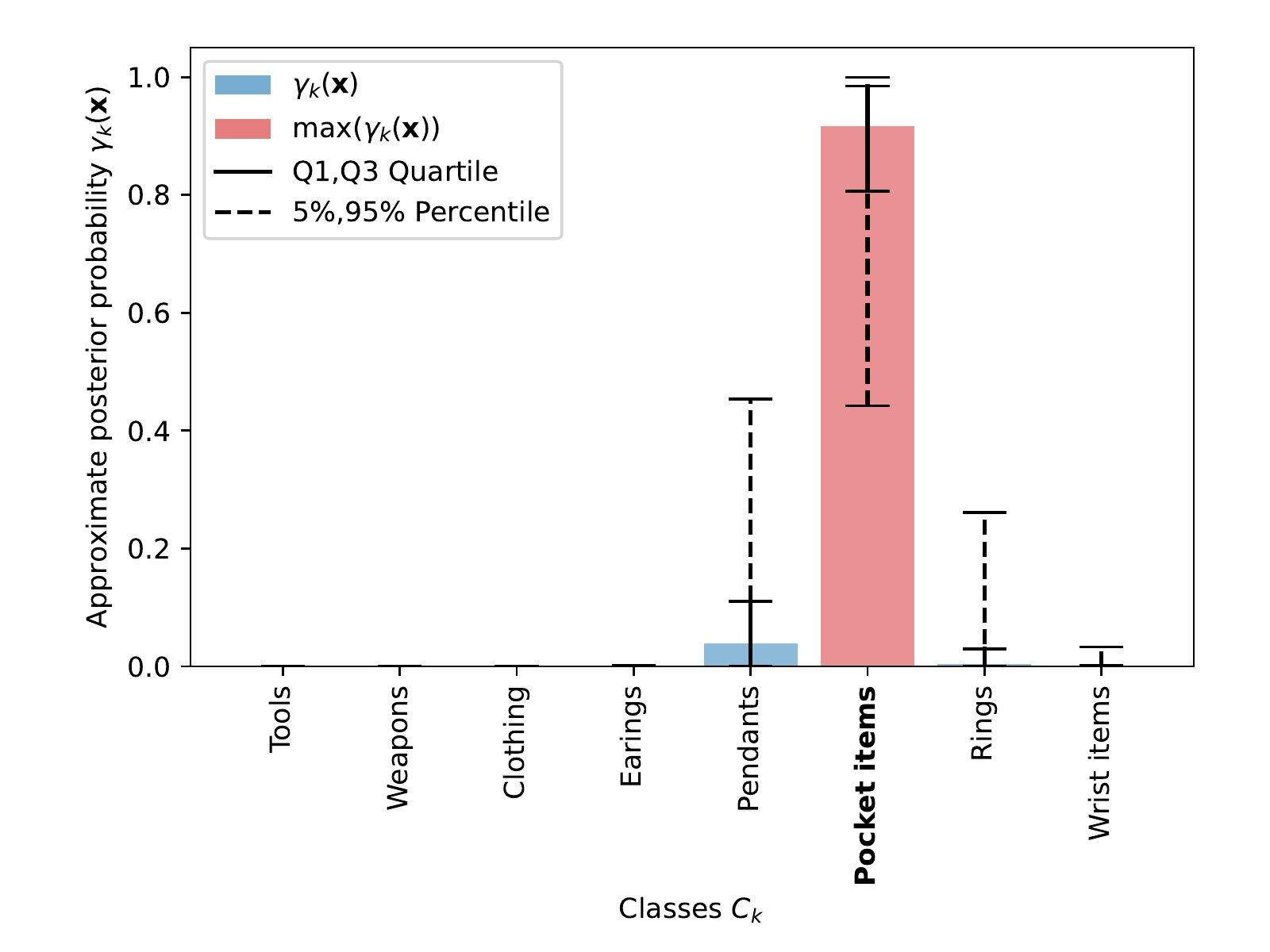} & \includegraphics[scale=0.5]{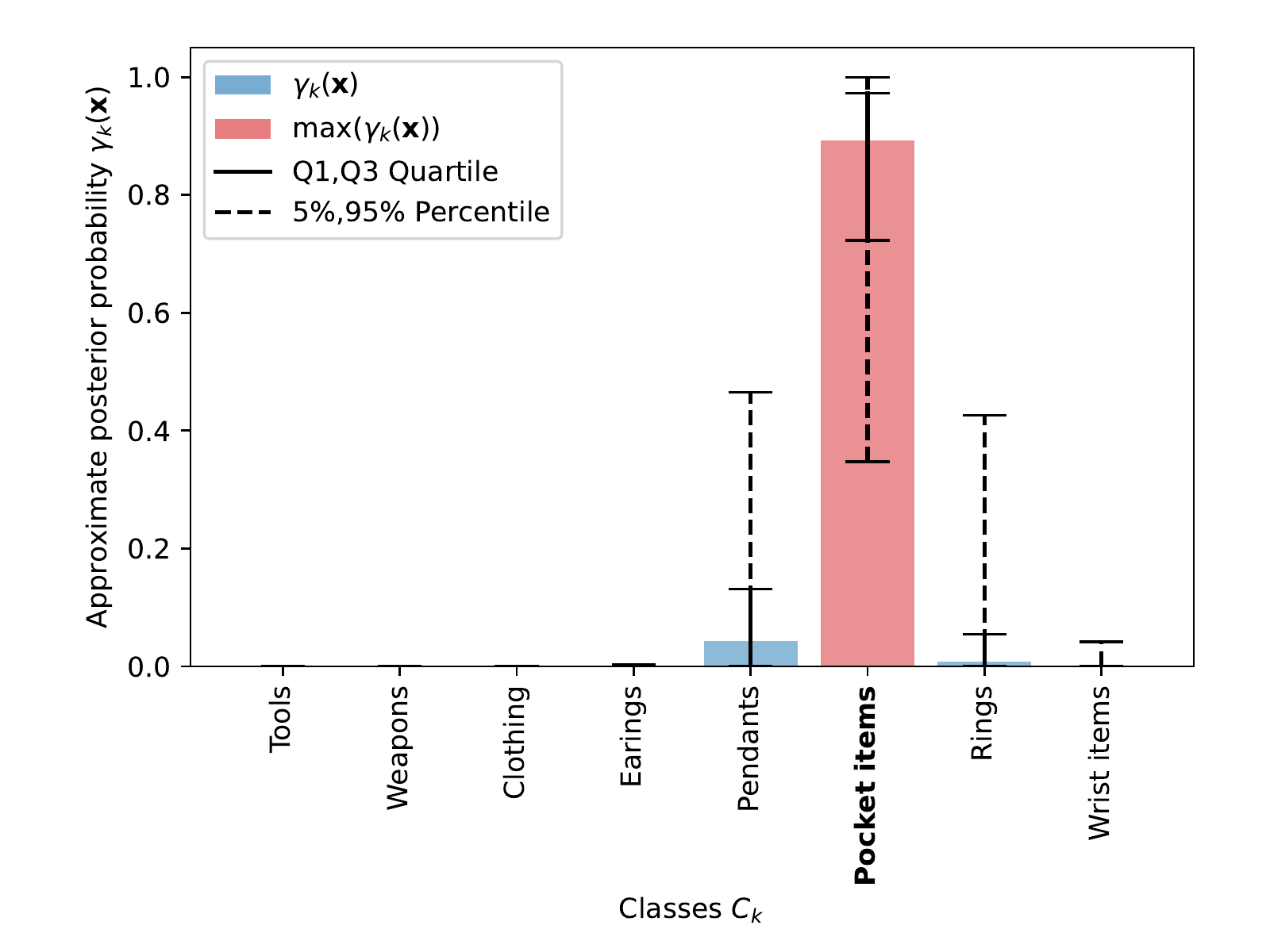}\\
\textrm{\footnotesize{(c) MLP SNR=40dB}} & \textrm{\footnotesize{(d) MLP SNR=20dB}}
\end{array}$$
\caption{Set of multiple threat and non-threat objects: 
Approximate posterior probabilities $\gamma_k({\mathbf x}) \approx p(C_k|{\mathbf x})$, $k=1,\ldots,K$, where $({\mathbf x}, {\mathbf t})\in D_8^{(\text{test},(2))}$ for $P^{(k)}=5000$ when $K=8$ showing the classifiers (a) gradient boost SNR=40dB, (b) gradient boost SNR=20dB, (c) MLP SNR=40dB and (d) MLP SNR=20dB. } \label{fig:postprob8class}
\end{figure}

The approximate posterior probability distributions 
$\gamma_k({\mathbf x})\approx p(C_k|{\mathbf x})$, $k=1,\ldots,K$, we obtain for gradient boost and MLP are shown in Figure~\ref{fig:postprob8class}. We have chosen  $({\mathbf x}, {\mathbf t})\in D_8^{(\text{test},(6))}$ so that the correct classification should be $C_6$ (ie a pocket item: a coin or key). Additionally, the bars we show are for the median value $\gamma_{k,50}$ of $\gamma_k({\mathbf x})$, obtained by considering all the samples  $({\mathbf x}, {\mathbf t})\in D_8^{(\text{test},(6))}$, and we also indicate the $Q_1$, $Q_3$ quartiles as well as $\gamma_{k,5}$ and $\gamma_{k,95}$, for different SNR, which have been obtained using (\ref{eqn:percentile}). 
The results for  SNR=40dB  strongly indicate that the most likely class is a pocket item for both classifiers,  since $\gamma_{6,50} \approx 1$. For the gradient boost classifier,  the inter quartile and inter percentile ranges are small and, so, we have high confidence in this prediction. However, we have less confidence in  the corresponding prediction for the MLP as both the inter quartile and inter percentile ranges are larger. For SNR=20dB, we see the median value $\gamma_{6,50}$ fall for both classifiers and we also have much greater uncertainty in the classification over the samples, as illustrated by the larger inter percentile ranges for the different object classes. Comparing the two classifiers, we have less confidence in the prediction with MLP than for gradient boost.

\begin{figure}[!h]
$$\begin{array}{cc}
\includegraphics[scale=0.5]{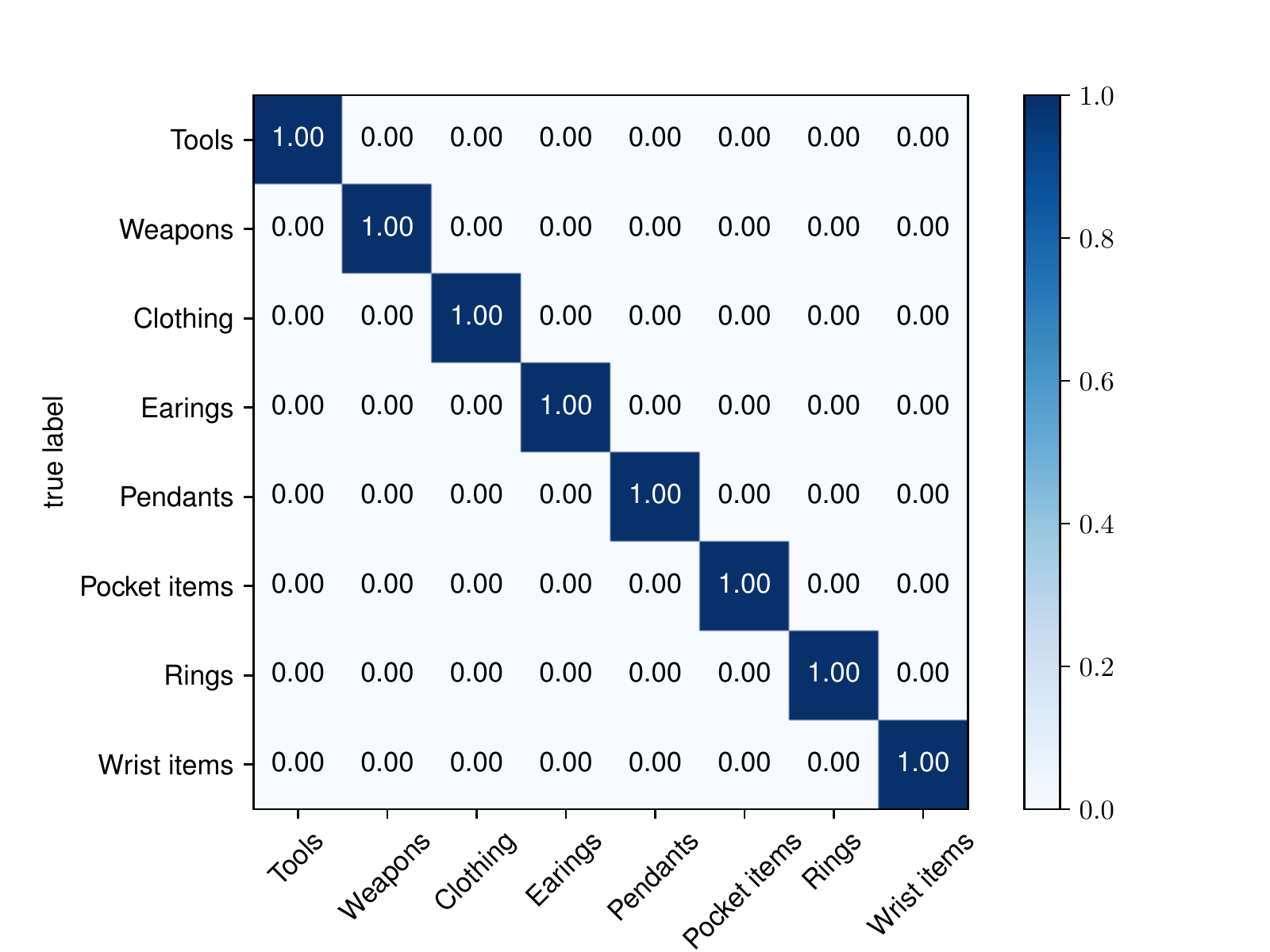} & \includegraphics[scale=0.5]{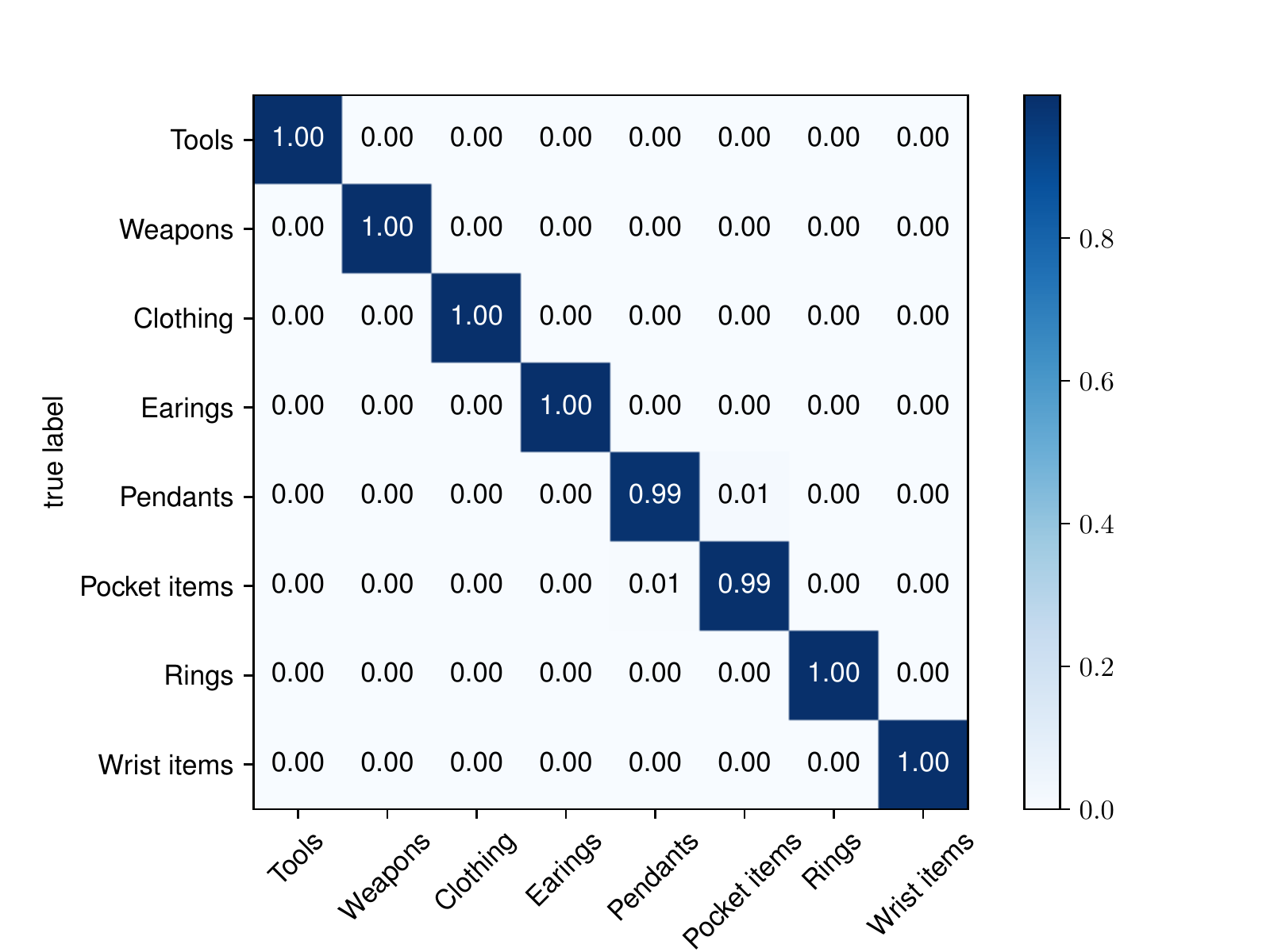}\\
\textrm{\footnotesize{(a) Gradient boost SNR=40dB}} & \textrm{\footnotesize{(b) Gradient boost SNR=20dB}}\\
\includegraphics[scale=0.5]{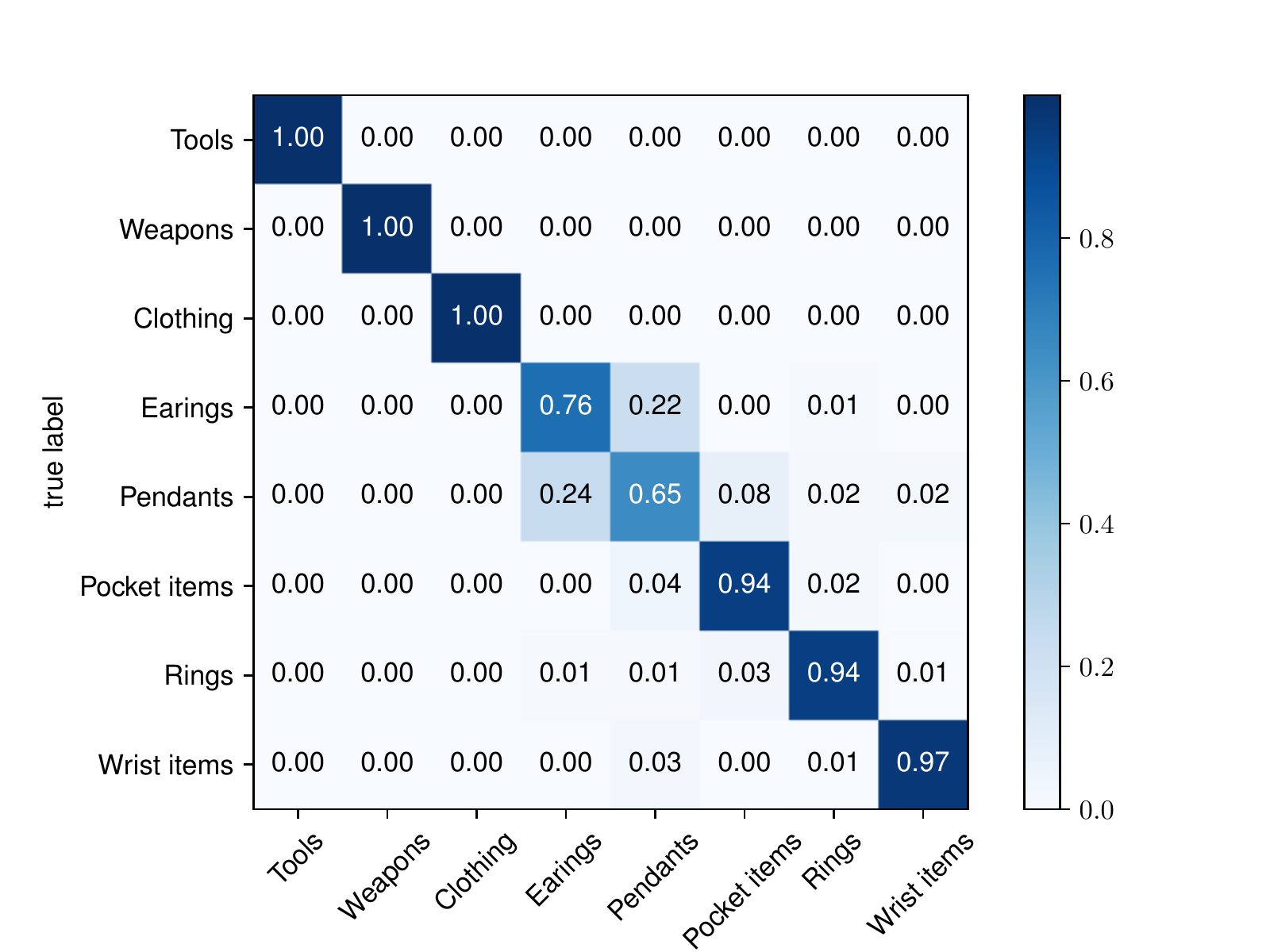} & \includegraphics[scale=0.5]{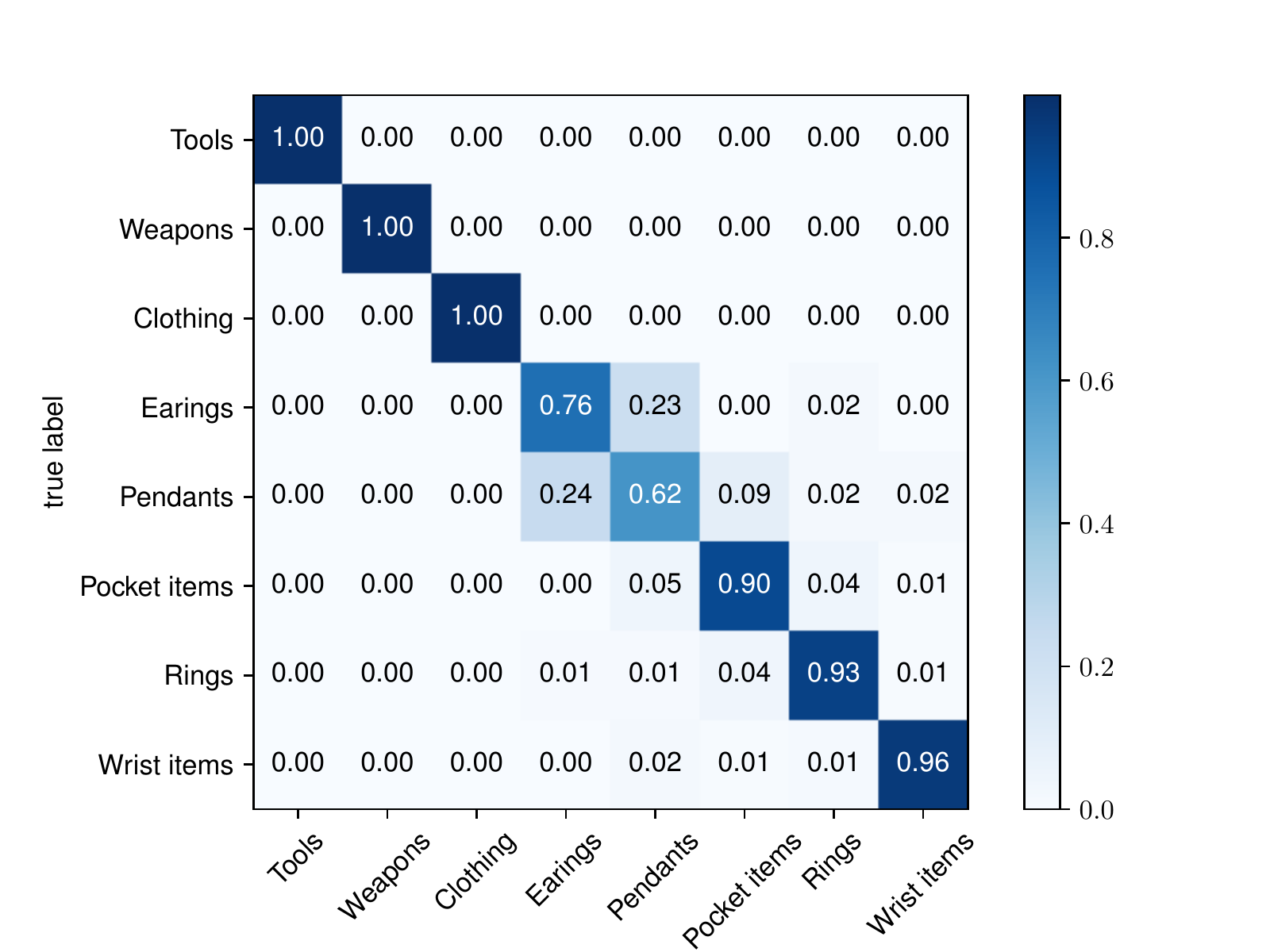}\\
\textrm{\footnotesize{(c) MLP SNR=40dB}} & \textrm{\footnotesize{(d) MLP SNR=20dB}}
\end{array}$$
\caption{Set of multiple threat and non-threat objects: Comparison of confusion matrices for  $P^{(k)}= 5000$ when $K=8$ showing the classifiers (a) gradient boost SNR=40dB, (b) gradient boost SNR=20dB, (c) MLP SNR=40dB and (d) MLP SNR=20dB.}\label{fig:confmatrix8class}
\end{figure}

Next, we consider the frequentist approximations to 
$p(C_j |{\mathbf x})$ for $({\mathbf x},{\mathbf t})\in D^{(\text{test},(i))}$ presented in the form of a confusion matrix with entries $({\mathbf C})_{ij}$, $i,j=1,\ldots,K$,
for the cases of SNR=40dB and SNR=20dB and the gradient boost and MLP classifiers in Figure~\ref{fig:confmatrix8class}.
 As expected, for SNR=20dB, we see increased misclassification amongst the classes compared to SNR=40dB with situation being worse for the MLP classifier compared to the gradient boost. Looking at the row corresponding to the true label for the $C_6$ (pocket items) class, for both SNR=40dB and SNR=20dB, we can see that the frequentist probability in column $j$ are approximately similar to the median approximate posterior probability $\gamma_j({\mathbf x})$ shown in Figure~\ref{fig:postprob8class}.  Also, while the gradient boost exhibits near perfect classification for SNR=40dB (and SNR=20dB), MLP does not perform as well, particularly among the earrings and pendents.

In Table~\ref{tab:8-classprecision}, we show the  precision, sensitivity and specificity for each of the different object classes $C_k$, $k=1,\ldots,K$, for the case of SNR=20dB and the MLP classifier. In general, we see the proportion of negatives that are correctly identified  is very high (as indicated by the specificity) and is close to $1$ in all cases, whereas the proportions of positives correctly identified (indicated by the precision and sensitivity) varies amongst the different object  classes, the best case being $C_1$  (weapons) and worst case $C_5$ (pendents).  The corresponding results for gradient boost are all close to $1$.

\begin{table}[!h]
\begin{center}
\begin{tabular}{|c|c|c|c|}
\hline
$C_k$ & Precision &  Sensitivity & Specificity \\
\hline
Tools $(C_1)$         & 0.97 & 0.98 & 1.00 \\\hline
Weapons $(C_2)$        & 0.99 & 0.99 & 1.00 \\\hline
\hline
Clothing   $(C_3)$    & 0.98 & 0.98 & 1.00 \\\hline
Earrings     $(C_4)$  & 0.64 & 0.84 & 0.93 \\\hline
Pendants    $(C_5)$   & 0.60 & 0.34 & 0.97 \\\hline
Pocket items $(C_6)$  & 0.75 & 0.73 & 0.96 \\\hline
Rings      $(C_7)$    & 0.70 & 0.82 & 0.95 \\\hline
Wrist items  $(C_8)$  & 0.90 & 0.87 & 0.99 \\\hline
\end{tabular}
\end{center}
\caption{Set of multiple threat and non-threat objects: Precision, sensitivity and specificity measures (to 2d.p.) for each of the classes $C_k$  when SNR=20dB and $P^{(k)}= P/K =5000$ for the MLP classifier.} \label{tab:8-classprecision}
\end{table}

%\textcolor{blue}{Next we look at the effect varying the hyperparameters for the different models has on the results.}

%\subsubsection{Hyperparameter tuning}\label{sect:Hyperparameter_turning}
%\textcolor{blue}{A hyperparameter is a model parameter which is defined by the user as opposed to derived in the training process. Each model has different hyperparameters which affect the training processes and therefore then end model produced. For the MLP, by varying the number of hidden layers in the network along with the number of neurons and connections between layers we may change the architecture of the network created. This in turn affects the overall performance of the produced model, in Figure ~\ref{fig:noiselvlvsinstance8mlp} we compare the performance of 3 MLP}

\begin{figure}[h]
\begin{center}
\includegraphics[width=0.50\textwidth, keepaspectratio]{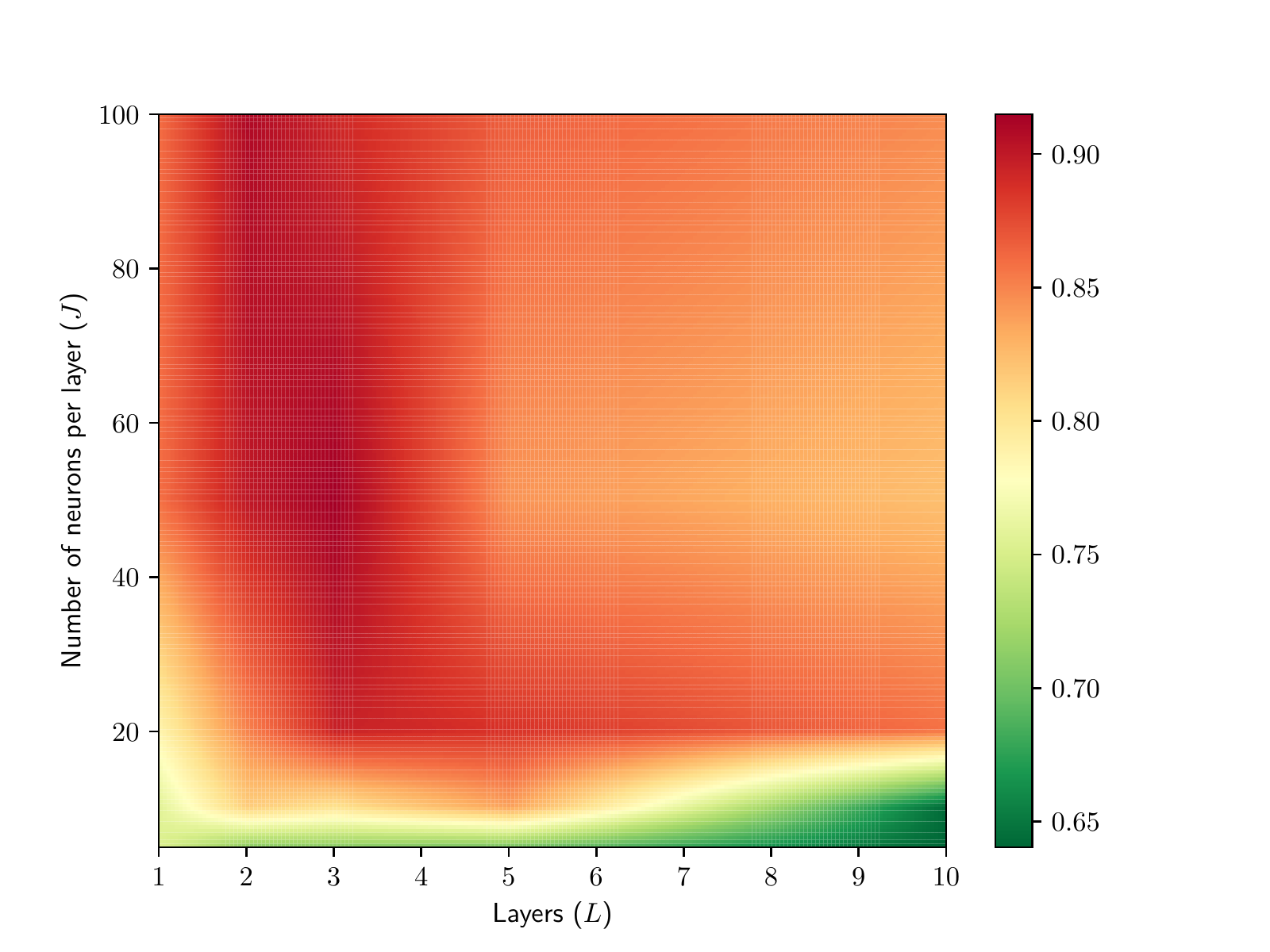}
\end{center}
\caption{Set of multiple threat and non-threat objects: Overall performance of MLP for different uniform network architectures, with $P^{(k)}=5000$ when $K=8$ and SNR=40dB, showing $\kappa$ score for different numbers of hidden layers $L$ and numbers of neurons per layer $J$.}
\label{fig:mlp_contour}
\end{figure}

To justify our choice of $L=3$ and $J=50$ for MLP, we investigate the $\kappa$ score for different choices of $L$ and $J$ in  Figure~\ref{fig:mlp_contour} for the case where SNR=40dB. For this result, we have assumed the same number of neurons in each layer. From this figure, we observe there are a range of different $L$ and $J$ that lead to a network with a similar level of accuracy.  As remarked in Section~\ref{sect:mlp}, for the type of network we are considering, the number of  variables grows quadratically with $J$ and linearly with $L$. Hence, from a computational cost perspective, choosing a network with a small $J$ and a large $L$ is generally preferable to a network with a large $J$ and a small $L$, if the cost of computing each variable is assumed the same. For this, reason, we adopt a network with 
$L=3$ and $J=50$ as MLP architectures in this range result in high $\kappa$ score, while minimising computational cost. Also, if desired, $J$ could be further reduced without comprising accuracy. Of course, this choice has been optimised for  $P^{(k)}=5000$, SNR=40dB and this $K=8$ class problem, for different  $P^{(k)}$ and SNR levels, as well as other classification problems, this choice may no longer be optimum. 

\subsubsection{Classification results using $D_{15 }$}\label{sect:Classification_Results15}

\begin{figure}[!h]
$$\begin{array}{cc}
\includegraphics[scale=0.5]{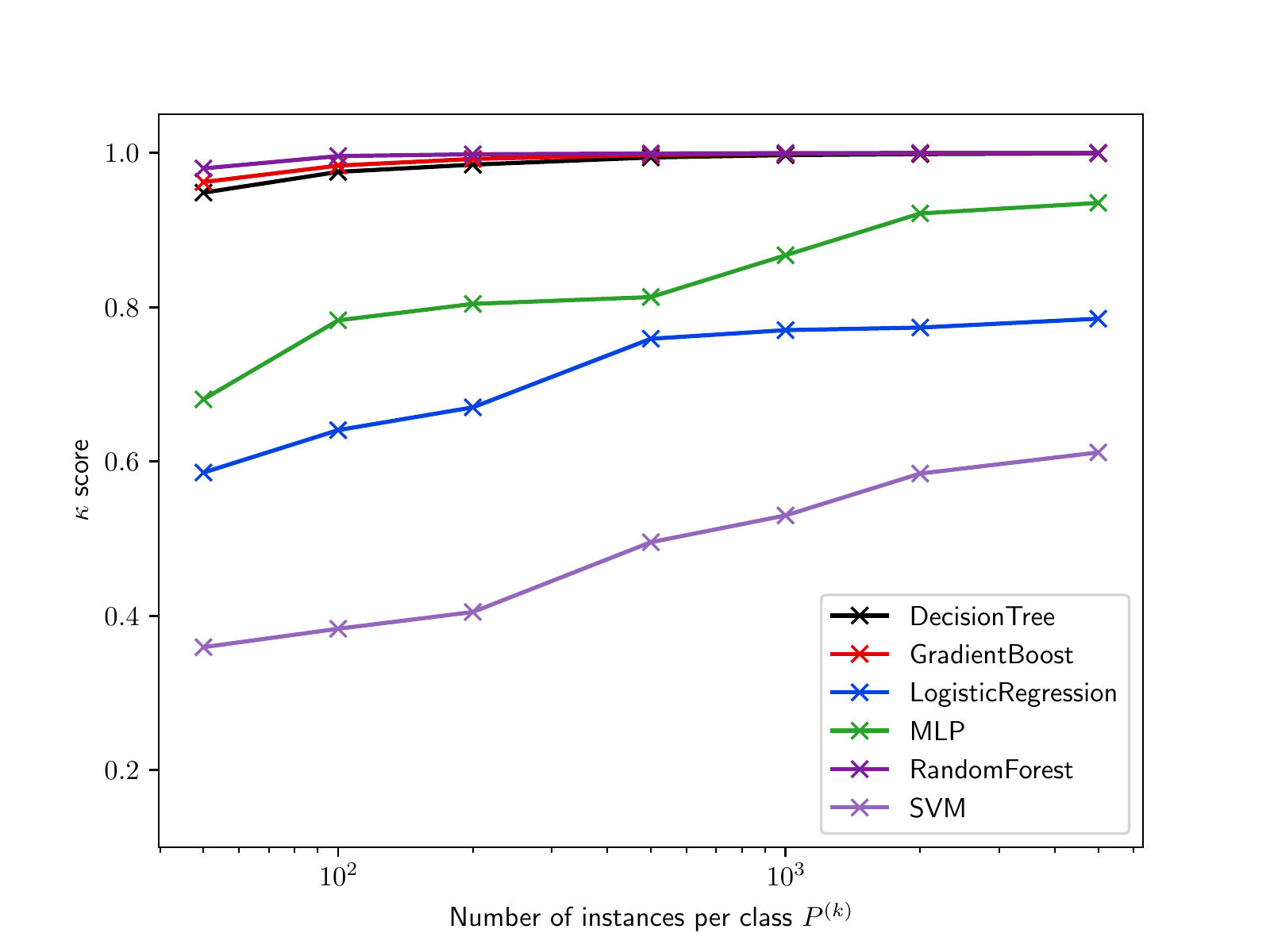}
& \includegraphics[scale=0.5]{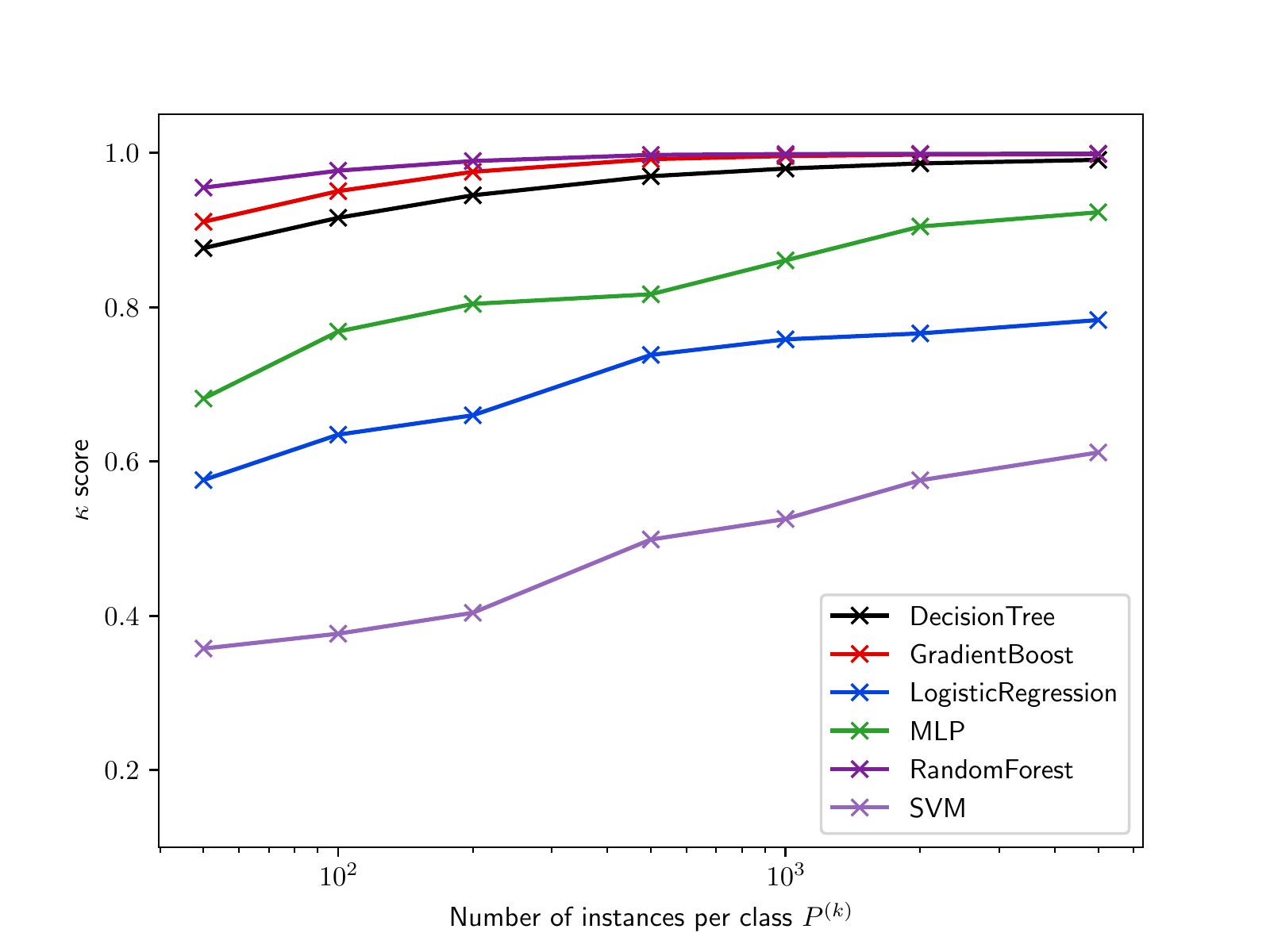} \\
\textrm{\footnotesize{(a) SNR=40dB}} & \textrm{\footnotesize{(b) SNR=20dB}}
\end{array}$$
\caption{Set of multiple threat and non-threat objects:  
Overall performance of different classifiers as a function of $P^{(k)}$ when $K=15$ using the $\kappa$ score (\ref{eqn:kappa}) showing (a) SNR=40dB and (b) SNR=20dB.} \label{fig:noiselvlvsinstance15}
\end{figure}

Figure~\ref{fig:noiselvlvsinstance15}
repeats the investigation shown in Figure~\ref{fig:noiselvlvsinstance8}  for $D_{15}$, instead of  $D_{8}$, using  the same classifier hyperparameters. The trends described previously again apply, except, with a further significant gain in the performance for all classifiers for the increased fidelity $K=15$ class problem compared to the previous $K=8$ class problem. This is because  each  class, for $K=15$, is comprised of objects that have increased similarity between their volumes, shapes and materials, and, hence, their MPT spectral signatures, compared to the $K=8$ problem. This, in turn, reduces each classifier's bias as it becomes easier to establish the relationship between the features and class. Nonetheless,
$X =I_i (\tilde{\mathcal R}[\alpha B^{(g_k)},\omega_m,\sigma_*,\mu_r) \sim p({\rm x}_{i+(m-1)M}  |C_k) $ 
and
$X=I_i( {\mathcal I}[\alpha B^{(g_k)},\omega_m,\sigma_*,\mu_r]) \sim p({\rm x}_{i+(m-1)M}  |C_k) $ are still far from normal and, so, logistic regression does not perform well. The best performance being again given by random forests, gradient boost and decision trees.  We also do not expect any signifiant improvement of logistic regression for different choices of hyper parameters, although SVM could possibly be improved. We focus on the gradient boost and MLP, which are the best two performing probabilistic classifiers in the following. 

\begin{figure}[!h]
$$\begin{array}{cc}
\includegraphics[scale=0.5]{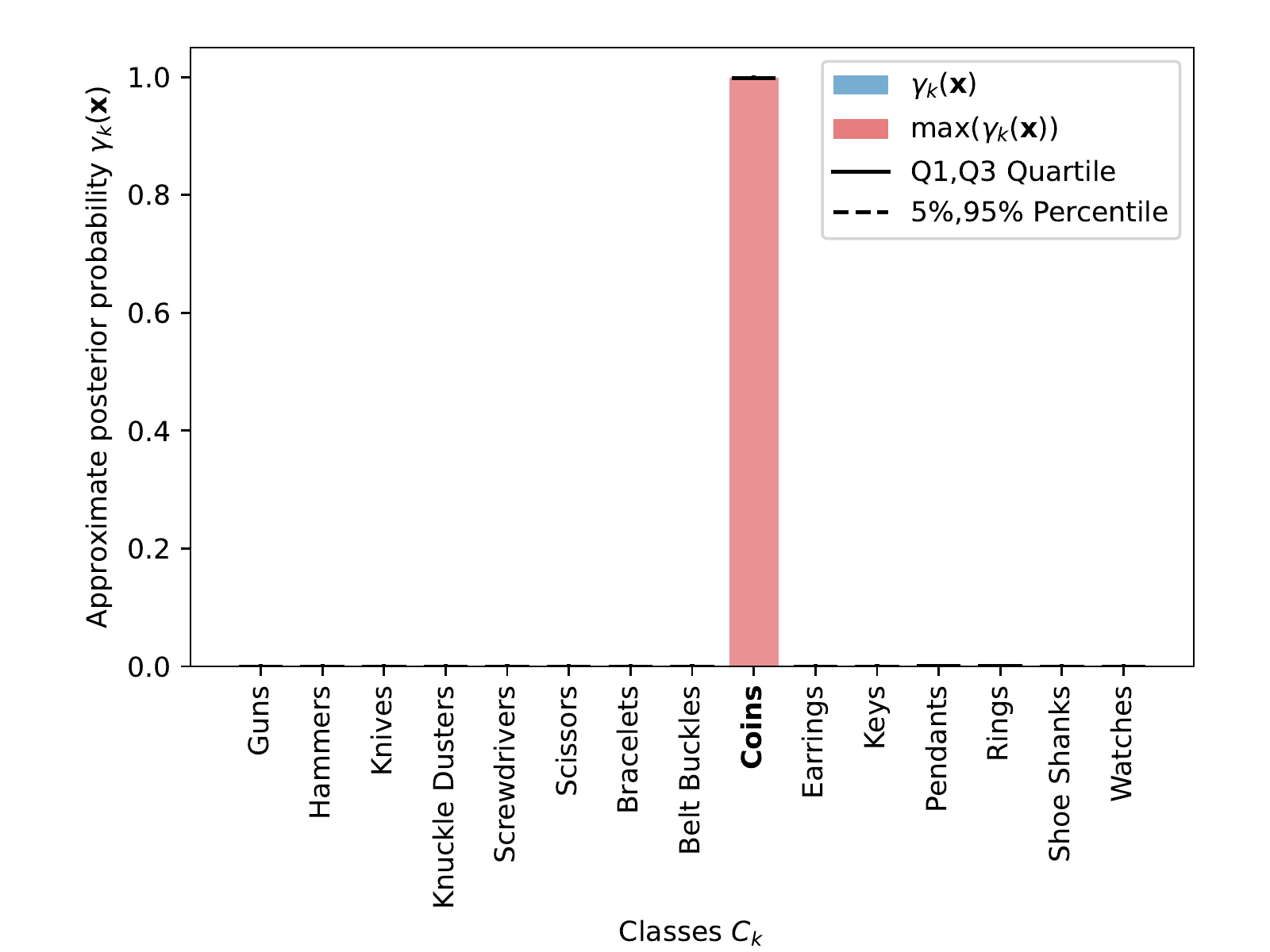} & 
\includegraphics[scale=0.5]{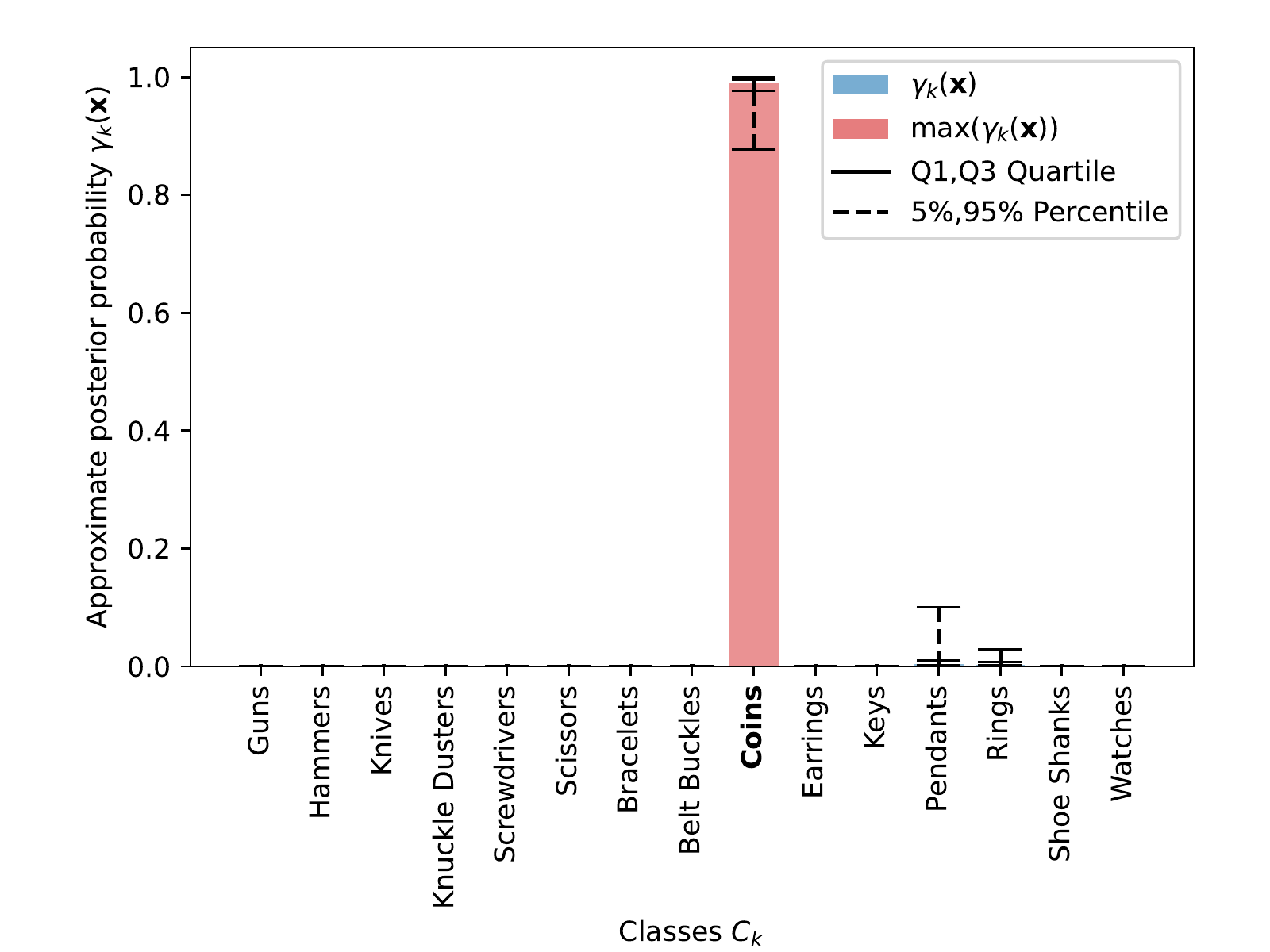}\\
\textrm{\footnotesize{(a) Gradient boost SNR=40dB}} & \textrm{\footnotesize{(b) Gradient boost SNR=20dB}}\\
\includegraphics[scale=0.5]{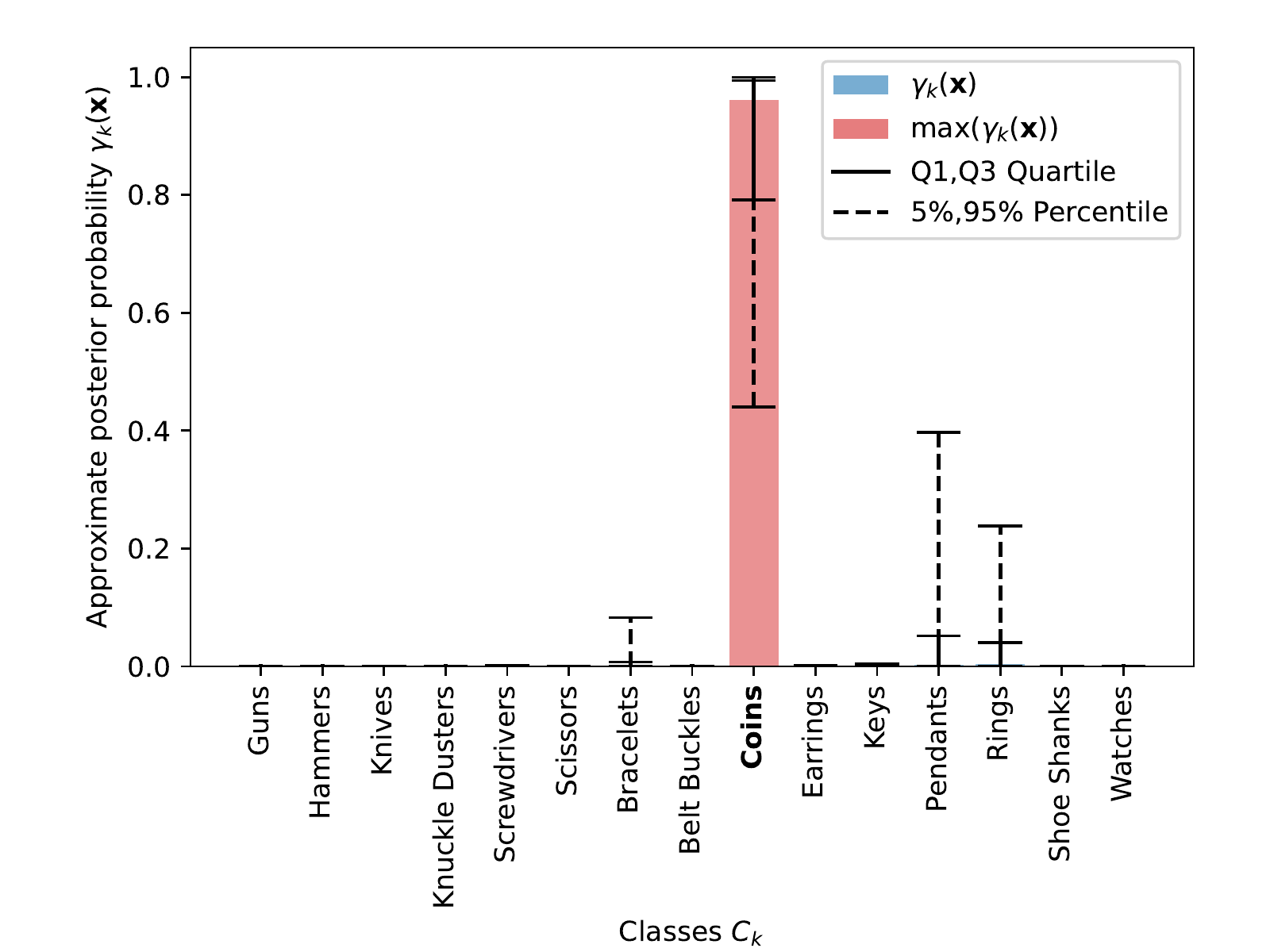} & 
\includegraphics[scale=0.5]{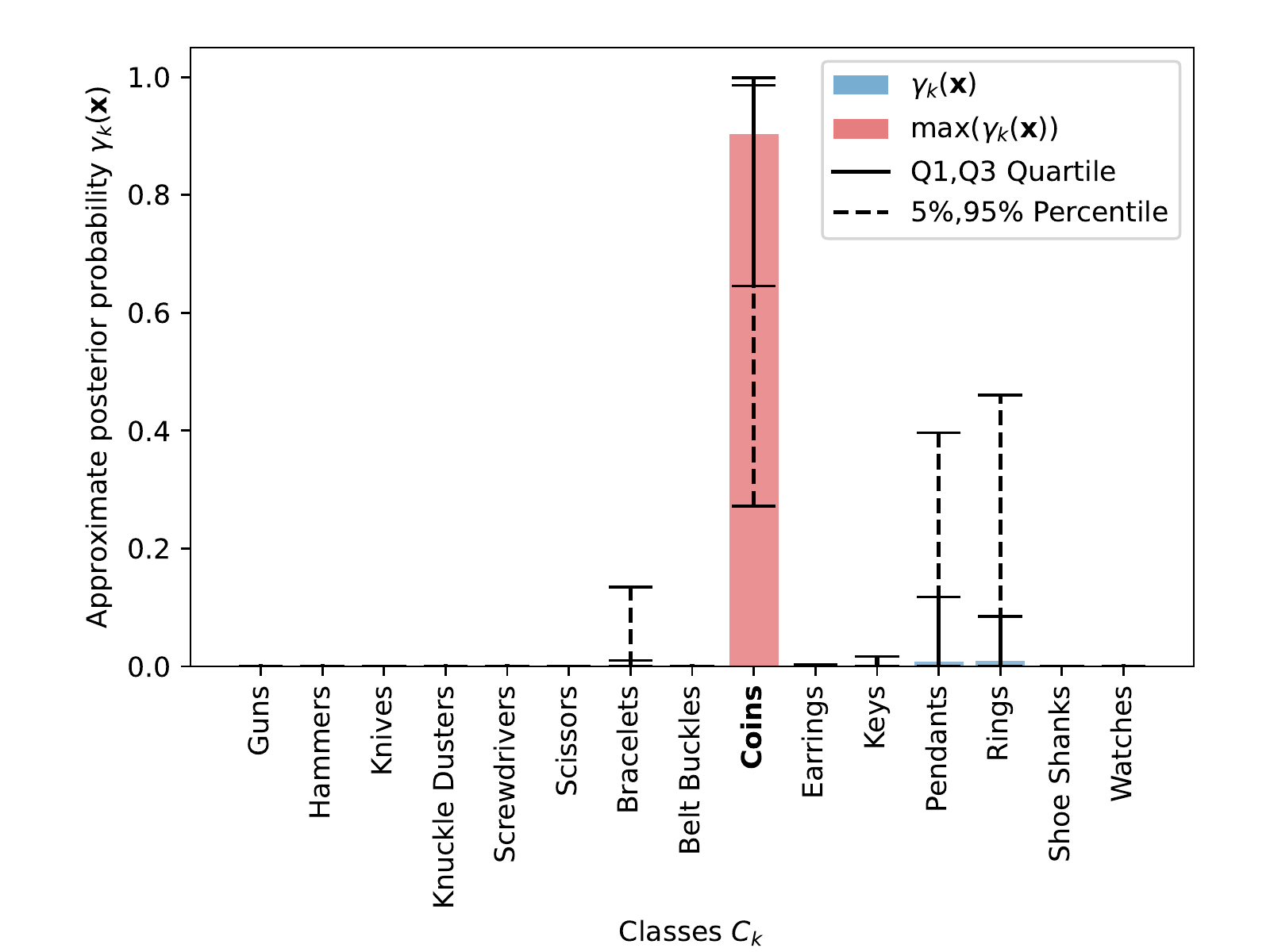}\\
\textrm{\footnotesize{(c) MLP SNR=40dB}} & \textrm{\footnotesize{(d) MLP SNR=20dB}}
\end{array}$$
\caption{Set of multiple threat and non-threat objects: 
Approximate posterior probabilities $\gamma_k({\mathbf x}) \approx p(C_k|{\mathbf x})$, $k=1,\ldots,K$, where $({\mathbf x}, {\mathbf t})\in D_{15}^{(\text{test},(9))}$ (Coins) for $P^{(k)}=5000$ when $K=15$ showing (a) gradient boost SNR=40dB, (b) gradient boost SNR=20dB, (c) MLP SNR=40dB and (d) MLP SNR=20dB.}\label{fig:postprob15class}
\end{figure}

The approximate posterior probability distributions 
$\gamma_k ({\mathbf x}) \approx p(C_k|{\mathbf x})$, $k=1,\ldots,K$, we obtain for gradient boost and MLP are shown in Figure~\ref{fig:postprob15class}. We have chosen  $({\mathbf x}, {\mathbf t})\in D_{15}^{(\text{test},(9))}$ so that the correct classification should be $C_{9}$. The bars are for $\gamma_{k,50}$, obtained by considering all the samples  $({\mathbf x}, {\mathbf t})\in D_{15}^{(\text{test},(9))}$, and we also indicate the $Q_1$, $Q_3$ quartiles as well as $\gamma_{k,5}$ and $\gamma_{k,95}$, for different SNR, which have been obtained using (\ref{eqn:percentile}). 
The results for SNR=40dB  strongly indicate that the most likely class is a pocket item for both classifiers, since $\gamma_{9,50} \approx 1$. For the gradient boost classifier,  the inter quartile and inter percentile ranges are very small and so we have very high confidence in this prediction; the MLP classifier has larger ranges and less confidence. For SNR=20dB, we see $\gamma_{9,50}$ fall slightly for gradient boost and by a larger amount for MLP. The gradient boost still shows a high degree of confidence in the prediction, but the MLP is more uncertain. Compared to the results shown in Figure~\ref{fig:postprob8class} for $D_8$, the performance in  Figure~\ref{fig:postprob15class} for  $D_{15}$ is improved for MLP and remains excellent for gradient boost (when considering the amalgamated pocket item class and the split coin and keys classes).

\begin{figure}[!h]
$$\begin{array}{cc}
\includegraphics[scale=0.35]{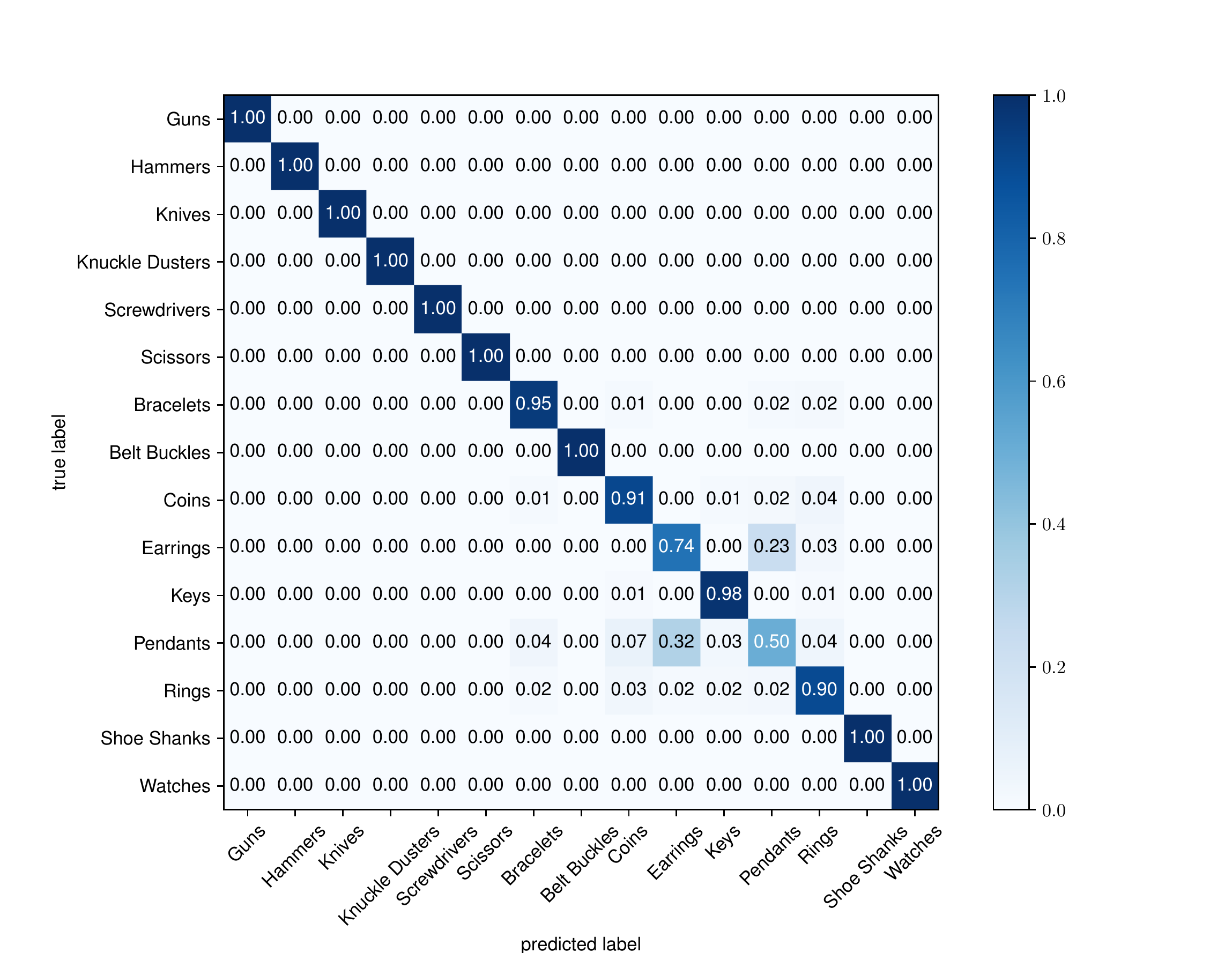} &
 \includegraphics[scale=0.35]{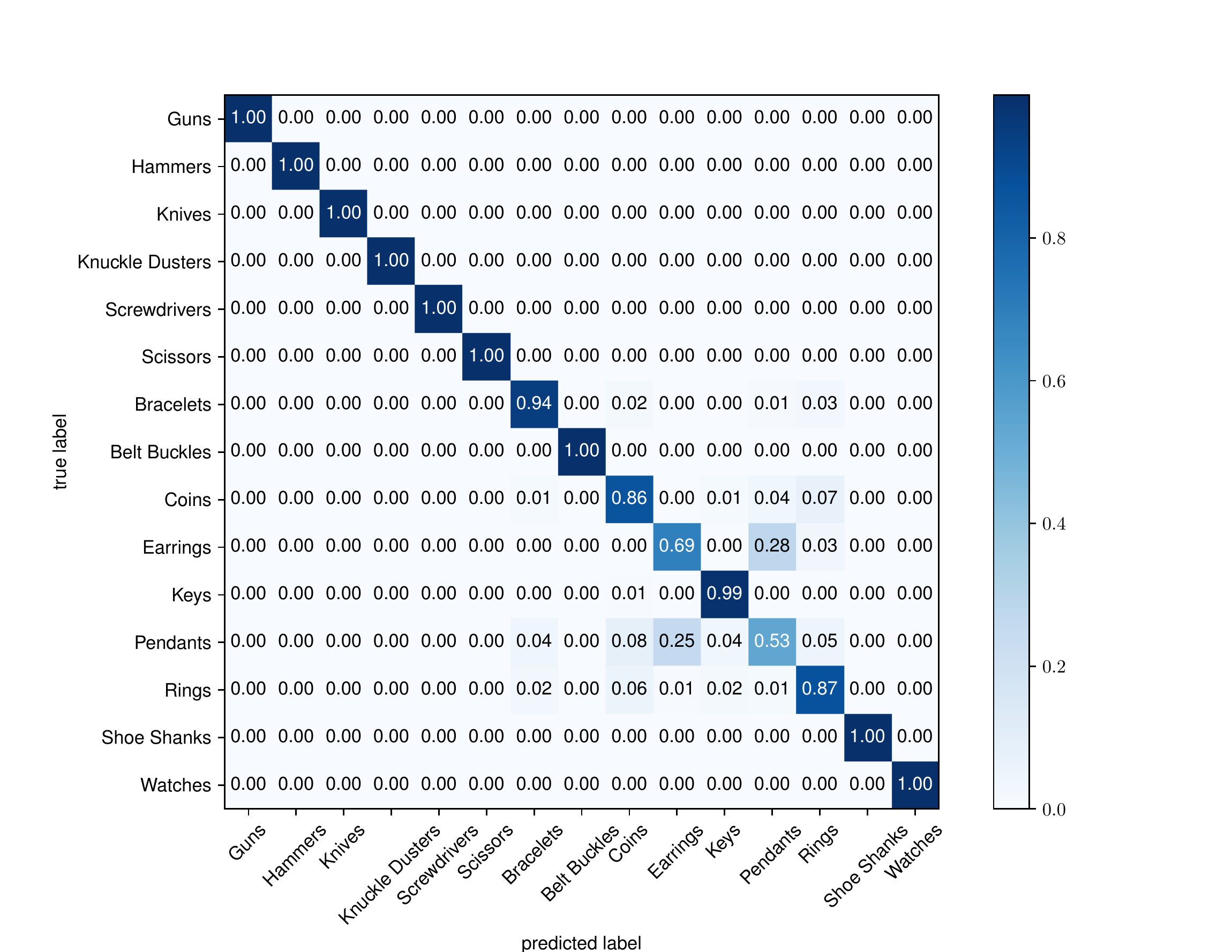}\\
\textrm{\footnotesize{(a) MLP SNR=40dB}} & \textrm{\footnotesize{(b) MLP SNR=20dB}}
\end{array}$$
\caption{Set of multiple threat and non-threat objects: Comparison of confusion matrices for  $P^{(k)}= 5000$ when $K=15$ showing (a) MLP SNR=40dB and (b) MLP SNR=20dB.} \label{fig:confmatrix15class}
\end{figure}

Next, we consider the frequentist
frequentist approximations to 
$p(C_j |{\mathbf x})$ for $({\mathbf x},{\mathbf t})\in D^{(\text{test},(i))}$ presented in the form of a confusion matrix with entries $({\mathbf C})_{ij}$, $i,j=1,\ldots,K$,
for the cases of SNR=40dB and SNR=20db and the MLP classifier, in Figure~\ref{fig:confmatrix15class}. We do not show the results for the gradient boost as it has a near perfect identity confusion matrix on this scale for these noise levels.
Compared to the corresponding results shown in  Figure~\ref{fig:confmatrix8class} for $D_8$, the results for $D_{15}$ show the ability of the classifier to better discriminate between different objects. However, MLP still shows significant misclassifications for pendents whereas gradient boost does not.

In Table~\ref{tab:15-classprecision}, we show the  precision, sensitivity and specificity for each of the different object classes $C_k$, $k=1,\ldots,K,$ for the case of SNR=20dB and the MLP classifier. In general, we see the proportion of negatives that are correctly identified (as indicated by the specificity) is very high and is close to $1$ in all cases. The proportions of positives correctly identified (indicated by the precision and sensitivity) varies amongst the different object  classes, but is generally much closer to 1 than shown in Table~\ref{tab:8-classprecision} for $D_8$. The classes $C_1$, $C_2$ and $C_3$ (guns, knives and knuckle dusters in $D^{(15)}$), which make up the amalgamated weapons class $C_1$ in  $D_8$, all perform very well, but the worst case still remains $C_{12}$ (the pendents). The corresponding results for precision, sensitivity and specificity for the gradient boost classifier are all close to $1$.

\begin{table}
\begin{center}
\begin{tabular}{|c|c|c|c|}
\hline
$C_k$ & Precision &  Sensitivity & Specificity \\
\hline
Guns       $(C_1)$      &  1.00  &  1.00  &  1.00 \\\hline
Hammers $(C_2)$         &  1.00  &  1.00  &  1.00 \\\hline
Knives     $(C_3)$      &  1.00  &  1.00  &  1.00 \\\hline
Knuckle dusters  $(C_4)$ &  1.00  &  1.00  &  1.00 \\\hline
Screwdrivers $(C_5)$    &  0.95  &  0.96  &  1.00 \\\hline
Scissors   $(C_6)$      &  1.00  &  1.00  &  1.00 \\\hline
\hline
Bracelets      $(C_7)$  &  0.83  &  0.85  &  0.99 \\\hline
Belt buckles  $(C_8)$   &  0.99  &  0.98  &  1.00 \\\hline
Coins       $(C_9)$     &  0.70  &  0.72  &  0.98 \\\hline
Earrings   $(C_{10})$      &  0.75  &  0.75  &  0.98 \\\hline
Keys          $(C_{11})$   &  0.90  &  0.95  &  0.99 \\\hline
Pendants     $(C_{12})$    &  0.67  &  0.53  &  0.98 \\\hline
Rings        $(C_{13})$    &  0.71  &  0.77  &  0.98 \\\hline
Shoe shanks $(C_{14})$     &  0.99  &  1.00  &  1.00 \\\hline
Watches      $(C_{15})$    &  0.99  &  0.99  &  1.00 \\\hline
\end{tabular}
\end{center}
\caption{Set of multiple threat and non-threat objects: Precision, sensitivity and specificity measures (to 2d.p.) for each of the classes $C_k$ when SNR=20db and $P^{(k)}= P/K =5000$ for the MLP classifier.} \label{tab:15-classprecision}
\end{table}

\subsubsection{Classification of unseen objects using $D_8$}\label{sect:Leave_One_Out_Results}

When testing the performance of classifiers in the previous sections, the construction of the dictionary, described in Section~\ref{sect:multidata}, means that $D^{(\text{train})}$ and $D^{(\text{test})}$ are both comprised of  samples that have MPT spectral signatures associated  with objects that share the same geometry and have similar object sizes and material parameters. To illustrate the ability of a classifier to recognise an unseen threat object, we construct  $D_8^{(\text{train})}$, as in Section~\ref{sect:multidata}, except, for one class $C_k$, where we replace 
 $D_8^{(\text{train},(k) )}$ with 
data that  is obtained from $G^{(k)}-1$ (instead of $G^{(k)}$) geometries  and $V^{(k)}$ samples. Also, we use $P^{(k)}=2000$, instead of  $P^{(k)}=5000$, due to the higher computational cost of the investigation  presented in the following. We proceed to  test the classifier using a sample that is constructed only from $V^{(k)}$ samples of the unseen $G^{(k)}$th geometry.

\begin{figure}[h]
\begin{center}
\includegraphics[scale=0.5]{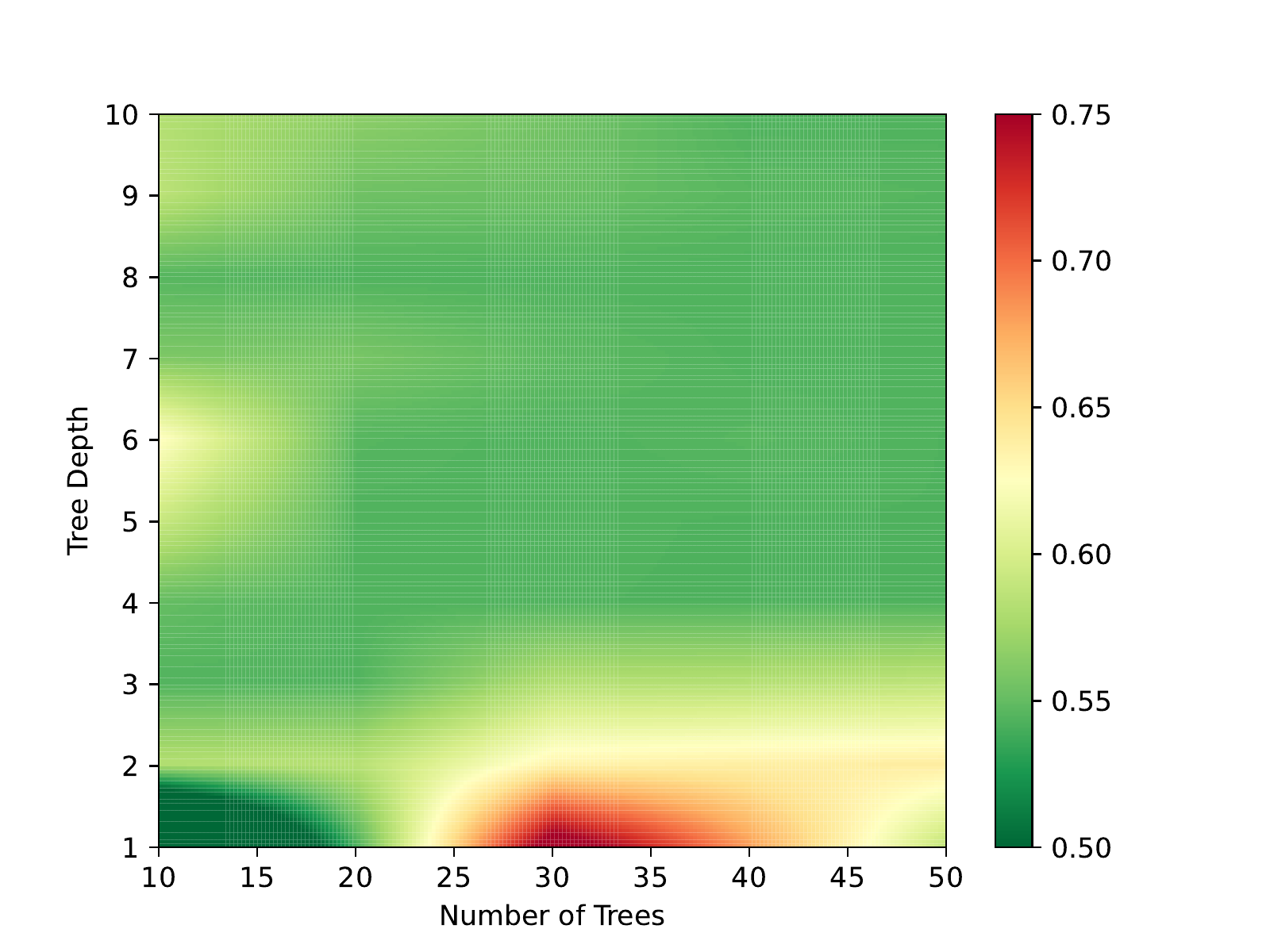}
\end{center}
\caption{Set of multiple threat and non-threat objects: Overall performance of the gradient boost classifier for different values in the hyperparameter space, with $P^{(k)}=2000$ when $K=8$ and SNR=40dB, showing average $\kappa$ score for different values of numbers of trees (\texttt{n\_estimators}) and tree depth (\texttt{max\_depth}).}
\label{fig:knife_parameter_space}
\end{figure}

We focus on removing a geometry from the $C_2$ class of weapons, which originally has $G^{(2)}=8$ geometries, and   we vary the unseen geometry  to be one of the  chef, cutlet, meat cleaver, Santoku, and Wusthof knives, shown in Figure~\ref{fig:egthreat}, where the naming convention from Section 6.4 of ~\cite{ledgerwilsonamadlion2021} is adopted.
We apply the gradient boost classifier to this problem, as it was seen  to perform best for both the $K=8$ and $K=15$ class problems. Previously, the hyperparameters \texttt{n\_estimators=50} and \texttt{max\_depth=2}  have been shown to lead to accurate results. However, this problem is more challenging,  as it involves attempting to classify data from the samples in $({\mathbf x},{\mathbf t}) \in D_8^{(\text{test},(2)) }$ that are only constructed from samples of the unseen $G^{(2)}$th geometry, and, therefore, the previous hyperparameters are no longer optimal. This is illustrated in Figure~\ref{fig:knife_parameter_space} for the case where SNR=40dB and, here, the average $\kappa$ score obtained  from considering the situations when instances of the  chef, cutlet, meat cleaver, Santoku, and Wusthof knife geometries as being unseen is presented. This suggests the optimal performance will be for a very limited region where \texttt{n\_estimators} $\approx 30$ and \texttt{max\_depth=1} and, away from this, the performance of the classifier will be poor.

The poor performance of the gradient boost classifier for this problem for a large range of hyperparameters  is due to its inability to correctly classify the cutlet knife geometry, with the classifier instead predicting this as a tool rather than a weapon in the majority of cases (indicating bias against this geometry) and, additionally, for other geometries, the relatively high degree of uncertainty that is associated with $\gamma_2( {\mathbf x})$ despite $\gamma_{2,50} $ being high (indicating a high variance). This can further be explained by the comparison of the knife volumes using a fixed $\alpha=0.001$m shown in Table~\ref{tab:knife_volumes}, where it can be seen that the cutlet knife geometry  has a volume that is an order of magnitude smaller than that of the knives. The MPT spectral signature depends on the object's volume as well as its materials and geometry and, as each cutlet knife  tends to be associated 
smaller volumes to those considered in $D_8^{(\text{train})}$ this has contributed to the classifier not being able to recognise it.
\begin{table}[!h]
\begin{center}
\begin{tabular}{!\vrule l!\vrule l!\vrule }
\hline
Knife & Volume ($\text{m}^3$)\\\hline
Chef & $1.46\times 10^{-5}$\\\hline%1.4594225029634278e-05
Cutlet & $3.28\times 10^{-6}$\\\hline%3.2778394362872175e-06
Meat cleaver & $6.50\times 10^{-5}$\\\hline%6.502589375583549e-05
Santoku & $2.51\times 10^{-5}$\\\hline%2.5057487068299797e-05
Wusthof & $3.48\times 10^{-5}$\\\hline%3.48280668249006e-05
\end{tabular}
\end{center}
\caption{Set of multiple threat and non-threat objects: Comparison of volumes for different knife models.}\label{tab:knife_volumes}
\end{table}

\begin{figure}[!h]
$$\begin{array}{cc}
\includegraphics[scale=0.5]{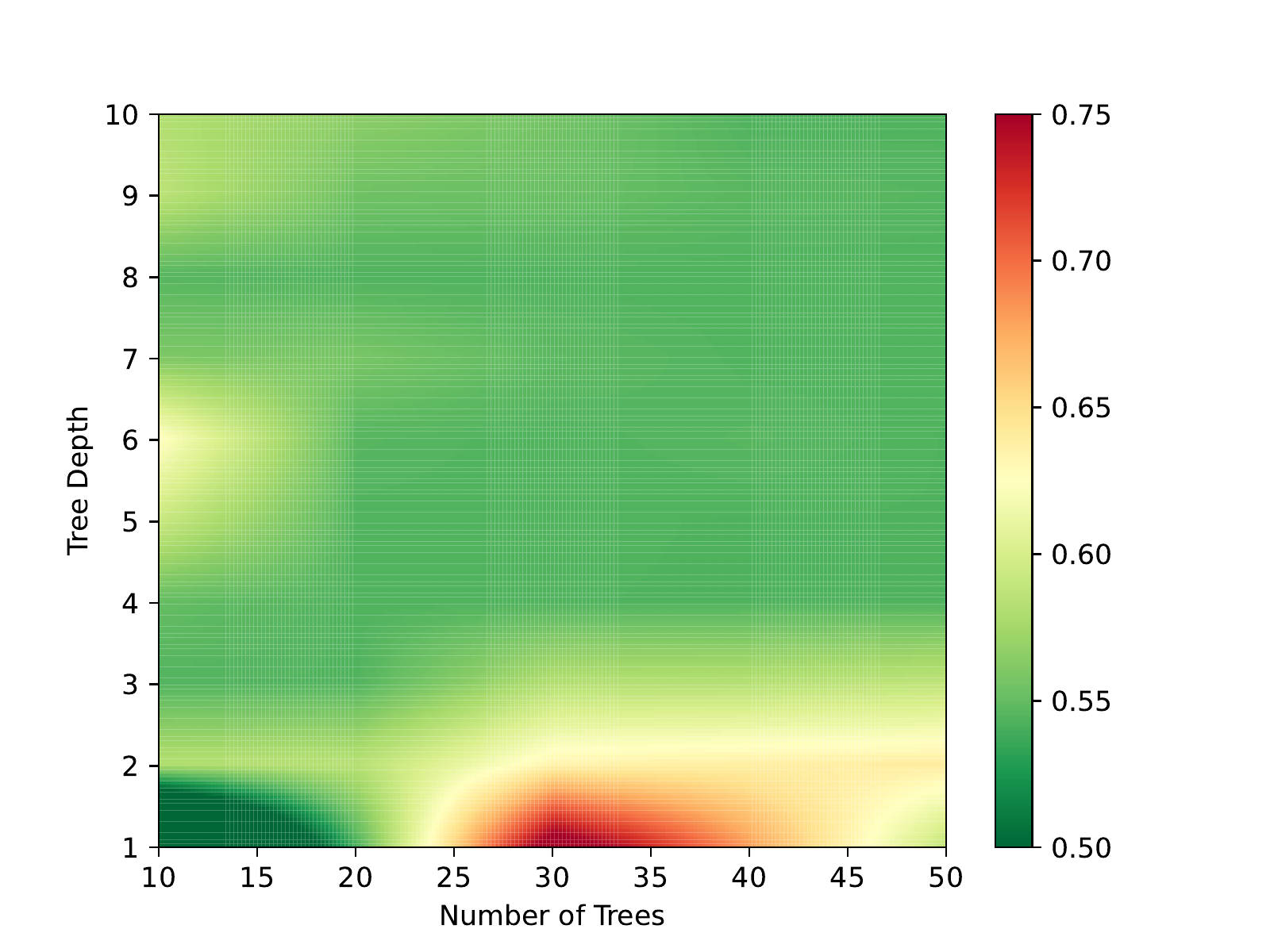} &
 \includegraphics[scale=0.5]{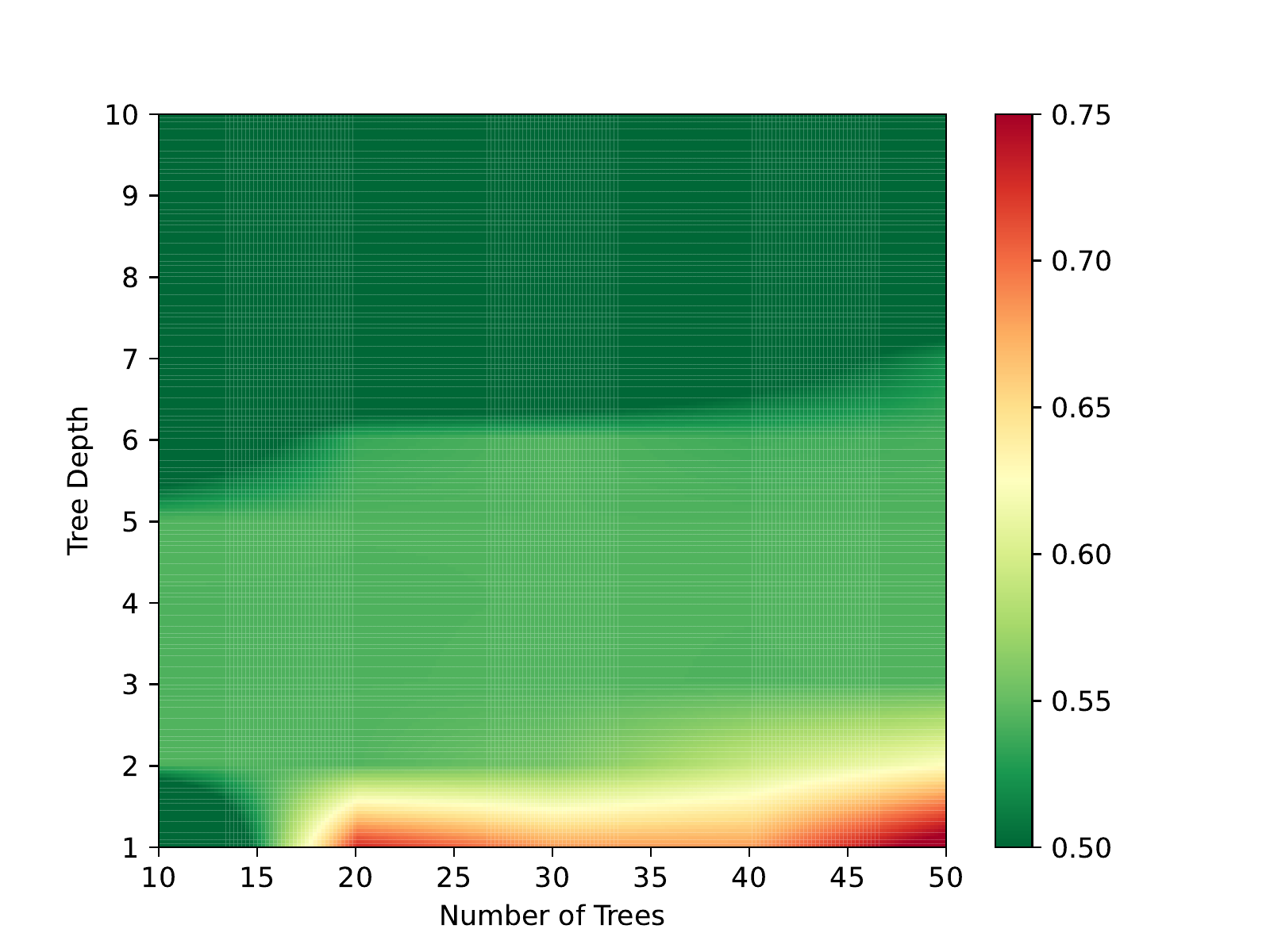}\\
\textrm{\footnotesize{(a) Control}} & \textrm{\footnotesize{(b) A}}\\
\includegraphics[scale=0.5]{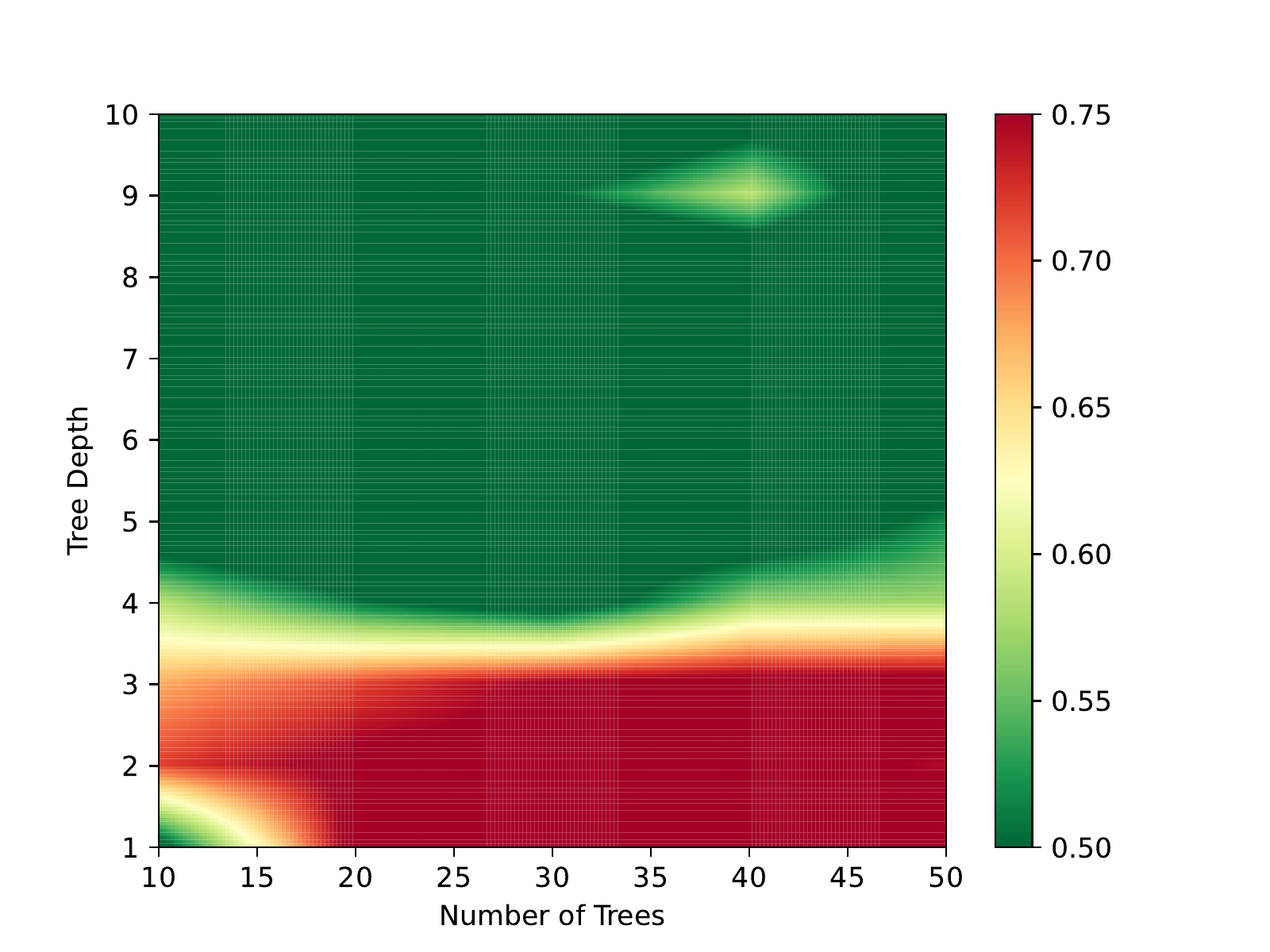} &
 \includegraphics[scale=0.5]{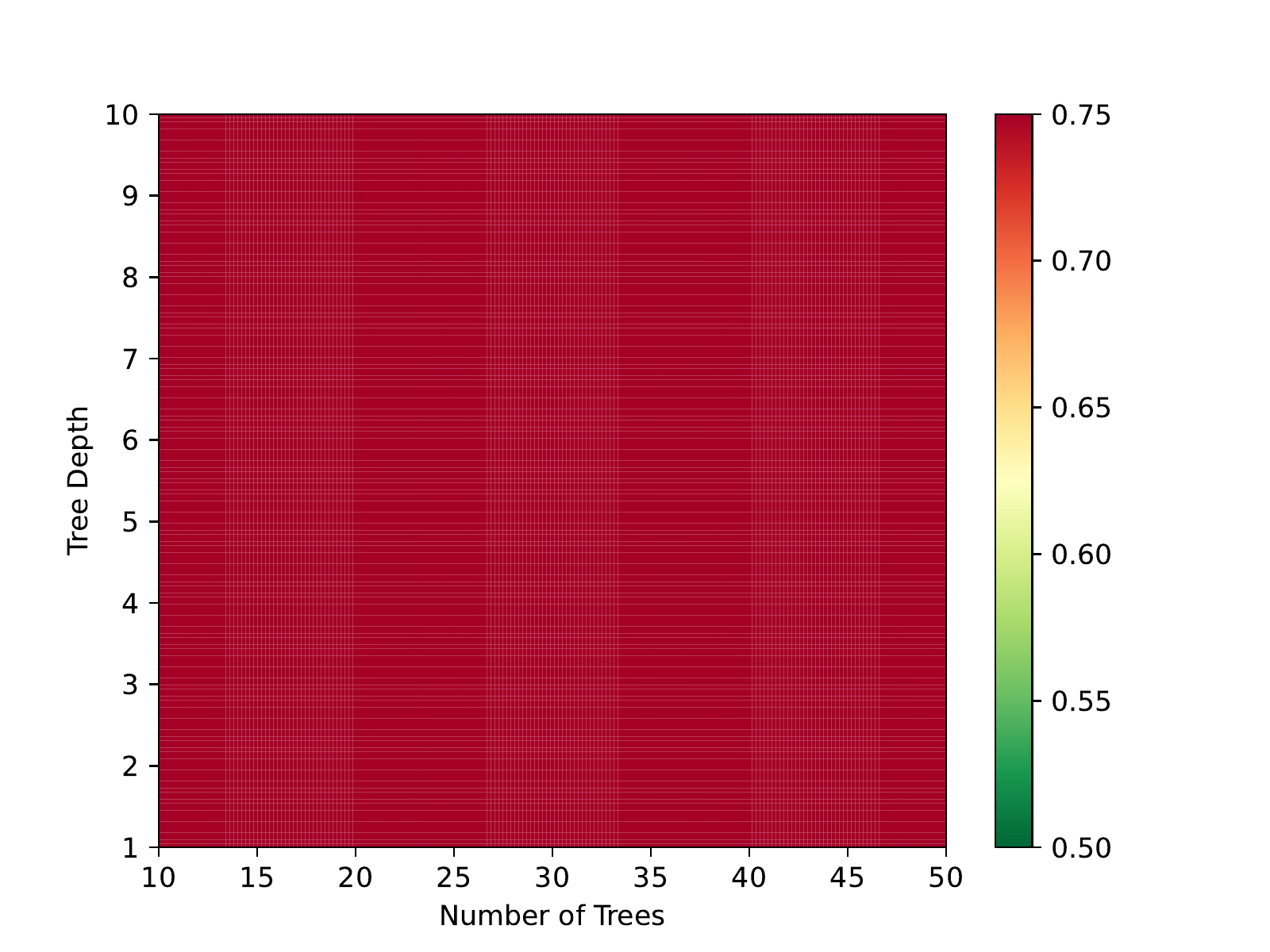}\\
\textrm{\footnotesize{(c) B}} & \textrm{\footnotesize{(d) C}}
\end{array}$$
\caption{Set of multiple threat and non-threat objects: Overall performance of the gradient boost classifier for different values in the hyperparameter sapce, with $P^{(k)}=2000$ with $K=8$ and SNR=40dB, showing average $\kappa$ score for different values \texttt{n\_estimators} and \texttt{max\_depth}, for different scaling regimes (a) Control, (b) A, (c) B and (d) C.} \label{fig:knife_scaling_parameter_space}
\end{figure}

\begin{figure}[!h]
\begin{center}
$\begin{array}{cc}
 \includegraphics[scale=0.5]{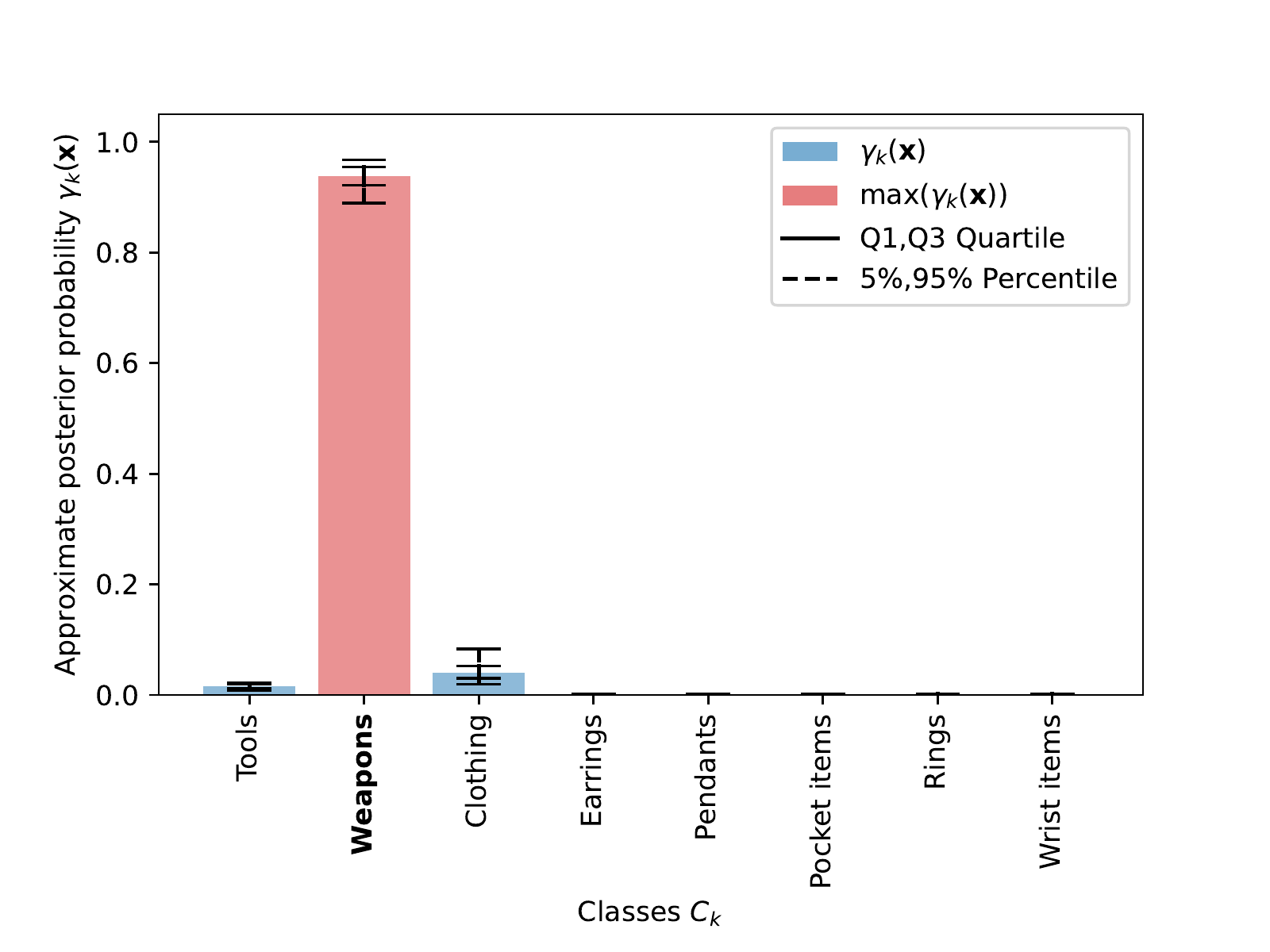} &
 \includegraphics[scale=0.5]{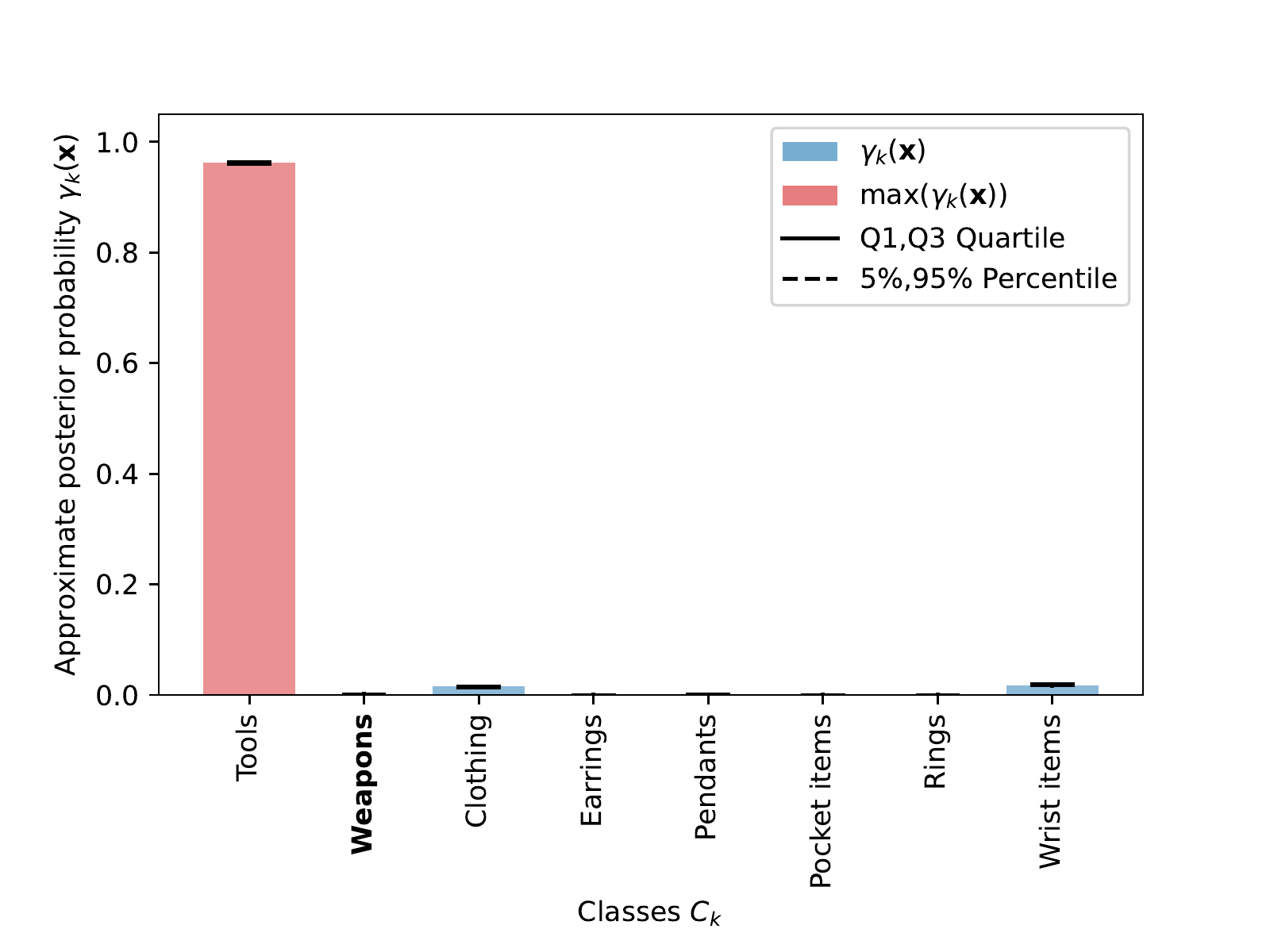}\\
\textrm{\footnotesize{(a) $D^{(\text{test})}=$ Chef}} & \textrm{\footnotesize{(b) $D^{(\text{test})}=$ Cutlet}}\\
 \includegraphics[scale=0.5]{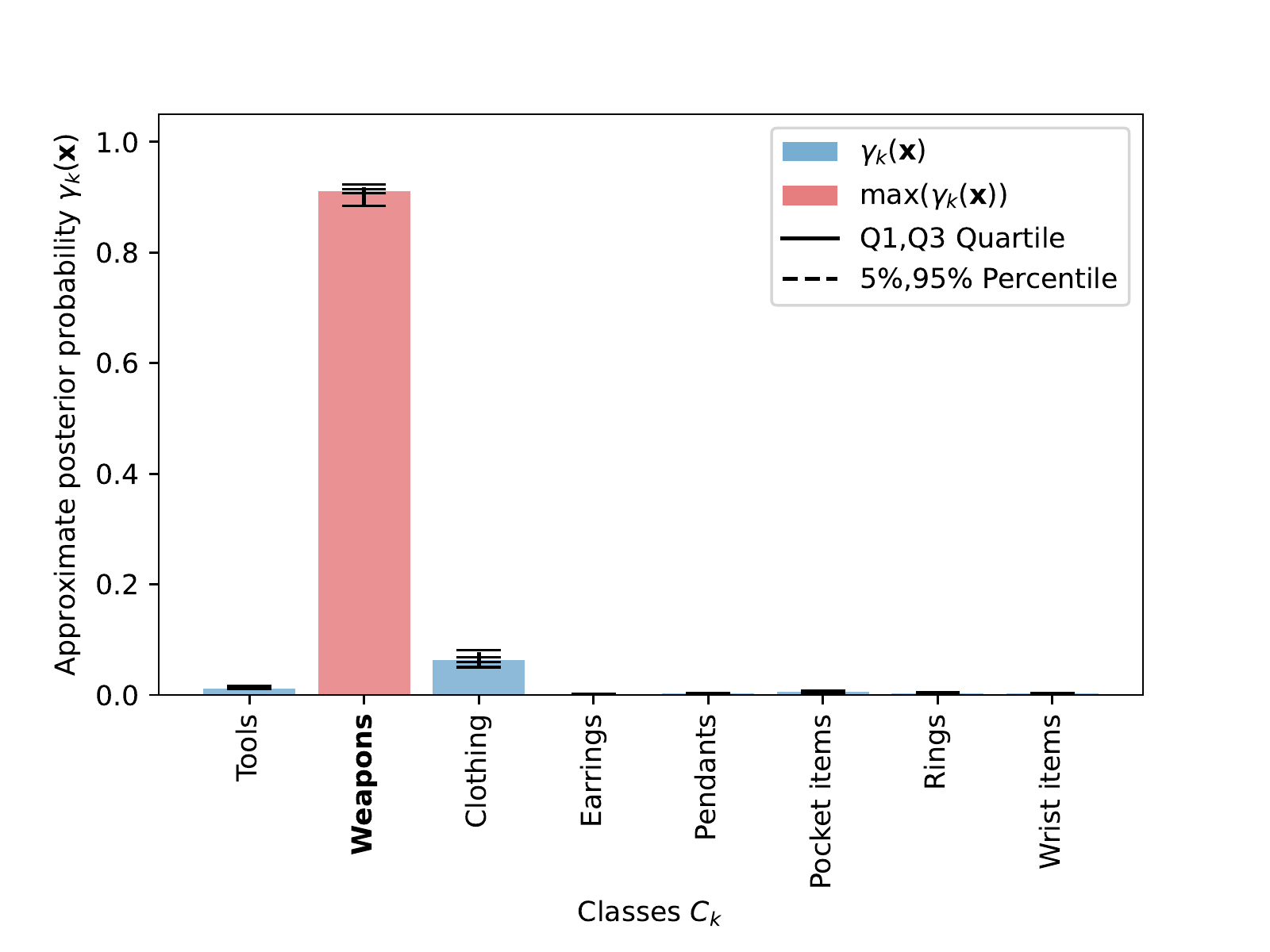} &
 \includegraphics[scale=0.5]{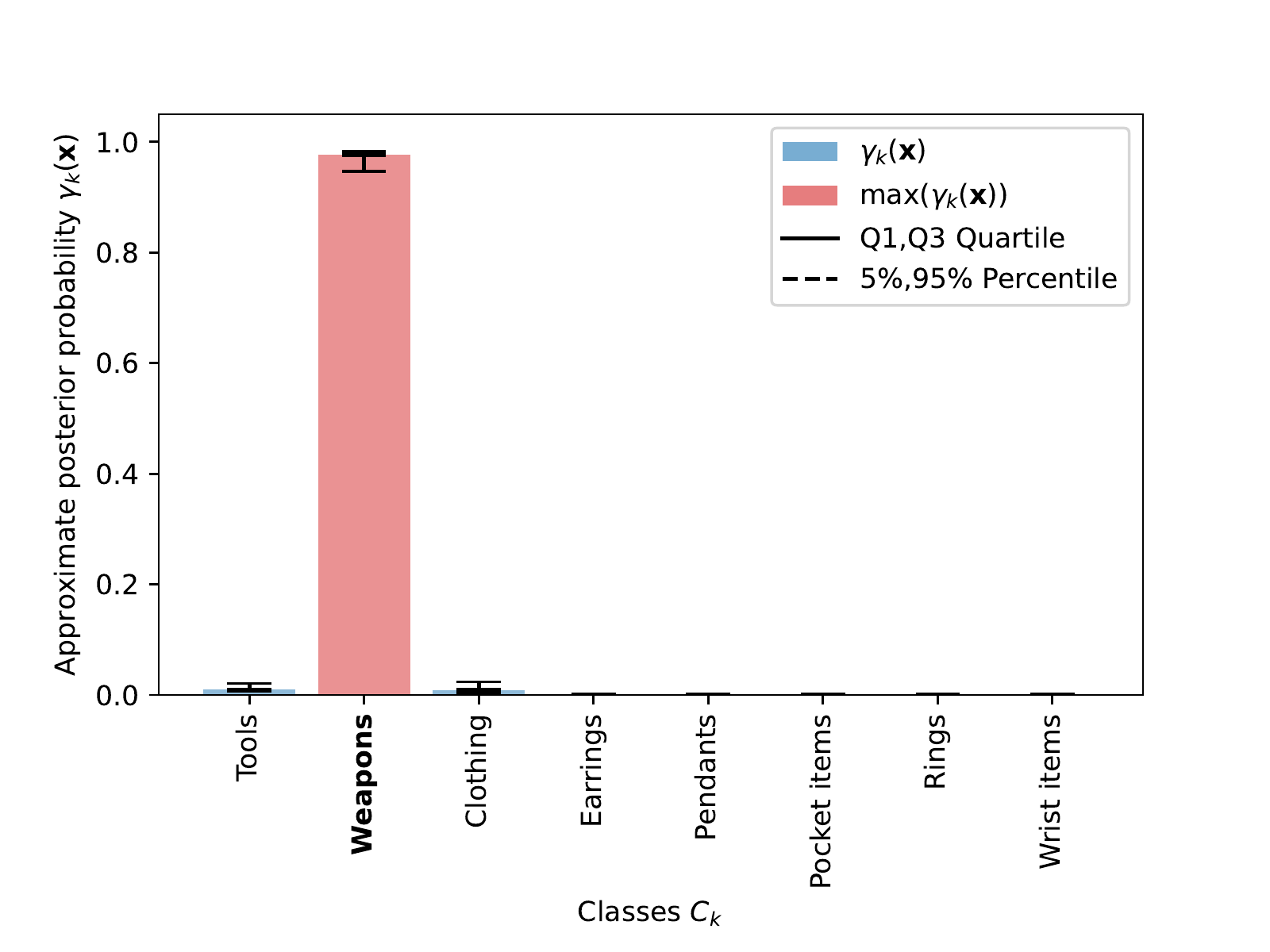}\\
\textrm{\footnotesize{(c) $D^{(\text{test})}=$ meat cleaver}} & \textrm{\footnotesize{(d) $D^{(\text{test})}=$ Santoku}}
 \end{array}$
 $\begin{array}{c}
 \includegraphics[width=0.5\textwidth, keepaspectratio]{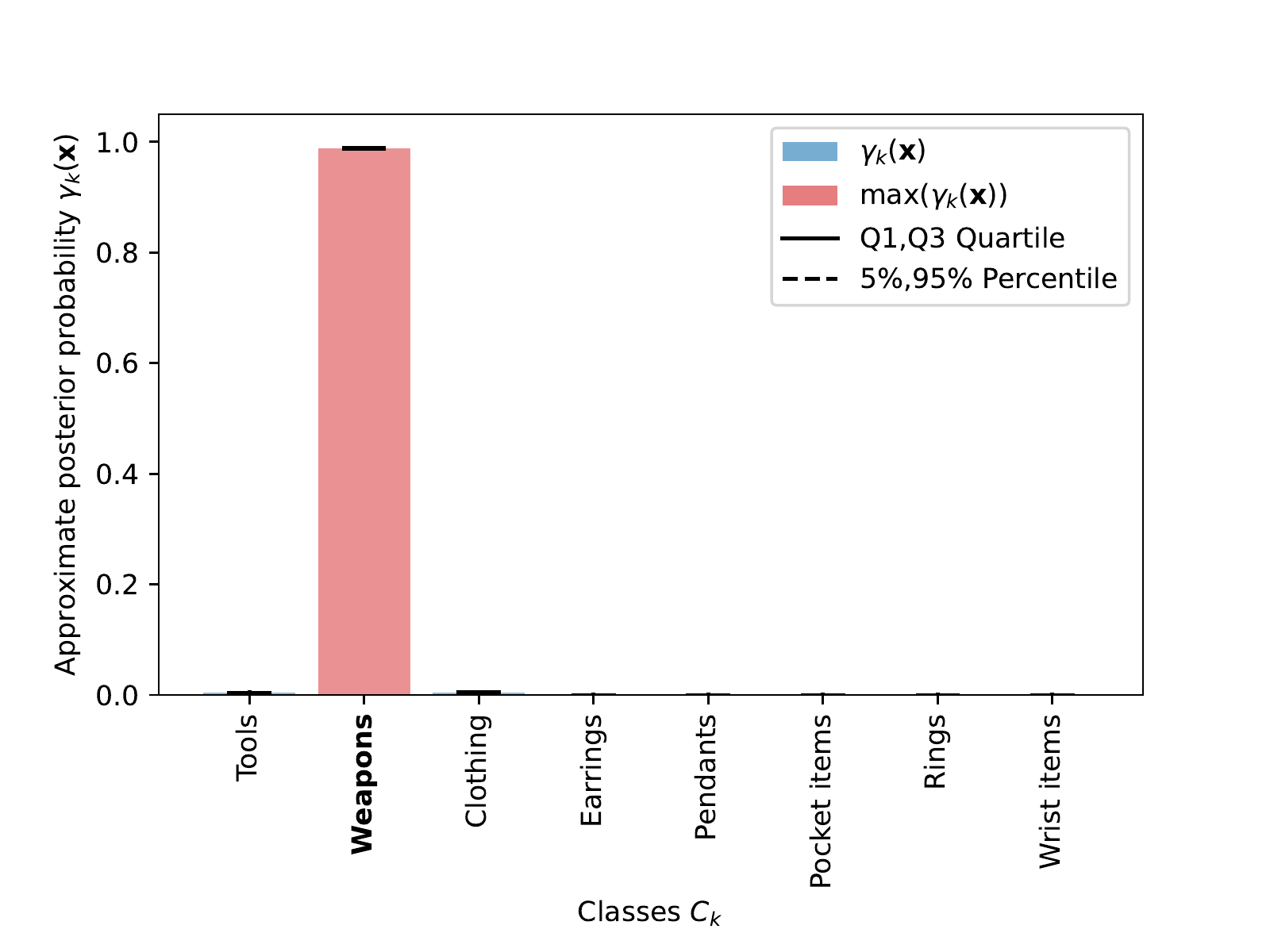}\\
\textrm{\footnotesize{(e) $D^{(\text{test})}=$ Wusthof}} 
 \end{array}$
 \end{center}
\caption{Set of multiple threat and non-threat objects: Approximate 
posterior probabilities $\gamma_k({\mathbf x}) \approx p(C_k|\mathbf{x})$, $k=1,\ldots,K,$ using the gradient boost classifier for $P^{(k)}=2000$ when $K=8$ and SNR=40dB showing the case when $D^{(\text{train})}$ is constructed using the scaling regime C and cases where $D^{(\text{test})}$ is constructed of instances (a) chef, (b) cutlet, (c) meat cleaver, (d) Santoku and (e) Wusthof.} \label{fig:knife_Posteriors_large_scaling}
\end{figure}

The situation can be improved by increasing the standard deviations $s_\alpha$ and $s_{\sigma_*}$,  so that $D_8^{(\text{train}, (2)) }$ includes MPT spectral signatures that are closer to that of the omitted $G^{(2)}$th geometry. In Table~\ref{tab:distribution_values}, we consider three alternatives A, B and C to the previous control choice. Then, in Figure~\ref{fig:knife_scaling_parameter_space}, we repeat the investigation shown in Figure~\ref{fig:knife_parameter_space} for cases A, B and C.  In this figure we observe that the classifier has less variability and performs increasingly well over a wide range of hyperparameters, as $s_\alpha$ and $s_{\sigma_*}$  are increased. In the limiting case of $C$, the overall performance of the classifier is uniform with  $\kappa=0.75$ over the complete space of hyperparameters considered. 
\begin{table}[!h]
\begin{center}
\begin{tabular}{!\vrule l!\vrule l!\vrule l!\vrule }
\hline
scaling regime & $s_{\alpha}$ & $s_{\sigma_*}$\\\hline
Control & $0.0084m_{\alpha}$ & $0.0236333m_{\sigma_*}$\\\hline
A & $0.02m_{\alpha}$ & $0.05m_{\sigma_*}$\\\hline
B & $0.05m_{\alpha}$ & $0.1m_{\sigma_*}$\\\hline
C & $0.1m_{\alpha}$ & $0.2m_{\sigma_*}$\\\hline
\end{tabular}
\end{center}
\caption{Set of multiple threat and non-threat objects: List of the parameters for the sampling distributions considered.}\label{tab:distribution_values}
\end{table}
Furthermore,  in Figure \ref{fig:knife_Posteriors_large_scaling}, we show, for case C, the approximate posterior probability distributions $\gamma_k ({\mathbf x}) \approx p(C_k|\mathbf{x})$, $k=1,\ldots,K,$ obtained for the case where the training samples are taken as $({\mathbf x}, {\mathbf t})\in D_8^{(\text{train}, (2))}$, with either the chef, cutlet, meat cleaver, Santoku or Wusthof  geometry being treated as unseen, in turn.  These results were obtained with  \texttt{n\_estimators}=50 and \texttt{max\_depth}=2 with SNR=40dB. From  this figure we observe that $\gamma_{2,50}\approx 1$ for the unseen chef, meat cleaver, Santoku or Wusthof knives suggesting the most likely class is $C_2$ (a weapon)
 with  small interpercentile and interquartile ranges, which indicates a high degree of certainty associated with the prediction and also a low variability.
  However, when the unseen object is a cutlet knife, $\gamma_{1,50} \approx1$ with small interpercentile and interquartile ranges, 
  which indicates that the classifier is still consistently misclassifying this object  as a tool, instead of a weapon, despite the classifier being trained over a wider range of object sizes and conductivities. Hence, the classifier remains  biased against this geometry. We conjecture this is due to the significant difference in the shape of the MPT spectral signature for the cutlet knife geometry shown in Figure 33 of~\cite{ledgerwilsonamadlion2021}, compared to the other knives and gun geometry on which the classifier is trained.

\section{Conclusion}

This paper has presented a  novel approach to training ML classifiers using our simulated  \text{MPT-Library}, which has been enhanced by simple scaling results to create large dictionaries of object characterisations at a low computational cost. We have employed  tensor invariants of MPT spectral signatures as novel object features for training ML classifiers and considered a range of both probabilistic and non-probabilistic classifiers. We have presented
a well-reasoned approach for justifying the performance of different ML classifiers for practical classifications problems using both uncertainty quantification using statistical analysis and ML metrics. Furthermore, we have explored the ability of our classification approaches to classify unseen threat objects.

For the $K=8$ class coin classification problem, we have found that the logistic regression classifier performs well to discriminate between different denominations of British coins. We have seen significant benefits from increasing the number of frequencies considered in the MPT spectral signature as well as increasing the size of the training data set, which all led to an improvement of the accuracy of the classifier by reductions in its variability.  A classifier of this type could help with the automated sorting of coins and in fraud detection. It also has potential applications in coin  counting. Our results are useful for coin designers as it could help them to design new coins that have greater differences in their spectral signature to make them easier to classify.

For the mutli-class problem, involving the discrimination between threat and non-threat objects, we have found improvements in the accuracy of the classification by using MPT spectral signatures over a larger range of frequencies compared to a narrow range, and, once again, also by  increasing the size of the training data set. The best performing probabilistic classifier for both the $K=8$ and $K=15$ class classification problems being the gradient boost algorithm. The gradient boost algorithm is also seen to perform well on the classification of unseen objects, provided the training set contains sufficiently similar MPT spectral signatures from other objects. This classifier could help in security screening applications such at transport hubs as well as in parcels transportation.

\section*{Acknowledgement}
Ben A. Wilson gratefully acknowledges the financial support received from EPSRC in the form of a DTP studentship with project reference number 2129099. 
Paul D. Ledger gratefully acknowledges the financial support received from EPSRC in the form of grants EP/R002134/2, EP/V049453/1 and EP/V009028/1. William R. B. Lionheart gratefully acknowledges the financial support received from EPSRC in the form of grants EP/R002177/1, EP/V049496/1 and EP/V009109/1 and would like to thank the Royal Society for the financial support received from a Royal Society Wolfson Research Merit Award.

\bibliographystyle{acm}
\bibliography{paperbib}
 
 \end{document}